\documentclass[10pt,journal,compsoc]{IEEEtran}

\ifCLASSOPTIONcompsoc
  \usepackage[nocompress]{cite}
\else
  \usepackage{cite}
\fi

\usepackage{amsfonts}       
\usepackage{nicefrac}       
\usepackage{microtype}      
\usepackage{xcolor}         
\usepackage{graphicx}
\usepackage{wrapfig}
\usepackage{tabu}
\usepackage{url}
\usepackage{amsmath}
\usepackage{amssymb}
\usepackage{amsthm}
\usepackage{algorithm}
\usepackage{algorithmic}
\usepackage{listings}
\usepackage{multirow}
\usepackage{hyperref}
\definecolor{codegreen}{rgb}{0,0.6,0}
\definecolor{codegray}{rgb}{0.5,0.5,0.5}
\definecolor{codepurple}{rgb}{0.58,0,0.82}
\definecolor{backcolour}{rgb}{0.95,0.95,0.92}
\lstdefinestyle{mystyle}{
    backgroundcolor=\color{backcolour},   
    commentstyle=\color{magenta},
    keywordstyle=\color{codegreen},
    keywordstyle = [2]{\color{magenta}},
    numberstyle=\tiny\color{codegray},
    stringstyle=\color{codepurple},
    basicstyle=\ttfamily\footnotesize,
    breakatwhitespace=false,         
    breaklines=true,                 
    captionpos=b,                    
    keepspaces=true,                 
    numbers=left,                    
    numbersep=5pt,                  
    showspaces=false,                
    showstringspaces=false,
    showtabs=false,                  
    tabsize=2,
    otherkeywords={numpy,scipy}
}

\newtheorem{theorem}{Theorem}
\newtheorem{observation}{Observation}

\newtheorem{corollary}{Corollary}

\begin{document}
%
\title{Dynamic Time Warping based Adversarial Framework for Time-Series Domain}
%
%
%
%

\author{Taha~Belkhouja,
        ~Yan~Yan,~\IEEEmembership{Member,~IEEE}
        and~Janardhan~Rao~Doppa,~\IEEEmembership{Senior Member,~IEEE}    
\IEEEcompsocitemizethanks{\IEEEcompsocthanksitem The authors are with the School of Electrical Engineering and Computer Science, Washington State University, Pullman, WA, 99164. E-mail: \{taha.belkhouja,yan.yan1,jana.doppa\}@wsu.edu}
\thanks{Manuscript received April 19, 2005; revised August 26, 2015.}}

\markboth{Journal of \LaTeX\ Class Files,~Vol.~14, No.~8, August~2015}%
{Shell \MakeLowercase{\textit{et al.}}: Bare Demo of IEEEtran.cls for Computer Society Journals}
%



\IEEEtitleabstractindextext{%
\begin{abstract}
Despite the rapid progress on research in adversarial robustness of deep neural networks (DNNs), there is little principled work for the time-series domain. Since time-series data arises in diverse applications including mobile health, finance, and smart grid, it is important to verify and improve the robustness of DNNs for the time-series domain. In this paper, we propose a novel framework for the time-series domain referred as {\em Dynamic Time Warping for Adversarial Robustness (DTW-AR)} using the dynamic time warping measure. Theoretical and empirical evidence is provided to demonstrate the effectiveness of DTW over the standard Euclidean distance metric employed in prior methods for the image domain. We develop a principled algorithm justified by theoretical analysis to efficiently create diverse adversarial examples using random alignment paths. Experiments on diverse real-world benchmarks show the effectiveness of DTW-AR to fool DNNs for time-series data and to improve their robustness using adversarial training. 
\end{abstract}

\begin{IEEEkeywords}
Time Series, Robustness, Deep Neural Networks, Adversarial Examples, Dynamic Time Warping.
\end{IEEEkeywords}

}

\maketitle

\IEEEdisplaynontitleabstractindextext

%
\IEEEpeerreviewmaketitle

\IEEEraisesectionheading{\section{Introduction}\label{sec:introduction}}

\IEEEPARstart{T}{o} deploy deep neural network (DNN) based systems in important real-world applications such as healthcare, we need them to be robust \cite{madry2017towards,fawaz2019adversarial,karim2020adversarial}. Adversarial methods expose the brittleness of DNNs \cite{karim2020adversarial,oregiSPL18} and motivate methods to improve their robustness. There is little principled work for the time-series domain \cite{TCAD} even though this type of data arises in many real-world applications including mobile health \cite{ignatov2018real}, finance \cite{ozbayoglu2020deep}, and smart grid analytics \cite{zheng2017wide}.  The time-series modality poses unique challenges for studying adversarial robustness that are not seen in images \cite{kolter2018materials} and text \cite{wang2019deep}. The standard approach of imposing an $l_p$-norm bound to create worst possible scenarios from a learning agent's perspective does not capture the true similarity between time-series instances. Consequently, $l_p$-norm constrained perturbations can potentially create adversarial examples that correspond to a completely different class label. There is no prior work on filtering methods in the signal processing literature to {\em automatically} identify such adversarial candidates. Hence, adversarial examples from prior methods based on $l_p$-norm will confuse the learner when they are used to improve the robustness of DNNs. In other words, the accuracy of DNNs will degrade on real-world data after adversarial training. 

This paper proposes a novel adversarial framework for time-series domain referred as {\em Dynamic Time Warping for Adversarial Robustness (DTW-AR)} to address the above-mentioned challenges. DTW-AR employs the dynamic time warping measure \cite{sakoe1971dynamic,muller2007dynamic} as it can be used to measure a realistic distance between two time-series signals (e.g., invariance to shift and scaling operations) \cite{berndt1994using,muller2007dynamic}. For example, a signal that has its frequency changed due to Doppler effect would output a small DTW measure to the original signal. However, if Euclidean distance is used, both signals would look very dissimilar, unlike the reality. We theoretically analyze the suitability of DTW measure over the Euclidean distance. Specifically, the space of candidate adversarial examples in the DTW space is a superset of those in Euclidean space for the same distance bound. Therefore, DTW measure provides a more appropriate bias than the Euclidean space for the time series domain and our experiments demonstrate practical benefits of DTW-based adversarial examples. While certain time-series classification tasks can be solved using low-complexity algorithms such as 1NN-DTW and avoid the adversarial setting, we find that deep models are better suited for multivariate time-series data. Due to the rising complexity of time-series data in several applications (e.g., mobile health \cite{ignatov2018real}, Human activity recognition \cite{ramanujam2021human}, or sleep monitoring \cite{zhao2017learning}), low-complexity algorithms such as kNN-DTW can potentially perform badly on high-dimensional multivariate data as we demonstrate in Section \ref{sec:knndtw}. Therefore, the adversarial setting remains applicable for time-series domain.

To create targeted adversarial examples, we formulate an optimization problem with the DTW measure bound constraint and propose to solve it using an iterative gradient-based approach. However, this simple method has two drawbacks. First, this method allows us to only find one valid adversarial example out of multiple solution candidates from the search space because it operates on a single optimal alignment. Second, we need to compute DTW measure in each iteration as the optimal DTW alignment path changes over iterations. Since the number of iterations are typically large and DTW computation is expensive, the overall algorithm becomes prohibitively slow. To successfully overcome these two drawbacks, {\em our key insight is to employ stochastic alignments to create adversarial examples}. We theoretically and experimentally show that a simpler distance measure based on random alignment path upper-bounds the DTW measure measure and that this {\em bound is tight}. This algorithm allows us to efficiently create many diverse adversarial examples using different alignment paths to improve the robustness of DNN models via adversarial training. Our experiments on real-world time-series datasets show that the DTW-AR creates more effective adversarial attacks to fool DNNs when compared to prior methods and enables improved robustness.

\vspace{1.0ex}

\noindent {\bf Contributions.} The key contribution of this paper is the development and evaluation of the DTW-AR framework for studying adversarial robustness of DNNs for time-series domain. Specific list includes:
\begin{itemize}
    \item Theoretical and empirical analysis to demonstrate the effectiveness of DTW over the standard $l_2$ distance metric for adversarial robustness studies.
    \item Principled algorithm using DTW measure to efficiently create diverse adversarial examples via random alignment paths justified by theoretical analysis.
    \item Experimental evaluation of DTW-AR on diverse real-world benchmarks and comparison with state-of-the-art baselines. The  source  code  of DTW-AR algorithms is available at \href{https://github.com/tahabelkhouja/DTW-AR}{https://github.com/tahabelkhouja/DTW-AR}
\end{itemize}

\section{Background and problem setup}

\label{sec:background}

Let $X\in \mathbb{R}^{n \times T}$ be a multi-variate time-series signal, where $n$ is the number of channels and $T$ is the window-size of the signal. We consider a DNN classifier $F_{\theta}: \mathbb{R}^{n \times T} \rightarrow \mathcal{Y}$, where $\theta$ stands for parameters and $\mathcal{Y}$ is the set of classification labels. Table \ref{tab:notation} summarizes the different mathematical notations used in this paper.

\begin{table}[!h]
\centering
\caption{Mathematical notations used in this paper.}
\begin{tabular}{|c|l|} 
\hline
\textbf{Variable}  & \textbf{Definition} \\ \hline 
$F_{\theta}$ & DNN classifier with parameters $\theta$ \\ \hline
$\mathbb{R}^{n \times T}$  & Time-series input space, where $n$ is the number \\ & of channels and $T$ is the window-size\\ \hline 
$X_{adv}$ & Adversarial example generated from time-series \\ & input $X \in \mathbb{R}^{n \times T}$ \\ \hline 
$\mathcal{Y}$ & Set of output class labels\\ \hline 
$DTW(\cdot, \cdot)$ & Dynamic time warping distance \\ \hline 
$P$ & Alignment path: a sequence of cost matrix cells \\ & $\{(i,j)\}_{i\le T, j\le T}$\\ \hline 
$C$ & Alignment cost matrix generated by dynamic \\ & programming with elements $C_{i,j}$ \\ \hline 
$\delta$ & Distance bound constraint \\ \hline 
\end{tabular}
\label{tab:notation}
\end{table}

$X_{adv}$ is called an adversarial example of $X$ if:
\begin{equation*}
\big\{X_{adv} ~\big|~ DIST(X_{adv},X) \le \delta \text{~and~} F_{\theta}(X) \neq F_{\theta}(X_{adv})\big\}
\end{equation*}
where $\delta$ defines the neighborhood of highly-similar examples for input $X$ using a distance metric $DIST$ to create worst-possible outcomes from the learning agent's perspective. Note that adversarial examples depend on the target concept because it defines the notion of invariance we care about.

\vspace{1.0ex}
\noindent \textbf{Challenges for time-series data.}
The standard $l_p$-norm distance does not capture the unique characteristics (e.g., fast-pace oscillations, sharp peaks) and the appropriate notion of invariance for time-series signals.
Hence, $l_p$-norm based perturbations can lead to a time-series signal that semantically belongs to a different class-label. Our experiments show that small perturbations result in adversarial examples whose $l_2$ distance from the original time-series signal is greater than the distance between time-series signals from two different class labels (see Section 5.2). Therefore, we need to study new methods by exploiting the structure and unique characteristics of time-series signals.

\begin{figure}[!h]
\centering
\includegraphics[width=.85\linewidth]{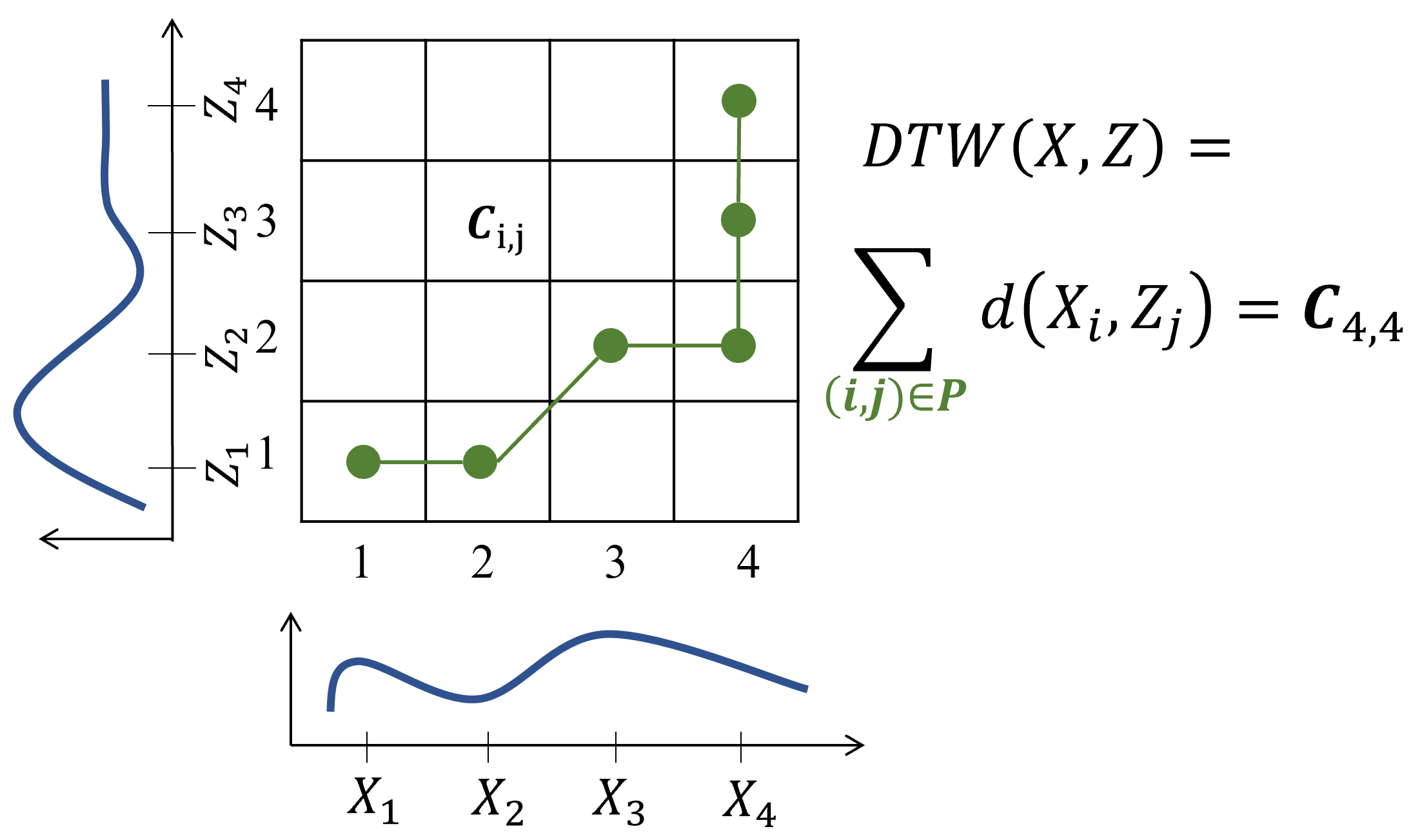}
\caption{Illustration of DTW alignment between two uni-variate signals $X$ and $Z$ of length 4. The optimal alignment path (shown in green color) is $P$ = $\{(1,1), (2,1), (3,2), (4,2), (4,3), (4,4)\}$.}
\label{fig:costmx}
\end{figure}

\vspace{1.0ex}
\noindent \textbf{DTW measure.}
The DTW measure between two uni-variate signals $X$ and $Z \in \mathbb{R}^{T}$ is computed via a cost matrix $C \in \mathbb{R}^{T \times T}$ using a dynamic programming (DP) algorithm with time-complexity $\mathcal{O}(T^2)$. The cost matrix is computed recursively using the following equation:
\begin{equation}
    C_{i,j} = d(X_i, Z_j) + \min \big\{C_{i-1,j}, C_{i,j-1}, C_{i-1,j-1} \big\}
    \label{eq:costmatrix}
\end{equation}
where $d(\cdot, \cdot)$ is any given distance metric (e.g., $\|\cdot\|_p$ norm). The DTW measure between signals $X$ and $Z$ is $DTW(X, Z)$ = $C_{T,T}$. The sequence of cells $P$ = $\{c_{i, j}$ = $(i, j)\}$ contributing to $C_{T,T}$ is the {\em optimal alignment path} between $X$ and $Z$. 
Figure \ref{fig:costmx} provides illustration for an optimal alignment path. We note that the diagonal path corresponds to the Euclidean distance metric.

For the multi-variate case, where $X$ and $Z \in \mathbb{R}^{n \times T}$, to measure the DTW measure using Equation \ref{eq:costmatrix}, we have $d(X_i, Z_j)$ with $X_i, Z_j\in \mathbb{R}^n$ \cite{shokoohi2017generalizing}.
We define the distance function $dist_P(X,Z)$ between time-series inputs $X$ and $Z$ according to an alignment path $P$ using the following equations:
\begin{equation}
    dist_P(X,Z) = \sum_{(i,j)\in P}d(X_i,Z_j)
    \label{eq:distp}
\end{equation}

Hence, the DTW measure between $X$ and $Z$ is given by:
\begin{equation}
    DTW(X,Z) = \displaystyle \min_P dist_P(X,Z)
\end{equation}



\section{Dynamic Time Warping based Adversarial Robustness framework}

\begin{figure*}[t]
    \centering
    \begin{minipage}[t]{\linewidth} 
        \includegraphics[width=\linewidth]{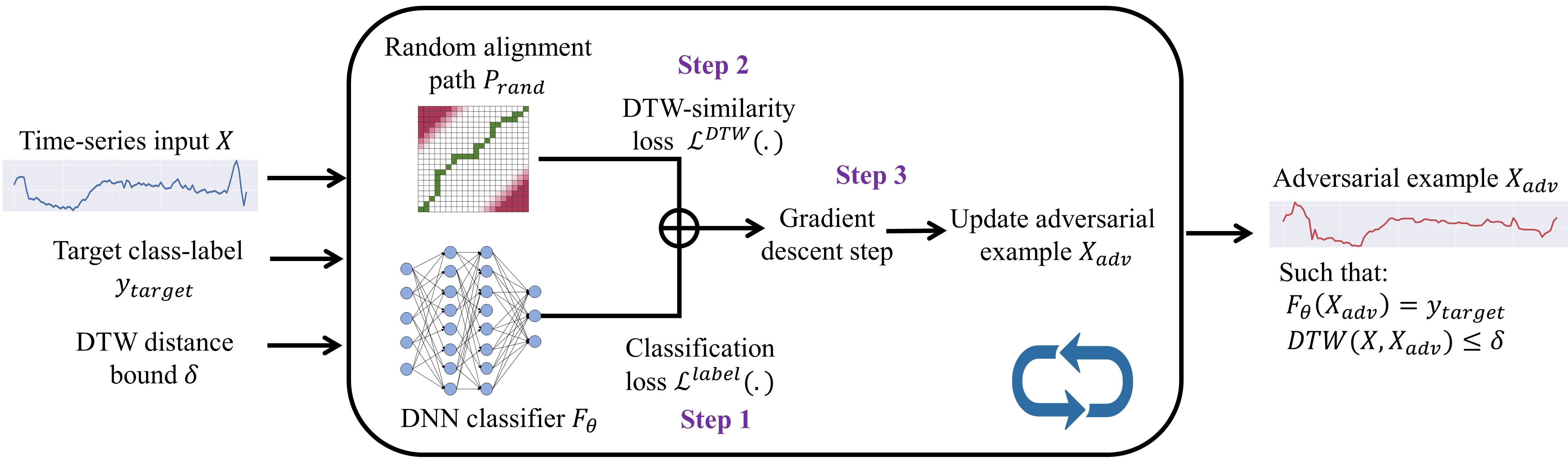}
    \end{minipage}
    \caption{Overview of the DTW-AR framework to create targeted adversarial examples. Given an input $X$, a target class-label $y_{target}$ and a distance bound $\delta$, DTW-AR aims to identify an adversarial example $X_{adv}$ using a random alignment path $P_{rand}$. DTW-AR solves an optimization problem involving a DTW-similarity loss and a classification loss using a random alignment path $P_{rand}$ and a DNN classifier $F_{\theta}$. Using different random alignment paths, DTW-AR will be able to create diverse adversarial examples which meet the DTW measure bound $\delta$.}
    \label{fig:highlevel}
\end{figure*}

The DTW-AR framework creates targeted adversarial examples for time-series domain using the DTW  measure as illustrated in Figure \ref{fig:highlevel}. For any given time-series input $X$, DNN classifier $F_\theta$, and distance bound $\delta$, we solve an optimization problem to identify an adversarial example $X_{adv}$ which is within DTW measure $\delta$ to the original time-series signal $X$. In what follows, we first provide empirical and theoretical results to demonstrate the suitability of DTW measure over Euclidean distance for adversarial robustness studies in the time-series domain (Section 3.1). Next, we introduce the optimization formulation based on the DTW measure to create adversarial examples and describe its main drawbacks (Section 3.2). Finally, we explain our key insight of using stochastic alignment paths to successfully overcome those drawbacks to efficiently create diverse adversarial examples and provide theoretical justification (Section 3.3).

\subsection{Effectiveness of DTW measure measure}

\vspace{1.0ex}

\begin{figure}[!h]
    \centering 
        \begin{minipage}{\linewidth} 
        \centering 
        \begin{minipage}{.48\linewidth}
                \centering
                \includegraphics[width=\linewidth]{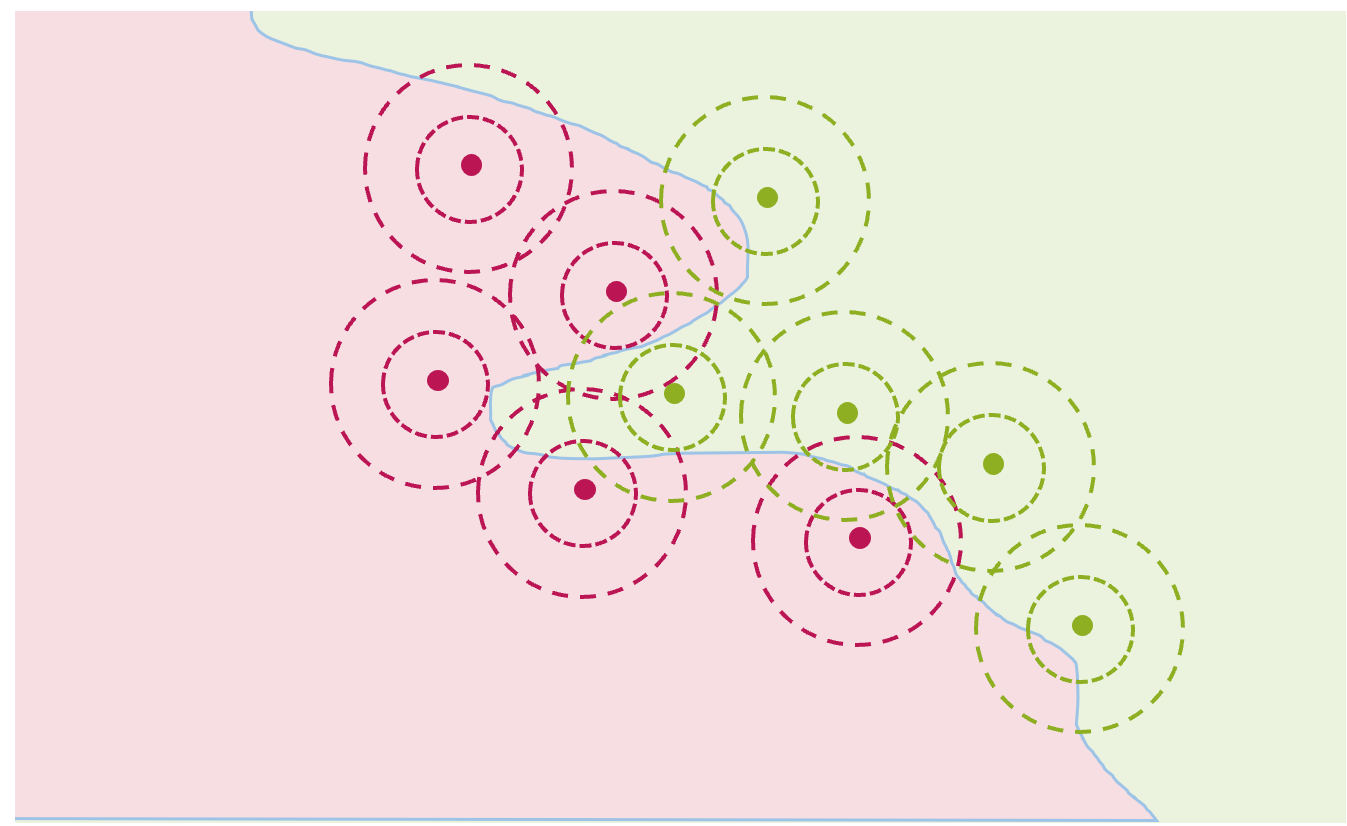}
            \end{minipage}%
            \hfill
        \begin{minipage}{.48\linewidth}
                \centering
                \includegraphics[width=\linewidth]{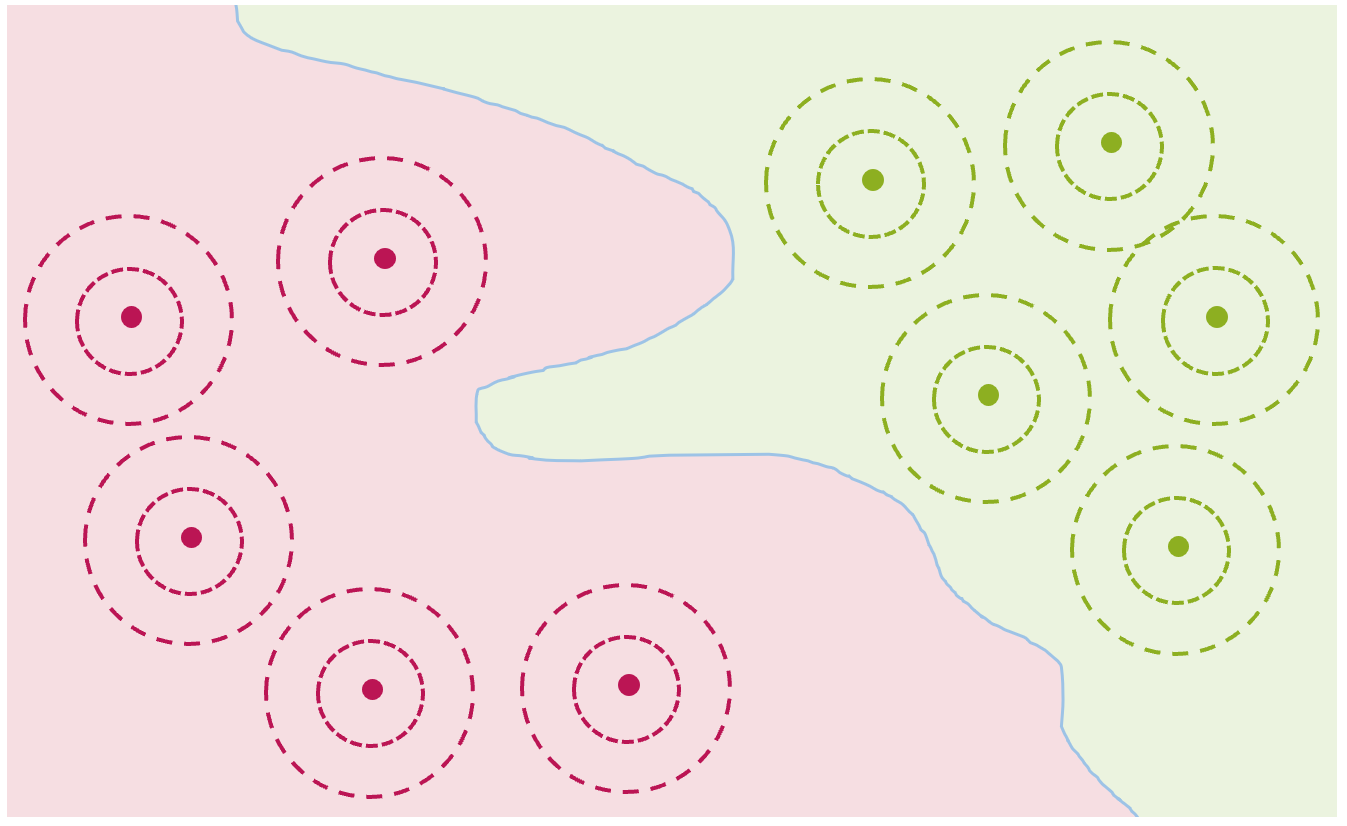}
            \end{minipage}
        \begin{minipage}{.48\linewidth}
        \vspace{1em}
                \centering
                Euclidean Space
            \end{minipage}%
        \begin{minipage}{.48\linewidth}
        \vspace{1em}
                \centering
                DTW Space
            \end{minipage}
    \end{minipage}
    \caption{Illustration of the suitability of DTW over Euclidean distance using the true data distribution from two classes shown in red and green colors. The concentric circles represent the close-similarity area of each input instance (i.e., center) using the corresponding distance measure.}
    \label{fig:thillustdtwspace}
\end{figure}

\begin{figure}[!h]
        \begin{minipage}{\linewidth} 
        \centering 
        \begin{minipage}{.48\linewidth}
                \centering
                \includegraphics[width=\linewidth]{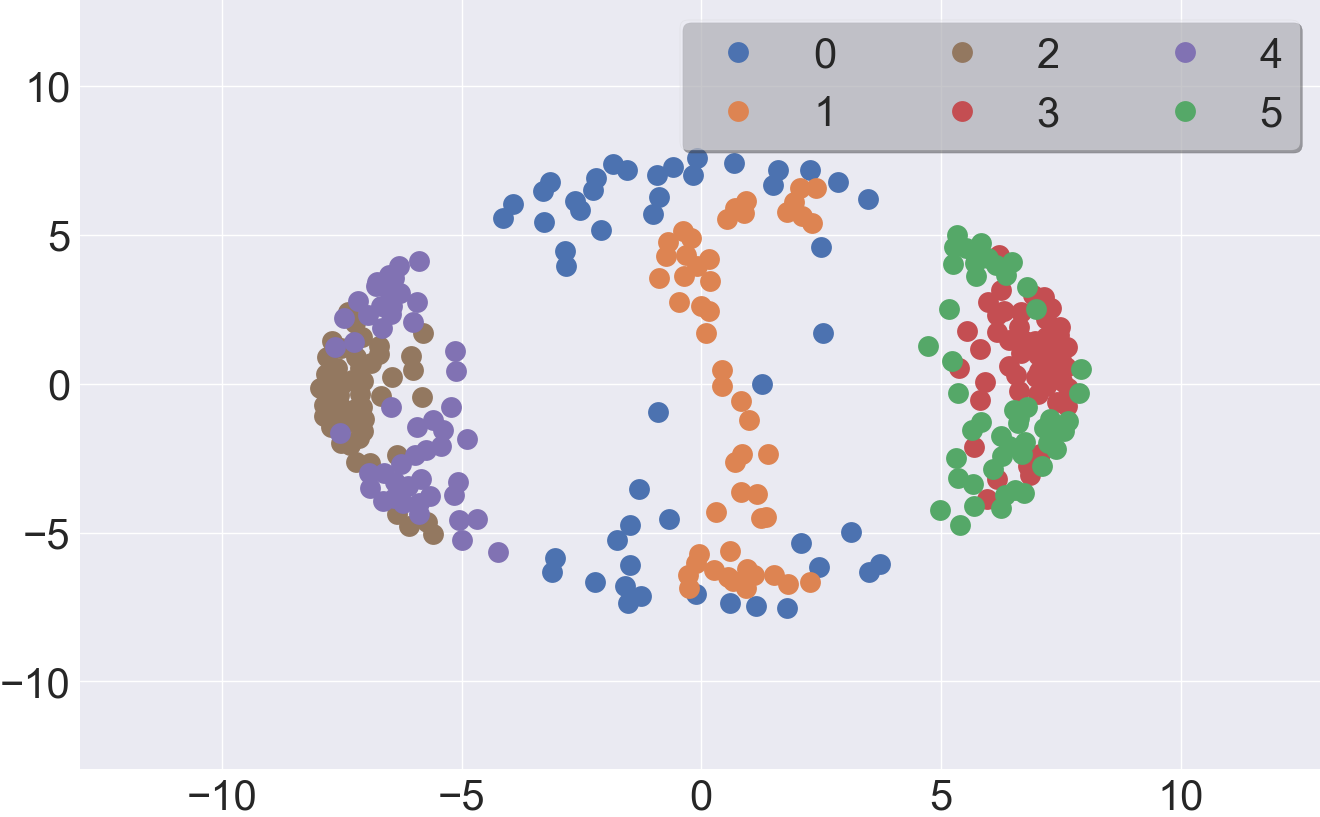}
            \end{minipage}%
            \hfill
        \begin{minipage}{.48\linewidth}
                \centering
                \includegraphics[width=\linewidth]{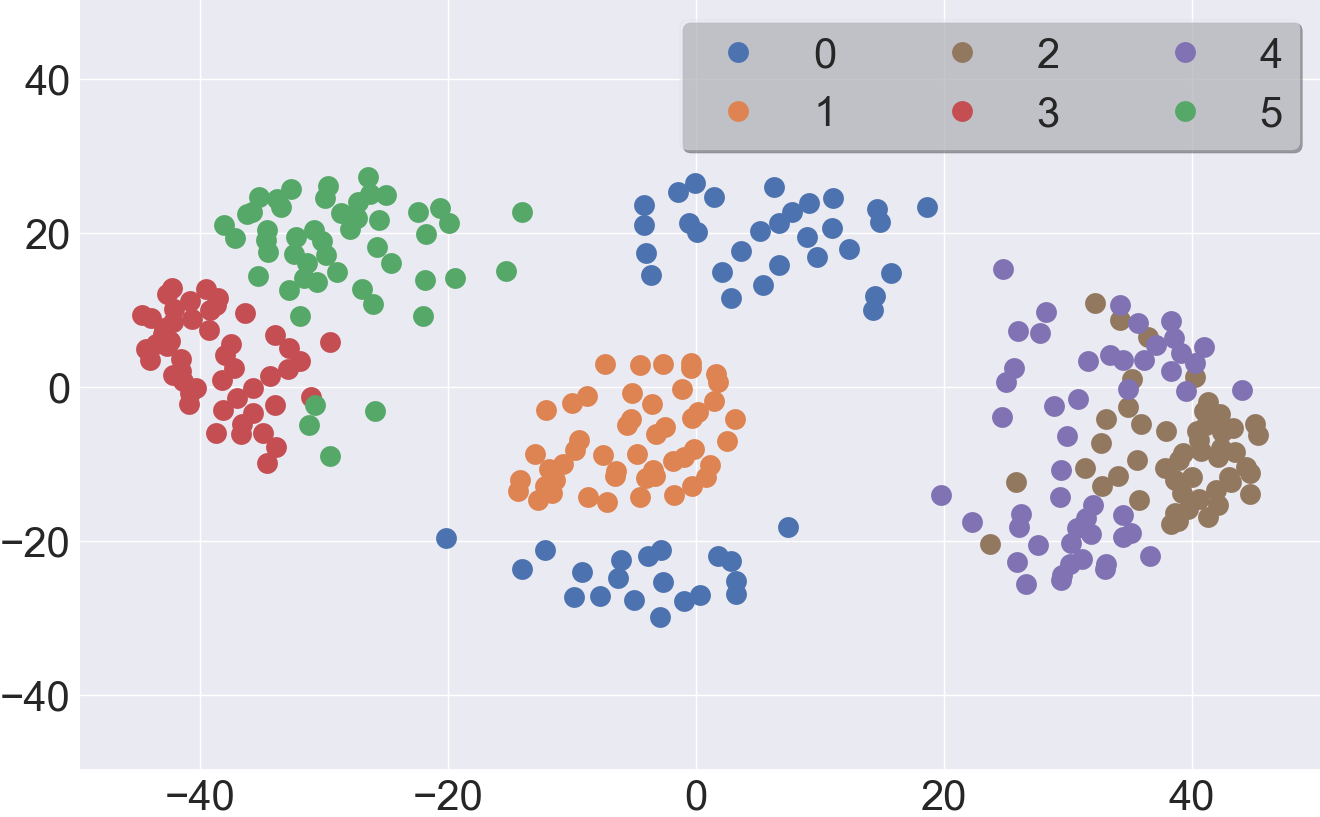}
            \end{minipage}
        \begin{minipage}{.48\linewidth}
        \vspace{1em}
                \centering
                Euclidean Space
            \end{minipage}%
        \begin{minipage}{.48\linewidth}
        \vspace{1em}
                \centering
                DTW Space
            \end{minipage}
    \end{minipage}
    \caption{Multi-dimensional scaling results showing the labeled data distribution in Euclidean space (left column) and DTW space (right column) for the SC dataset. DTW space exhibits better clustering for same-class data than Euclidean space.}
    \label{fig:dtwl2space}
\end{figure}

\noindent {\bf Empirical justification.} As we argued before, the standard $l_2$ distance is impractical for adversarial learning in time-series domain. Perturbations based on Euclidean distance can result in adversarial time-series signals which semantically belong to a different class-label. Based on the real-world data representation provided in Figure \ref{fig:dtwl2space}, we create and show in Figure \ref{fig:thillustdtwspace} an intuitive illustration of suitability of DTW over $l_2$ distance to explain the advantages of DTW as a similarity measure.
It shows the difference in the true data distribution in Euclidean space (i.e., $l_2$ is used as the similarity measure) and in DTW space (i.e., DTW is used as the similarity measure) for two classes shown in red and green colors. The concentric circles represent the close-similarity area around each input instance (i.e., center) where adversarial examples are considered. We can observe that in the Euclidean space, adversarial example of an input instance can belong to another class label, which is not the case in the DTW space. This simple illustration shows how DTW-AR can generate effective adversarial examples due to the appropriate bias of DTW for time-series domain. 

\vspace{1ex}
This abstraction is tightly based on the observations made on real-world data. We employ multi-dimensional scaling (MDS), a visual representation of dissimilarities between sets of data points \cite{buja2008data}, to compare DTW and Euclidean spaces. MDS is a dimensionality reduction method that preserves the distances between data points in the original space. Figure \ref{fig:dtwl2space} shows MDS results of \texttt{SC} dataset.
We can clearly see how the data from different class labels are better clustered in the DTW space compared to the Euclidean space, as provided in Figure \ref{fig:dtwl2space}. An adversarial example for an \texttt{SC} data point in the green-labeled class is more likely to semantically belong to the red-labeled distribution in the Euclidean space. However, in the DTW space, the adversarial example is more likely to remain in the green-labeled space, while only being misclassified by the DNN classifier due the adversarial problem.

\vspace{1.0ex}
\noindent {\bf Theoretical justification.} We prove that the DTW measure allows DTW-AR to explore a larger space of candidate adversarial examples when compared to perturbations based on the Euclidean distance, i.e., identifies blind spots of prior methods. This result is based on the fact that the point-to-point alignment (i.e., Euclidean distance) between two time-series signals is not always the optimal alignment. Hence, the existence of adversarial examples which are similar based on DTW and may not be similar based on the Euclidean distance. To formalize this intuition, we provide Observation 1. We characterize the effectiveness of DTW-AR based attack as better for their ability to extend the space of attacks based on the Euclidean distance and their potential to fool DNN classifiers that rely on Euclidean distance for adversarial training. Our experimental results demonstrate that DTW-AR generates effective adversarial examples to fool the target DNN classifiers by leveraging the appropriate bias of DTW for time-series data.

\begin{observation}
Let $l_2$ be the equivalent of Euclidean distance using the cost matrix in the DTW space. $\forall X \in \mathbb{R}^{n \times T}$ ($n>1, T>1$), there exists $\epsilon \in \mathbb{R}^{n \times T} $ and an alignment path $P$ such that $ dist_P(X,X+\epsilon) \le \delta$ and $l_2(X,X+\epsilon) > \delta$.
\end{observation}

\begin{theorem}
For a given input space $\mathbb{R}^{n \times T}$, a constrained DTW space for adversarial examples is a strict superset of a constrained euclidean space for adversarial examples. If $X\in \mathbb{R}^{n\times T}$:
\begin{equation}
\footnotesize
    \bigg\{  X_{adv} \big|  DTW(X,X_{adv}) \le \delta\bigg\} \supset  \bigg\{ X_{adv} \big| \|X-X_{adv}\|_2^2 \le \delta \bigg\}
\label{eq:dtwspaceth}
\end{equation}
\label{th:advspace}
\end{theorem}

As an extension of Observation 1, the above theorem states that in the space where adversarial examples are constrained using a DTW measure bound, there exists more adversarial examples that are not part of the space of adversarial examples based on the Euclidean distance for the same bound (i.e., blind spots). This result implies that DTW measure has an appropriate bias for the time-series domain. We present the proofs of both Observation 1 and Theorem 1 in the {\bf Appendix}. Hence, our DTW-AR framework is {\em potentially} capable of creating more effective adversarial examples than prior methods based on $l_2$ distance for the same distance bound constraint. These adversarial examples are potentially more effective as they are able to break deep models by leveraging the appropriate bias of DTW measure. 

However, to convert this potential to reality, we need an algorithm that can efficiently search this larger space of attacks to identify most or all adversarial examples which meet the DTW measure bound. Indeed, developing such an algorithm is one of the key contributions of this paper.

\subsection{Naive optimization based formulation and challenges to create adversarial examples} 

To create adversarial examples to fool the given DNN $F_{\theta}$, we need to find an optimized perturbation of the input time-series $X$ to get $X_{adv}$. Our approach is based on minimizing a {\em loss function} $\mathcal{L}$ using gradient descent that achieves two goals. 1) Misclassification goal: Adversarial example $X_{adv}$ to be mis-classified by $F_{\theta}$ as a target class-label $y_{target}$; and 2) DTW similarity goal: close DTW-based similarity between time-series $X$ and adversarial example $X_{adv}$.

To achieve the mis-classification goal, we employ the formulation of \cite{carlini2017towards} to define a loss function:

\begin{equation}
\begin{split}
    \mathcal{L}^{label}(X_{adv}) = \max \Big[ \displaystyle \max_{y \neq y_{target}} & \left(  \mathcal{S}_y \left(X_{adv}\right) \right) \\&-
  \mathcal{S}_{y_{target}}\left( X_{adv}\right)\textbf{,} ~~\rho \Big]
\end{split}
\label{eq:classloss}
\end{equation}
where $\rho < 0$. It ensures that the adversarial example will be classified by the DNN as class-label $y_{target}$ with a confidence $|\rho|$ using the output of the pre-softmax layer $\{\mathcal{S}_y\}_{y \in Y}$.

To achieve the DTW similarity goal, we need to create $X_{adv}$ for a given time-series input $X$ such that $DTW(X,X_{adv}) \le \delta$. We start by a naive optimization over the DTW measure using the Soft-DTW measure SDTW$(X,X_{adv})$ \cite{cuturi2017soft}. Hence, the DTW similarity loss function is:
\begin{equation}
    \mathcal{L}^{DTW}(X_{adv})  = \text{SDTW} (X,X_{adv})
    \label{eq:dtwloss1}
\end{equation}
The final loss function $\mathcal{L}$ we want to minimize  to create optimized adversarial example $X_{adv}$ is:
\begin{equation}
    \mathcal{L}(X_{adv}) = \mathcal{L}^{label}(X_{adv}) +  \mathcal{L}^{DTW}(X_{adv})
    \tag{$\star$}
    \label{eq:floss}
\end{equation}

We operate under white-box setting and can employ gradient descent to minimize the loss function in Equation \ref{eq:floss} over $X_{adv}$. This approach works for black-box setting also. In this work, we consider the general case where we do not query the black-box target DNN classifier. We show through experiments that the created adversarial examples can generalize to fool other black-box DNNs.

\vspace{1.0ex}
\noindent \textbf{Challenges of Naive approach.} Recall that our overall goal is to identify most or all targeted adversarial time-series examples that meet the DTW measure bound. This will allow us to improve the robustness of DNN model using adversarial training. This naive approach has two main drawbacks.

\begin{figure}[!h]
    \centering
        \centering
        \begin{minipage}[t]{0.43\linewidth}
                \centering
                \includegraphics[width=\linewidth]{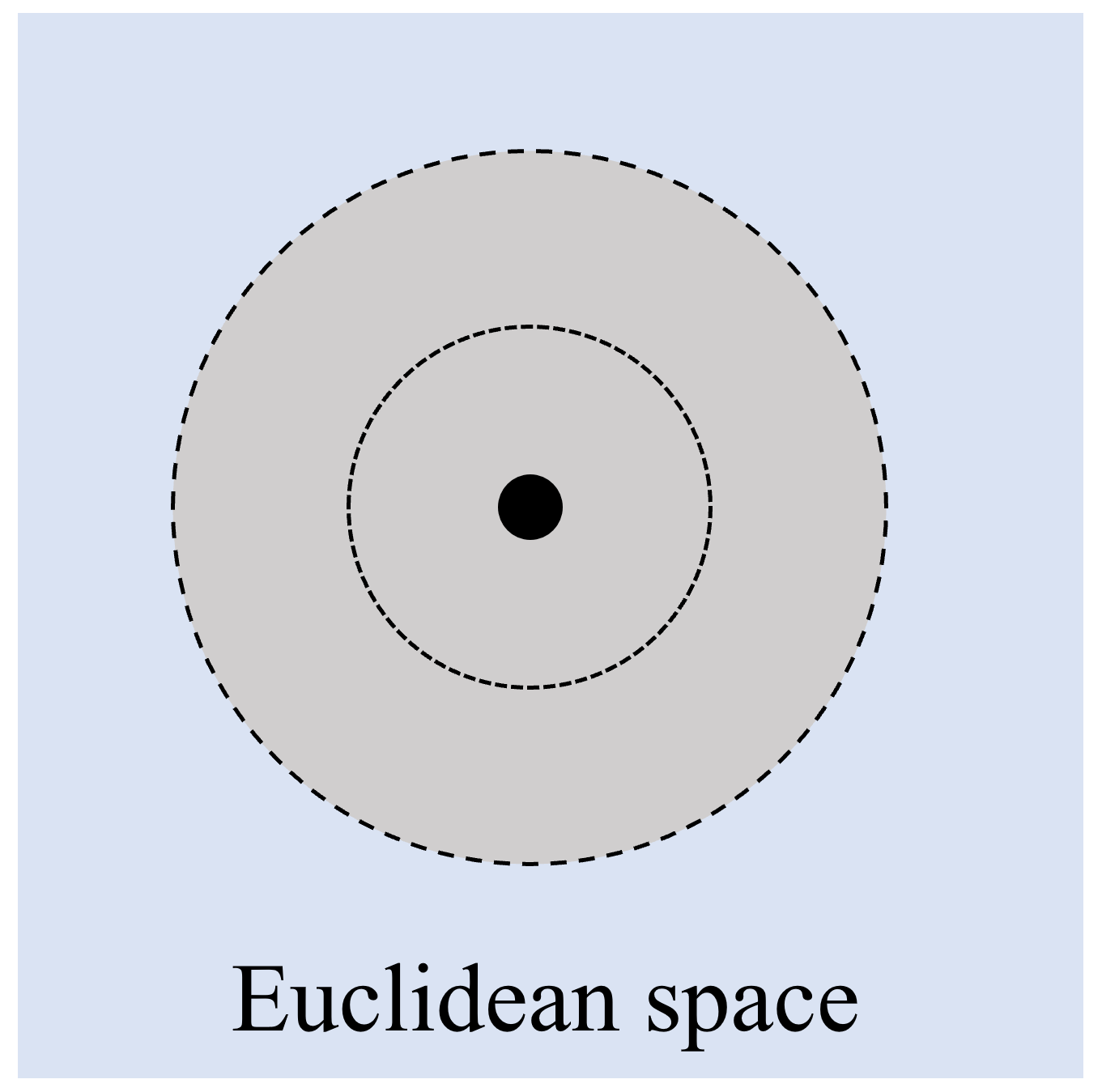}
            \end{minipage}%
            \hspace{2ex}
        \begin{minipage}[t]{.43\linewidth}
                \centering
                \includegraphics[width=\linewidth]{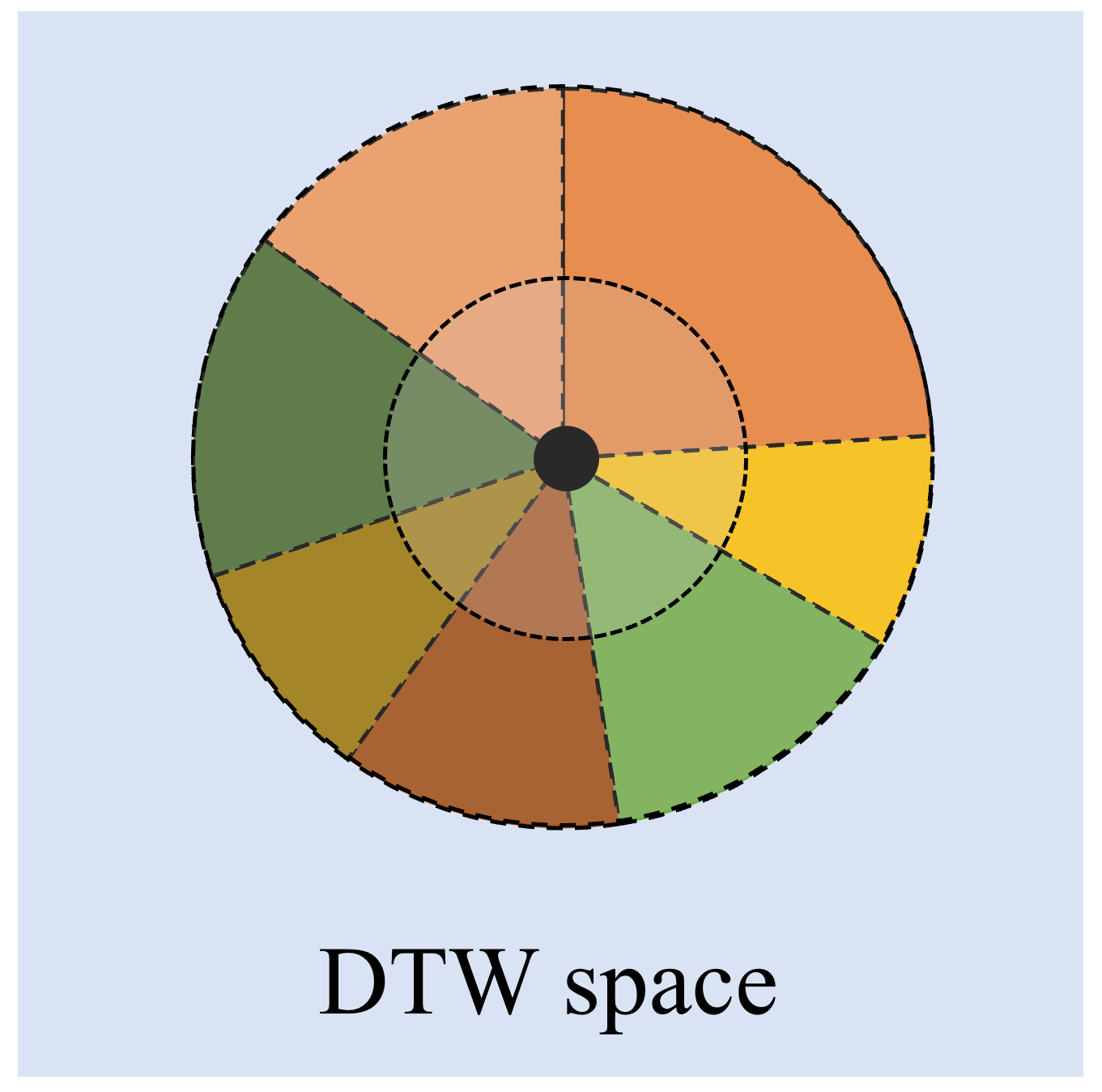}
            \end{minipage}
            
        \begin{minipage}[t]{0.43\linewidth}
                \centering
                (a) 
            \end{minipage}%
        \begin{minipage}[t]{.43\linewidth}
                \centering
                (b)
            \end{minipage}
    \caption{Illustration of the close-similarity space around a given time-series signal (black center) in the Euclidean and DTW space. Using $l_2$ norm is sufficient to explore the entire Euclidean space around the input. However, in the DTW space, each colored section corresponds to one adversarial example that meets the DTW measure bound constraint. Each of them can be found using only a subset of candidate alignment paths.} 
    \label{fig:dtwsubspace}
\end{figure}

\hspace{0.5em} $\bullet$ {\em Single adversarial example.} The method allows us to only find one valid adversarial example out of multiple solution candidates from the search space because it operates on a single optimal alignment path. 
Using a single alignment path (whether the diagonal path for Euclidean distance or the optimal alignment path generated by DTW), the algorithm will be limited to the adversarial examples which use that single alignment. In Figure \ref{fig:dtwsubspace}, we provide a conceptual illustration of $S_{ADV}(X)$, the set of all adversarial examples $X_{adv}$ which meet the distance bound constraint $DTW(X, X_{adv}) \le \delta$. In the Euclidean space, using $l_2$ norm is sufficient to explore the entire search space around the original input to create adversarial examples. However, in the DTW space, each colored section in $S_{ADV}(X)$ can only be found using a subset of candidate alignment paths.

\hspace{0.5em} $\bullet$ {\em High computational cost.} DTW is non-differentiable and approximation methods are often used in practice. These methods require $\mathcal{O}(n.T^2)$ to fill the cost matrix and $\mathcal{O}(T)$ to backtrack the optimal alignment path. These steps are computationally-expensive. Gradient-based optimization iteratively updates the adversarial example $X_{adv}$ to achieve the DTW similarity goal, i.e., $DTW(X, X_{adv}) \le \delta$, and the mis-classification goal, i.e., $F_{\theta}(X_{adv})$=$y_{target}$. Standard algorithms such as projected gradient descent (PGD) \cite{madry2017towards} and Carloni \& Wagner (CW) \cite{carlini2017towards} require a large number of iterations to generate valid adversarial examples. This is also true for the recent computer vision specific adversarial algorithms \cite{laidlaw2019adv, Shafahi2020uni}. For time-series signals arising in many real-world applications, the required number of iterations to create successful attacks can grow even larger. 
We need to compute DTW measure in each iteration as the optimal DTW alignment path changes over iterations. Therefore, it is impractical to use the exact DTW computation algorithm to create adversarial examples. We also show that the existing optimized approaches to estimate the DTW measure remain computationally expensive for an adversarial framework. We provide results to quantify the runtime cost in our experimental evaluation.

\begin{algorithm}[!h]
\caption{DTW-AR based Adversarial Algorithm}
\label{alg:univ}
\textbf{Input}:  time-series $X$; DNN classifier $F_{\theta}$; target class-label $y_{target}$;  learning rate $\eta$; maximum iterations \textsc{Max}\\
\textbf{Output}: adversarial example $X_{adv}$
\begin{algorithmic}[1] 
\STATE $P_{rand} \leftarrow$ random alignment path
\STATE Initialization: $X_{adv} \leftarrow X $
\FOR{$i$=1 to \textsc{Max}}
\STATE $\mathcal{L}(X_{adv}) \leftarrow \mathcal{L}^{label}(X_{adv}) +  \mathcal{L}^{DTW}(X_{adv}, P_{rand})$
\STATE Compute gradient $\nabla_{X_{adv}}  \mathcal{L}(X_{adv})$
\STATE Perform gradient descent step: \\ $X_{adv}\leftarrow X_{adv} - \eta \times \nabla_{X_{adv}}  \mathcal{L}(X_{adv})$ 
\ENDFOR
\STATE \textbf{return} optimized adversarial example $X_{adv}$
\end{algorithmic}
\label{alg:adv}
\end{algorithm}

\subsection{Stochastic alignment paths for the DTW similarity goal and theoretical justification}

\vspace{1.0ex}
In this section, we describe the key insight of DTW-AR to overcome the above-mentioned two challenges and provide theoretical justification. 

To overcome the above-mentioned two challenges of the naive approach, we propose the use of a random alignment path to create adversarial attacks on DNNs for time-series domain. The key idea is to select a random alignment path $P$ and to execute our adversarial algorithm while constraining over $dist_{P}(X,X_{adv})$ instead of $DTW(X,X_{adv})$. This choice is justified from a theoretical point-of-view due to the special structure in the problem to create DTW based adversarial examples. Using the distance function $dist_{P}(X,X_{adv})$, we redefine Equation \ref{eq:dtwloss1} as follows:
\begin{equation}
\begin{split}
    \mathcal{L}^{DTW}(X_{adv}, P)  = \; &  \; \alpha_1 \times dist_{P}(X,X_{adv}) \\ & -\alpha_2 \times dist_{P_{diag}}(X,X_{adv})
\end{split}
    \label{eq:dtwloss}
\end{equation}
where $\alpha_1>0,~\alpha_2\ge 0$, $P_{diag}$ is  the diagonal alignment path equivalent to the Euclidean distance, and $P$ is a given alignment path ($P \neq P_{diag}$).
The first term of Equation \ref{eq:dtwloss} is defined to bound the DTW similarity of adversarial example $X_{adv}$ to a threshold $\delta$ as stated in Observation \ref{th:optalignment}. The second term represents a penalty term to account for adversarial example with close Euclidean distance to the original input $X$ and pushes the algorithm to look beyond adversarial examples in the Euclidean space. The coefficients $\alpha_1$ and $\alpha_2$ contribute in defining the position of the adversarial output in the DTW and/or Euclidean space. If $\alpha_2 \rightarrow 0$, the adversarial example $X_{adv}$ will be highly similarity to the original input $X$ in the DTW space with no consideration to the Euclidean space. Hence, the adversarial example may be potentially adversarial in the Euclidean space also. However, if $\alpha_2 > 0$, the adversarial output will be highly similar to the original input in the DTW space but out of the scope of adversarial attacks in the Euclidean space (i.e., a blind spot). Recall from Theorem 1 that DTW space allows more candidate adversarial examples than Euclidean space. Hence, this setting allows us to find blind spots of Euclidean space based attacks. 

The DTW-AR approach to create adversarial examples is shown in Algorithm \ref{alg:adv}. We note that the naive approach that uses Soft-DTW with the Carlini \& Wagner loss function is a sub-case of DTW-AR as shown below:
\begin{equation}
\begin{split}
    &\text{SDTW} (X,X_{adv}) = \mathcal{L}^{DTW}(X_{adv}, P_{DTW})  = \\ &  1 \times dist_{P_{DTW}}(X,X_{adv})  -0  \times dist_{P_{diag}}(X,X_{adv})
\end{split}
    \label{eq:sdtwloss}
\end{equation}

where $P_{DTW}$ is the optimal DTW alignment path.

\begin{observation}
Given any alignment path $P$ and two multivariate time-series signals $X,Z \in \mathbb{R}^{n \times T}$. If we have $dist_P(X,Z) \le  \delta$, then $DTW(X,Z) \le \delta$.
\label{th:optalignment}
\end{observation}

Observation \ref{th:optalignment} states that $dist_P(X, Z)$ defined with respect to a path $P$ is always an upper bound for $DTW(X, Z)$, since DTW uses the optimal alignment path. Hence, when the alignment path is fixed, the time-complexity is reduced to a simpler similarity measure that requires only $\mathcal{O}(n.T)$, which results in significant computational savings due to repeated calls within the adversarial algorithm.

Our stochastic alignment method also improves the search strategy for finding multiple desired adversarial examples. Suppose $S_{ADV}(X)$ is the set of all adversarial examples $X_{adv}$ which meet the distance bound constraint $DTW(X, X_{adv}) \le \delta$. Each adversarial example in $S_{ADV}(X)$ can be found using only a subset of candidate alignment paths.  By using a stochastic alignment path, we can leverage the large pool of different alignment paths to uncover more than one adversarial example from $S_{ADV}(X)$. On the other hand, if the exact DTW computation based algorithm was feasible, we would only find a {\em single} $X_{adv}$, as DTW based algorithm operates on a single optimal alignment path.  

\vspace{1.0ex}
\noindent {\bf Theoretical tightness of bound.} While Observation \ref{th:optalignment} provides an upper bound for the DTW measure, it does not provide any information about the tightness of the bound. To analyze this gap, we need to first define a similarity measure between two alignment paths to quantify their closeness. We define $\texttt{PathSim}$ as a similarity measure between two alignment paths $P_1$ and $P_2$ in the DTW cost matrix of time-series signals $X,Z \in \mathbb{R}^{n \times T}$. Let $P_1 = \{c^1_1,..., c^1_k\}$ and $P_2 = \{c^2_1,..., c^2_l\}$ represent the sequence of cells for paths $P_1$ and $P_2$ respectively.
\begin{equation}
\begin{split}
    \texttt{PathSim}& ~(P_1, P_2) = \\ & \dfrac{1}{2T}\left(\sum_{c^1_i} \min_{c^2_j}\|c^1_i-c^2_j\|_1 + \sum_{c^2_i} \min_{c^1_j}\|c^2_i-c^1_j\|_1\right)
\end{split}
\label{eq:pathsim}
\end{equation}

As $\texttt{PathSim}(P_1, P_2)$ approaches 0, $P_1$ and $P_2$ are very similar, and they will be the exact same path if $\texttt{PathSim}(P_1, P_2)=0$. For $X,Z \in \mathbb{R}^{n \times T}$, two very similar alignment paths corresponds to a similar feature alignment between $X$ and $Z$. Theorem \ref{th:dtwgap} shows the tightness of the bound given in Observation \ref{th:optalignment} using the path similarity measure defined above.

\begin{theorem}
For a given input $X \in \mathbb{R}^{n \times T}$ and a random alignment path $P_{rand}$, the resulting adversarial example $X_{adv}$ from the minimization over $dist_{P_{rand}}(X,X_{adv})$ is equivalent to minimizing over $DTW(X,X_{adv})$. For any $X_{adv}$ generated by DTW-AR using $P_{rand}$, we have:
\begin{equation}
\begin{cases}
    \texttt{PathSim}(P_{rand}, P_{DTW}) = 0 \\ ~~\& \\
    dist_{P_{rand}}(X,X_{adv}) = DTW(X,X_{adv})
\end{cases}
\label{eq:dtwgap}
\end{equation}
where $P_{DTW}$ is the optimal alignment path found using DTW computation between $X$ and $X_{adv}$.
\label{th:dtwgap}
\end{theorem}

\noindent {\bf Similarity measure $\texttt{PathSim}$ definition}.
\begin{figure*}[!h]
    \centering
        \begin{minipage}{\linewidth}
         \centering
            \begin{minipage}{.19\linewidth}
                \centering               
                \includegraphics[width=\linewidth]{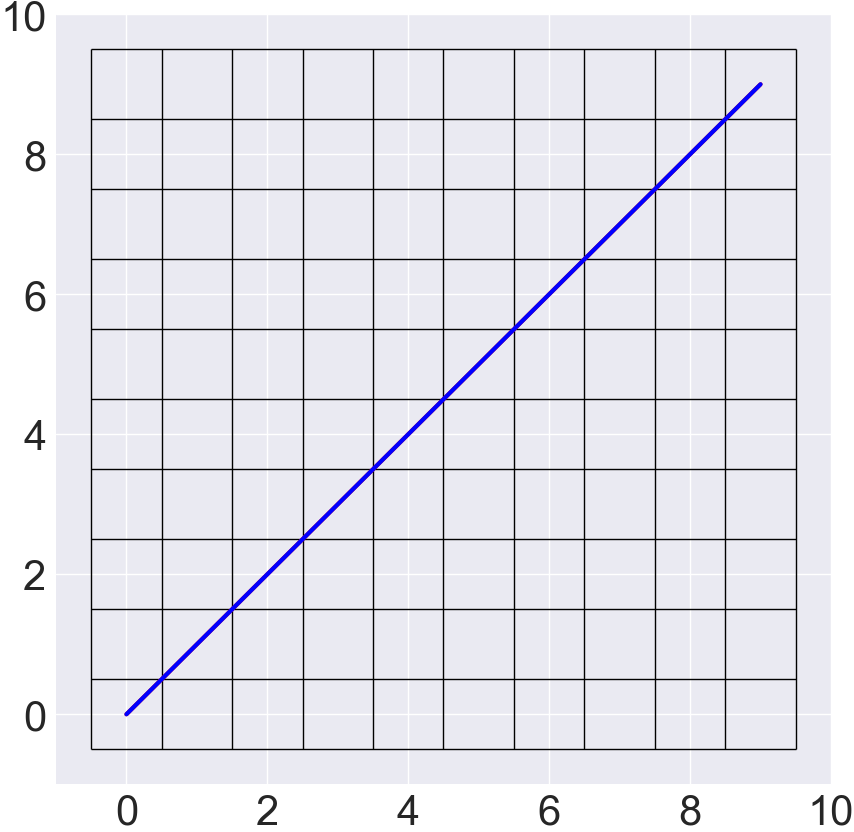}
            \end{minipage}%
            \begin{minipage}{.19\linewidth}
                \centering                
                \includegraphics[width=\linewidth]{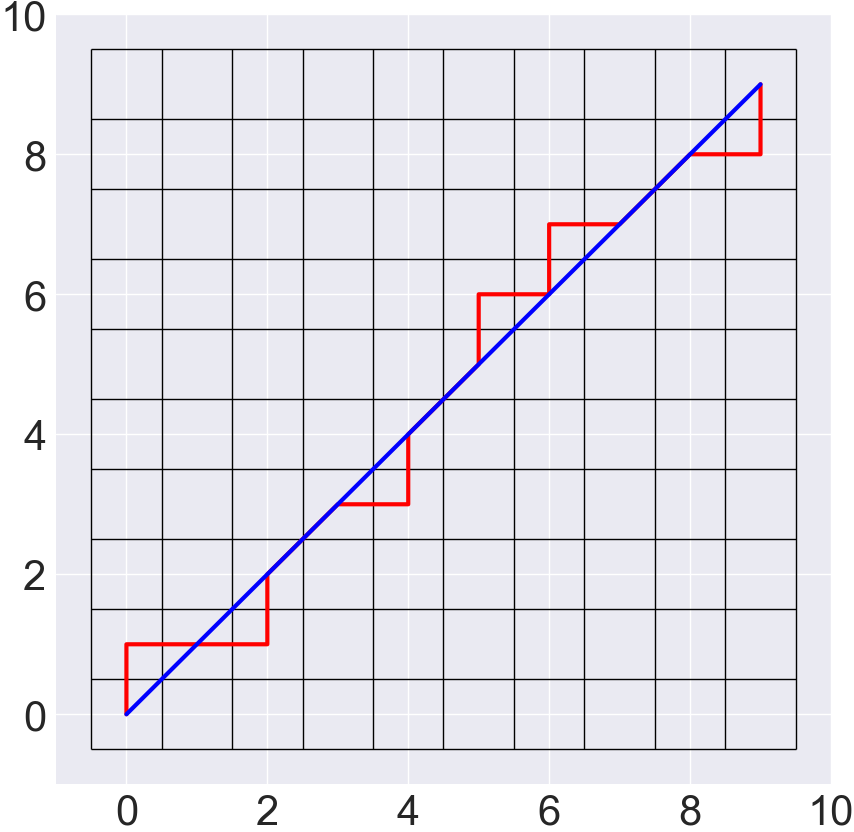}
            \end{minipage}%
            \begin{minipage}{.19\linewidth}
                \centering                
                \includegraphics[width=\linewidth]{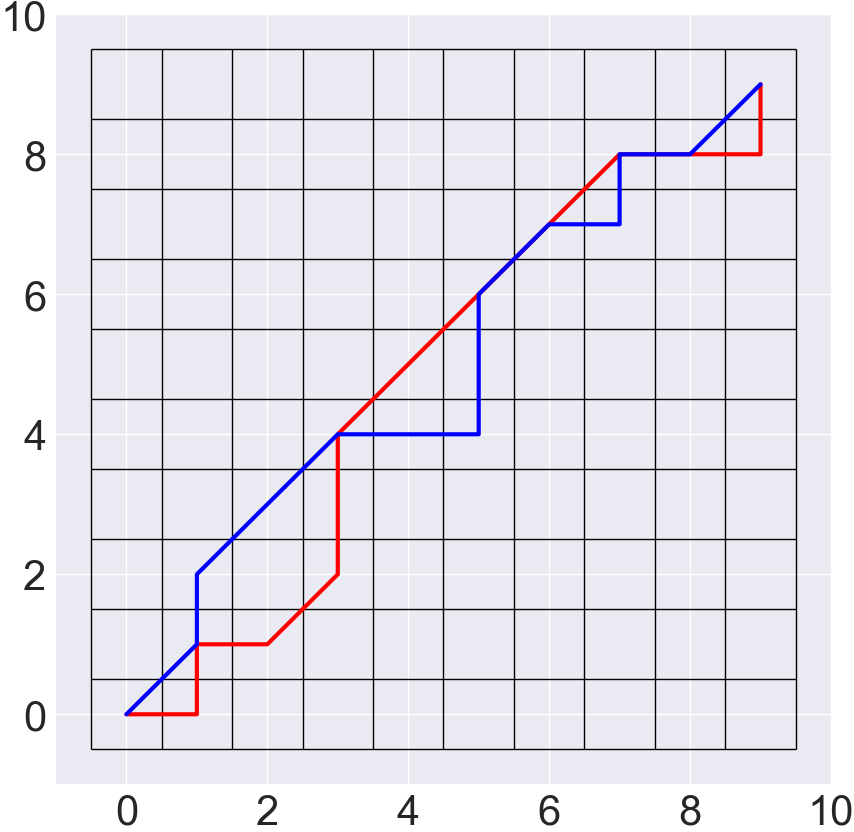}
            \end{minipage}%
            \begin{minipage}{.19\linewidth}
                \centering                
                \includegraphics[width=\linewidth]{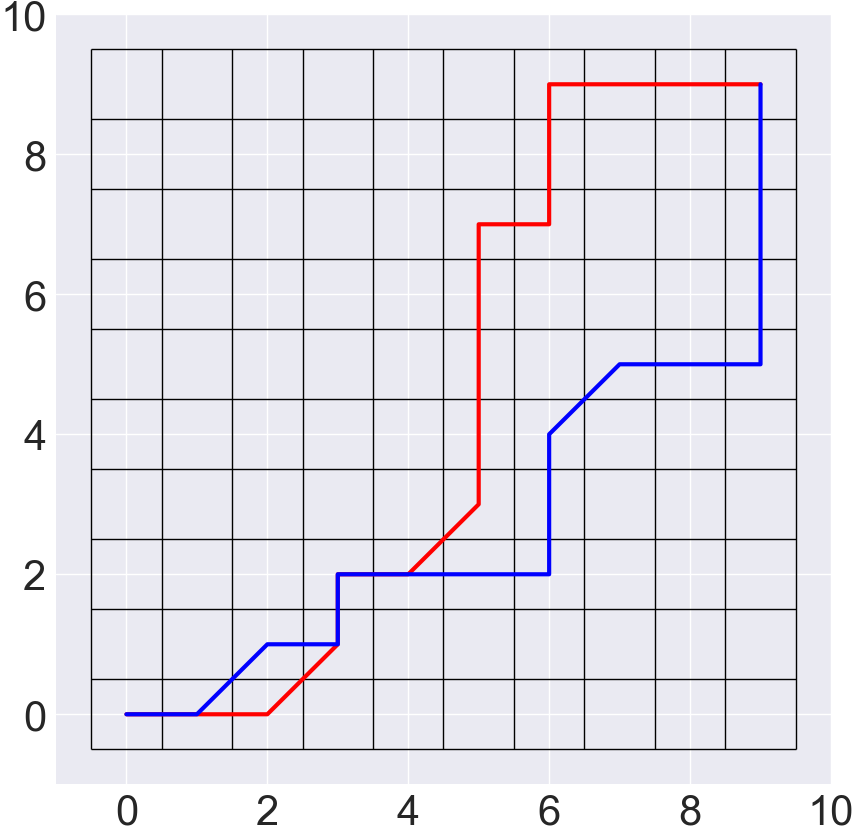}
            \end{minipage}%
            \begin{minipage}{.19\linewidth}
                \centering                
                \includegraphics[width=\linewidth]{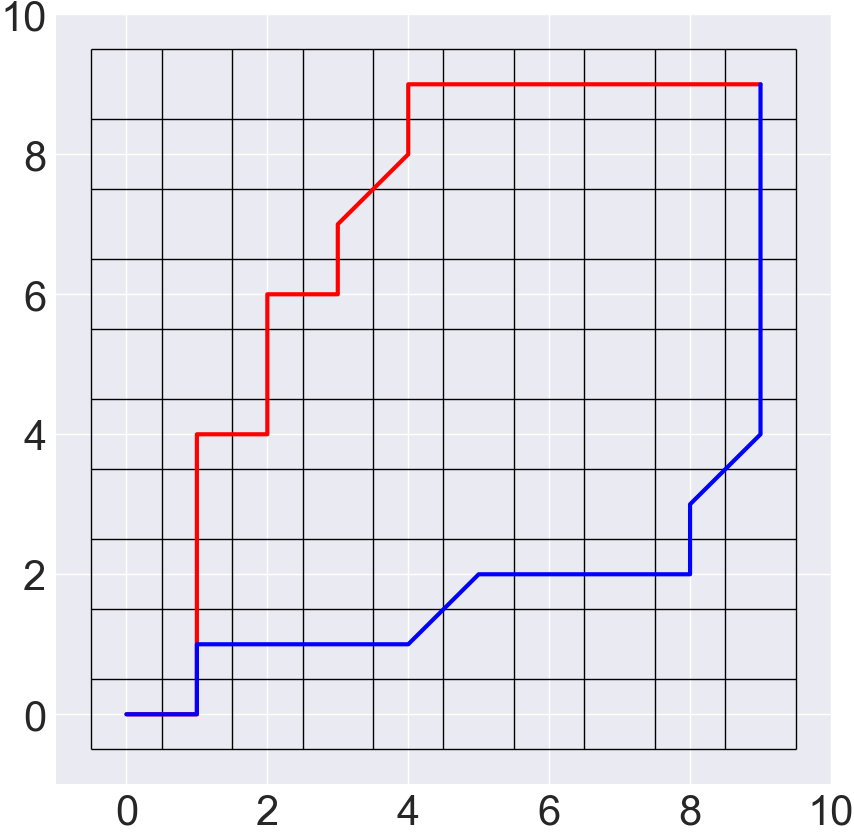}
            \end{minipage}
            
            \begin{minipage}{.19\linewidth}
                \centering              
                $\texttt{PathSim}=0$
            \end{minipage}%
            \begin{minipage}{.19\linewidth}
                \centering                              
                $\texttt{PathSim}=0.25$
            \end{minipage}%
            \begin{minipage}{.19\linewidth}
                \centering                              
                $\texttt{PathSim}=0.5$
            \end{minipage}%
            \begin{minipage}{.19\linewidth}
                \centering                              
                $\texttt{PathSim}=1.25$
            \end{minipage}%
            \begin{minipage}{.19\linewidth}
                \centering                              
                $\texttt{PathSim}=2.5$
            \end{minipage}
        \end{minipage}
        \begin{minipage}{.1\linewidth}
        \text{ }
        \end{minipage}
        \begin{minipage}{\linewidth}
        \centering
            \begin{minipage}{.19\linewidth}
                \centering                \includegraphics[width=\linewidth]{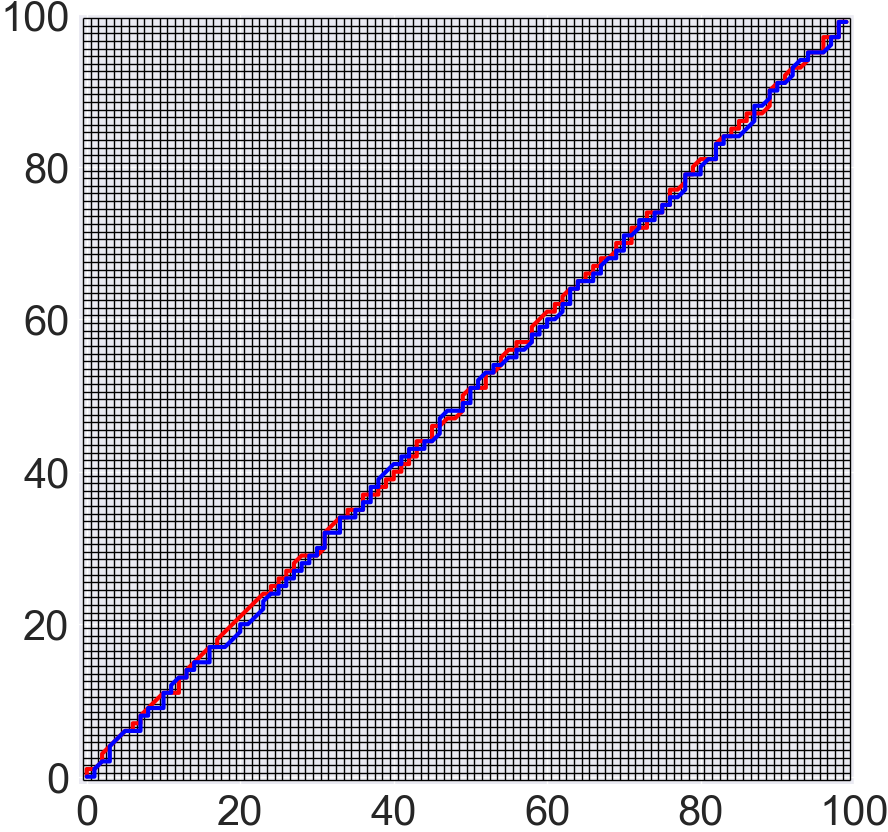}
            \end{minipage}%
            \begin{minipage}{.19\linewidth}
                \centering                \includegraphics[width=\linewidth]{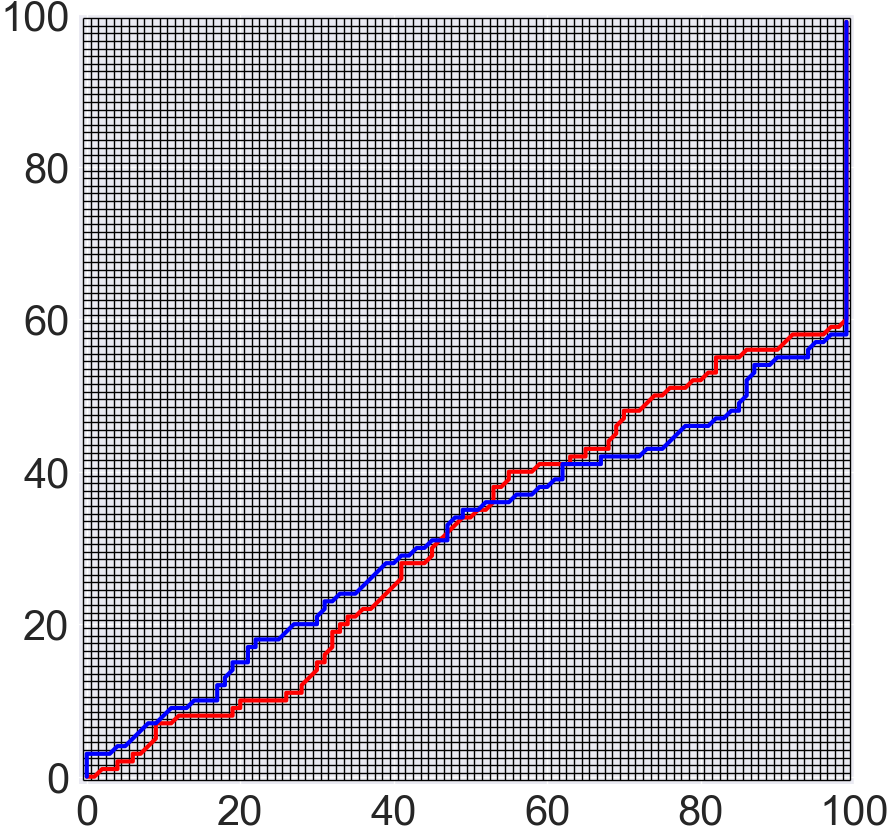}
            \end{minipage}%
            \begin{minipage}{.19\linewidth}
                \centering                \includegraphics[width=\linewidth]{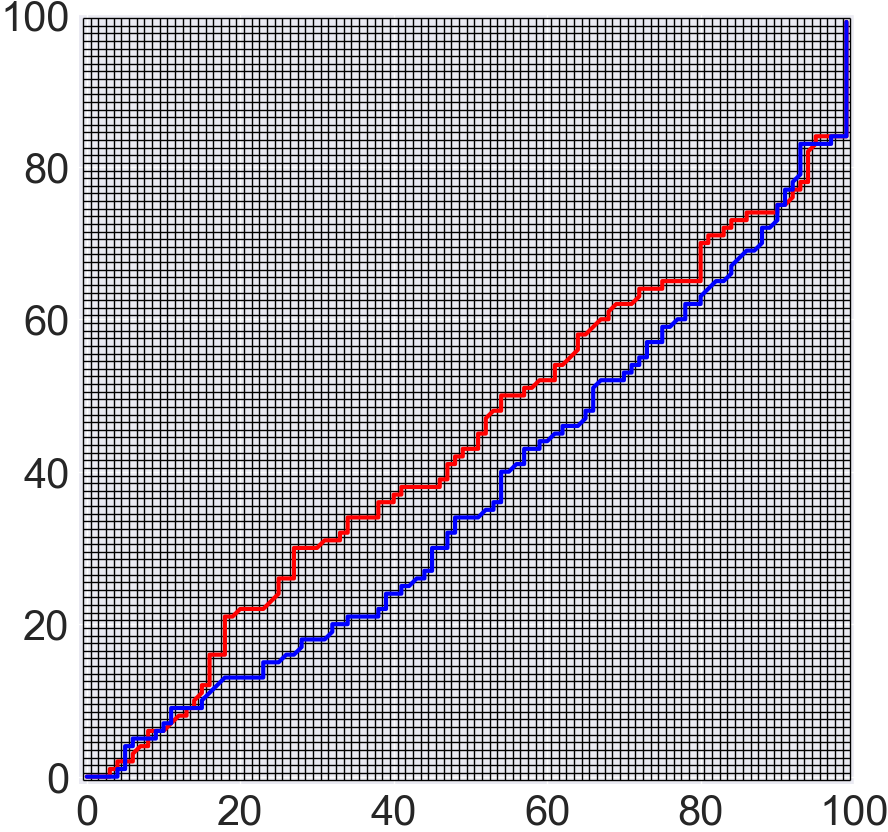}
            \end{minipage}%
            \begin{minipage}{.19\linewidth}
                \centering                \includegraphics[width=\linewidth]{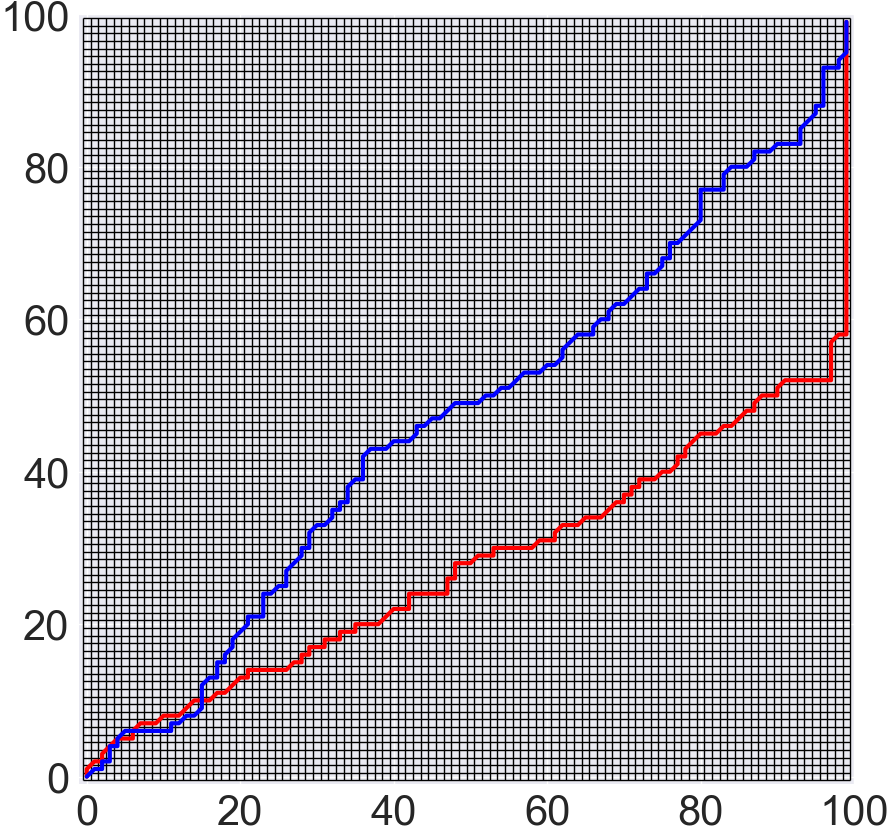}
            \end{minipage}%
            \begin{minipage}{.19\linewidth}
                \centering                \includegraphics[width=\linewidth]{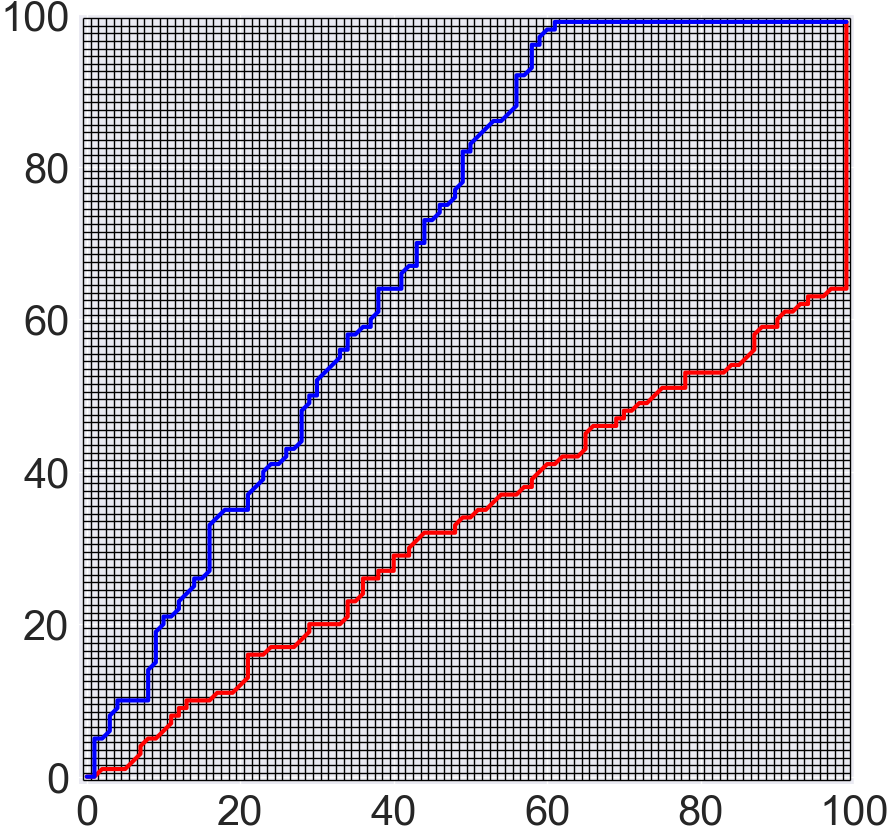}
            \end{minipage}
            
            \begin{minipage}{.19\linewidth}
                \centering              
                $\texttt{PathSim}=0.5$
            \end{minipage}%
            \begin{minipage}{.19\linewidth}
                \centering                              
                $\texttt{PathSim}=2.5$
            \end{minipage}%
            \begin{minipage}{.19\linewidth}
                \centering                              
                $\texttt{PathSim}=5$
            \end{minipage}%
            \begin{minipage}{.19\linewidth}
                \centering                              
                $\texttt{PathSim}=15$
            \end{minipage}%
            \begin{minipage}{.19\linewidth}
                \centering                              
                $\texttt{PathSim}=25$
            \end{minipage}
        \end{minipage}
\caption{Visualization of $\texttt{PathSim}$ values along different example alignment paths in $\mathbb{R}^{n \times 10}$ (First row) and $\mathbb{R}^{n \times 100}$ (Second row) spaces.}
\label{fig:pathfig}
\end{figure*}
For DTW-AR, we rely on a  stochastic alignment path to compute $dist_P$ defined in Equation \ref{eq:distp}. To improve our understanding of the behavior of DTW-AR framework based on stochastic alignment paths, we propose to define a similarity measure that we call $\texttt{PathSim}$. This measure quantifies the similarities between two alignment paths $P_1$ and $P_2$ in the DTW cost matrix for two time-series signals $X,Z \in \mathbb{R}^{n \times T}$. If we denote the alignment path sequence $P_1$ = $\{c^1_1,\cdots, c^1_k\}$ and $P_2$ = $\{c^2_1,\cdots,c^2_l\}$, then we can measure their similarity as defined in Equation \ref{eq:pathsim}.

This definition is a {\em valid} similarity measure as it satisfies all the distance axioms \cite{cullinane2011metric}:

\vspace{1.0ex}

{\em Non-negativity:} By definition, $\texttt{PathSim}(P_1, P_2)$ is a sum of $l_1$ distances, which are all positives. Hence, $\texttt{PathSim}(P_1, P_2) \ge 0$.

\vspace{1.0ex}
    
{\em Unicity:} $\texttt{PathSim}(P_1, P_2)=0$
    \begin{equation*}
        \begin{split}
            &\iff \dfrac{1}{2T}(\sum_{c^1_i} \min_{c^2_j}\|c^1_i-c^2_j\|_1 \\&~~~~~~~~~~~~~~~+ \sum_{c^2_i} \min_{c^1_j}\|c^2_i-c^1_j\|_1) = 0\\
            &\iff \sum_{c^1_i} \min_{c^2_j}\|c^1_i-c^2_j\|_1 + \sum_{c^2_i} \min_{c^1_j}\|c^2_i-c^1_j\|_1=0 
        \end{split}
    \end{equation*}
As we have a sum equal to 0 of all positive terms, we can conclude that each term ($\min\|\cdot\|_1$) is equal to 0: $\texttt{PathSim}(P_1, P_2)=0$
\begin{equation*}
        \begin{split}
            &\iff \forall i: \|c^1_i-c^2_i\|_1=0\\
            &\iff \forall i: ~~c^1_i = c^2_i
        \end{split}
    \end{equation*}
As both paths have the same sequence of cells, we can safely conclude that $\texttt{PathSim}(P_1, P_2)=0 \iff P_1=P_2$:
    \vspace{1.0ex}

{\em Symmetric Property:} 
    \begin{equation*}
    \begin{split}
        &\texttt{PathSim}(P_1, P_2) =\\& \dfrac{1}{2T}\left(\sum_{c^1_i} \min_{c^2_j}\|c^1_i-c^2_j\|_1 + \sum_{c^2_i} \min_{c^1_j}\|c^2_i-c^1_j\|_1\right) \\&
        = \dfrac{1}{2T}\left(\sum_{c^2_i} \min_{c^1_j}\|c^2_i-c^1_j\|_1 + \sum_{c^1_i} \min_{c^2_j}\|c^1_i-c^2_j\|_1 \right) \\&
        =\texttt{PathSim}(P_2, P_1)
    \end{split}
    \end{equation*}

Note that the triangle inequality is  not applicable as the alignment path spaces does not support additive operations.

This similarity measure quantifies the similarity between two alignment paths as it measures the $l_1$ distance between the different cells of each path. The multiplication factor ${1}/{2T}$ is introduced to prevent scaling of the measure for large $T$ values for a given time-series input space $\mathbb{R}^{n \times T}$.

In Figure \ref{fig:pathfig}, we visually show the relation between $\texttt{PathSim}$ measure and the alignment path for a given cost matrix. We observe that when $\texttt{PathSim}(P_1, P_2) \rightarrow 0$, $P_1$ and $P_2$ are very similar, and they will be the exact same path if $\texttt{PathSim}(P_1, P_2)=0$. For $\texttt{PathSim}(P_1, P_2)\gg0$, the alignment path will go through different cells which are far-placed from each other in the cost matrix.

\vspace{1.0ex}

\begin{figure}[!h]
        \centering
        \begin{minipage}[t]{0.43\linewidth}
                \centering
                \includegraphics[width=\linewidth]{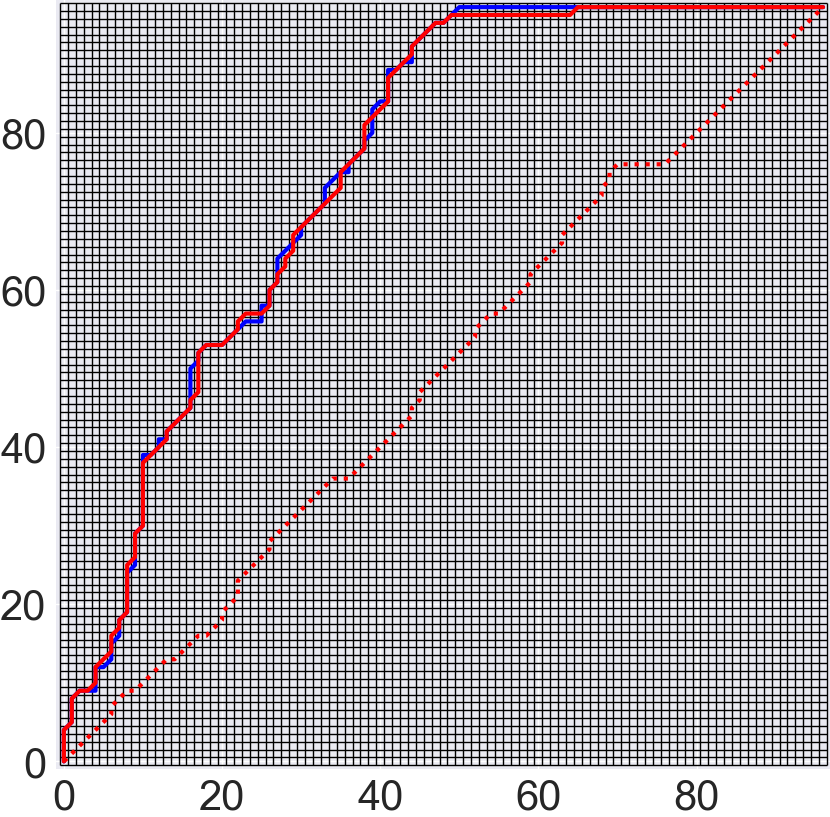}
            \end{minipage}%
            \hspace{2ex}
        \begin{minipage}[t]{.43\linewidth}
                \centering
                \includegraphics[width=\linewidth]{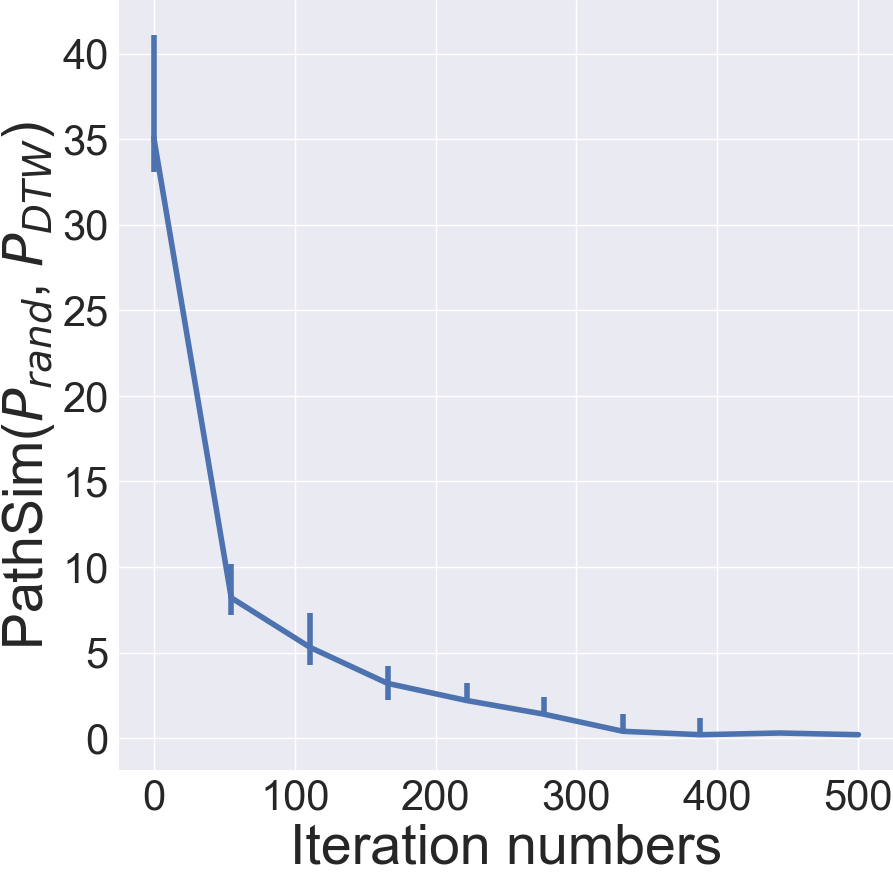}
            \end{minipage}
            
        \begin{minipage}[t]{0.43\linewidth}
                \centering
                (a) 
            \end{minipage}%
        \begin{minipage}[t]{.43\linewidth}
                \centering
                (b)
            \end{minipage}
    \caption{(a) Example of the convergence of the optimal alignment path between the adversarial example and the original example at the start of the algorithm (dotted red path) and at the end (red path) to the given random alignment path (blue path). (b) $\texttt{PathSim}$ score of the optimal alignment path between the adversarial example and the original example and the given random path for the ECG200 dataset averaged over multiple random alignment paths.}
    \label{fig:pathconvplot}
\end{figure}

\noindent {\bf Empirical tightness of bound.} Figure \ref{fig:pathconvplot} shows that over the iterations of the DTW-AR algorithm, the updated adversarial example yields to an optimal alignment path that is more similar to the input random path. This result strongly demonstrate that Theorem 2 holds empirically. 

\begin{corollary}
Let $P_1$ and $P_2$ be two alignment paths such that $\texttt{PathSim}(P_1, P_2) > 0$. If $X_{adv}^1$ and $X_{adv}^2$ are the adversarial examples generated using DTW-AR from any given time-series $X$ using paths $P_1$ and $P_2$ respectively such that $DTW(X, X_{adv}^1) = \delta$ and $DTW(X, X_{adv}^2) = \delta$, then $X_{adv}^1$ and $X_{adv}^2$ are not necessarily the same.
\end{corollary}

Theorem \ref{th:dtwgap} shows that the adversarial example generation using DTW-AR is equivalent to the ideal setting where it is possible to optimize $DTW(X,X_{adv})$. The above corollary extends Theorem \ref{th:dtwgap} to show that if we employ different alignment paths within Algorithm 1, we will be able to find more adversarial examples which meet the distance bound in contrast to the naive approach. 

\section{Related work}

\indent {\bf Adversarial methods.} Prior work on adversarial examples mostly focus on image and text domains \cite{kolter2018materials,wang2019deep}. Such methods include  Carlini \& Wagner attack \cite{carlini2017towards}, boundary attack \cite{brendel2017decision}, and universal attacks \cite{moosavi2017universal}. Recent work focuses on regularizing adversarial example generation methods to obey intrinsic properties of images \cite{laidlaw2019functional,xiao2018spatially,hosseini2017limitation}. In NLP domain, methods to fool text classifiers employ the saliency map of input words to generate adversarial examples while preserving meaning to a human reader in white-box setting \cite{samanta2017towards}. DeepWordBug \cite{gao2018black} employs a black-box strategy to fool classifiers with simple character-level transformations. Since characteristics of time-series (e.g., fast-pace oscillations, sharp peaks) are different from images and text, prior methods are not suitable to capture the appropriate notion of invariance for time-series domain.

\vspace{1ex}
\noindent {\bf Adversarial robustness.} 
Adversarial training is one of the strongest empirical defense methods against adversarial attacks \cite{tramer2020adaptive, madry2017towards}. This involves employing attack methods to create adversarial examples to augment the training data for improving robustness. Stability training \cite{zheng2016improving} is an alternative method that explicitly optimizes for robustness by defining a loss function that evaluates the classifier on small perturbations of clean examples. This method yield to a deep network that is stable against natural and adversarial distortions in the visual input.
There are other defense methods which try to overcome injection of adversarial examples \cite{athalye2018obfuscated, papernot2016distillation, kurakin2018ensemble}. However, for time-series domain, as $l_p$-norm based perturbations may not guarantee preserving the semantics of true class label, adversarial examples may mislead DNNs during adversarial training resulting in accuracy degradation.

\vspace{1.0ex}
\noindent {\bf Adversarial attacks for time-series domain.} There is little to no principled prior work on adversarial methods for time-series\footnote{In a concurrent work, Belkhouja et al., developed an adversarial framework for using statistical features \cite{TSA-STAT,ICCAD} and another min-max optimization methods \cite{RO-TS} to explicitly train robust deep models for time-series domain using global alignment kernels}. Fawaz et al., \cite{fawaz2019adversarial} employed the standard Fast Gradient Sign method \cite{kurakin2016adversarial} to create adversarial noise with the goal of reducing the confidence of deep convolutional models for classifying {\em uni-variate} time-series. Network distillation is employed to train a student model for creating adversarial attacks \cite{karim2020adversarial}. However, this method is severely limited: it can generate adversarial examples for only a small number of target labels and cannot guarantee generation of adversarial example for every input. \cite{oregiSPL18} tried to address adversarial examples with elastic similarity measures, but does not propose any elastic-measure based attack algorithm.

\vspace{1.0ex}
\noindent \textbf{Time-series pre-processing methods.} A possible solution to overcome the Euclidean distance concerns is to introduce pre-processing steps that are likely to improve the existing frameworks. Simple pre-processing steps such as MinMax-normalization or z-normalization only solves problems such as scaling problem. However, they do not address any concern about signal-warping or time-shifts. Other approaches rely on learning feature-preserving representations. A well-known example is the GRAIL\cite{papa2019grail} framework. This framework aims to learn compact time-series representations that preserve the properties of a pre-defined comparison function such as DTW. The main concern about feature-preserving pre-processing steps is that the representation learnt is not reversible. In other words, a real-world time-series signal cannot be generated from the estimated representation. The goal of adversarial attacks is to create real-world time-series that can be used to fool any DNN. Such challenges would limit the usability and the generality of methods based on pre-processing steps to study the robustness of DNNs for time-series data.

\vspace{1.0ex}
In summary, existing methods for time-series domain are lacking in the following ways: 1) they do not create targeted adversarial attacks; and 2) they employ $l_p$-norm based perturbations which do not take into account the unique characteristics of time-series data.

\section{Experiments and Results}

\label{sec:dtwspace}
\label{sec:experiments}

We empirically evaluate the DTW-AR framework and discuss the results along different dimensions.

\subsection{Experimental setup}

\vspace{1.0ex}
\noindent \textbf{Datasets.} We employ the UCR datasets benchmark \cite{ucrdata}. We present the results on five representative datasets (\texttt{AtrialFibrillation}, \texttt{Epilepsy}, \texttt{ERing}, \texttt{Heartbeat}, \texttt{RacketSports}) from diverse domains noting that our findings are general as shown by the results on remaining UCR datasets in the \textbf{Appendix}. We employ the standard training/validation/testing splits from these benchmarks.

\vspace{1.0ex}
\noindent \textbf{Configuration of algorithms.} We employ a 1D-CNN architecture for the target DNNs. We operate under a white-box (WB) setting for creating adversarial examples to fool $WB$ model. To assess the effectiveness of attacks, we evaluate the attacks under the black-box (BB) setting and to fool $BB$ model. Neural architectures of both $WB$ and $BB$ models are in the  Appendix. The adversarial algorithm has no prior knowledge/querying ability of target DNN classifiers. Target DNNs include: 1) DNN model with a different architecture trained on clean data ($BB$); 2) DNNs trained using augmented data from baselines attacks that are not specific to image domain: Fast Gradient Sign method ($FGS$) \cite{kurakin2016adversarial}, Carlini \& Wagner ($CW$) attack \cite{carlini2017towards}, and Projected Gradient Descent ($PGD$) \cite{madry2017towards}; and 3) DNN models trained using stability training \cite{zheng2016improving} ($STN$) for learning robust classifiers. 

\vspace{1.0ex}
\noindent \textbf{Evaluation metrics.} We evaluate attacks using the efficiency metric $\alpha_{Eff} \in [0,1]$ over the created adversarial examples. $\alpha_{Eff}$ (higher means better attacks) measures the capability of adversarial examples to fool a given DNN $F_{\theta}$ to output the target class-label.  $\alpha_{Eff}$ is calculated as the fraction of adversarial examples that are predicted correctly by the classifier: $\alpha_{Eff} = \frac{\text{\# Adv. examples s.t.}F(X)==y_{target}}{\text{\# Adv. examples}}$.
We evaluate adversarial training by measuring the accuracy of the model to predict ground-truth labels of adversarial examples. A DNN classifier is robust if it is successful in predicting the true label of any given adversarial example.

\label{sec:results}
\subsection{Results and Discussion}

\noindent \textbf{Spatial data distribution with DTW.}  
We have shown in Figure \ref{fig:dtwl2space} how the data from different class labels are better clustered in the DTW space compared to the Euclidean space. These results demonstrate that DTW suits better the time-series domain as generated adversarial examples lack true-label guarantees. Moreover, Euclidean distance based attacks can potentially create adversarial examples that are {\em inconsistent} for adversarial training. Our analysis showed that for datasets such as \texttt{WISDM}, there are time-series signals from different classes with $l_2$-distances $\le 2$, while PGD or FGS require $\epsilon \ge 2$ to create successful adversarial examples for more than $70\%$ time-series instances. We provide in the Appendix an additional visualization of the adversarial examples using DTW.

\begin{figure}[!h]
    \centering
    \begin{minipage}{.29\linewidth}
        \centering
        \includegraphics[width=\linewidth]{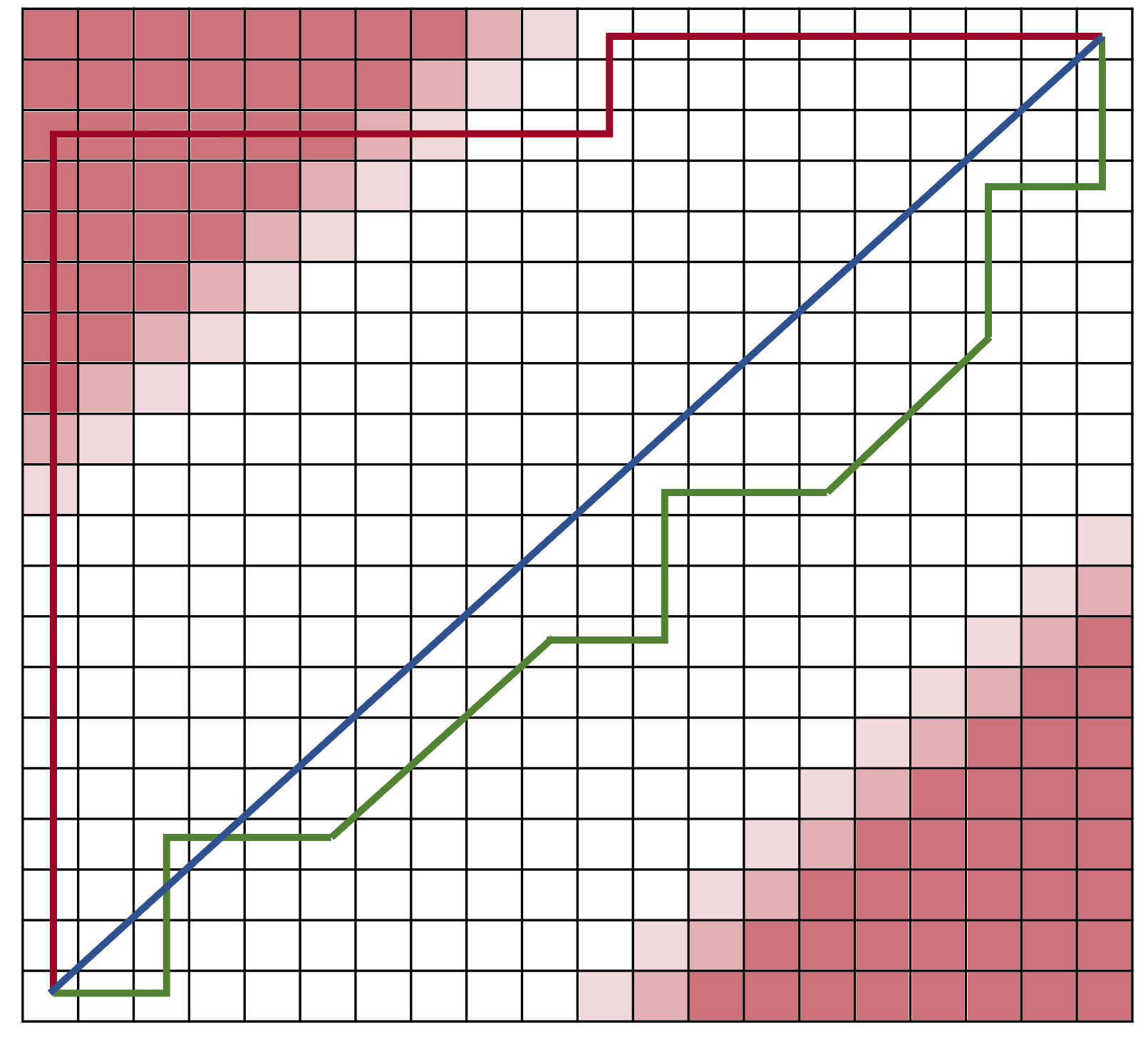}
    \end{minipage}
    \hfill
    \begin{minipage}{.6\linewidth}
        \centering
        \includegraphics[width=\linewidth]{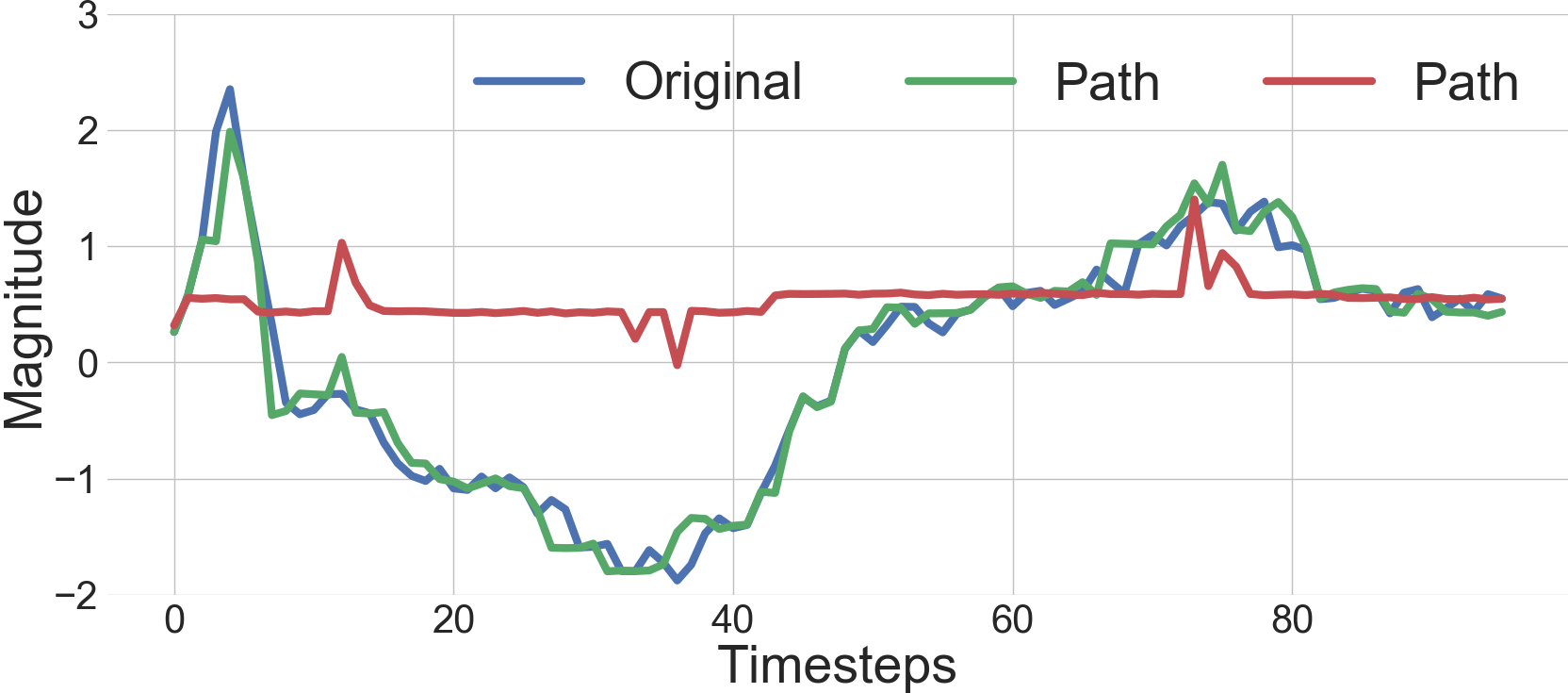}
    \end{minipage}
    \caption{Effect of alignment path on adversarial example.}
    \label{fig:goodpath}
\end{figure}

\vspace{1.0ex}
\noindent\textbf{Admissible alignment paths.}  The main property of DTW alignment is the one-to-many match between time-steps to identify similar warped pattern. Intuitively, if an alignment path matches few time-steps from the first signal with too many steps in the second signal, both signals are not considered similar. Consequently, the optimal path would be close to the corners of the cost matrix. Figure \ref{fig:goodpath} provides a comparison between two adversarial signals generated using a green colored path closer to the diagonal vs. a red colored path that is close to the corners. We can see that the red path produces an adversarial example that is not similar to the original input. Hence, we limit the range of the random path $P_{rand}$ used  to a safe range omitting the cells at the top and bottom halves of the top-left and bottom-right corners. 

\vspace{1.0ex}
\noindent {\bf Multiple diverse adversarial examples using DTW-AR.} In section 3.2, we argued that using stochastic alignment paths, we can create multiple diverse adversarial time-series examples within the same DTW measure bound. DTW-AR method leverages the large pool of candidate alignment paths to uncover more than one adversarial example as illustrated in Figure \ref{fig:dtwsubspace}. To further test this hypothesis, we perform the following experiment. We sample a subset of different (using $\texttt{PathSim}$) alignment paths $\{P_{rand}\}_{i}$ and execute DTW-AR algorithm to create adversarial examples for the same time-series $X$. Let $X_{adv, i}$ be the adversarial example generated from $X$ using $P_{rand,i}$. We measure the similarities between the generated $\{X_{adv}\}_{i}$ using DTW and $l_2$ distance. If the distance between two adversarial examples is less than a threshold $\epsilon_{sim}$, then they are considered the same adversarial example. 
\begin{table}[t]
    \centering
    \caption{Average percentage of dissimilar adversarial examples created by DTW-AR using stochastic alignment paths for a given time-series. The  threshold $\epsilon_{sim}$ determines whether two adversarial examples are dissimilar or not based on $l_2$ and DTW measures.}
    \begin{tabular}{|l|c|c|c|c|c|c|} 
    \hline
      & \multicolumn{3}{c|}{$\epsilon_{sim} ~l_2$ norm} & \multicolumn{3}{c|}{$\epsilon_{sim}$ DTW} \\ \cline{2-7}
     & 0.01 & 0.05 & 0.1 & 0.01 & 0.05 & 0.1\\ \hline
     Atrial Fibrillation&  98\% & 90\% & 87\% & 100\% & 100\%& 98\% \\ \hline
     Epilepsy &  99\% & 96\% & 93\% & 100\% & 100\%& 97\% \\ \hline
     ERing &  99\% & 95\% & 93\% & 100\% & 100\%& 98\% \\ \hline
     Heartbeat &  99\% & 94\% & 92\% & 100\% & 99\%& 98\% \\ \hline
     RacketSports &  99\% & 94\% & 92\% & 100\% & 100\%& 96\% \\ \hline
    \end{tabular}
    \label{tab:advpiechart}
\end{table}
Table \ref{tab:advpiechart} shows the percentage of adversarial examples generated using different alignment paths from a given time-series signal that are not similar to any other adversarial example. We conclude that DTW-AR algorithm indeed creates multiple different adversarial examples from a single time-series signal for the same DTW measure bound.

\vspace{1.0ex}
\noindent {\bf Empirical justification for Theorem 2.}
We provided a proof for the gap between creating an adversarial example using the proposed DTW-AR algorithm and an ideal DTW algorithm. In Figure \ref{fig:pathconvplot}(a), we provide an illustration of the optimal alignment path update using DTW-AR. This experiment was performed on the ECG200 dataset as an example (noting that we observed similar patterns for other datasets as well): the blue path represents the selected random path to be used by DTW-AR and the red path represents the optimal alignment path computed by DTW. At the beginning, the optimal alignment path (dotted path) and the random path are dissimilar. However, as the execution of DTW-AR progresses, the updated adversarial example yields to an optimal alignment path similar to the random path. In Figure \ref{fig:pathconvplot}(b), we show the progress of the $\texttt{PathSim}$ score as a function of the iteration numbers of Algorithm 1. 
\begin{figure}[b]
    \centering
    \begin{minipage}{\linewidth}
        \begin{minipage}{.46\linewidth}
                \centering
                \includegraphics[width=\linewidth]{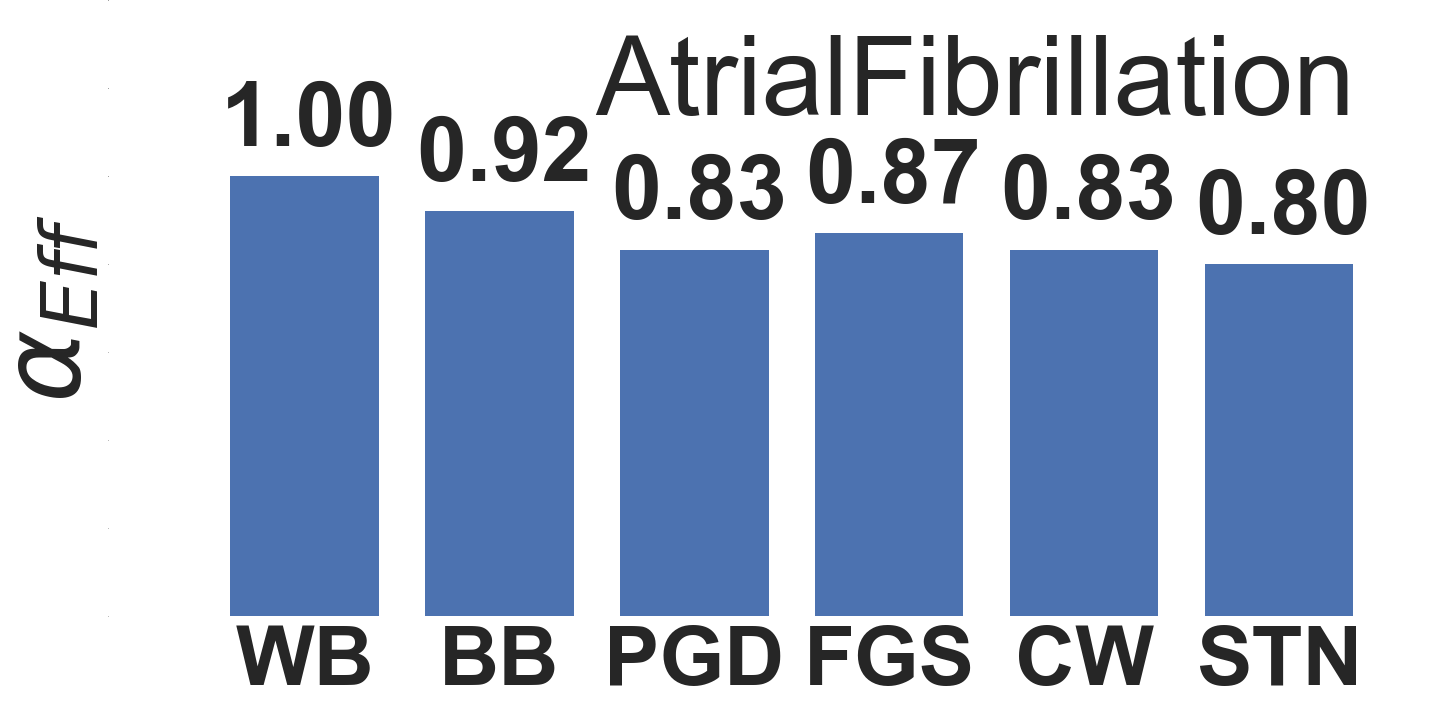}
            \end{minipage}%
            \hfill
        \begin{minipage}{.46\linewidth}
                \centering
                \includegraphics[width=\linewidth]{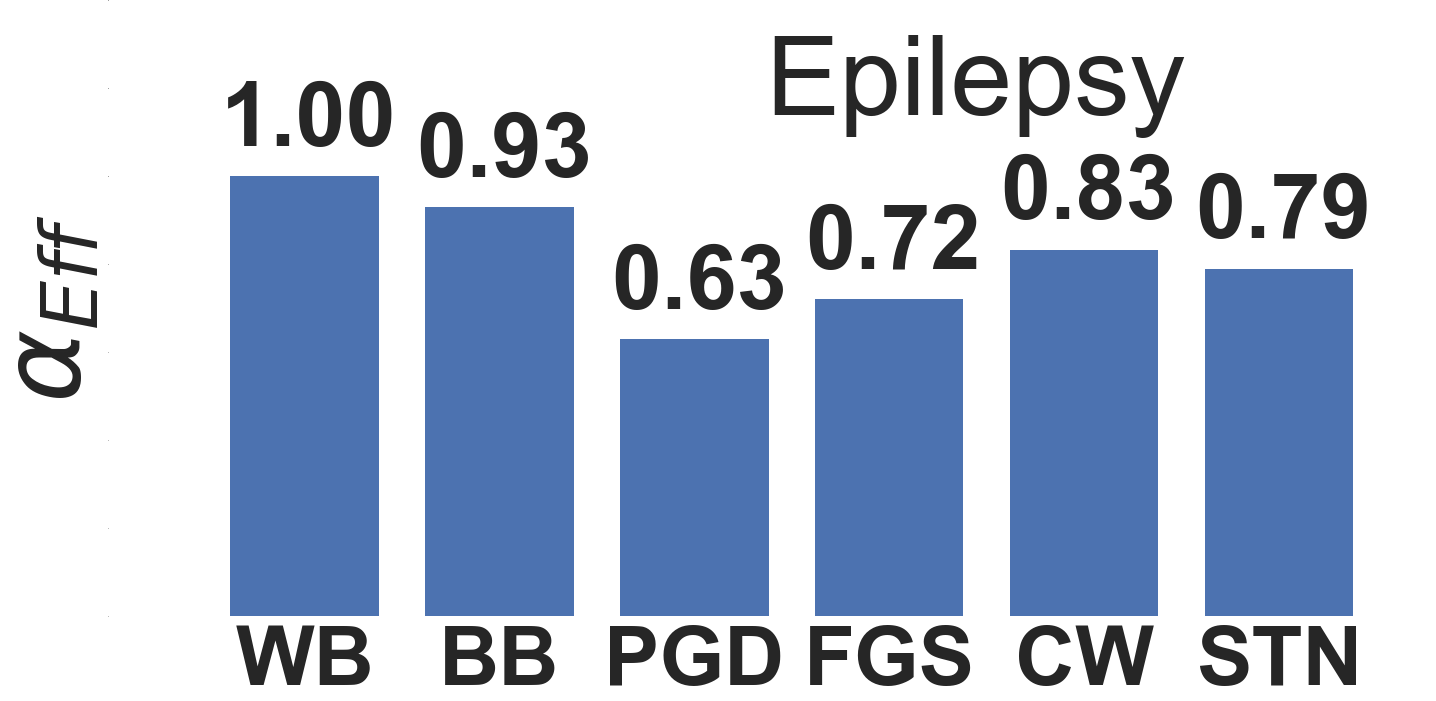}
            \end{minipage}
        \begin{minipage}{.46\linewidth}
                \centering
                \includegraphics[width=\linewidth]{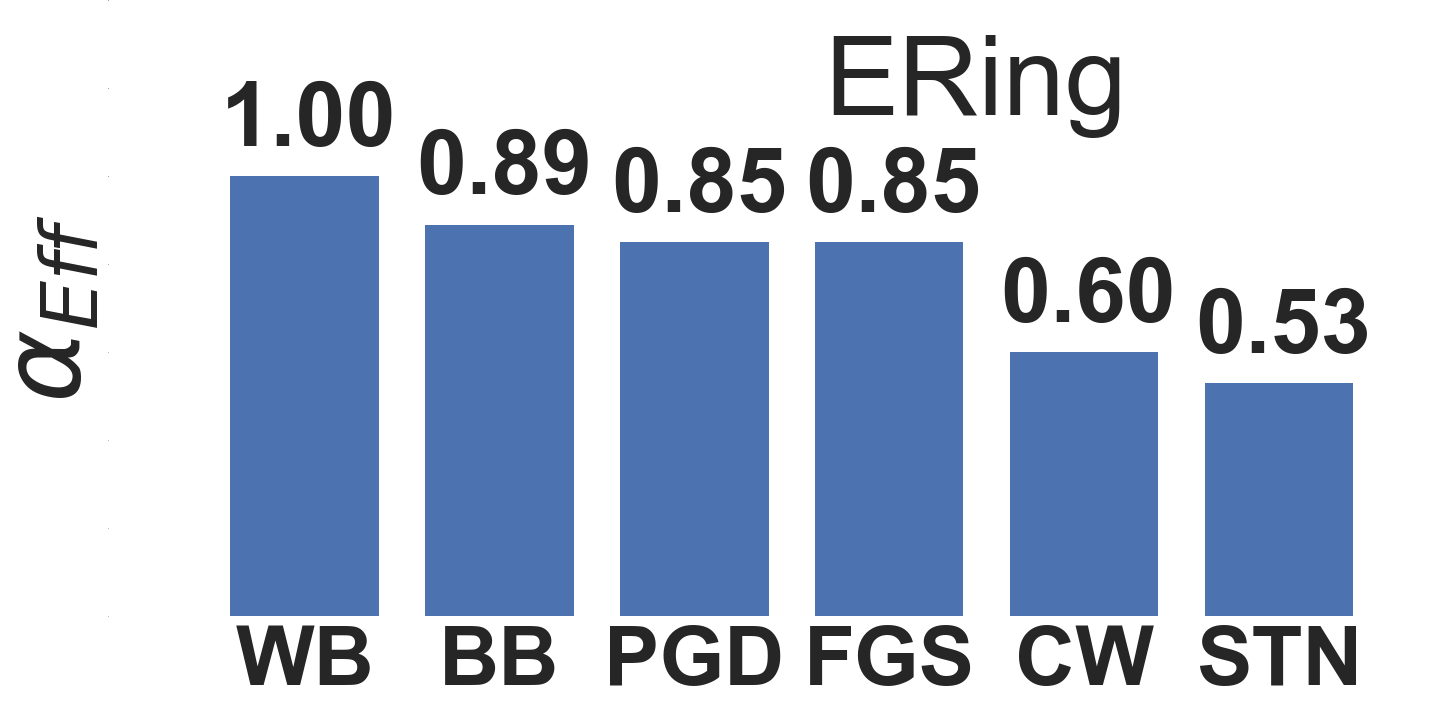}
            \end{minipage}%
            \hfill
        \begin{minipage}{.46\linewidth}
                \centering
                \includegraphics[width=\linewidth]{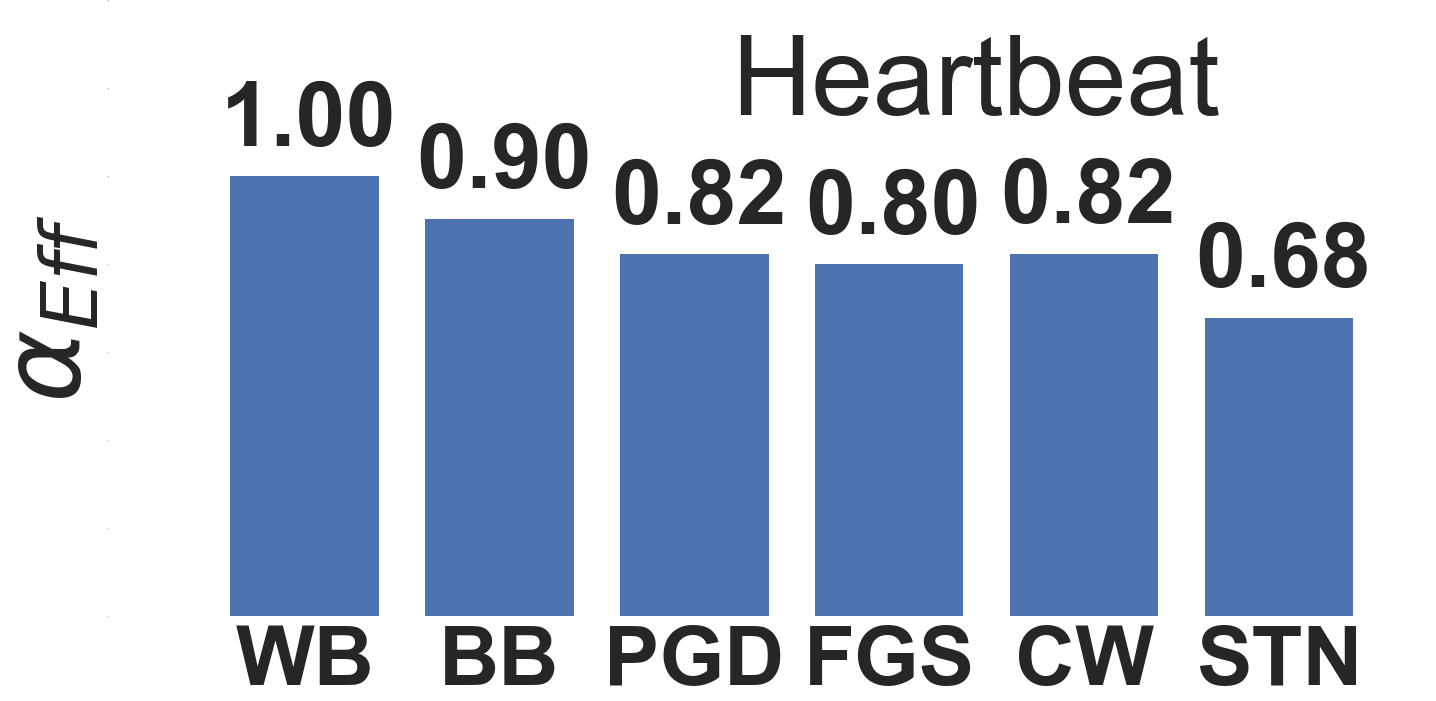}
            \end{minipage}
            \centering
        \begin{minipage}{.46\linewidth}
                \centering
                \includegraphics[width=\linewidth]{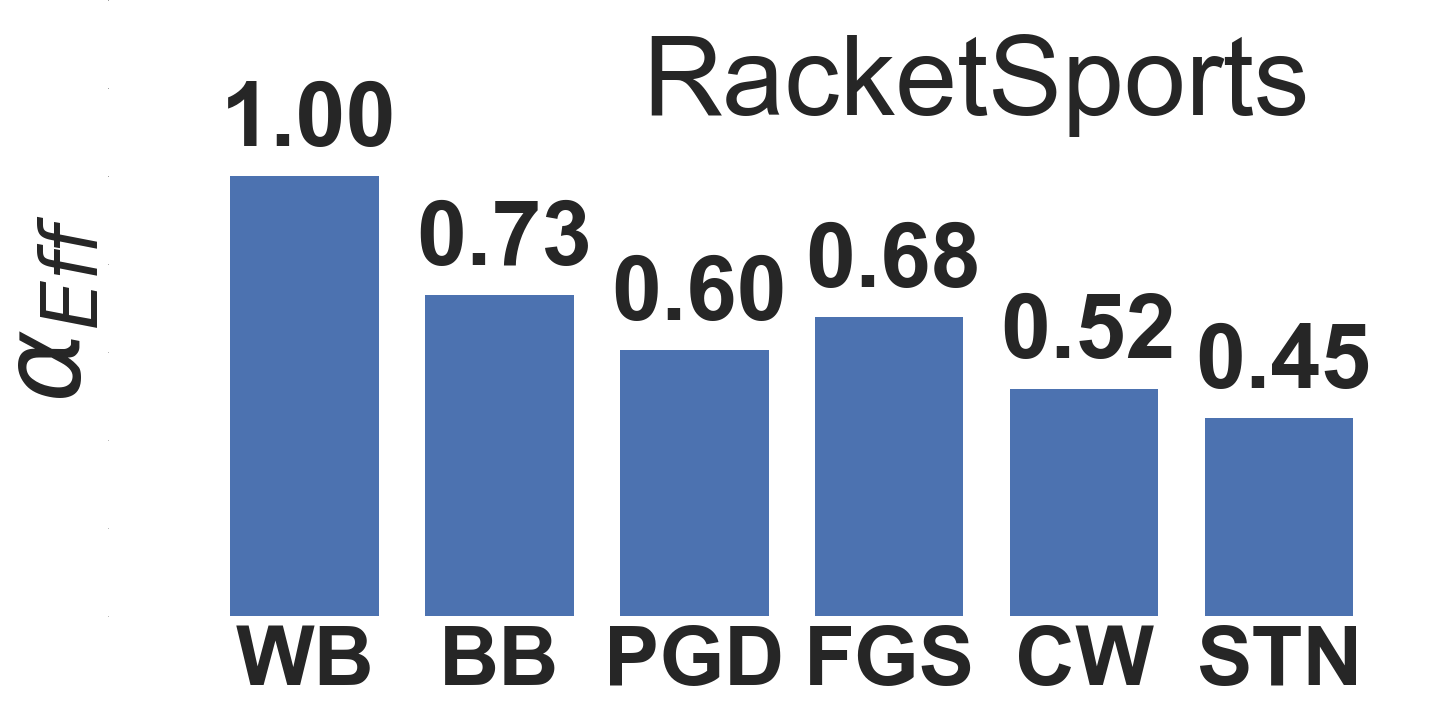}
            \end{minipage}
    
    \end{minipage}
\caption{Results for the effectiveness of adversarial examples from DTW-AR on different DNNs under white-box (WB) and black-box (BB) settings, and using adversarial training baselines (PGD, FGS, CW and STN) on different datasets.} 
\label{fig:advatk}
\end{figure}
This figure shows the convergence of the $\texttt{PathSim}$ score to 0. These strong results confirm the main claim of Theorem 2 that the resulting adversarial example $X_{adv}$ from the minimization over $dist_{P_{rand}}(X,X_{adv})$ is equivalent to minimizing over $DTW(X,X_{adv})$!

\begin{figure}[!h]
    \centering
    \includegraphics[width=.6\linewidth]{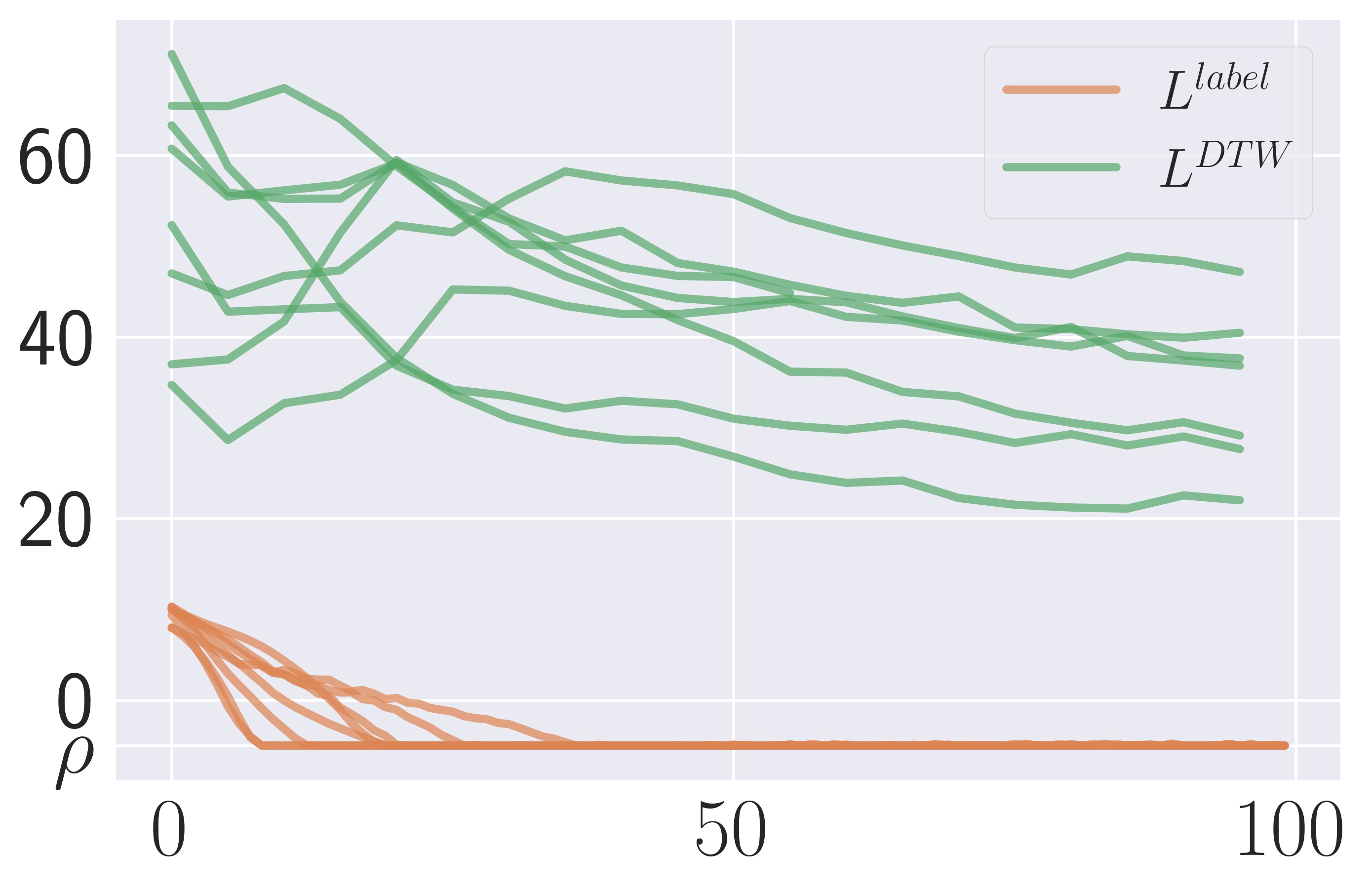}
    \caption{The progress of loss function values over the first 100 iterations of DTW-AR on different examples from \textsc{AtrialFibrillation} datasets. We observe that empirically, the Equation $\star$ plateaus at $\rho$ before minimizing $L^{DTW}$.}
    \label{fig:losscurve}
    \end{figure}

\vspace{1.0ex}
\noindent\textbf{Loss function scaling.}As the final loss function is using two different terms to create adversarial attacks, the absence of a scaling parameter can affect the optimization process. In Figure \ref{fig:losscurve}, we demonstrate that empirically, the first term of the Equation $\star$ plateaus at $\rho$ before minimizing $\mathcal{L}^{DTW}$. The figure shows the progress of both $\mathcal{L}^{label}$ and $\mathcal{L}^{DTW}=\alpha_1 \times dist_{P}(X,X_{adv})$ over the first 100 iterations of DTW-AR algorithm. We conclude that there is no need to scale the loss function noting that our findings were similar for other time-series datasets. In the general case, if a given application requires attention to scaling both terms ($\mathcal{L}^{DTW}$ and $\mathcal{L}^{label}$), the learning rate can be adjusted to two different values: Instead of having $\eta*\nabla L = \eta*\nabla L^{label}+\eta*\nabla L^{DTW}$, we can use a learning rate pair $\eta=(\eta_1, \eta_2)$ and gradient descent step becomes $\eta*\nabla L = \eta_1*\nabla L^{label}+\eta_2*\nabla L^{DTW}$.

\begin{figure}[!h]
    \centering
    \begin{minipage}{\linewidth}
        \begin{minipage}{.46\linewidth}
                \centering
                \includegraphics[width=\linewidth]{PerfFigures/AtkPeroformanceAtrialFibrillation.png}
            \end{minipage}%
            \hfill
        \begin{minipage}{.46\linewidth}
                \centering
                \includegraphics[width=\linewidth]{PerfFigures/AtkPeroformanceEpilepsy.png}
            \end{minipage}
        \begin{minipage}{.46\linewidth}
                \centering
                \includegraphics[width=\linewidth]{PerfFigures/AtkPeroformanceERing.png}
            \end{minipage}%
            \hfill
        \begin{minipage}{.46\linewidth}
                \centering
                \includegraphics[width=\linewidth]{PerfFigures/AtkPeroformanceHeartbeat.png}
            \end{minipage}
            \centering
        \begin{minipage}{.46\linewidth}
                \centering
                \includegraphics[width=\linewidth]{PerfFigures/AtkPeroformanceRacketSports.png}
            \end{minipage}
    \vspace{.5em}
    \end{minipage}
    \begin{minipage}{\linewidth}
            \centering
            {\small DTW-AR Setting: $\alpha_2 \neq 0$}
    \vspace{.5em}
    \end{minipage}
    \begin{minipage}{\linewidth}
        \begin{minipage}{.46\linewidth}
                \centering
                \includegraphics[width=\linewidth]{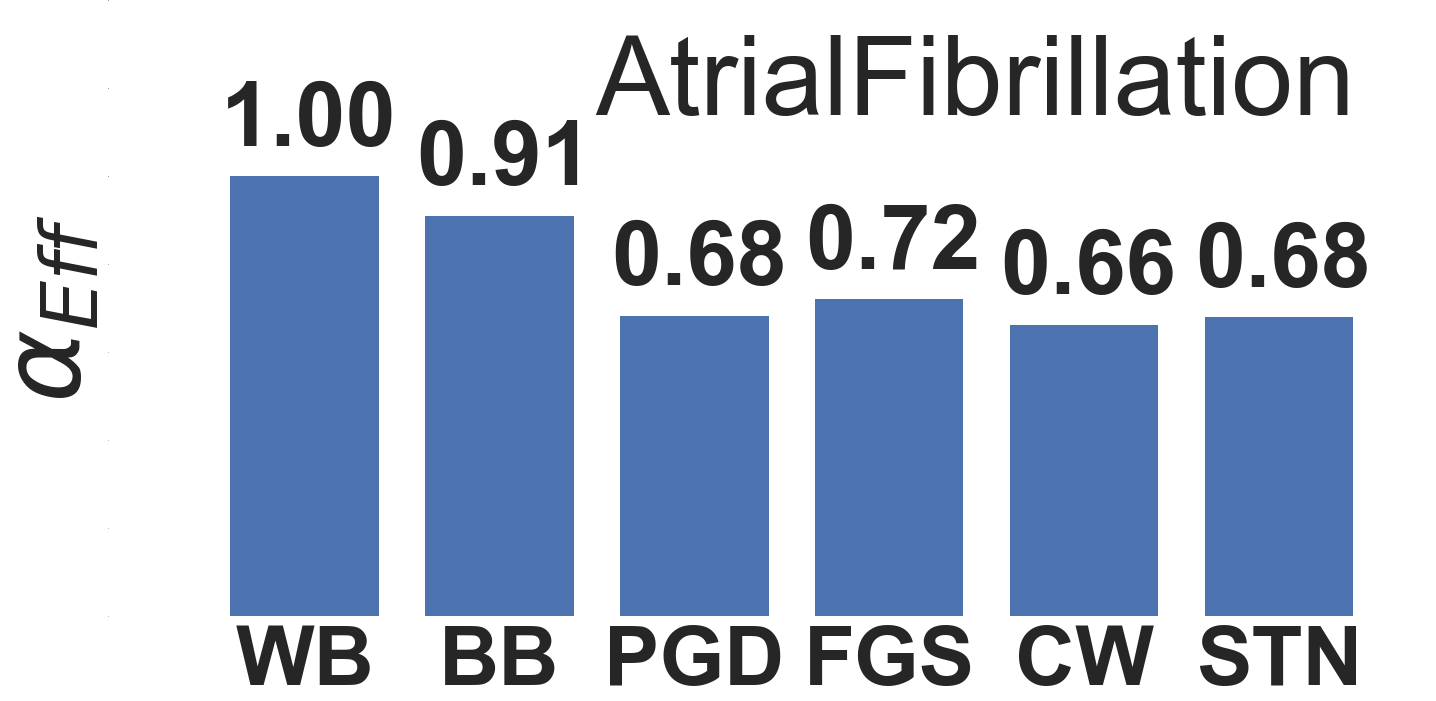}
            \end{minipage}%
            \hfill
        \begin{minipage}{.46\linewidth}
                \centering
                \includegraphics[width=\linewidth]{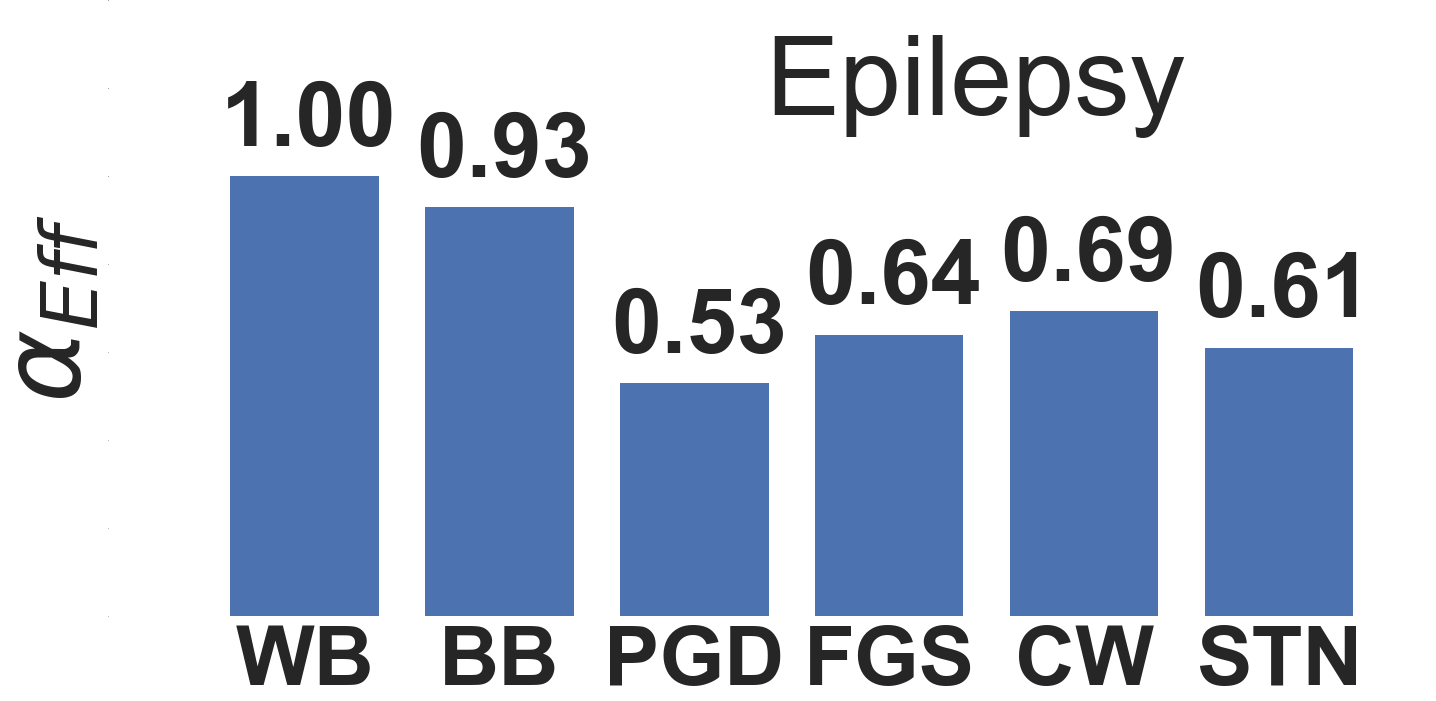}
            \end{minipage}
        \begin{minipage}{.46\linewidth}
                \centering
                \includegraphics[width=\linewidth]{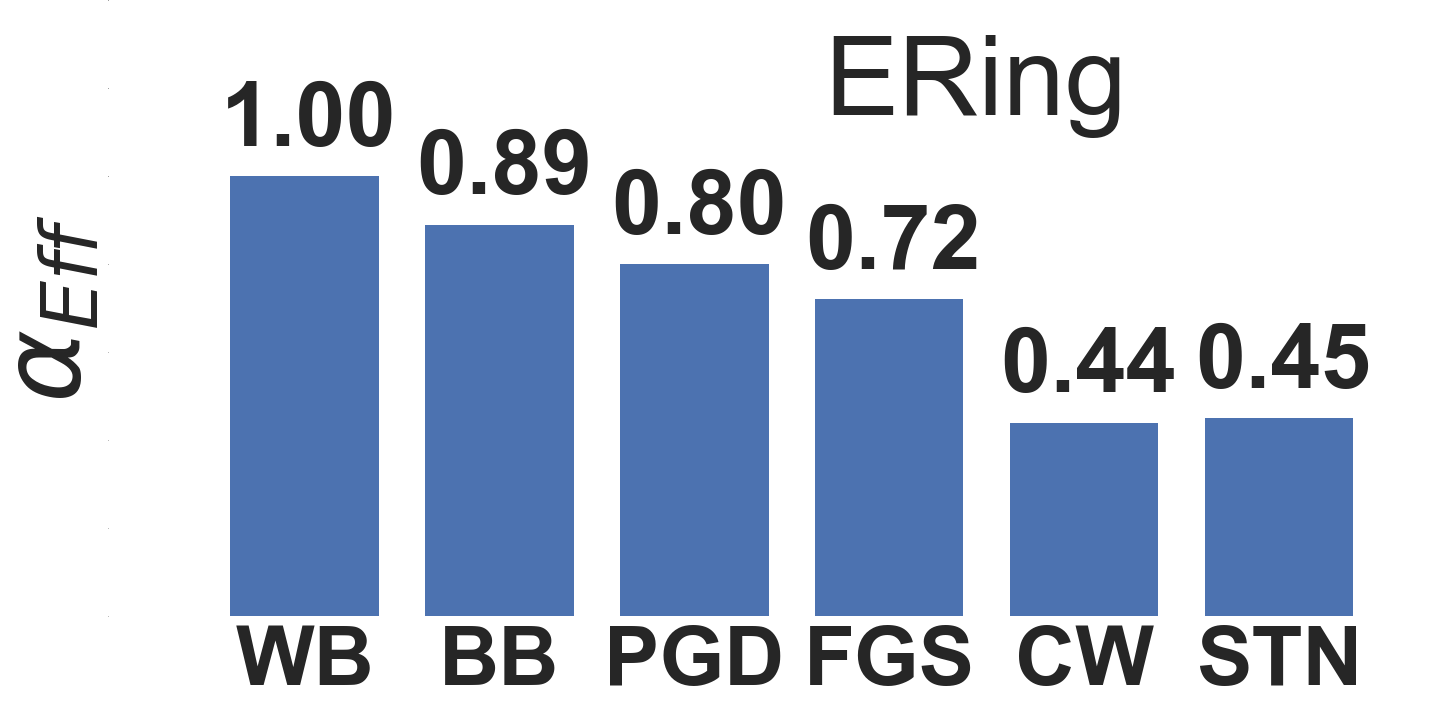}
            \end{minipage}%
            \hfill
        \begin{minipage}{.46\linewidth}
                \centering
                \includegraphics[width=\linewidth]{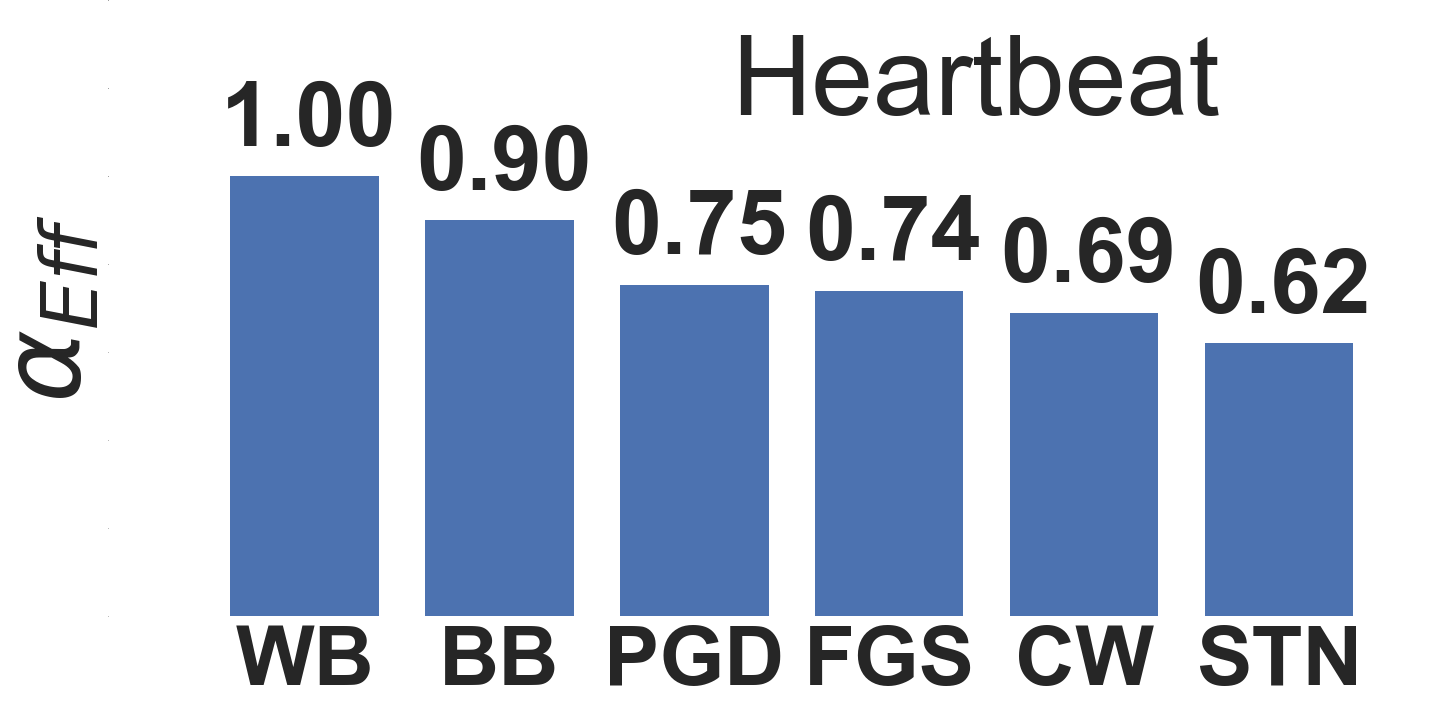}
            \end{minipage}
            \centering
        \begin{minipage}{.46\linewidth}
                \centering
                \includegraphics[width=\linewidth]{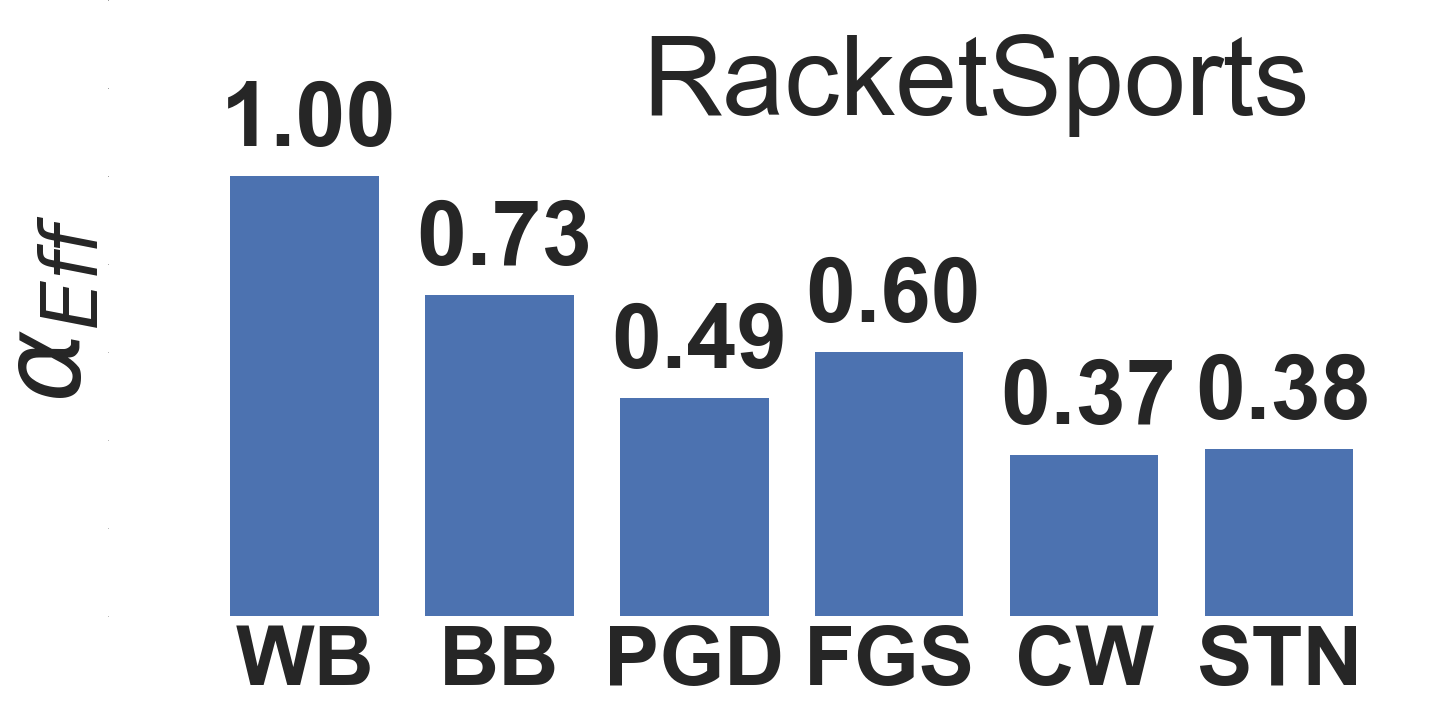}
        \end{minipage}
    \vspace{.5em}
    \end{minipage}
    \begin{minipage}{\linewidth}
            \centering
            {\small DTW-AR Setting: $\alpha_2 = 0$}
    \end{minipage}
\caption{Results for the effectiveness of adversarial examples from DTW-AR on different DNNs under white-box (WB) and black-box (BB) settings, and using adversarial training baselines (PGD, FGS, CW and STN) on different datasets under two attack settings:$\alpha_2 \neq 0$ and $\alpha_2=0$.} 
\label{fig:alphaadvatk}
\end{figure}

\vspace{1.0ex}
\noindent\textbf{Effectiveness of adversarial attacks.} Results of the fooling rate of DTW-AR generated attacks for different models are shown in Figure \ref{fig:advatk}. We observe that under the white-box setting (WB model), we have $\alpha_{Eff}$=1. This shows that for any $y_{target}$, DTW-AR successfully generates an adversarial example for every input in the dataset. For black-box setting (BB model) and other models using baseline attacks for adversarial training, we see that DTW-AR attack is highly effective for most cases. We conclude that these results support the theoretical claim made in Theorem 1 by showing that standard $l_2$-norm based attacks have blind spots and the DTW bias is appropriate for time-series. The importance of $\alpha_2$ in Equation \ref{eq:dtwloss} is shown to improve the fooling rate of adversarial examples. To implement DTW-AR adversarial attacks, we have fixed $\alpha_1=0.5$ and $\alpha_2=0.5$ for Equation \ref{eq:dtwloss}. The importance of $\alpha_2$ is to push the algorithm to create adversarial examples out of the scope of the Euclidean space as shown in Figure \ref{fig:alphaadvatk}. 
The adversarial examples with $\alpha_2 \neq 0$ evade DNNs with adversarial training baselines better than the examples with $\alpha_2=0$.

\begin{figure}[!h]
    \centering
        \begin{minipage}{\linewidth}
        \begin{minipage}{.46\linewidth}
                \centering
                \includegraphics[width=\linewidth]{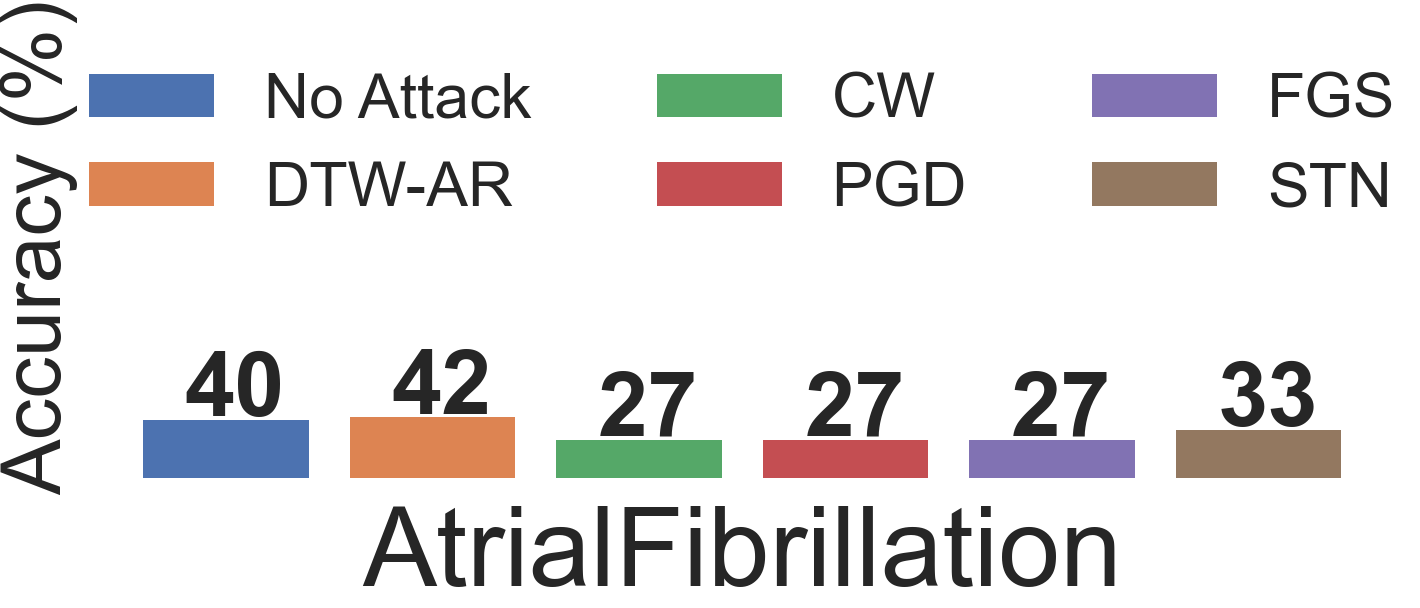}
            \end{minipage}%
            \hfill
        \begin{minipage}{.46\linewidth}
                \centering
                \includegraphics[width=\linewidth]{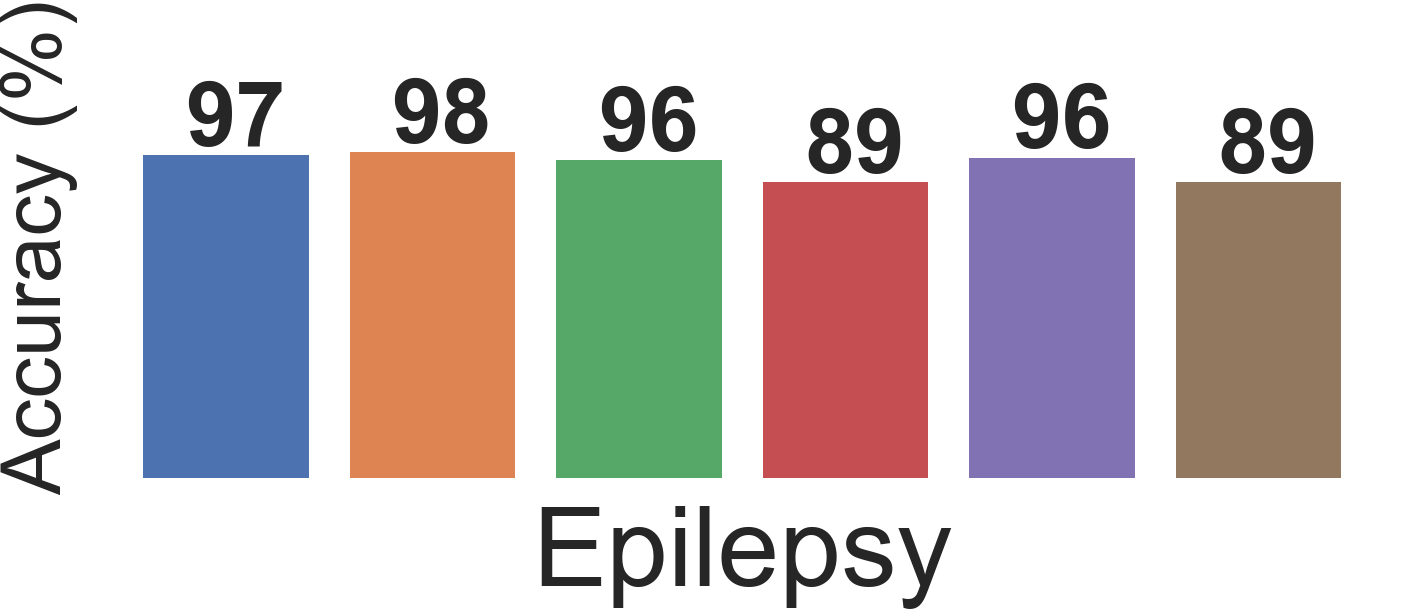}
            \end{minipage}
        \begin{minipage}{.46\linewidth}
                \centering
                \includegraphics[width=\linewidth]{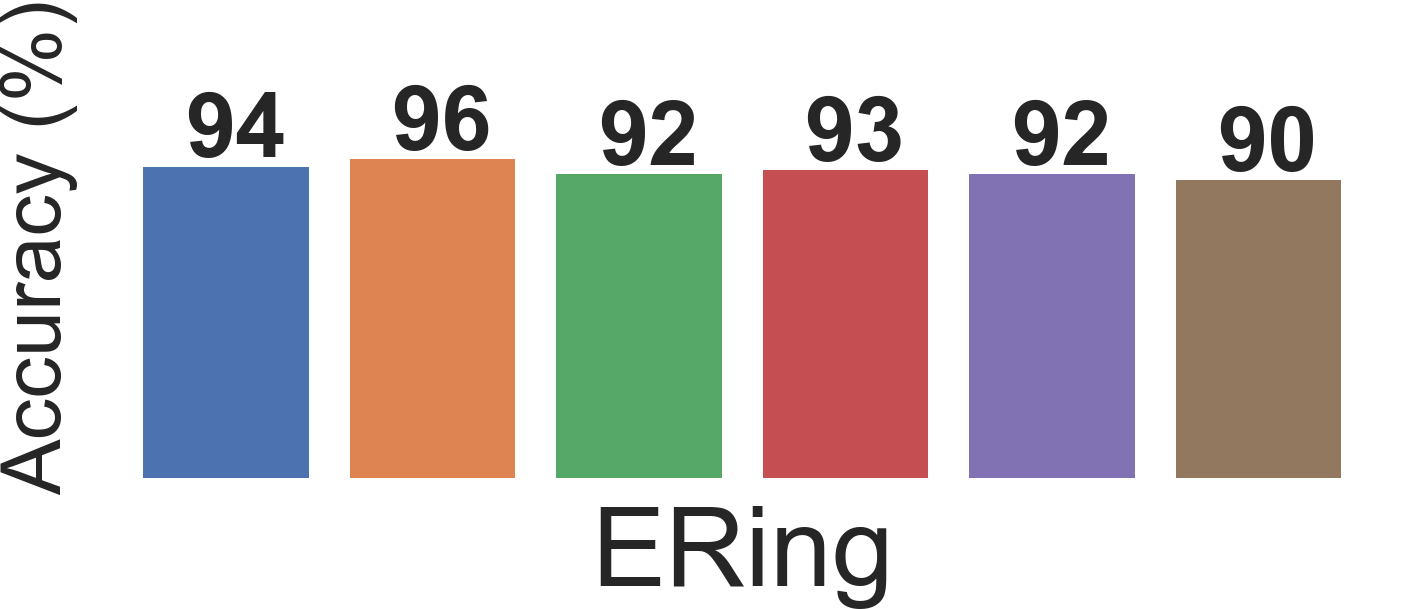}
            \end{minipage}%
            \hfill
        \begin{minipage}{.46\linewidth}
                \centering
                \includegraphics[width=\linewidth]{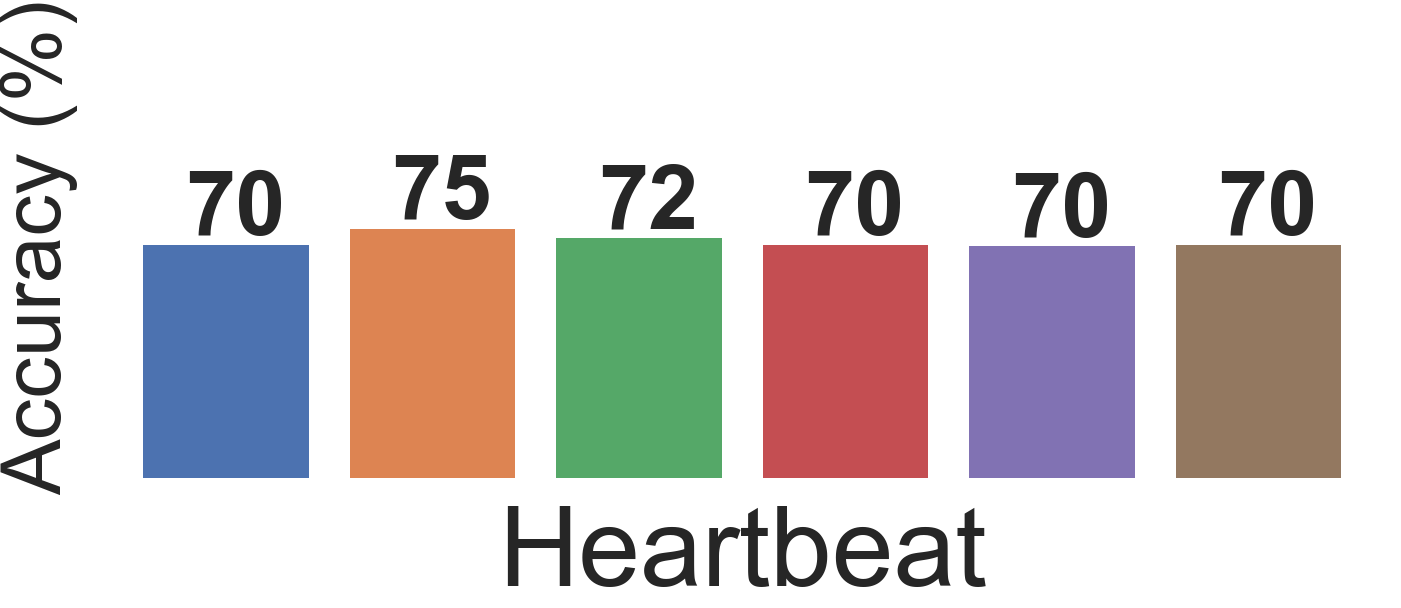}
            \end{minipage}
            \centering
        \begin{minipage}{.46\linewidth}
                \centering
                \includegraphics[width=\linewidth]{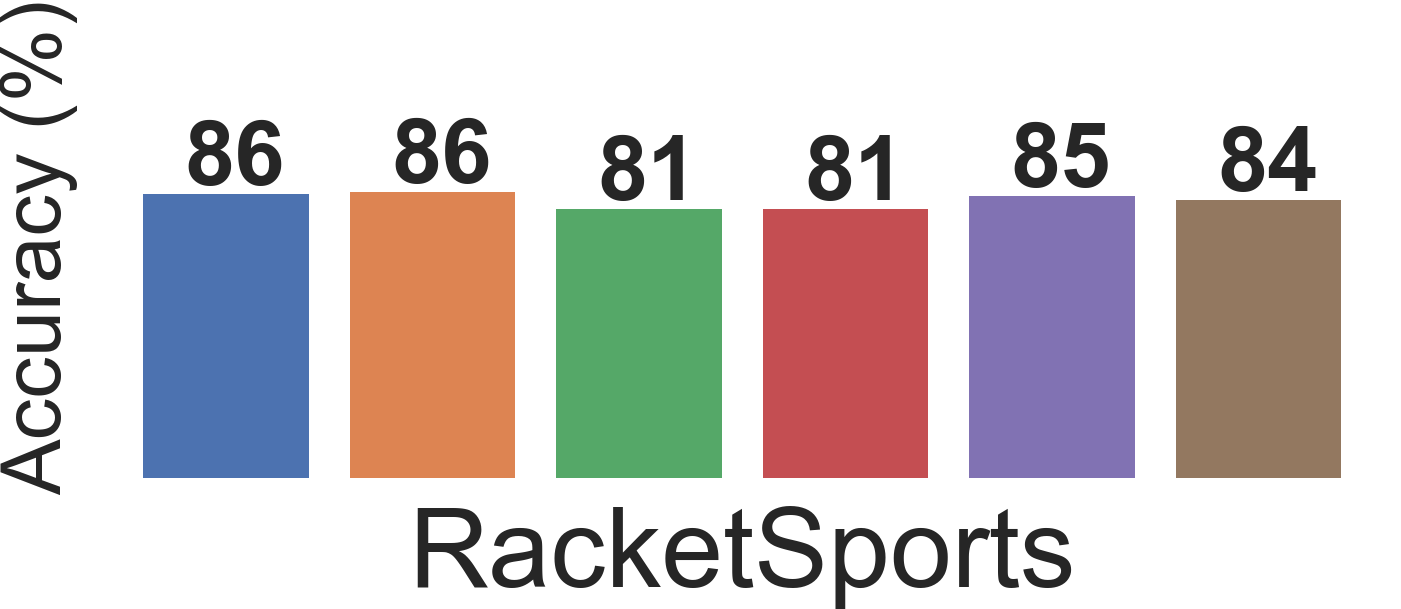}
            \end{minipage}
    \end{minipage}
\caption{Results of adversarial training using baseline attacks and DTW-AR, and comparison with standard training without adversarial examples (No Attack) to classify clean data.}
\label{fig:cleanadvdef}
\end{figure}

\begin{figure}[!h]
    \centering
    \begin{minipage}{\linewidth}
        \begin{minipage}{.46\linewidth}
                \centering
                \includegraphics[width=\linewidth]{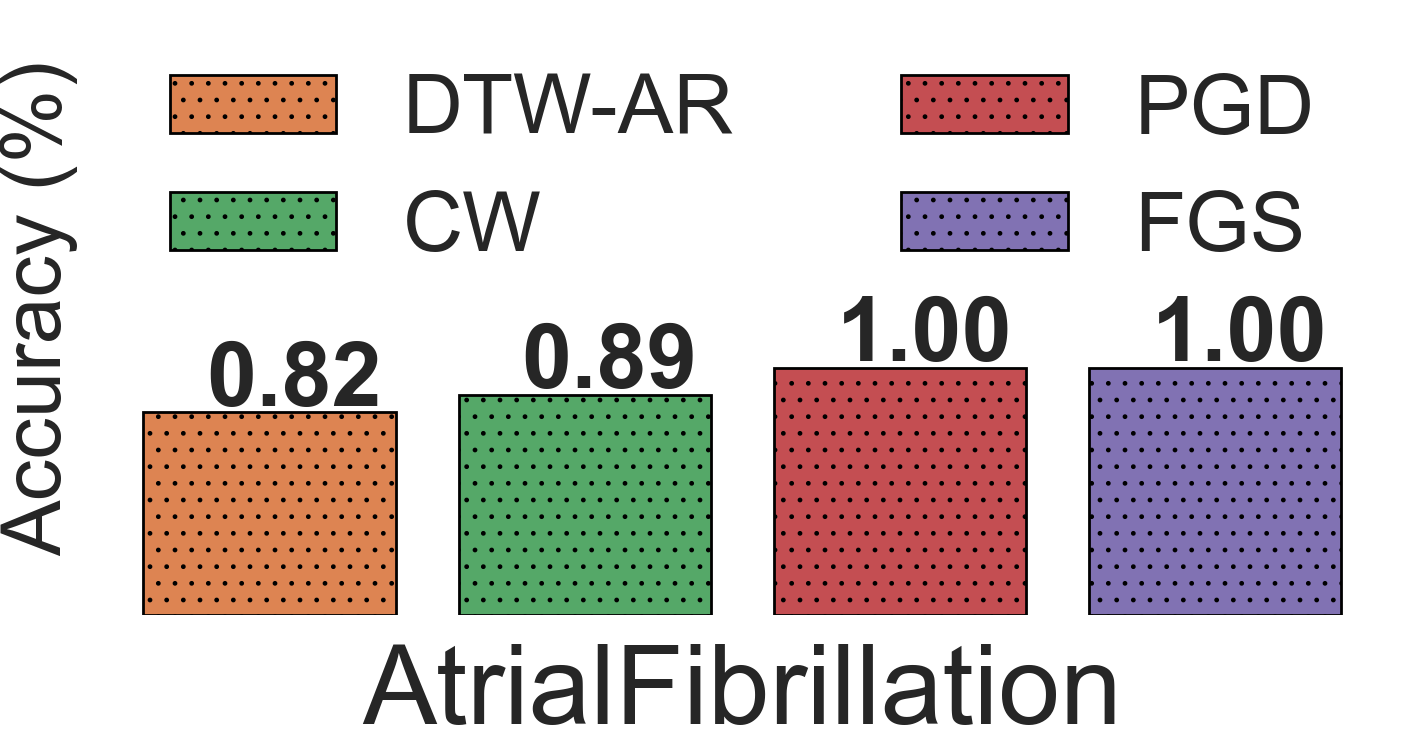}
            \end{minipage}%
            \hfill
        \begin{minipage}{.46\linewidth}
                \centering
                \includegraphics[width=\linewidth]{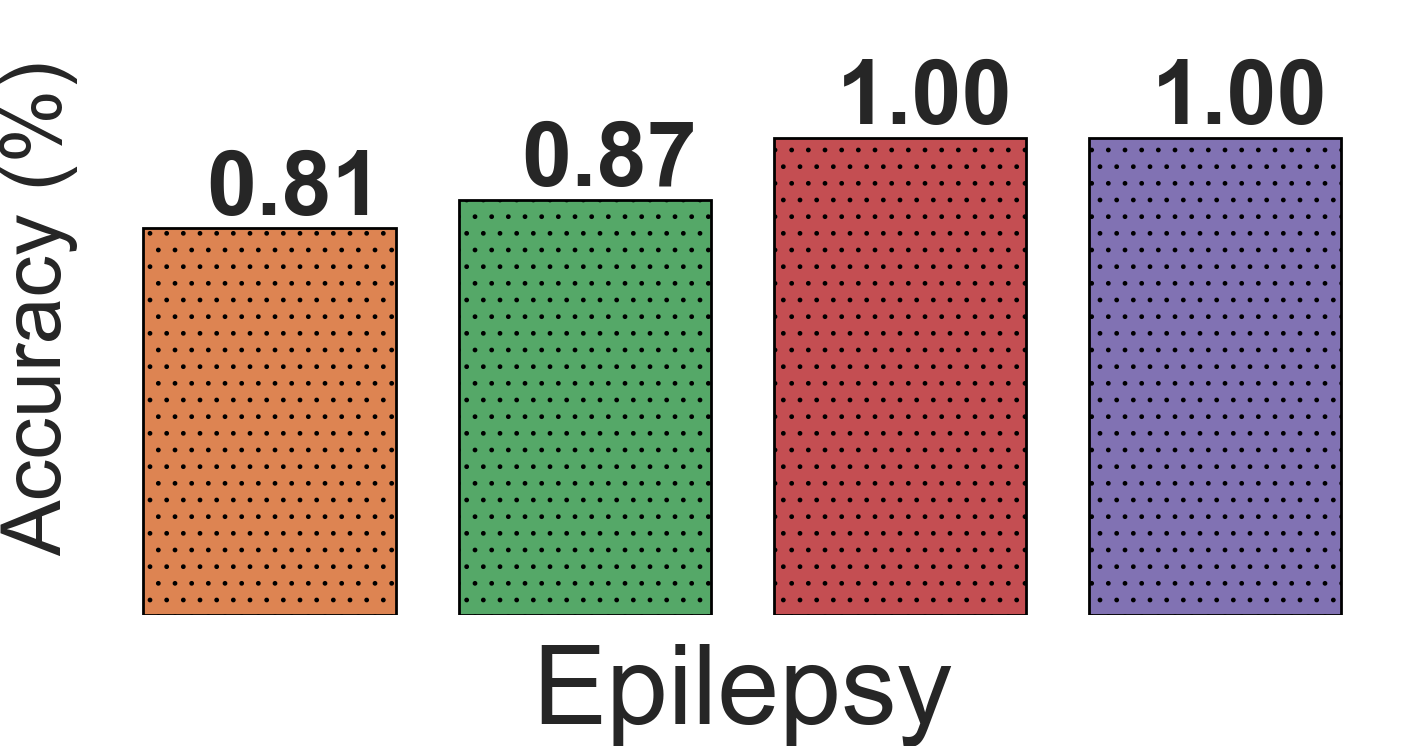}
            \end{minipage}
        \begin{minipage}{.46\linewidth}
                \centering
                \includegraphics[width=\linewidth]{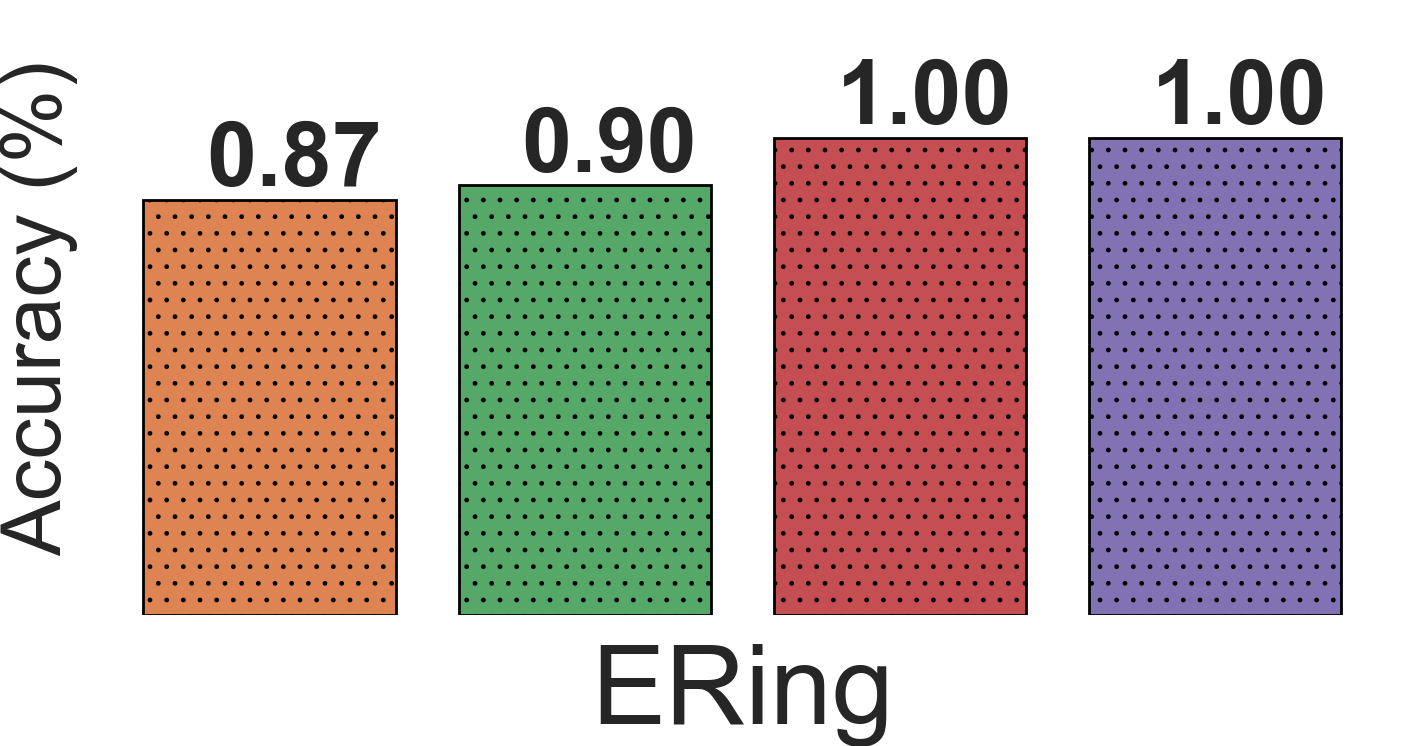}
            \end{minipage}%
            \hfill
        \begin{minipage}{.46\linewidth}
                \centering
                \includegraphics[width=\linewidth]{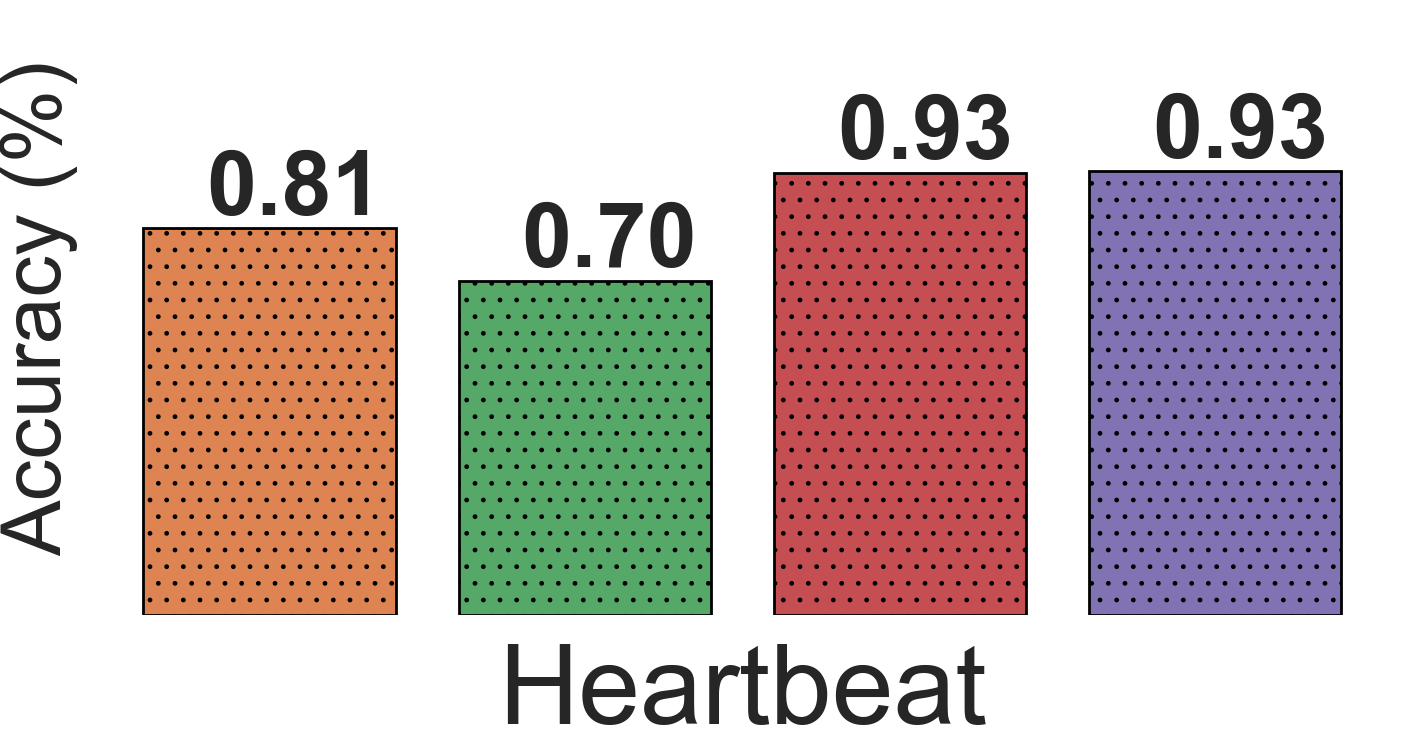}
            \end{minipage}
            \centering
        \begin{minipage}{.46\linewidth}
                \centering
                \includegraphics[width=\linewidth]{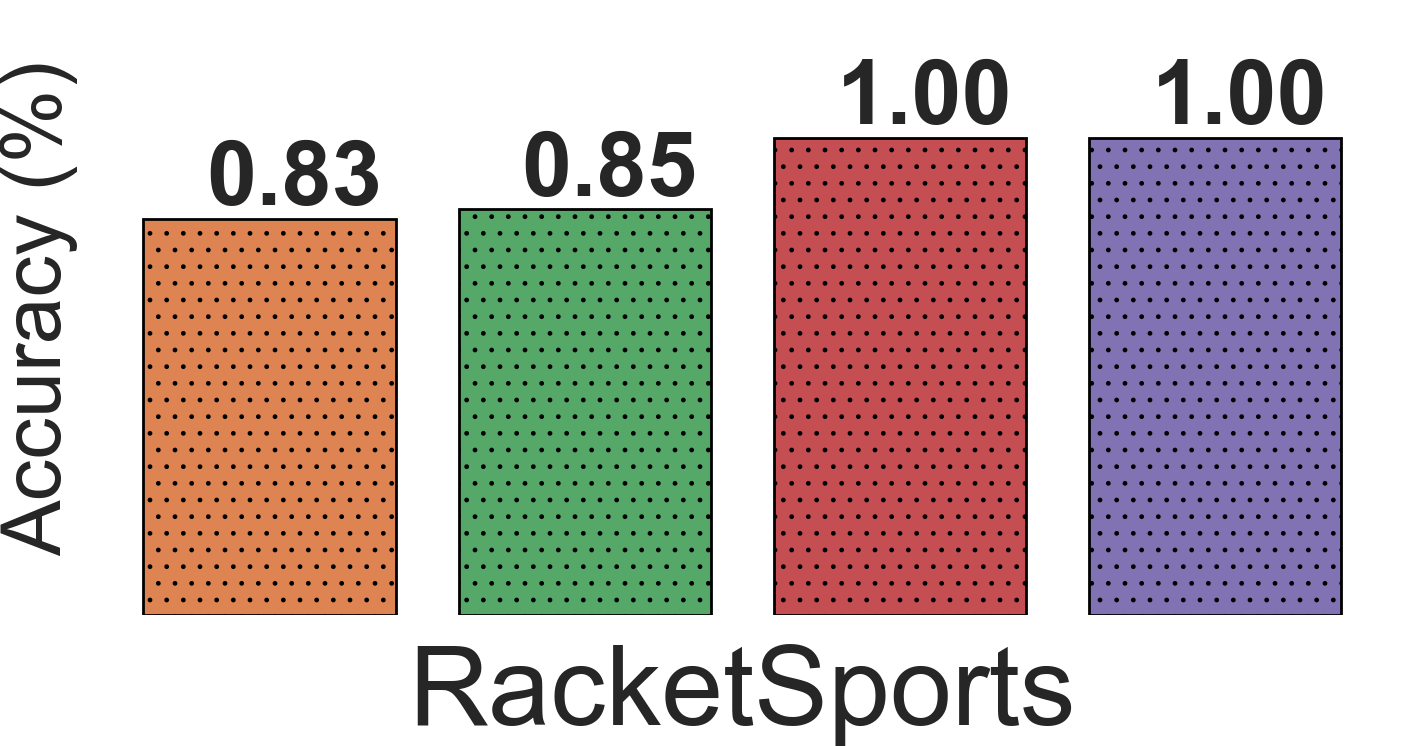}
            \end{minipage}
    \end{minipage}
\caption{Results of DTW-AR based adversarial training to predict the true labels of adversarial examples generated by DTW-AR and the baseline attack methods. The adversarial examples considered are those which successfully fooled DNNs that do not use adversarial training.}
\label{fig:advdef}
\end{figure}

\vspace{1.0ex}
\noindent\textbf{DTW-AR based adversarial training.} Our hypothesis is that $l_2$-based perturbations lack true-label guarantees and can degrade the overall performance of DNNs. Figure \ref{fig:cleanadvdef} shows the accuracy of different DNNs after adversarial training on clean data. This performance is relative to the clean testing set of each dataset. We observe that all $l_2$-based methods degrade the performance using adversarial training for at least one dataset. However, for many datasets, the performance is visibly improved using DTW-AR based adversarial training. Compared to standard training (i.e., no augmented adversarial examples), the performance on \texttt{AtrialFibrillation} improved using DTW-AR while it declined with other methods; and on \texttt{HeartBeat}, DTW-AR based training improves from 70\% to 75\%.
To evaluate the accuracy of DTW-AR in predicting the ground-truth label of adversarial examples, we create adversarial examples using a given attack algorithm and label each example with the true class-label of the corresponding clean time-series input. Figure \ref{fig:advdef} shows the results of DTW-AR based adversarial training using $WB$ architecture. 
In this experiment, we consider the adversarial examples that have successfully fooled the original DNN (i.e., no adversarial training). 
\begin{figure}[!h]
    \centering
    \begin{minipage}{\linewidth}
        \begin{minipage}{.46\linewidth}
                \centering
                \includegraphics[width=\linewidth]{PerfFigures/AugPeroformanceAtrialFibrillationagainstAtk.png}
            \end{minipage}%
            \hfill
        \begin{minipage}{.46\linewidth}
                \centering
                \includegraphics[width=\linewidth]{PerfFigures/AugPeroformanceEpilepsyagainstAtk.png}
            \end{minipage}
        \begin{minipage}{.46\linewidth}
                \centering
                \includegraphics[width=\linewidth]{PerfFigures/AugPeroformanceERingagainstAtk.png}
            \end{minipage}%
            \hfill
        \begin{minipage}{.46\linewidth}
                \centering
                \includegraphics[width=\linewidth]{PerfFigures/AugPeroformanceHeartbeatagainstAtk.png}
            \end{minipage}
            \centering
        \begin{minipage}{.46\linewidth}
                \centering
                \includegraphics[width=\linewidth]{PerfFigures/AugPeroformanceRacketSportsagainstAtk.png}
            \end{minipage}
    \vspace{.5em}
    \end{minipage}
    \begin{minipage}{\linewidth}
            \centering
            {\small DTW-AR Setting: $\alpha_2 \in [0, 1]$}
    \vspace{.5em}
    \end{minipage}
    
    \begin{minipage}{\linewidth}
        \begin{minipage}{.46\linewidth}
                \centering
                \includegraphics[width=\linewidth]{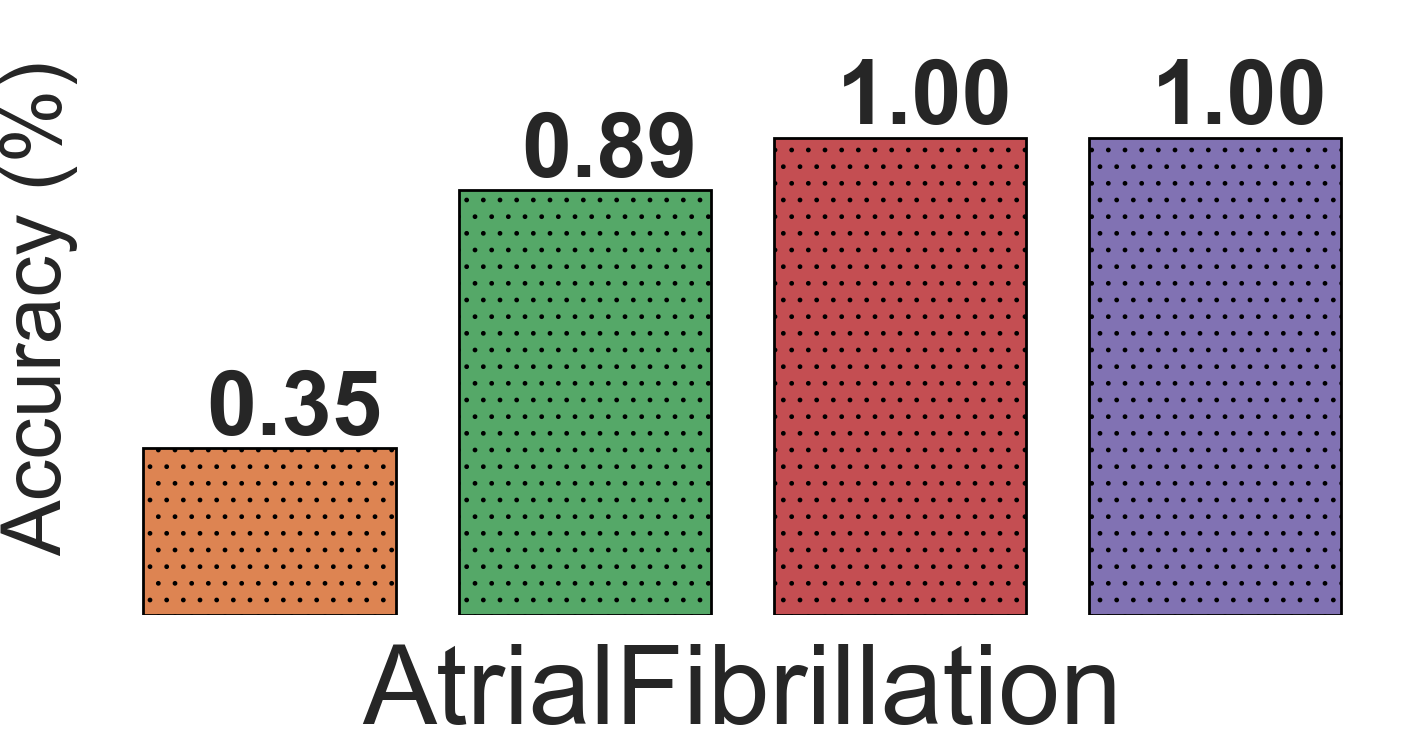}
            \end{minipage}%
            \hfill
        \begin{minipage}{.46\linewidth}
                \centering
                \includegraphics[width=\linewidth]{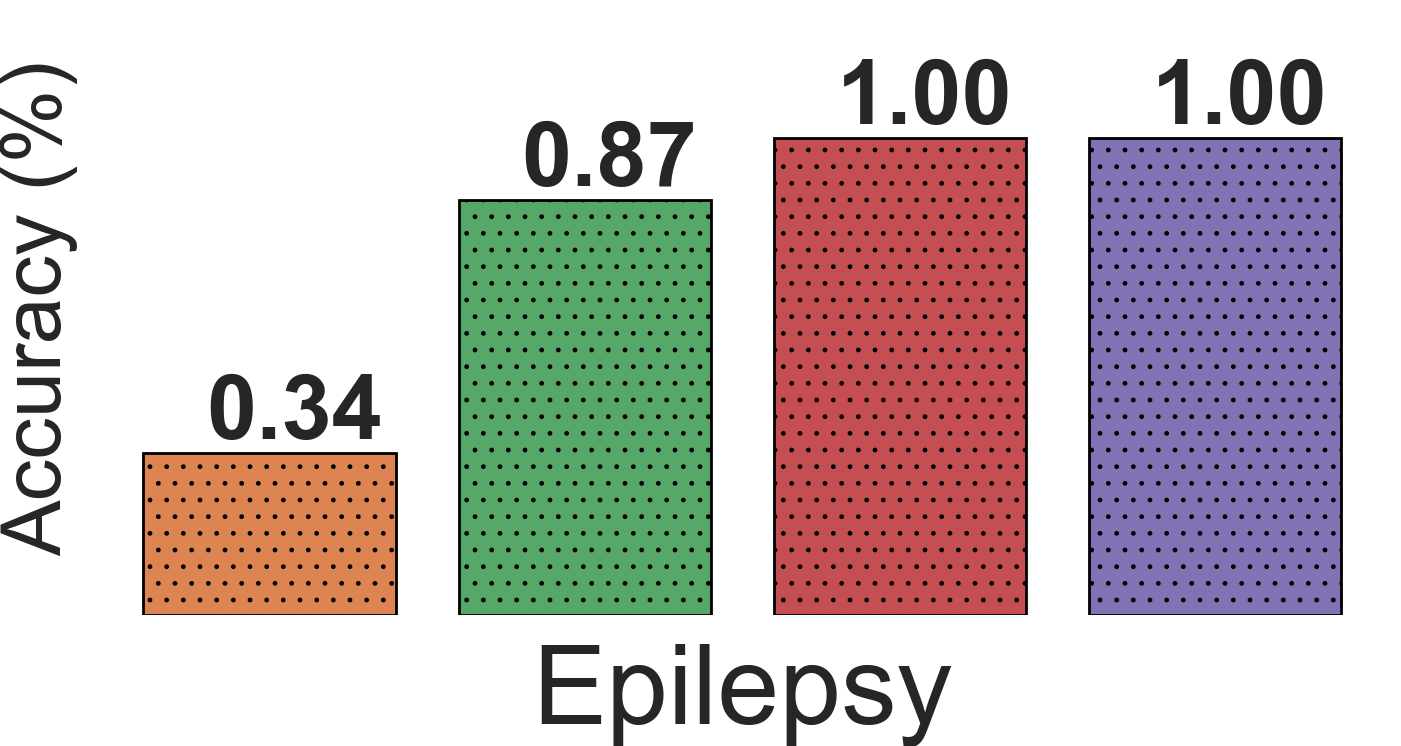}
            \end{minipage}
        \begin{minipage}{.46\linewidth}
                \centering
                \includegraphics[width=\linewidth]{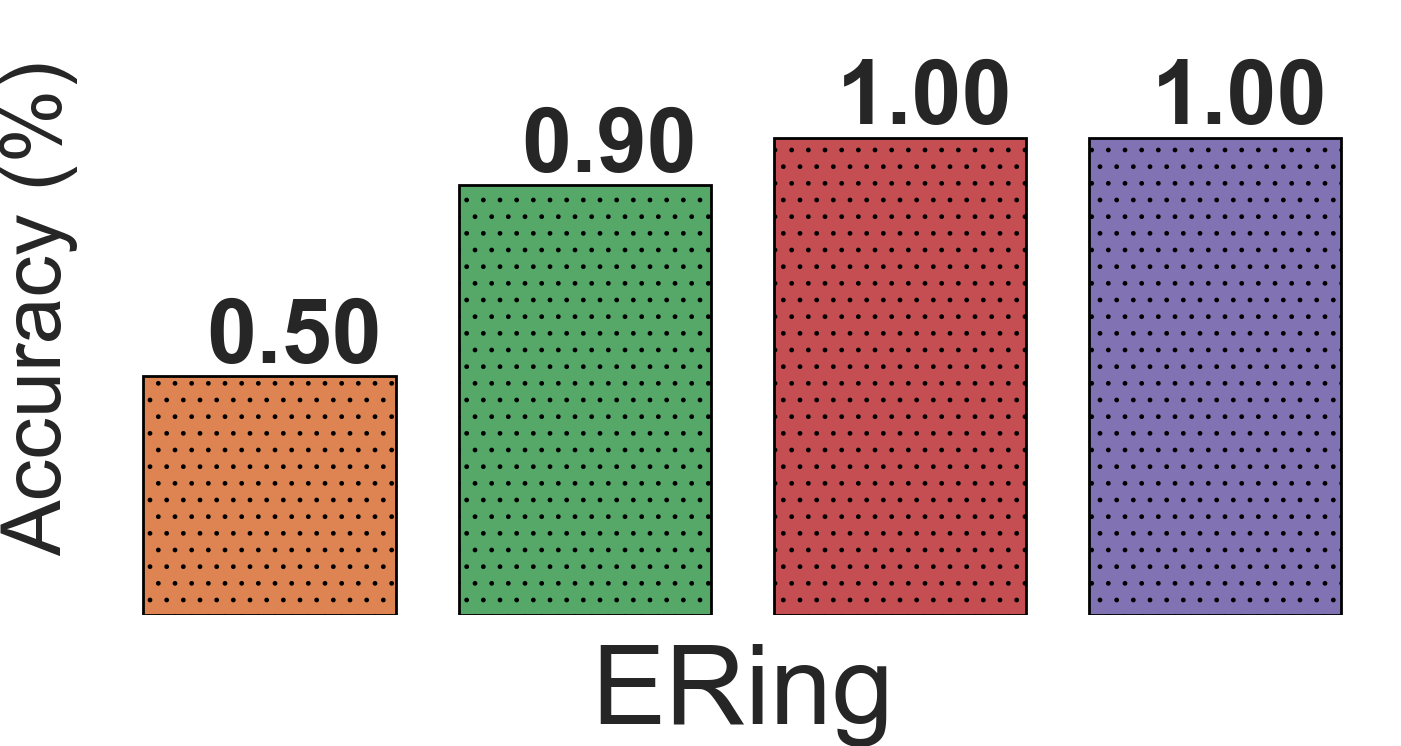}
            \end{minipage}%
            \hfill
        \begin{minipage}{.46\linewidth}
                \centering
                \includegraphics[width=\linewidth]{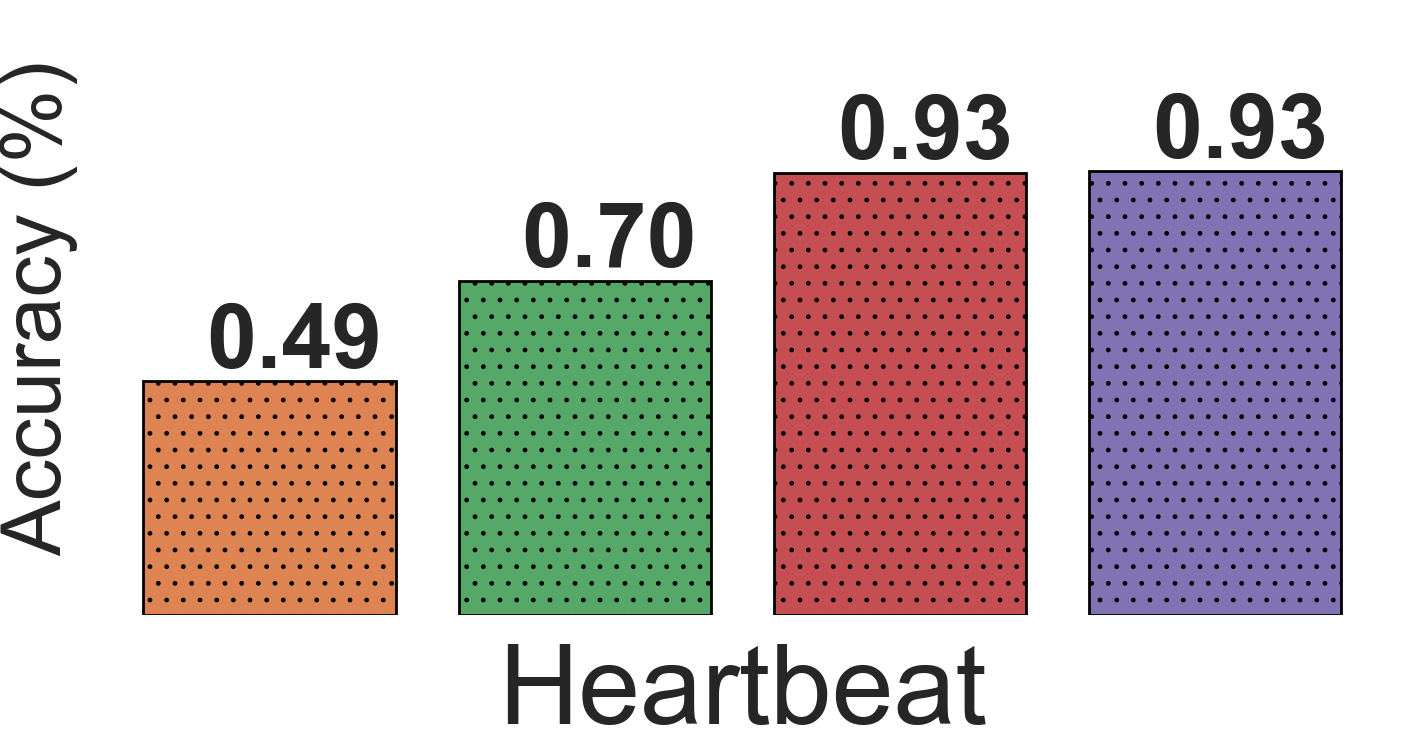}
            \end{minipage}
            \centering
        \begin{minipage}{.46\linewidth}
                \centering
                \includegraphics[width=\linewidth]{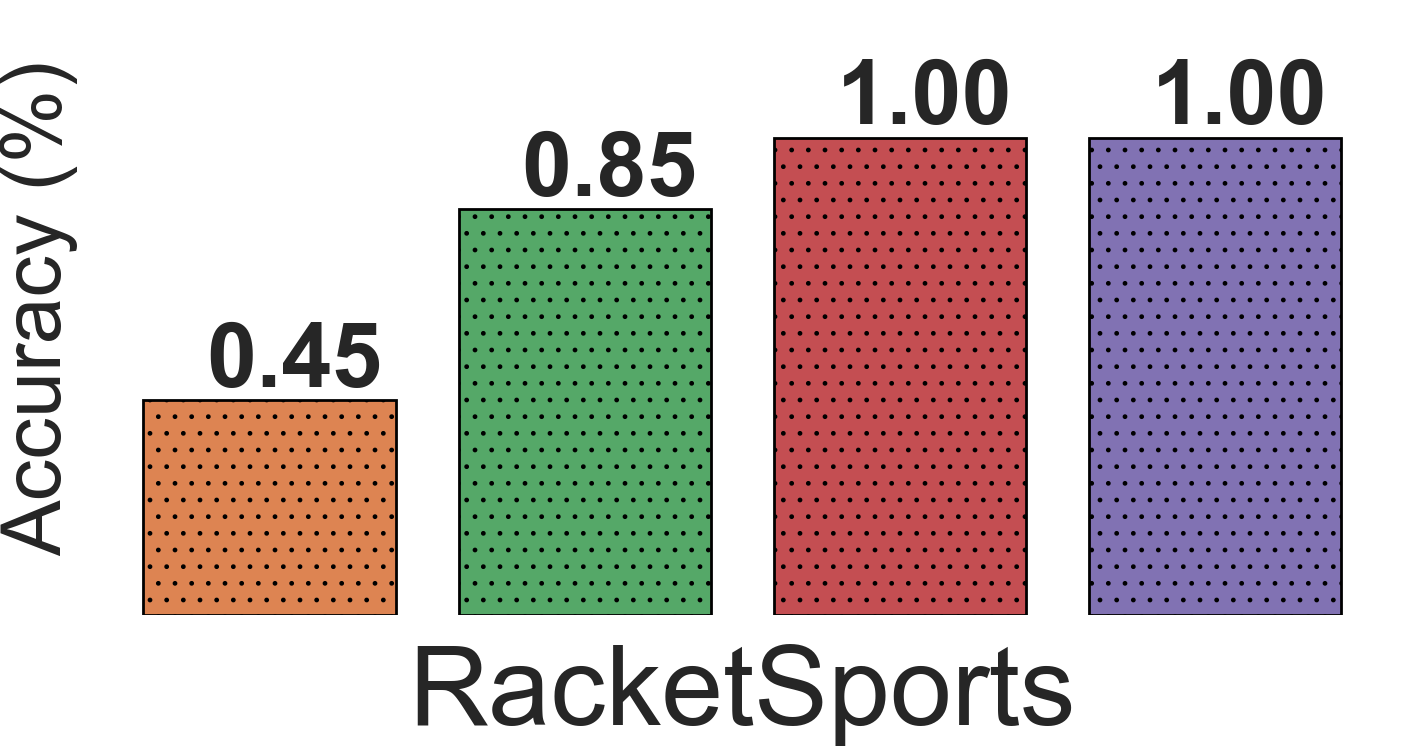}
            \end{minipage}
    \vspace{.5em}
    \end{minipage}
    \begin{minipage}{\linewidth}
            \centering
            {\small DTW-AR Setting: $\alpha_2=0$}
    \vspace{.5em}
    \end{minipage}
    
    \begin{minipage}{\linewidth}
        \begin{minipage}{.46\linewidth}
                \centering
                \includegraphics[width=\linewidth]{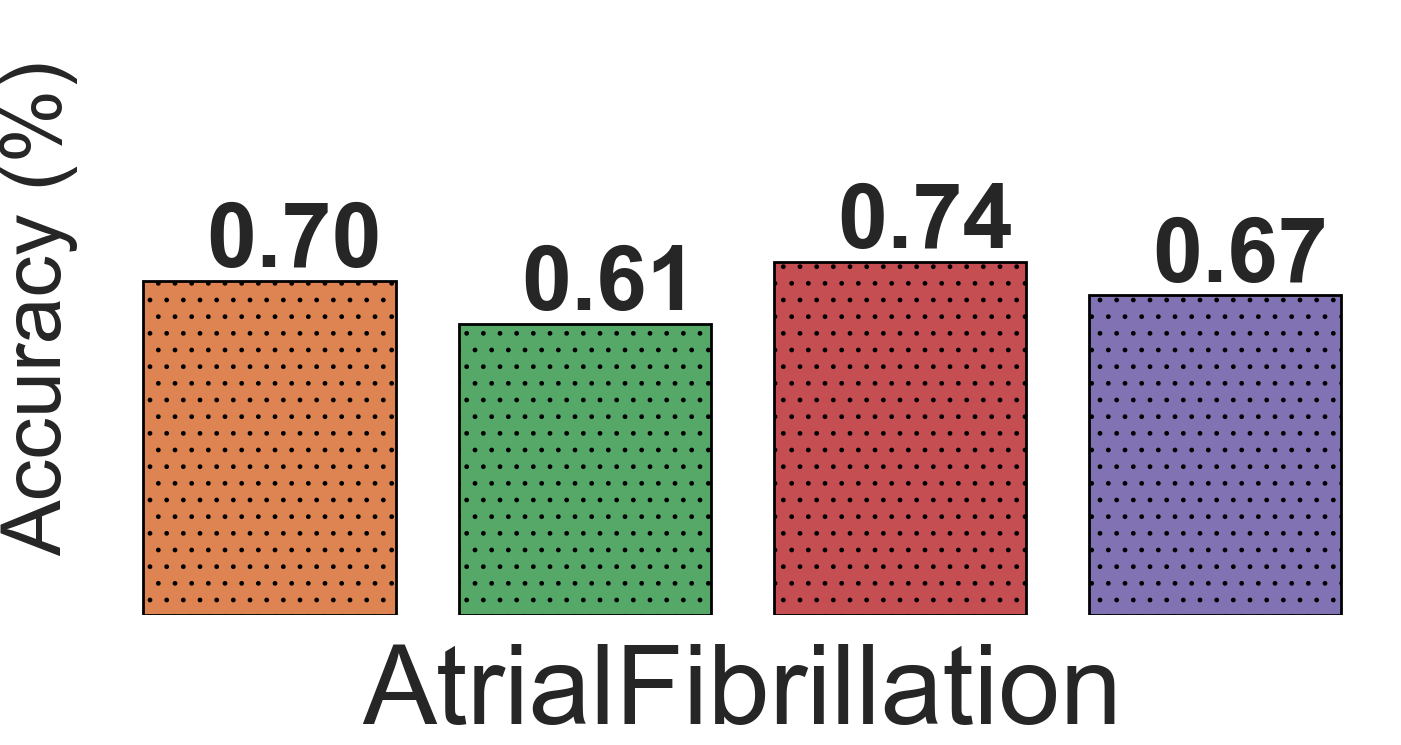}
            \end{minipage}%
            \hfill
        \begin{minipage}{.46\linewidth}
                \centering
                \includegraphics[width=\linewidth]{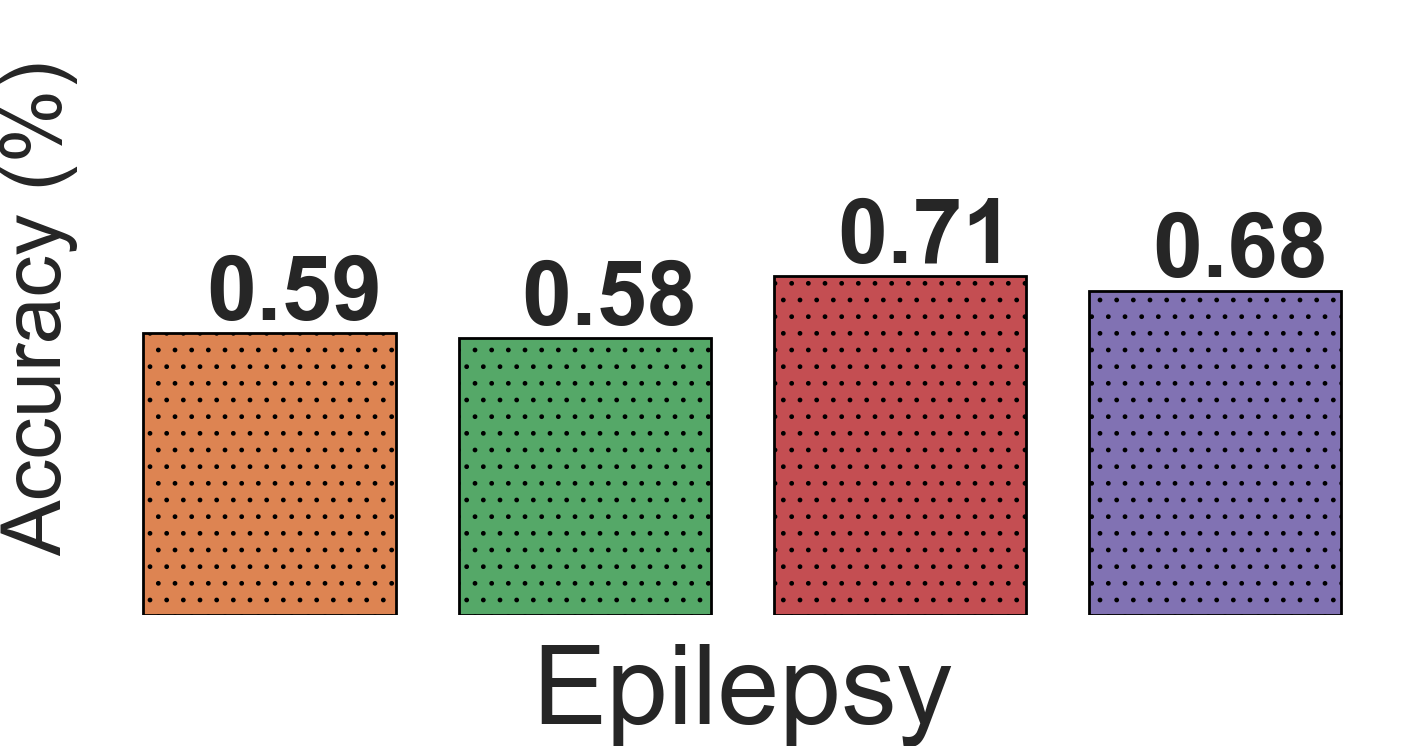}
            \end{minipage}
        \begin{minipage}{.46\linewidth}
                \centering
                \includegraphics[width=\linewidth]{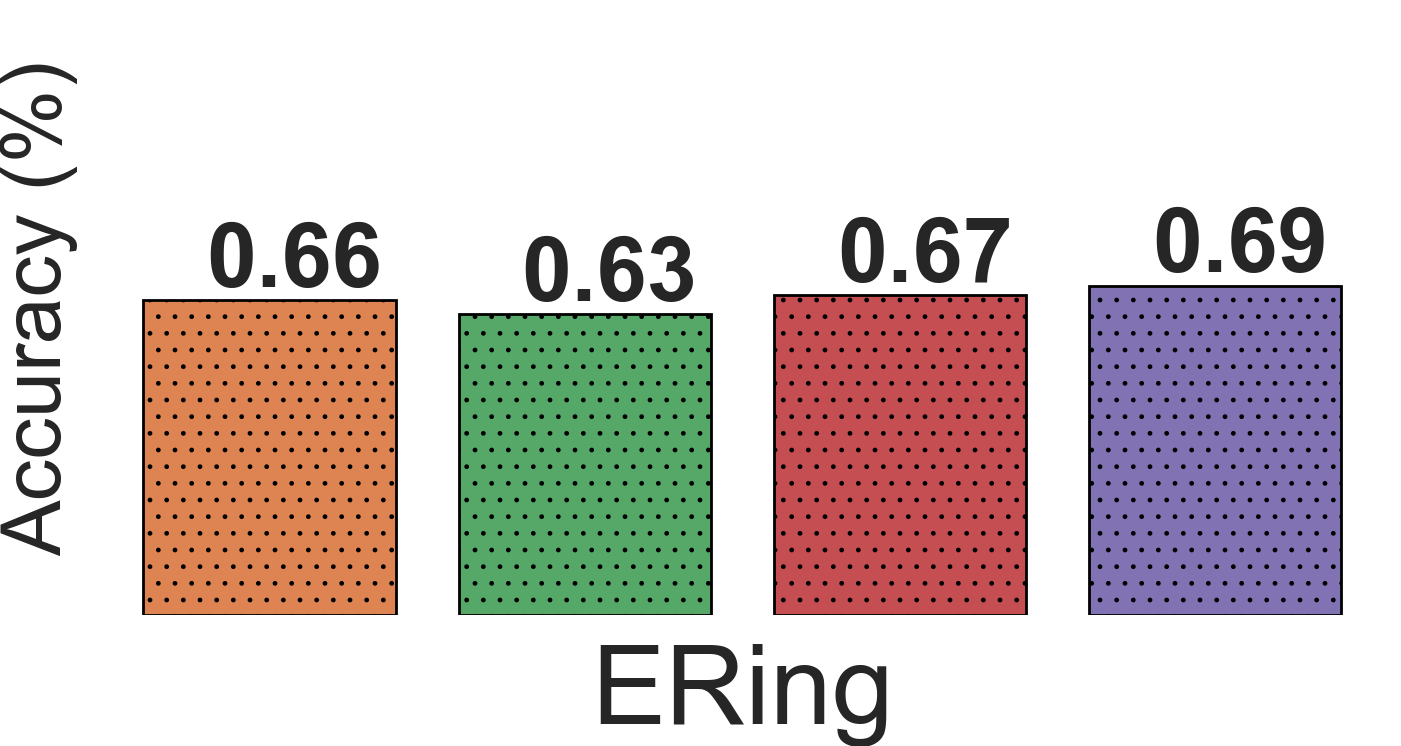}
            \end{minipage}%
            \hfill
        \begin{minipage}{.46\linewidth}
                \centering
                \includegraphics[width=\linewidth]{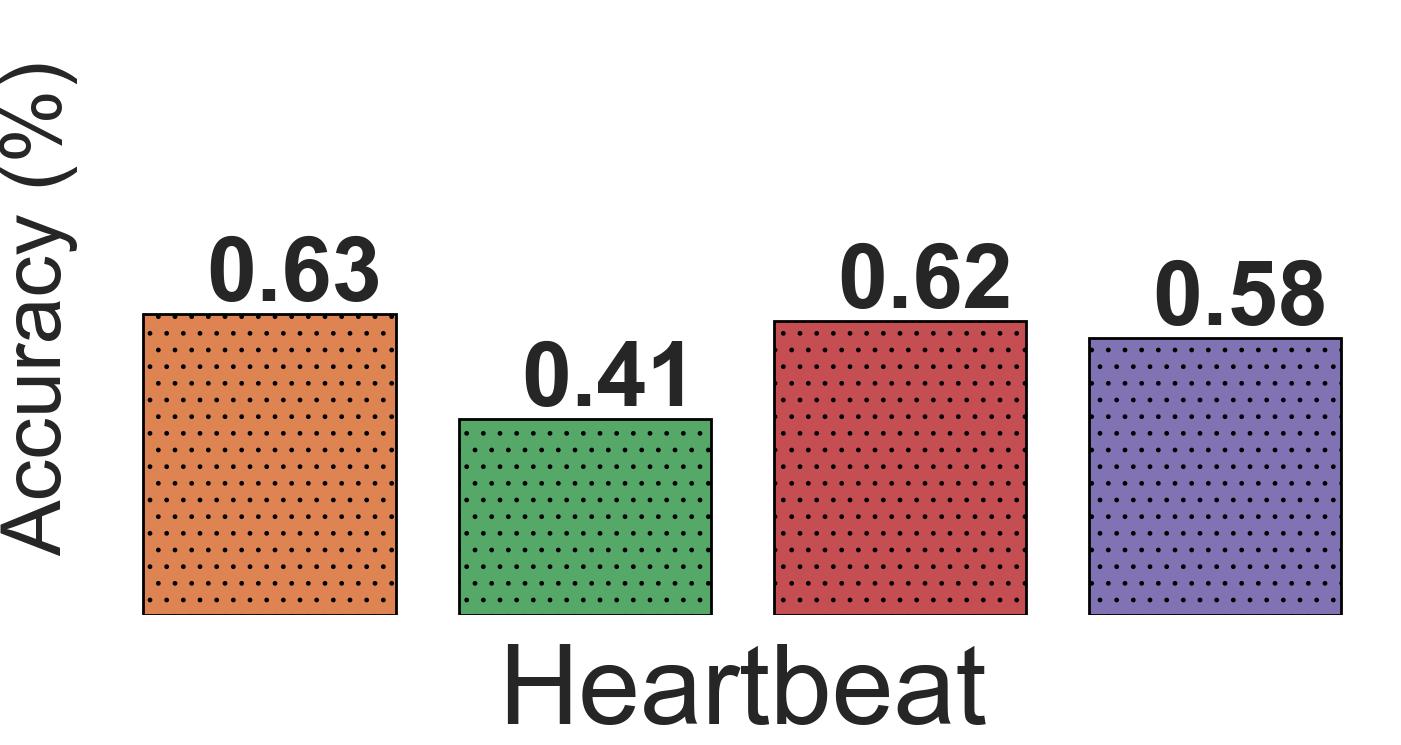}
            \end{minipage}
            \centering
        \begin{minipage}{.46\linewidth}
                \centering
                \includegraphics[width=\linewidth]{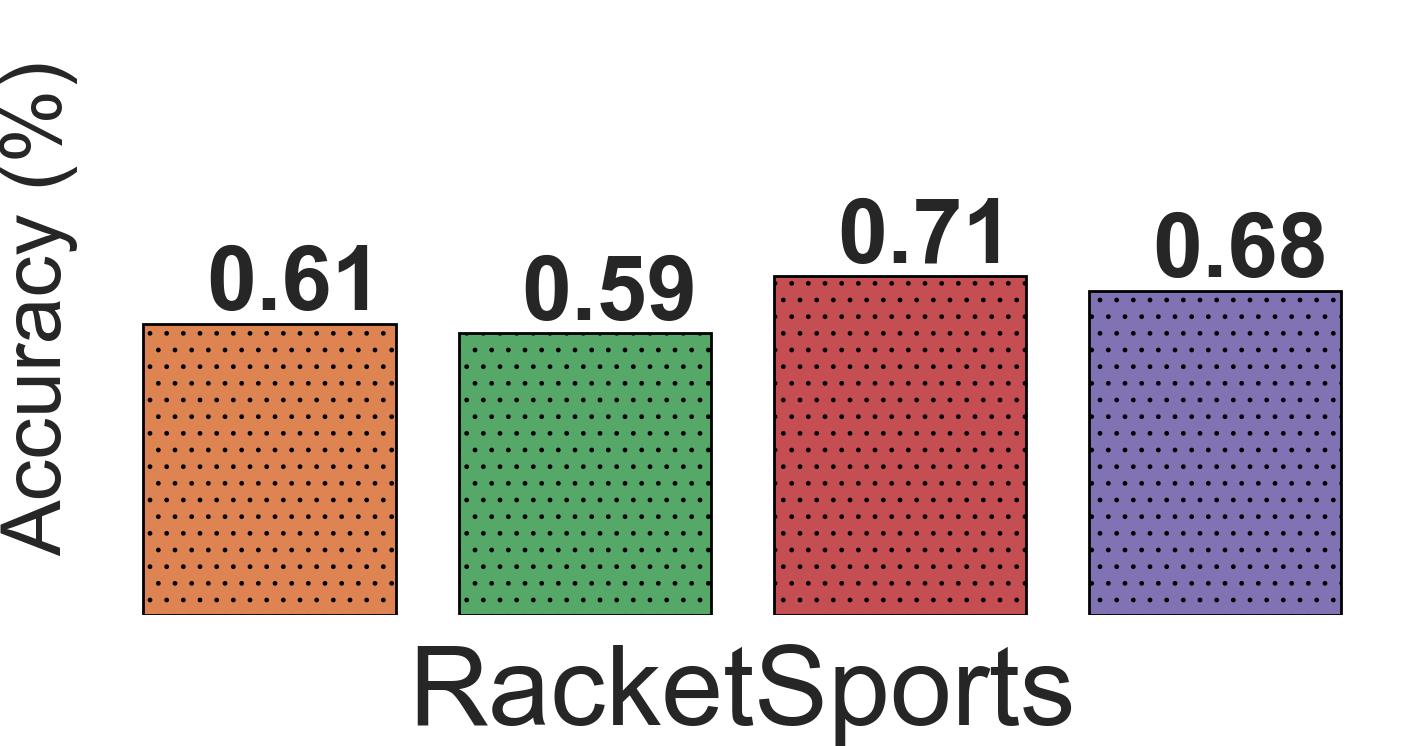}
            \end{minipage}
    \vspace{.5em}
    \end{minipage}
    \begin{minipage}{\linewidth}
            \centering
            {\small DTW-AR Setting: $\alpha_2 \neq 0$}
    \end{minipage}
\caption{Results of DTW-AR based adversarial training to predict the true labels of adversarial examples generated by DTW-AR and the baseline attack methods. The adversarial examples considered are those that successfully fooled DNNs that do not use adversarial training.}
\label{fig:alphaadvdef}
\end{figure}
We observe that DNNs using DTW-AR for adversarial training are able to predict the original label of  adversarial examples with high accuracy. We can see how FGS and PGD attacks cannot evade the DTW-AR based trained deep model for almost any dataset. These results show that DTW-AR significantly improves the robustness of deep models for time-series data to evade attacks generated by DTW-AR and other baseline attacks. For the adversarial training, we employ several values to create adversarial examples to be used in the training phase. We have set $\alpha_1 \in [0.1, 1]$ and $\alpha_2 \in [0, 1]$. In Figure \ref{fig:alphaadvdef}, we show the role of the term $\alpha_2$ of Equation \ref{eq:dtwloss} in the robustness of the DNN. $\alpha_2 \in [0, 1]$ ensures diverse DTW-AR examples to increase the robustness of a given DNN. When set to 0, we see that there is no significant difference in the performance against baseline attacks. However, the DNN cannot defend against all DTW-AR attacks. We can also observe that the setting where $\alpha_2$ is strictly different than 0 is the worst, as the DNN does not learn from the adversarial examples that are found in the Euclidean space by DTW-AR or the given baselines.

\vspace{1.0ex}
\noindent \textbf{Naive approach: Carlini \& Wagner with Soft-DTW.} Recall that naive approach uses DTW measure within the Carlini \& Wagner loss function. SoftDTW \cite{cuturi2017soft}  allows us to create a differentiable version of DTW measure. Hence, we provide results for this naive approach to verify if the the use of Soft-DTW with existing Euclidean distance based methods can solve the challenges for the time-series domain mentioned in this paper. We compare the DTW-AR algorithm with the CW-SDTW that plugs Soft-DTW within the Carlini \& Wagner algorithm instead of the standard $l_2$ distance. CW-SDTW has the following limitations when comapred against DTW-AR:
\begin{itemize}
    \item The time-complexity of Soft-DTW is quadratic in the dimensionality of time-series input space, whereas the distance computation in DTW-AR is linear.
    \item The CW-SDTW attack method is a sub-case of the DTW-AR algorithm. If DTW-AR algorithm uses the optimal alignment path instead of a random path, the result will be equivalent to a CW-SDTW attack.
    \item For a given time-series signal, CW-SDTW will output one single adversarial example and cannot uncover multiple adversarial examples which meet the DTW measure bound. However, DTW-AR algorithm gives the user control over the alignment path and can create multiple diverse adversarial examples. 
\end{itemize}
\begin{figure}[!h]
    \centering
    \includegraphics[width=.8\linewidth]{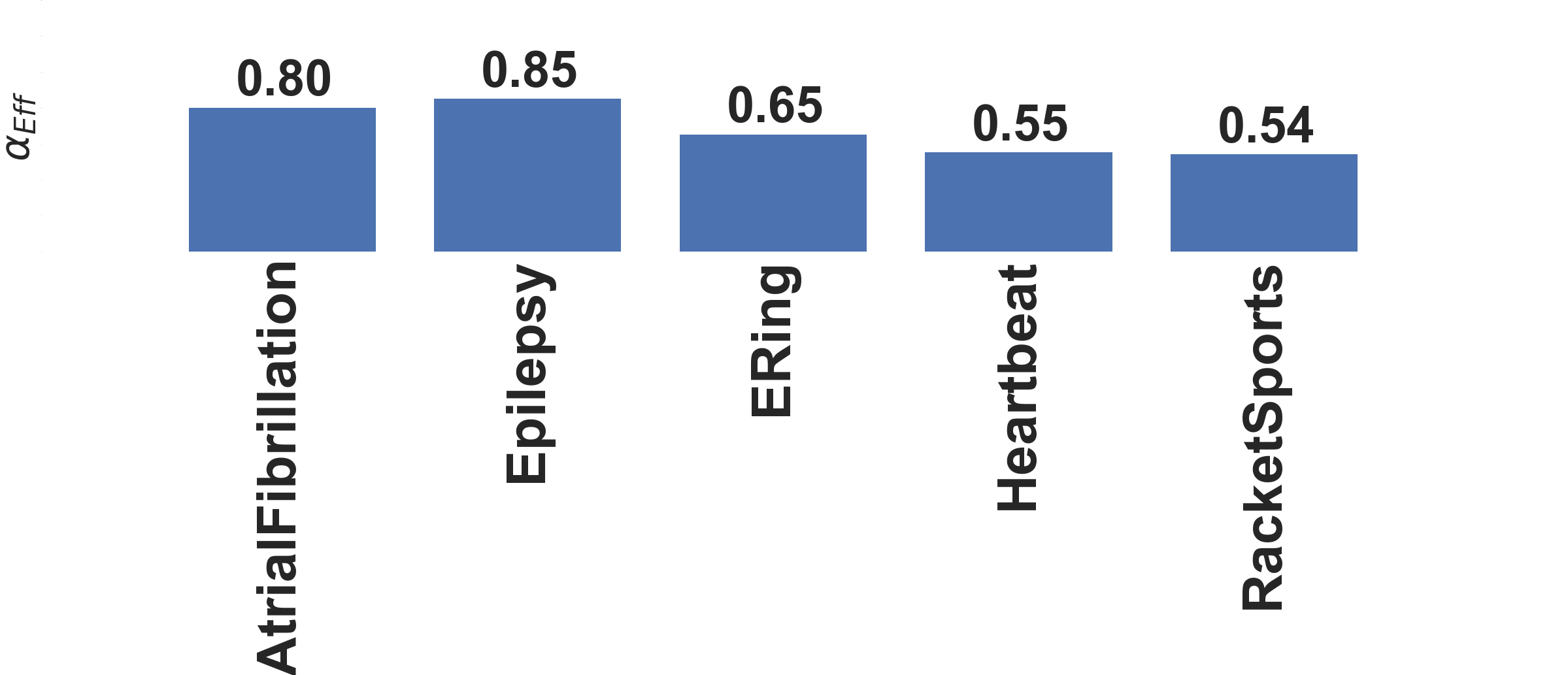}
    \caption{Results for the effectiveness of adversarial examples from DTW-AR against adversarial training using examples created by CW-SDTW on different datasets.}
    \label{fig:EffCWSDTW}
\end{figure}

\begin{figure}[!h]
    \centering
    \includegraphics[width=.8\linewidth]{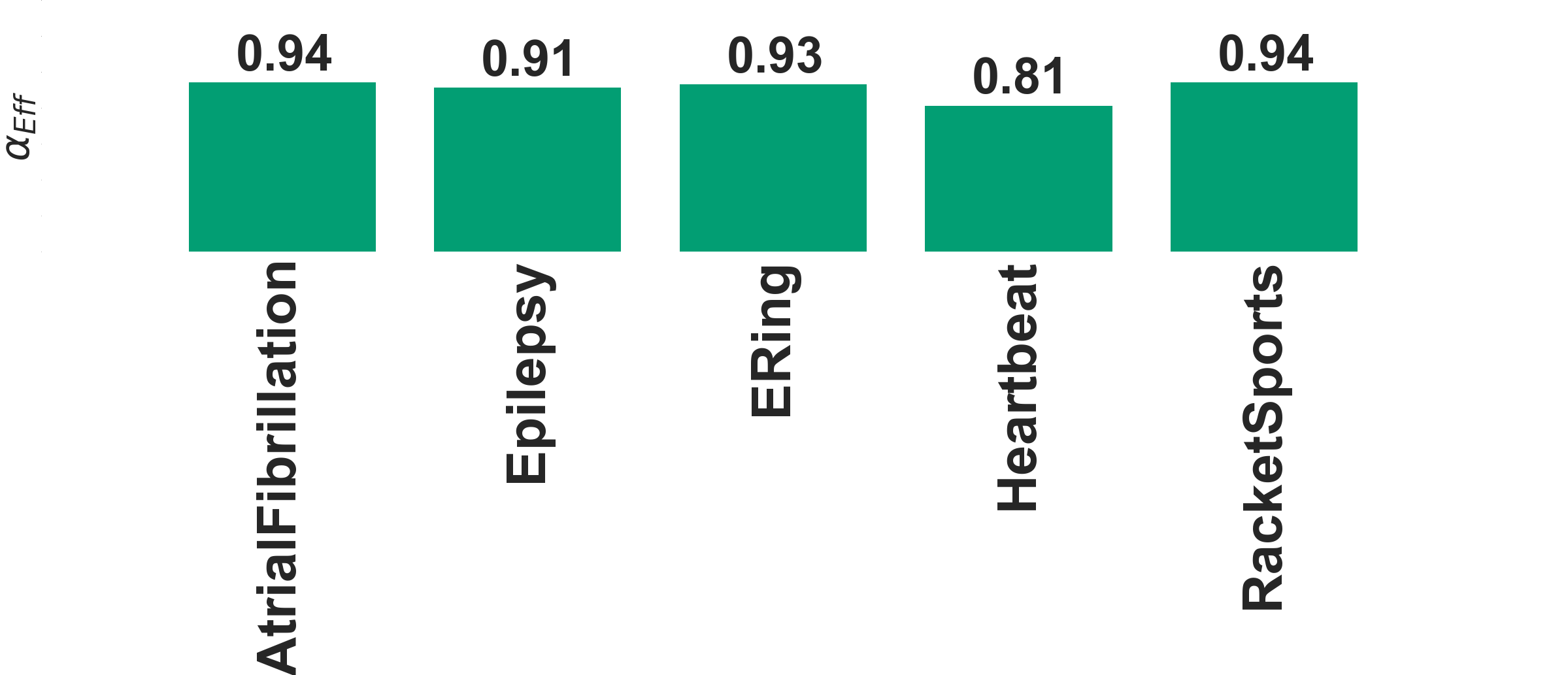}
    \caption{Results for the effectiveness of adversarial training using DTW-AR based examples against adversarial attacks from CW-SDTW on different datasets.}
    \label{fig:advTrnCWSDTW}
\end{figure}

In conclusion, both challenges that were explained in the \textit{Challenges of Naive approach} in Section 3.1 cannot be solved using CW-SDTW. As a consequence, the robustness goal aimed by this paper cannot be achieved using solely CW-SDTW. Indeed, our experiments support this hypothesis. Figure \ref{fig:EffCWSDTW} shows that DTW-AR is successful to fool a DNN that uses adversarial examples from CW-SDTW for adversarial training. This shows that our proposed framework is better than this naive baseline. Figure \ref{fig:advTrnCWSDTW} shows that DTW-AR significantly improves the robustness of deep models for time-series as it is able to evade attacks generated by CW-SDTW. Both these  experiments demonstrate that CW-SDTW is neither able to create stronger attacks nor a more robust deep model when compared to DTW-AR.

\vspace{1.0ex}
\noindent \textbf{Comparison with Karim et al., \cite{karim2020adversarial}.}
The approach from Karim et al., \cite{karim2020adversarial} employs network distillation to train a student model for creating adversarial attacks. However, this method is severely limited: only a small number of target classes yield adversarial examples and the method does not guarantee the generation of an adversarial example for every input. Karim et al., have shown that for many datasets, this method creates a limited number of adversarial examples in the white-box setting. To test the effectiveness of this attack against DTW-AR, Figure \ref{fig:karim} shows the success rate of deep model from DTW-AR based adversarial training to predict the true labels of the attacks generated by the method from Karim et al., on different datasets. 

\begin{figure}[!h]
    \centering
    \includegraphics[width=.8\linewidth]{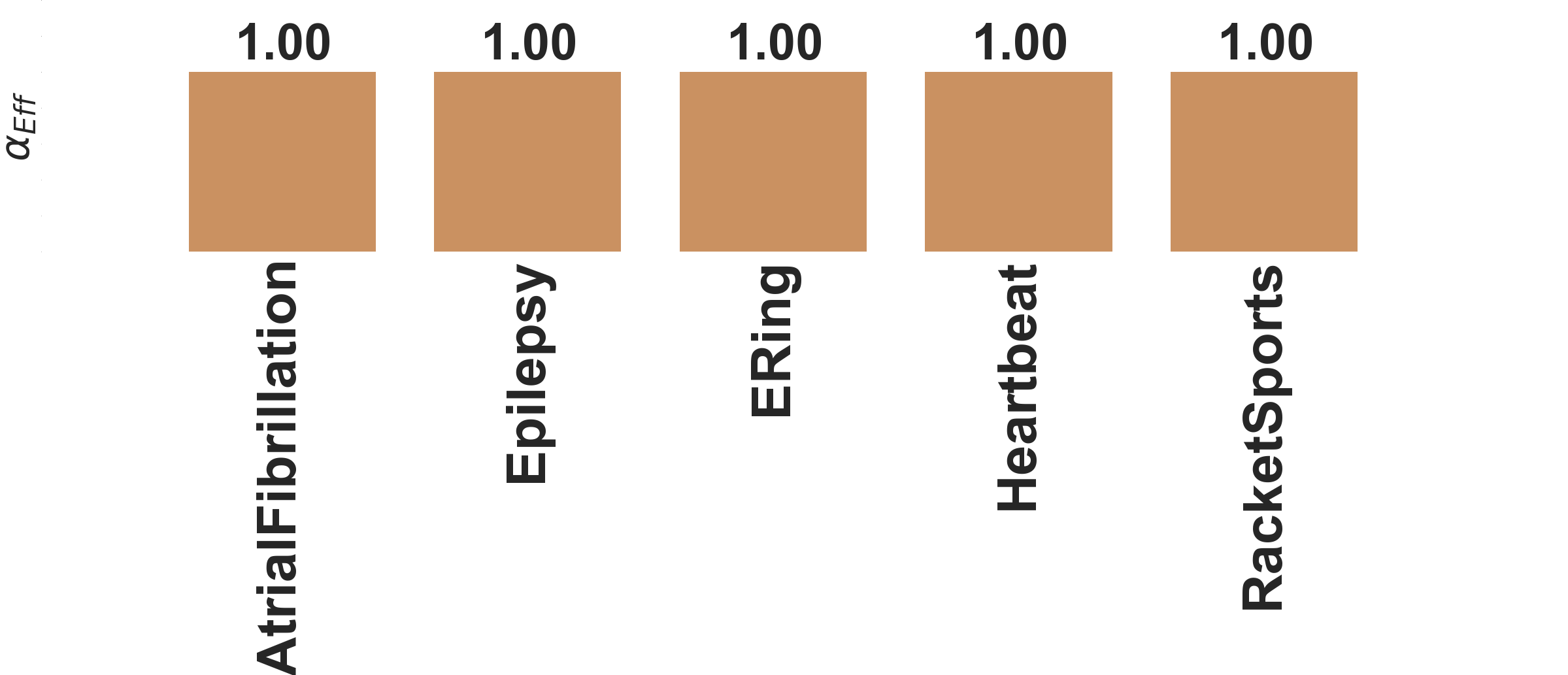}
    \caption{Results of the success rate of deep model from DTW-AR based adversarial training to predict the true label of adversarial attacks generated using method in \cite{karim2020adversarial}.}
    \label{fig:karim}
\end{figure}

DTW-AR outperforms \cite{karim2020adversarial} due to following reasons:
\begin{itemize}
    \item DTW-AR generates at least one adversarial example for every input $X \in \mathbb{R}^{n \times T}$ as shown in our experiments.
    \item Adversarial examples created by DTW-AR are highly effective against deep models relying on \cite{karim2020adversarial} for adversarial training as this baseline fails to create adversarial examples for many inputs and target classes (shown in \cite{karim2020adversarial}).
    \item Adversarial examples created by the method from \cite{karim2020adversarial} does not evade deep models from DTW-AR based adversarial training.
\end{itemize}

\begin{figure}[t]
    \centering
    \begin{minipage}{\linewidth}
        \begin{minipage}{.46\linewidth}
                \centering
                \includegraphics[width=\linewidth]{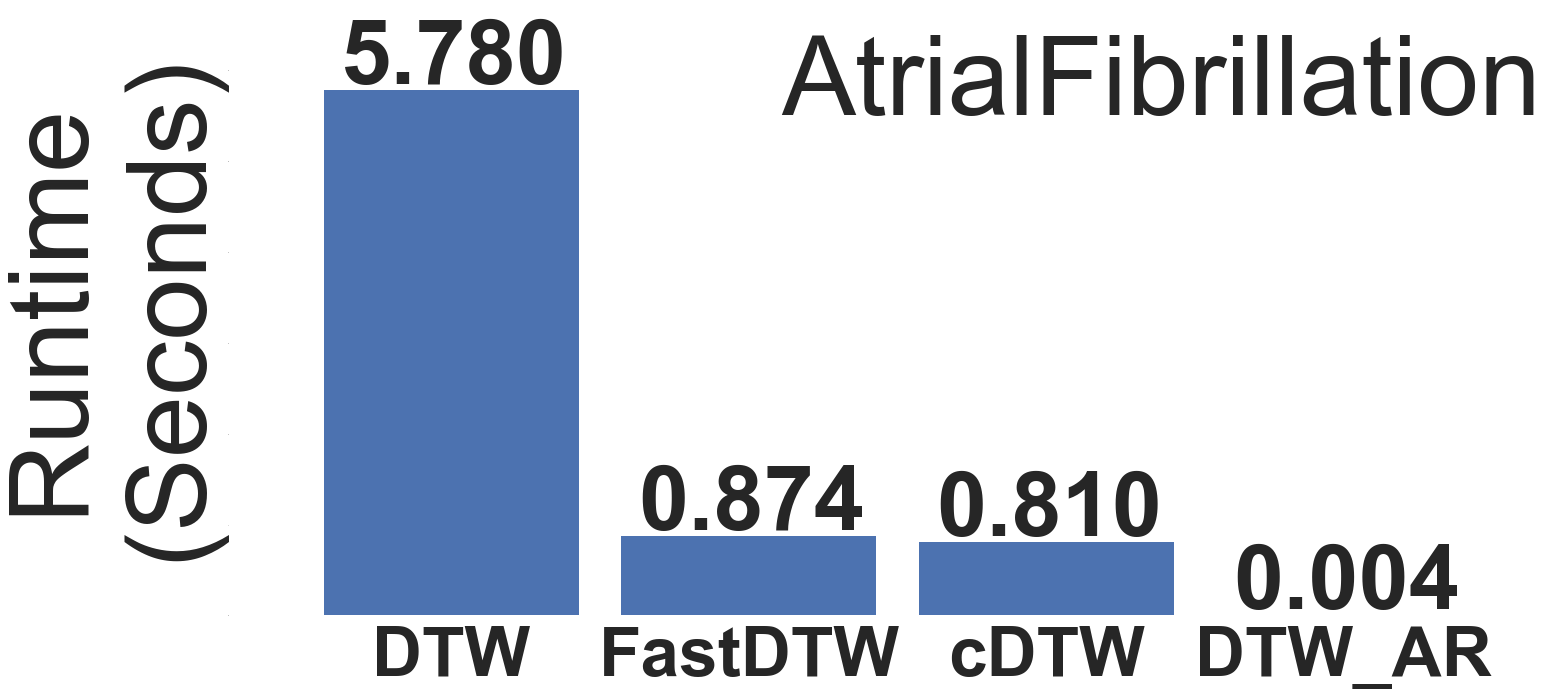}
            \end{minipage}%
            \hfill
        \begin{minipage}{.46\linewidth}
                \centering
                \includegraphics[width=\linewidth]{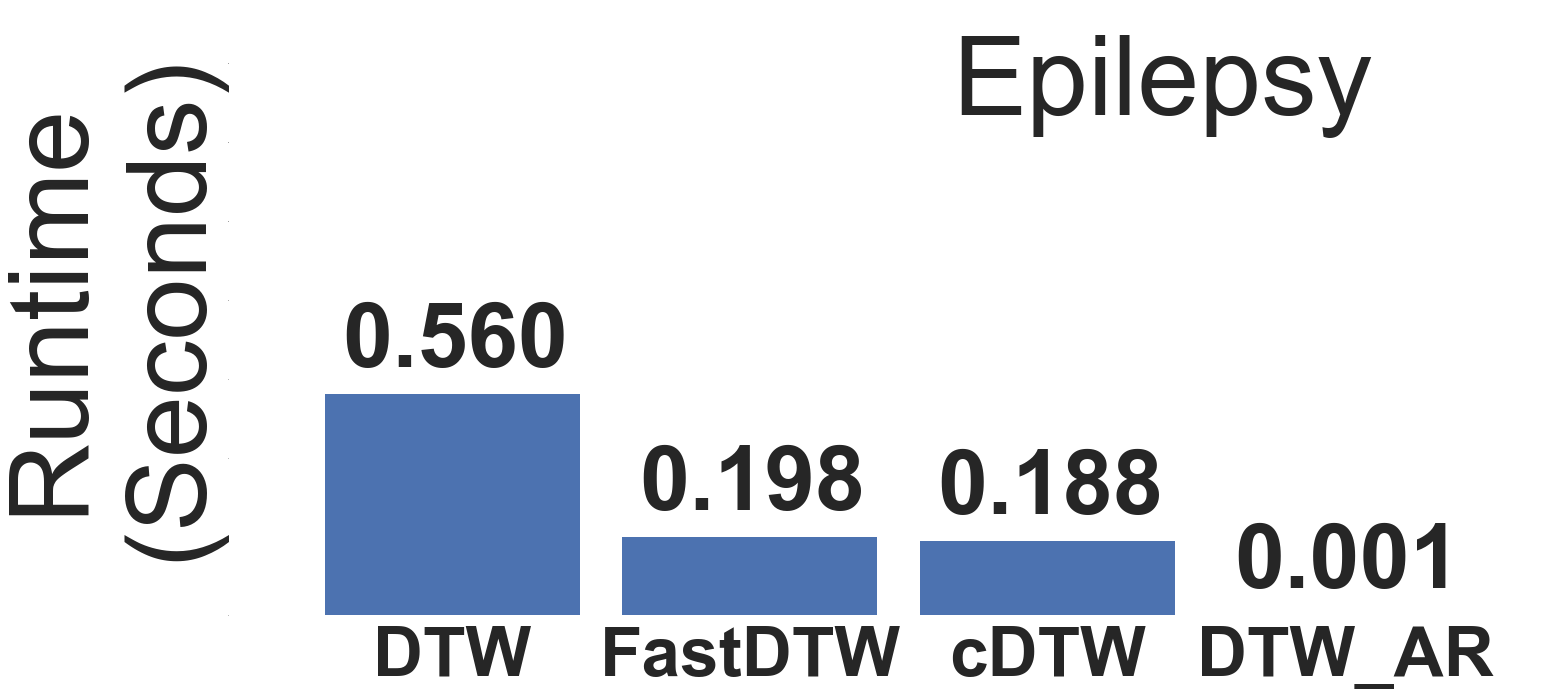}
            \end{minipage}
        \begin{minipage}{.46\linewidth}
                \centering
                \includegraphics[width=\linewidth]{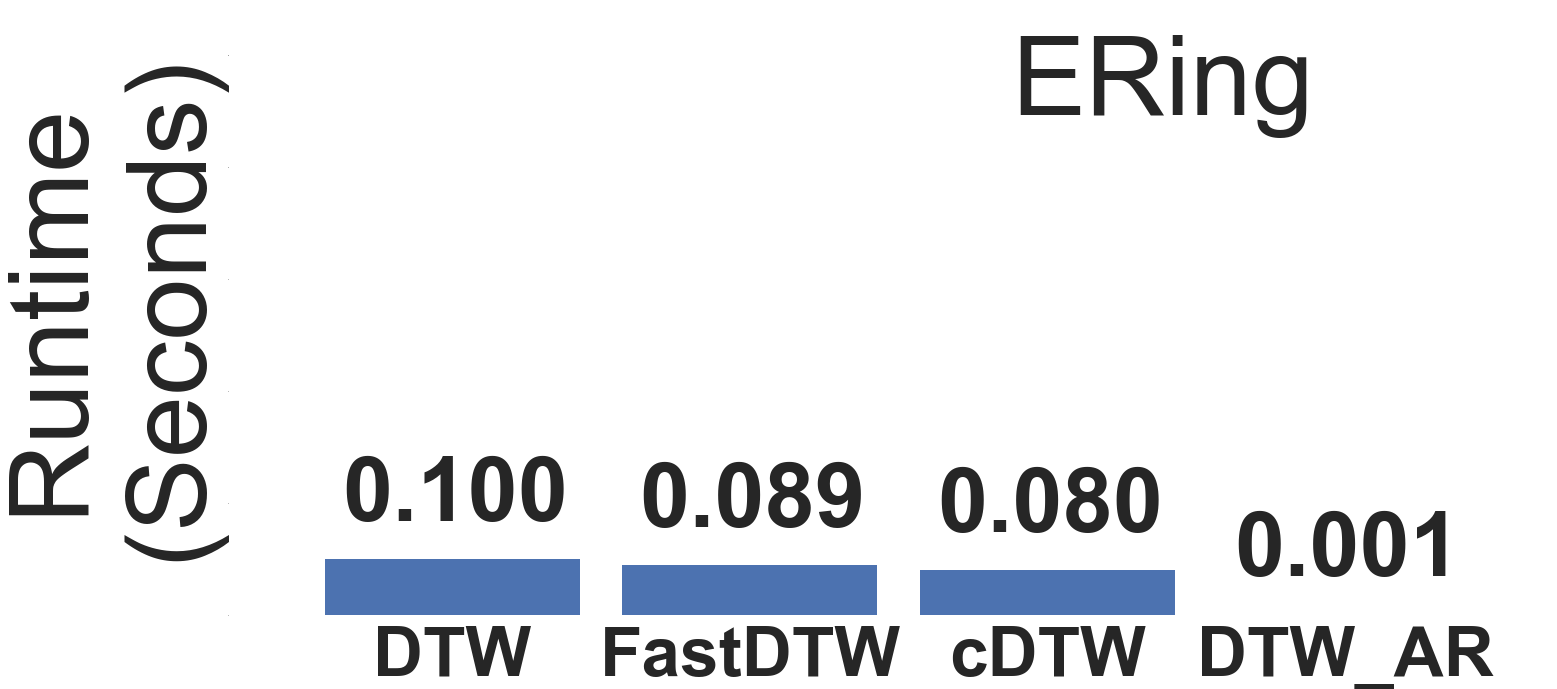}
            \end{minipage}%
            \hfill
        \begin{minipage}{.46\linewidth}
                \centering
                \includegraphics[width=\linewidth]{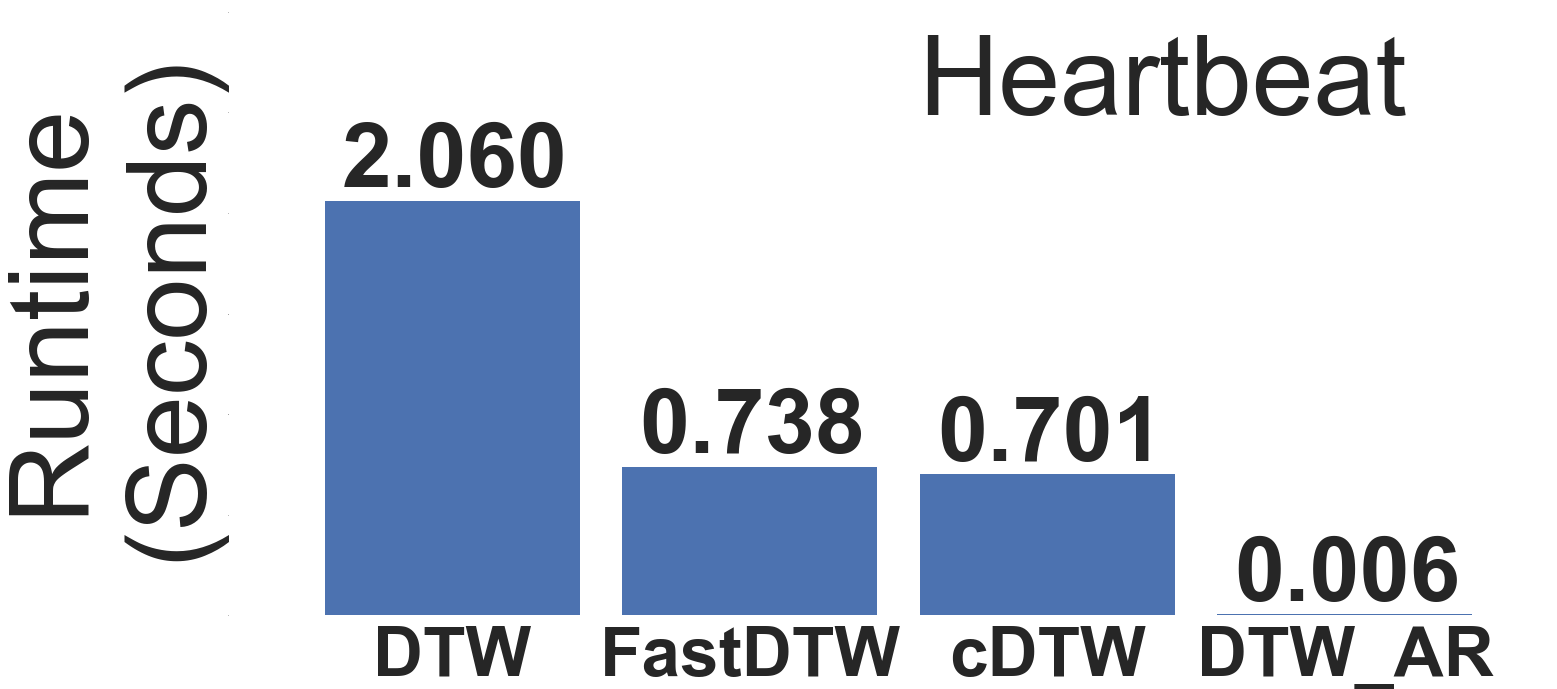}
            \end{minipage}
            \centering
        \begin{minipage}{.46\linewidth}
                \centering
                \includegraphics[width=\linewidth]{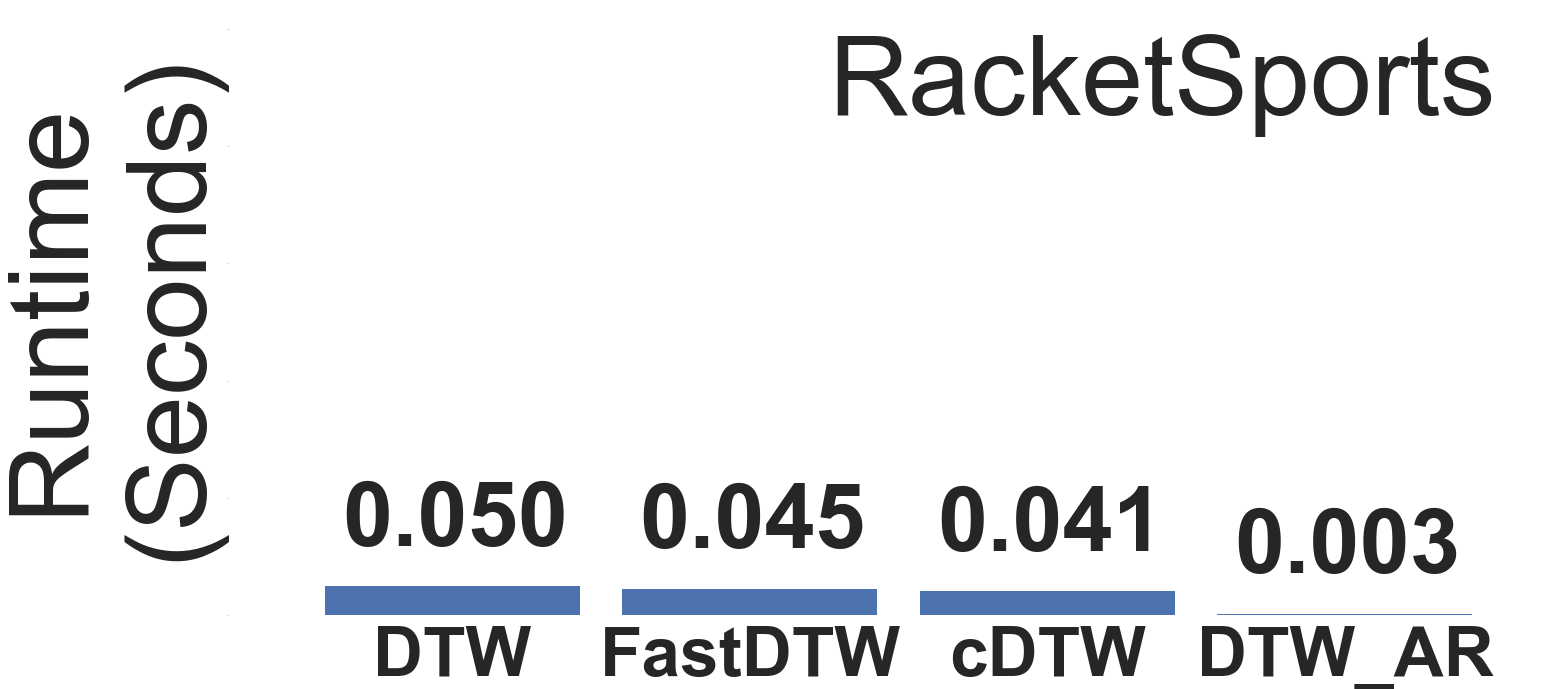}
            \end{minipage}
    \end{minipage}
    \caption{Average runtime \textbf{per iteration} for standard DTW, FastDTW, cDTW, and DTW-AR (on NVIDIA Titan Xp GPU).}
    \label{fig:runtime}
\end{figure}

\vspace{1.0ex}
\noindent \textbf{Computational runtime of DTW-AR vs. DTW.} As explained in the technical section, optimization based attack algorithm requires a large number of iterations to create a highly-similar adversarial example. For example, $10^3$ iterations is the required default choice for CW to create successful attacks, especially, for large time-series in our experiments. The exact DTW method is non-differentiable, thus, it is not possible to perform experiments to compare DTW-AR method to the exact DTW method. Hence, we assume that each iteration will compute the optimal DTW path and use it instead of the random path. To assess the runtime of computing the DTW measure, we employ three different approaches: 1) The standard DTW algorithm, 2) The FastDTW \cite{salvador2007fastdtw} that was introduced to overcome DTW computational challenges, and 3) cDTW \cite{dau2018optimizing} that measures DTW in a constrained manner using warping windows. We note that FastDTW was proven to be inaccurate, and cDTW is faster and more accurate for computing DTW measure \cite{wu2020cdtw}. We show both baselines for the sake of completeness. We provide the runtime of performing each iteration using the different algorithms in Figure \ref{fig:runtime}. We can clearly observe that DTW-AR is orders of magnitude faster than the standard DTW and the accelerated DTW algorithms. The overall computational cost will be significantly reduced using DTW-AR compared to exact DTW or soft-DTW \cite{cuturi2017soft} for large-size time-series signals.

\vspace{1.0ex}

\begin{figure}[!h]
    \centering
    \begin{minipage}{\linewidth}
        \begin{minipage}{.46\linewidth}
                \centering                \includegraphics[width=\linewidth]{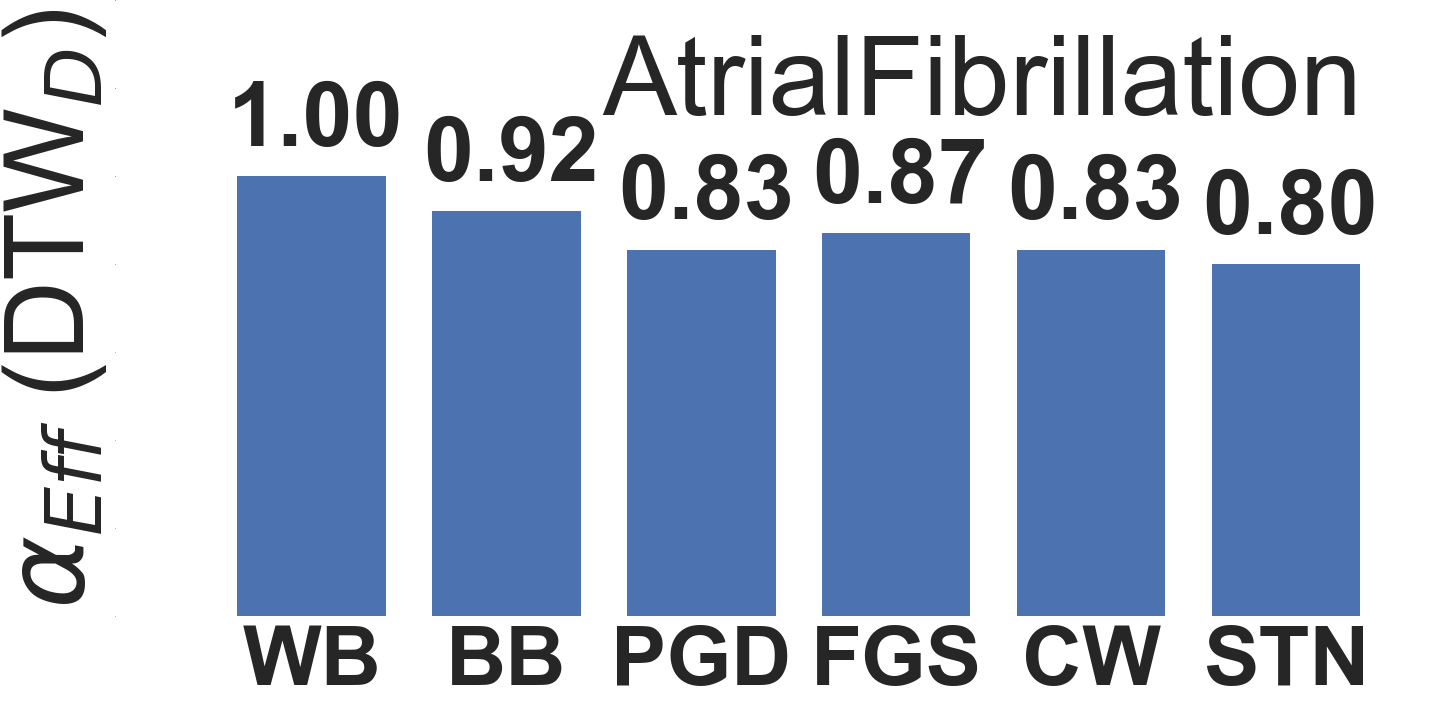}
            \end{minipage}%
            \hfill
        \begin{minipage}{.46\linewidth}
                \centering                \includegraphics[width=\linewidth]{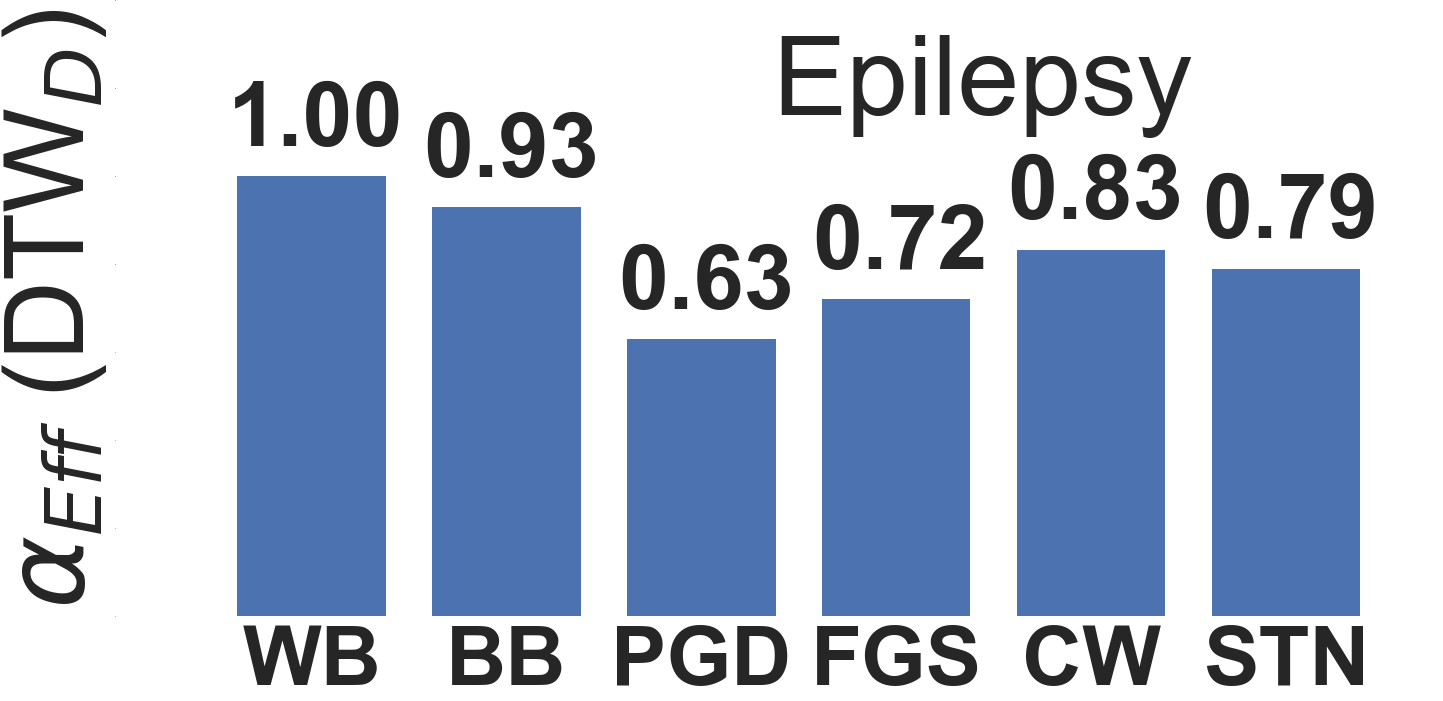}
            \end{minipage}
        \begin{minipage}{.46\linewidth}
                \centering                \includegraphics[width=\linewidth]{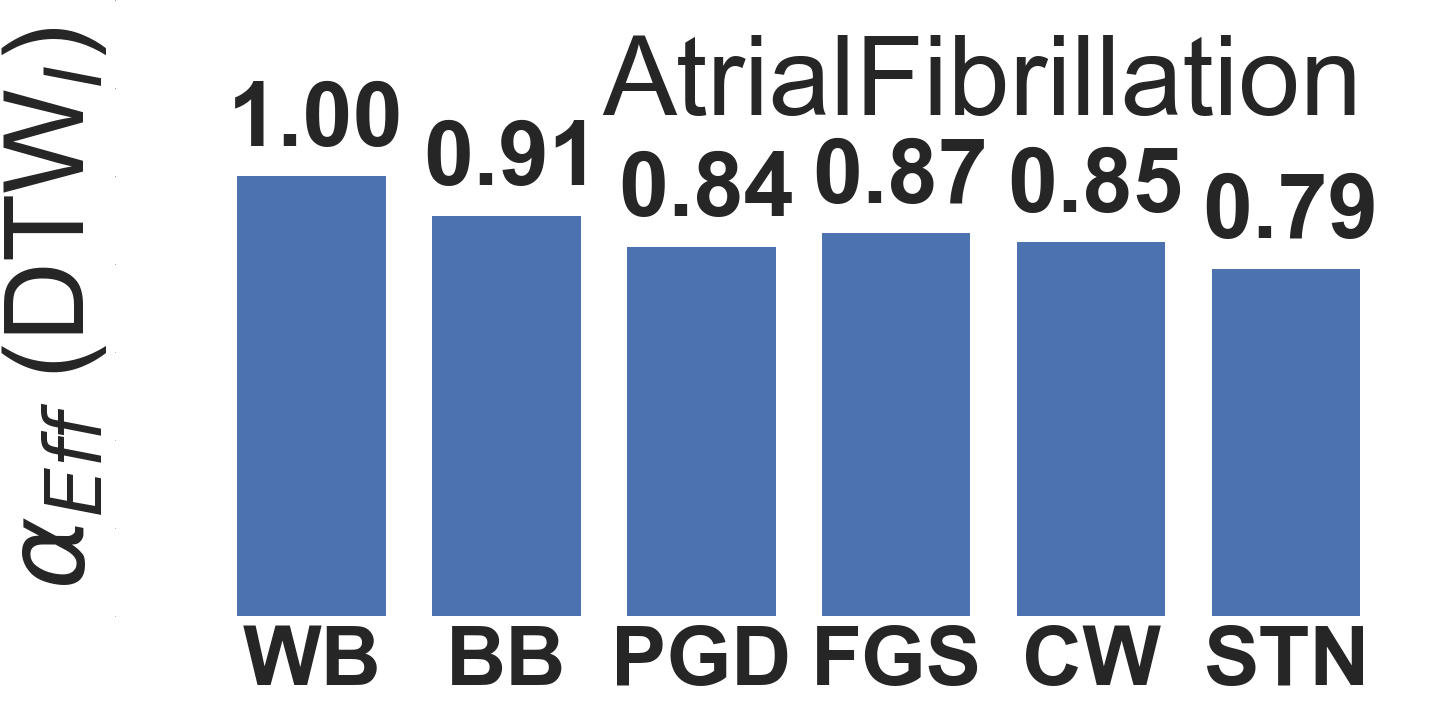}
            \end{minipage}%
            \hfill
        \begin{minipage}{.46\linewidth}
                \centering                \includegraphics[width=\linewidth]{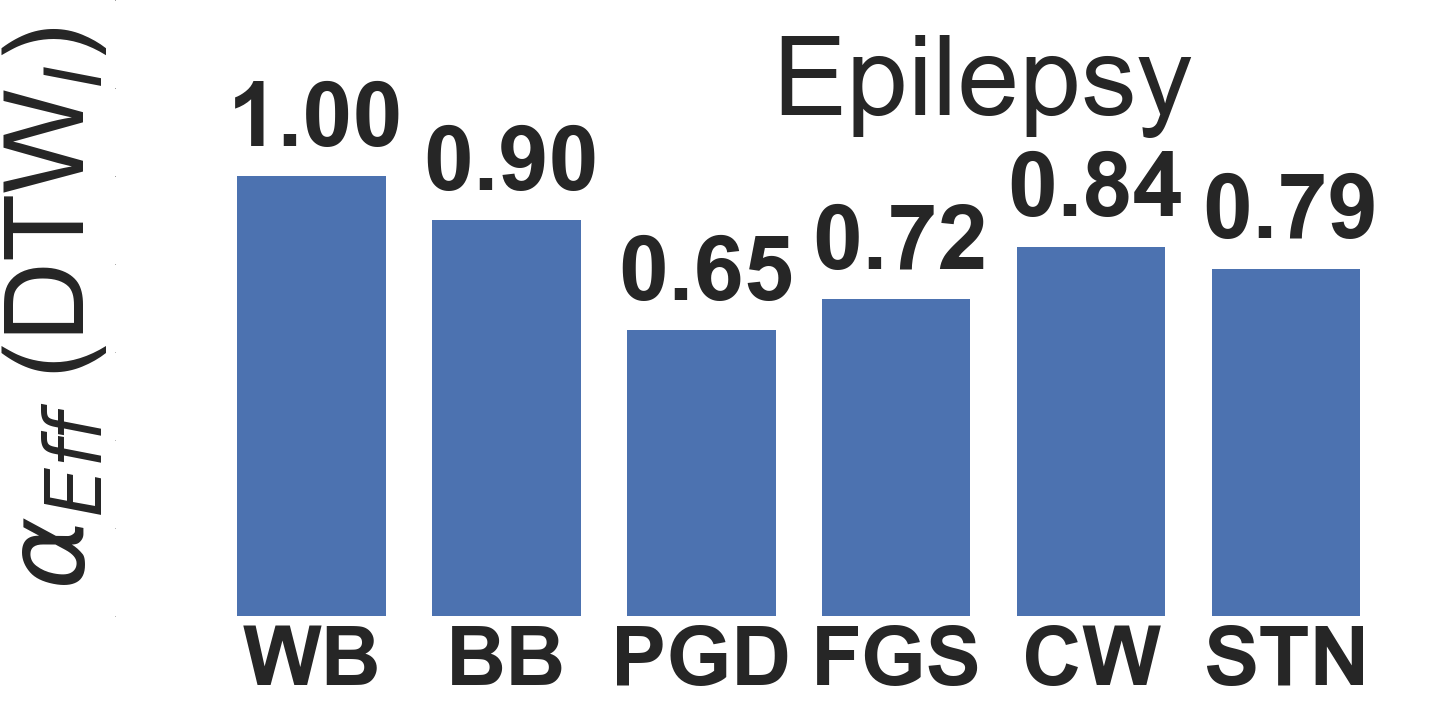}
            \end{minipage}
    \end{minipage}
\caption{Results for the effectiveness of adversarial examples from DTW-AR using DTWAdaptive\cite{shokoohi2017generalizing} (DTW$_D$ top row, DTW$_I$ bottom row) on different DNNs under different settings.}
\label{fig:advatkDTWA}
\end{figure}

\noindent\textbf{DTW-AR extension to other multivariate DTW measures.}The DTW-AR framework relies on the distance function $dist_P(X,Z) = \sum_{(i,j)\in P}d(X_i,Z_j)$ between two time-series signals $X$ and $Z$ according to an alignment path $P$ to measure their similarity. Extending the DTW notion from univariate to multivariate is a known problem, where depending on the application, researchers' suggest to change the definition of $dist_P(X,Z)$ to better fit the characteristics of the application at hand. In all cases, DTW-AR relies on using the final cost matrix $ DTW(X,Z) = \displaystyle \min_P dist_P(X,Z)$ using dynamic programming $C_{i,j} = d(X_i, Z_j) + \min \big\{C_{i-1,j}, C_{i,j-1}, C_{i-1,j-1} \big\}$. Therefore, the use of different variants of $dist_P(X,Z)$ (e.g., $DTW_I$ or $DTW_D$ \cite{shokoohi2017generalizing}) will only affect the cost matrix values, but will not change the assumptions and applicability of DTW-AR. Therefore, DTW-AR is general and can work with any variant of DTW. In Figure \ref{fig:advatkDTWA}, we demonstrate that using a different family of DTW ($DTW_I$) does not have a major impact on DTW-AR's performance and effectiveness. The performance of DTW-AR framework using both alternative measures of multi-variate DTW does not affect the overall performance. Therefore, for a given specific application, the practitioner can configure DTW-AR appropriately.
\vspace{1.0ex}

\subsection{Summary of Experimental Results}
Our experimental results supported all the claims made in Section 3. The summary list includes: 

\begin{itemize}
\setlength\itemsep{0em}
    \item Figure \ref{fig:dtwl2space} showed that DTW space is more suitable for adversarial studies in the time-series domain than Euclidean distance to support Theorem \ref{th:advspace}.
    \item Using stochastic alignment paths, DTW-AR creates multiple diverse adversarial examples to support Corollary 1 (Table \ref{tab:advpiechart}), which is impossible using the optimal alignment path.
    \item Figure \ref{fig:pathconvplot} provides empirical justification for Theorem \ref{th:dtwgap} showing that minimizing over a given alignment path is equivalent to minimizing using exact DTW method (bound is tight).
    \item Figure \ref{fig:advatk} shows that adversarial examples created by DTW-AR have higher potential to break time-series DNN classifiers.
    \item Figures \ref{fig:cleanadvdef} and \ref{fig:advdef} show that DTW-AR based adversarial training is able to improve the robustness of DNNs against baseline adversarial attacks.
    \item Figure \ref{fig:EffCWSDTW} and \ref{fig:advTrnCWSDTW} shows that DTW-AR outperforms the naive approach CW-SDTW that uses SoftDTW with the Carlini \& Wagner loss function. We also demonstrated several limitations of CW-SDTW to achieve the robustness goal aimed by this paper.
    \item Figure \ref{fig:runtime} clearly demonstrates that DTW-AR significantly reduces the computational cost compared to existing approaches of computing the DTW measure for creating adversarial examples.
    \item Figure \ref{fig:advatkDTWA} demonstrates that DTW-AR can generalize to any multivariate DTW measure (such as DTWAdaptive \cite{shokoohi2017generalizing}) without impacting on its performance and effectiveness.
\end{itemize}

\section{Conclusions}
We introduced the DTW-AR framework to study adversarial robustness of deep models for the time-series domain using dynamic time warping measure. This framework creates effective adversarial examples by overcoming the limitations of prior methods based on Euclidean distance. 
We theoretically and empirically demonstrate the effectiveness of DTW-AR to fool deep models for time-series data and to improve their robustness. We conclude that the time-series domain needs focused investigation for studying robustness of deep models by shedding light on the unique challenges.

\bibliographystyle{IEEEtran}
\bibliography{z_references}



\ifCLASSOPTIONcaptionsoff
  \newpage
\fi


%


\begin{IEEEbiography}[{\includegraphics[width=1in,height=1.25in,clip,keepaspectratio]{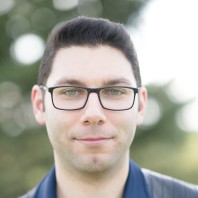}}]{Taha Belkhouja}
(S’19) is a PhD Candidate in Computer Science at Washington State University, USA. His general research interest include robust and trustworthy machine learning for safe deployment in real-world applications. His current research focuses on robustness of machine learning models for time-series domain and theoretically sound uncertainty quantification. He won a Outstanding Teaching Assistant Award from the College of Engineering, Washington State University.
\end{IEEEbiography}

\begin{IEEEbiography}[{\includegraphics[width=1in,height=1.25in,clip,keepaspectratio]{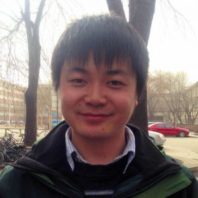}}]{Yan Yan}
(Senior Member from 2022) is an Assistant Professor of Computer Science at Washington State University, USA. He received his PhD in information systems from the University of Technology Sydney. His current research interest includes developing robust machine learning systems and developing sample-efficient learning algorithms. He has published papers in top-tier conferences including NeurIPS, AAAI, IJCAI, ICCV, CVPR, and ECCV.
\end{IEEEbiography}

\begin{IEEEbiography}[{\includegraphics[width=1in,height=1.25in,clip,keepaspectratio]{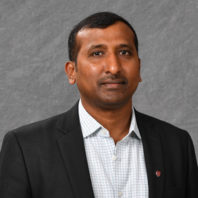}}]{Janardhan Rao Doppa}  (Senior Member, IEEE) is the Huie-Rogers Endowed Chair Associate Professor at Washington State University. He received his PhD in computer science from Oregon State University. His research interests include both foundations of machine learning and applications to real-world problems. He won NSF CAREER award, Outstanding Paper Award from AAAI conference (2013), Best Paper Award from ACM Transactions on Design Automation of Electronic Systems (2021), IJCAI Early Career Award (2021), Best Paper Award from Embedded Systems Week Conference (2022),  Outstanding Junior Faculty in Research Award (2020) and Reid-Miller Teaching Excellence Award (2018) from the College of Engineering, Washington State University.
\end{IEEEbiography}




\newpage
\appendices

\section{Experimental and Implementation Details}

\vspace{1.0ex}

\noindent {\bf Datasets.} We have employed the standard benchmark training, validation, and testing split on the datasets. All datasets are publicly available from the UCR repository \cite{ucrdata}. We employ the datasets in the main paper on which the classifier is able to have a performance better than random guessing. Experiments on the effectiveness of adversarial attack that aim to fool poor classifiers would not exhibit trust-worthy results, as the classifier originally is unable to predict clean data. 

\vspace{1.0ex}

\noindent {\bf DNN architectures.} To evaluate the DTW-AR framework, we employ two different 1D-CNN architectures --- $A_0$ and $A_1$ --- to create two DNNs: $WB$ uses $A_0$ to evaluate the adversarial attack under a white-box setting, and is trained using clean training examples. $BB$ uses the architecture $A_1$ to evaluate the black-box setting for a model trained using clean examples. The architecture details of the deep learning models are presented in Table~\ref{tab:archs}.
\begin{table}[!h]
\centering
\caption{Details of DNN architectures. C: Convolutional layers, K: kernel size, P: max-pooling kernel size, and R: rectified linear layer.}
\begin{tabular}{|l|ccccccc|c}  
\hline
 & C & K & C & K & P & R & R\\
\hline
$A_0$   & x & x & 66 & 12 & 12 & 1024 & x \\
$A_1$   & 100 & 5 & 50 & 5 & 4 & 200 & 100 \\
\hline
\end{tabular}
\label{tab:archs}
\end{table}

\vspace{1.0ex}

\begin{figure}[!h]
    \centering
    \includegraphics[width=.8\linewidth]{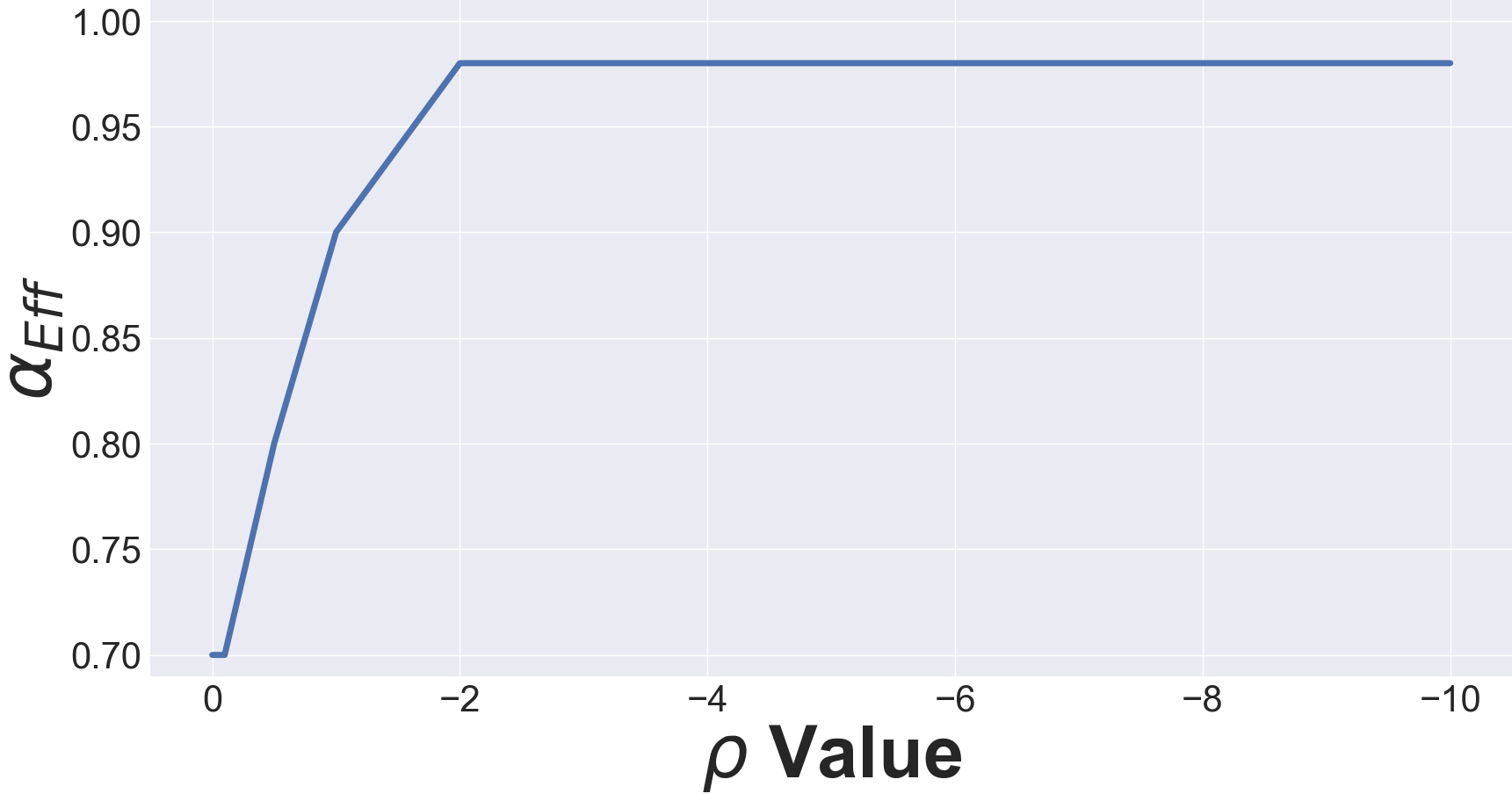}
    \caption{Results for the fooling rate on \texttt{ECG200} dataset w.r.t different $\rho$ values for a black-box attack setting.}
    \label{fig:rho}
\end{figure}

\noindent {\bf DTW-AR implementation.} We implemented the DTW-AR framework using TensorFlow 2 \cite{tensorflow2015whitepaper}.  The parameter $\rho$ that was introduced in Equation \ref{eq:classloss} plays an important role in the algorithm.
\begin{equation}
    \mathcal{L}^{label}(X_{adv}) = \max [ \max_{y \neq y_{target}}  \left( \mathcal{S}_y\left(X_{adv}\right) \right) -
  \mathcal{S}_{y_{target}}\left( X_{adv}\right)\textbf{,} ~~\rho ]
\tag{\ref{eq:classloss}}
\end{equation}
$\rho$ will push gradient descent to minimize mainly the second term ($\mathcal{L}^{DTW}$) when the first term plateaus at $\rho$. Otherwise, the gradient can minimize the general loss function by pushing $\mathcal{L}^{label}$ to $-\infty$, which is counter-productive for our goal. In all our experiments, we employ $\rho=-5$ for $\mathcal{L}^{label}$ in Equation \ref{eq:classloss} for a good confidence in the classification score. A good confidence score is important for the attack's effectiveness in a black-box setting. Black-box setting assumes that information about the target deep model including its parameters $\theta$ are not accessible. In general, the attacker will create a proxy deep model to mimic the behavior of the target model using regular queries. This technique can be more effective when a target scenario is well-defined \cite{tramer2020adaptive,papernot2017practical}. However, in this work, we consider the general case where we do not query the black-box target DNN classifier for a better assessment of the proposed framework.  Figure \ref{fig:rho} shows the role of $\rho$ value in enhancing DTW-AR attacks in a black-box setting on \texttt{ECG200} dataset noting that we see similar patterns for other datasets as well.
Adversarial examples are generated using a maximum of $5 \times 10^3$ iterations of gradient descent with the fixed learning rate $\eta$=0.01. After all the iterations, the final adversarial output is chosen from the iteration with the lowest DTW loss provided from Equation \ref{eq:dtwloss}.

\begin{equation}
\begin{split}
    \mathcal{L}^{DTW}(X_{adv}, P)  =&  \alpha_1 \times dist_{P}(X,X_{adv}) \\ &-\alpha_2 \times dist_{P_{diag}}(X,X_{adv})
\end{split}
\tag{\ref{eq:dtwloss}}
\end{equation}

Experimentally, we notice that for $d(\cdot,\cdot)$ in Equation 1 of the main paper, there is no influence on the performance between choosing $p=1,2$ or $\infty$ for $d(\cdot,\cdot) = \|\cdot\|_p$. However, for datasets in $\mathbb{R}^{n \times T}$ with $n\ge2$, we do not use $p=\infty$ as the dimensions are not normalized and data points will be compared only along the dimension with the greater magnitude.

\vspace{1.0ex}
\noindent {\bf Implementation of baselines.} The baseline methods for CW, PGD ,and FGS were implemented using the CleverHans library \cite{papernot2018cleverhans} with updates to TensorFlow 2. For FGS and PGD algorithms, we employed a minimal perturbation factors ($\epsilon < 1$ ) for two main reasons. First, larger perturbations significantly degrade the overall performance of the adversarial training and potentially creates adversarial signals that are semantically different than the original time-series input. Second, we want to avoid the risk of leaking label information \cite{madry2017towards}. STN was implemented using the code provided with the paper \cite{zheng2016improving}.

\section{Additional Experimental Results}
\vspace{1.0ex}
\label{sec:knndtw}
\noindent \textbf{Discussion on kNN-DTW based classification.} It has been shown previously \cite{dau2018optimizing,shokoohi2017generalizing} that the Nearest-Neighbour (NN) algorithm is suitable for time-series classification using DTW measure. Nevertheless, DNN classifiers show promising results (e.g., high accuracy, ease of deployment) in their use for the time-series domain:
\begin{itemize}
    \item While kNN-DTW can be effective in many settings, as demonstrated by the results in Table \ref{tab:knnPerf}, the accuracy of kNN-DTW algorithm remains lower when compared to the deep models considered in our study on the real-world {\em multivariate} datasets used for evaluation in the main paper. To implement kNN-DTW, we consider the implementation of a regular kNN algorithm, and we use cDTW [36] with its provided public python implementation (\href{https://wu.renjie.im/research/fastdtw-is-slow/#cdtw-alone}{https://wu.renjie.im/research/fastdtw-is-slow/\#cdtw-alone} Copyright 2020 Renjie Wu and Sara Alaee) with $c=10$ (This choice is based on the empirical evaluation in the cDTW paper \cite{dau2018optimizing}).
    \item Using kNN-DTW ($k>1$) instead of 1NN-DTW is not appropriate and principled for multivariate time-series data. By definition, DTW is an elastic similarity measure (the triangle inequality does not hold for DTW). Hence, if we have
    \begin{itemize}
        \item $X$ and $X_1$ are DTW-similar according to a warping path $Path_1$
        \item $X$ and $X_3$ are DTW-similar according to a warping path $Path_2$,
    \end{itemize} 
    then we cannot draw a straightforward conclusion that $X_1$ and $X_2$ are DTW-similar. Such an assumption yields a weak voting accuracy on the predicted label for high-dimensional and multivariate data. Consequently, kNN-DTW with $k>1$ for multivariate data can fail. Our empirical results in Table \ref{tab:knnPerf} corraborate this hypothesis by showing that 1NN-DTW performs better then kNN-DTW ($k>1$) in most cases.
    \item To be able to use kNN-DTW algorithms, the training data must be stored and accessible by the end-user. This rises many concerns including
    \begin{itemize}
        \item Data privacy: For applications such as healthcare and finance, the data is privileged and needs to be secured. A deployed classification model should not have direct access to the data.
        \item Storage space and scalability: For applications such as Human Activity Recognition where the models are deployed on resource-constrained hardware platforms, the use of resources for merely storing the data is inefficient. The data cannot be stored along with the classifier.
    \end{itemize}
\end{itemize}
Therefore, the use of DNNs for time-series data is well-motivated. Hence, there is a clear need for focused investigation to study robustness of deep models for the time-series domain.

\begin{table}[!h]
    \centering
    \caption{Accuracy (\%) of kNN-DTW classifier vs. 1D-CNN classifier task on the clean multivariate time-series data.}
    \resizebox{\linewidth}{!}{%
    \begin{tabular}{|l|c|c|c|c|c|} 
    \hline
     & Atrial Fibrillation & Epilepsy & ERing & Heartbeat & RacketSports \\ \hline
     1NN-DTW & 38  & 56  & 85  & 63 & 75 \\ \hline
     5NN-DTW & 13  & 40  & 86  & 67  & 70  \\ \hline
     10NN-DTW & 36  & 32  & 75  & 68  & 65  \\ \hline
     1D-CNN & 40  & 95  & 94  & 70  & 86  \\ \hline
    \end{tabular}
    \label{tab:knnPerf}
    }
\end{table}

\begin{figure}
    \centering
        \begin{minipage}{.8\linewidth}
                \centering                    \includegraphics[width=\linewidth]{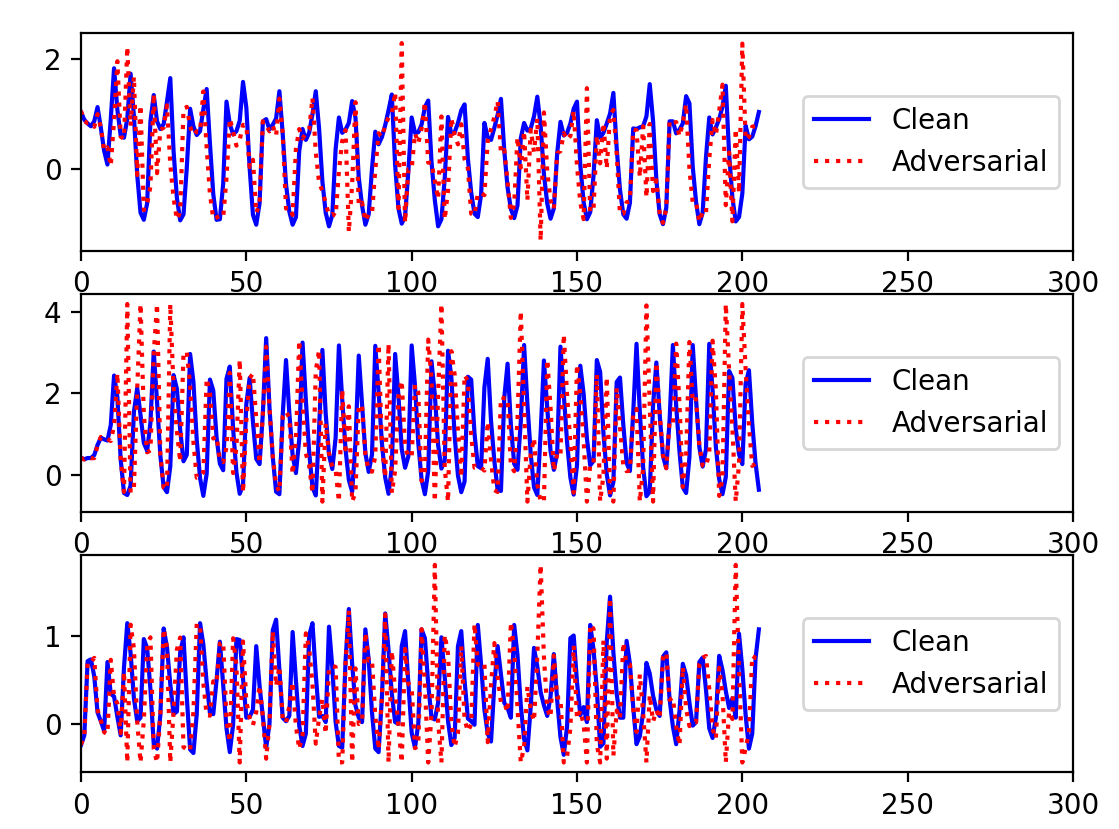}
            \end{minipage}
        \begin{minipage}{.8\linewidth}
                \centering
                \includegraphics[width=\linewidth]{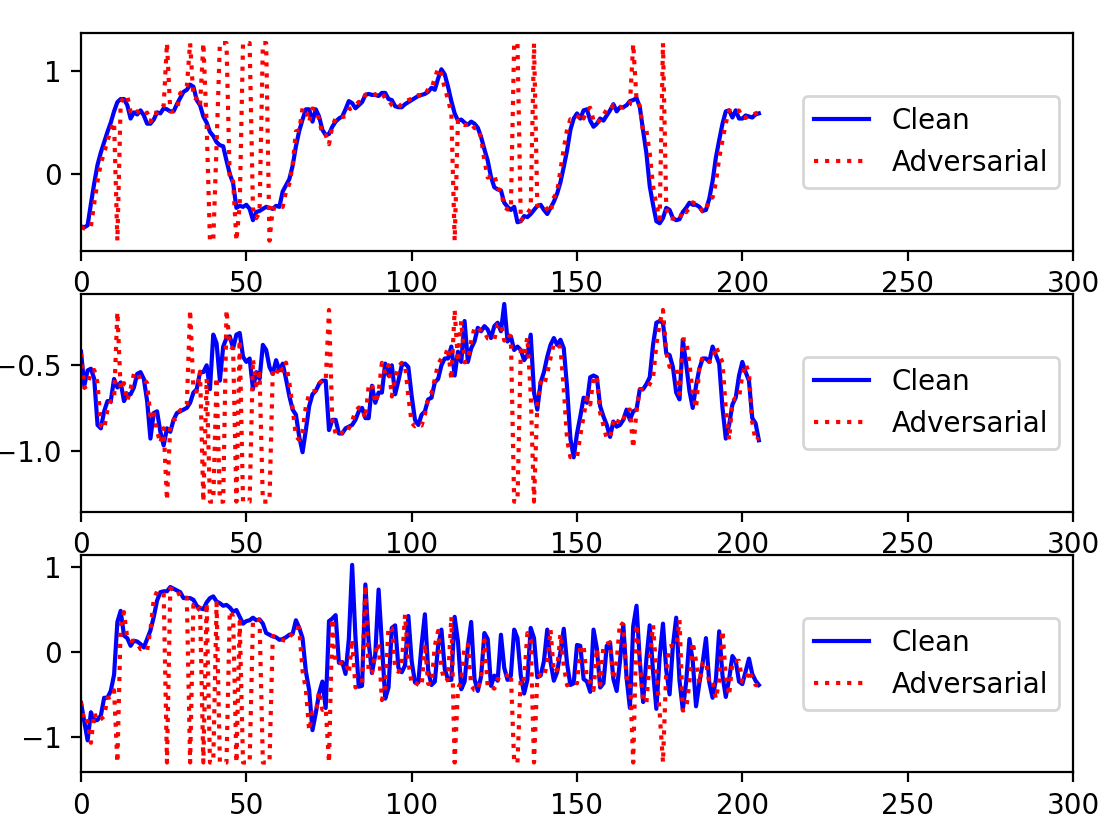}
            \end{minipage}
    \caption{DTW-AR adversarial examples from Epilepsy dataset using pre-defined warping path from user.}
    \label{fig:expEpilepsy}
\end{figure}

\noindent \textbf{Comparison of DTW-based adversarial example generation.} A different approach to create new examples that can be used for data augmentation is proposed by using resampling perturbations \cite{dau2018optimizing}. This approach ensures  a close DTW measure to the original example. However, we would like to clarify that the algorithm proposed in \cite{dau2018optimizing} and our DTW-AR have different goals. While the method in \cite{dau2018optimizing} aims for simple and effective DTW-based data augmentation, these examples cannot be considered as adversarial. Additionally, it does not provide control to the user over the warping path. Using DTW-AR and the tightness guarantees provided in Section 3.3, the user can generate several warped examples with full control over the warped timesteps. Finally, using Equation (5), the generated example is adversarial by definition. As a result, we obtain adversarial examples that remain close to the clean time-series input DTW-wise as illustrated in Figure \ref{fig:expEpilepsy}.

For a fair comparison, we implemented the method proposed by \cite{dau2018optimizing} in python as follows:
\begin{lstlisting}[language=Python]
def add_warping(T,p):
    len_T = len(T)
    i = numpy.arange(len_T)
    numpy.random.shuffle(i)
    i = sorted(i[:-int(np.floor(len_T * p))])
    warped_T = scipy.signal.savgol_filter(scipy.signal.resample(T[i], len_T),1,0)
    return warped_T
\end{lstlisting}

We use the method from \cite{dau2018optimizing} to generate additional examples to be used for adversarial training. Next, we compare the accuracy of the learned predictive models from adversarial training using DTW-AR and method in \cite{dau2018optimizing}. Table \ref{tab:cDTWadv} shows the results on multiple real-world time-series datasets. The accuracy is evaluated on 1) Clean examples, 2) Examples generated by the warping function provided by \cite{dau2018optimizing}, and 3) DTW-AR adversarial examples. Table \ref{tab:cDTWadv} clearly show that adversarial training based on \cite{dau2018optimizing} does not yield a robust model that can improve the performance of deep models against perturbations. Additionally, the same table explains that examples generated by method in \cite{dau2018optimizing} cannot be considered as adversarial. We clearly observe that these examples have low effect on decreasing the models' classification accuracy performance (unlike DTW-AR based examples). Hence, we can conclude that the method in \cite{dau2018optimizing} and DTW-AR are complementary for DTW-based data generation tasks, where \cite{dau2018optimizing} aims for simple and training-effective warped examples for standard classification tasks, and DTW-AR aims for improving the robustness of deep classifiers over time-series data.

\begin{table*}[!h]
    \centering
    \caption{Accuracy (\%) of method in \cite{dau2018optimizing} and DTW-AR based adversarial training on testing examples from different datasets.}
    \resizebox{\linewidth}{!}{%
    \begin{tabular}{|l|c|c|c|c|c|c|c|c|c|} 
    \hline
     & \multicolumn{3}{|c|}{Atrial Fibrillation} & \multicolumn{3}{|c|}{Epilepsy} & \multicolumn{3}{|c|}{ERing}  \\ \hline
     & Clean Exp. & \cite{dau2018optimizing} Exp. & DTW-AR Exp. & Clean Exp. & \cite{dau2018optimizing} Exp. & DTW-AR Exp. & Clean Exp. & \cite{dau2018optimizing} Exp. & DTW-AR Exp.  \\ \hline
     \cite{dau2018optimizing} Adv. training 
     & 42  & 42  &  29 
     & 90  & 90 & 26  
     & 90  & 89  &  43 \\ \hline
     DTW-AR Adv. training 
     & 42  & 38  &  82 
     & 98  & 93  & 96 
     & 96  & 90  & 99  \\ \hline
     &  \multicolumn{3}{|c|}{Heartbeat} & \multicolumn{3}{|c|}{RacketSports} & \multicolumn{3}{|c}{\multirow{4}{*}{ }} \\ \cline{1-7}
     &  Clean Exp. & \cite{dau2018optimizing} Exp. & DTW-AR Exp. & Clean Exp. & \cite{dau2018optimizing} Exp. & DTW-AR Exp. \\ \cline{1-7}
     \cite{dau2018optimizing} Adv. training 
     
     & 70  & 68  &  20 
     & 86  & 83  & 52 \\ \cline{1-7}
     DTW-AR Adv. training 
      
     & 75  & 72  &  96 
     & 86 & 80 & 92  \\ \cline{1-7}
    \end{tabular}
    }
    \label{tab:cDTWadv}
\end{table*}

\vspace{1.0ex}

\begin{figure*}[!h]
    \centering
        \begin{minipage}{\linewidth}
        \begin{minipage}{.19\linewidth}
                \centering
                \includegraphics[width=\linewidth]{PerfFigures/AtkPeroformanceAtrialFibrillation.png}
            \end{minipage}%
        \begin{minipage}{.19\linewidth}
                \centering
                \includegraphics[width=\linewidth]{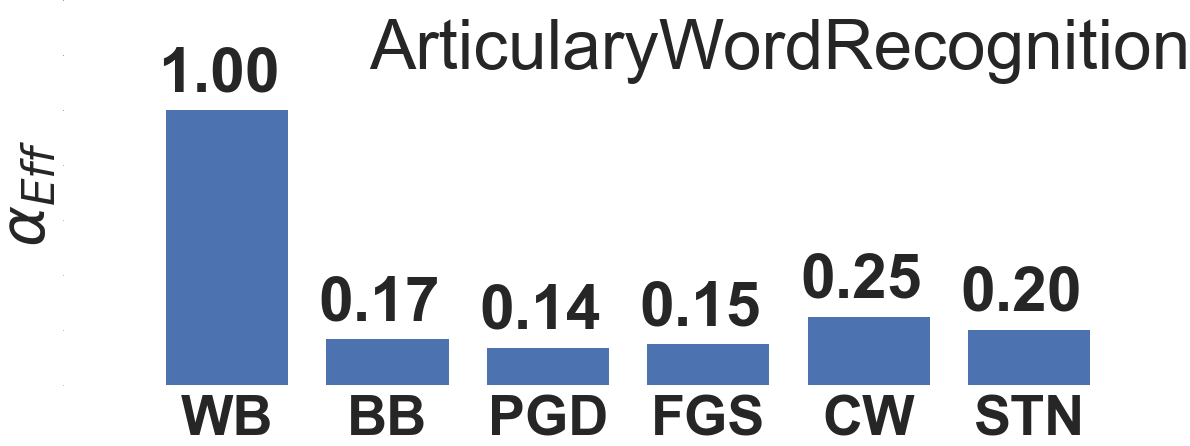}
            \end{minipage}%
        \begin{minipage}{.19\linewidth}
                \centering
                \includegraphics[width=\linewidth]{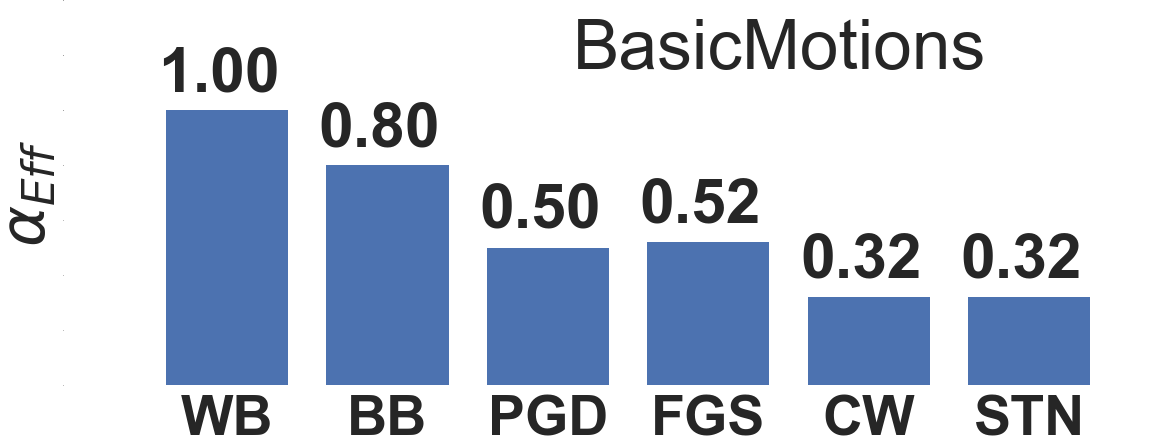}
            \end{minipage}%
        \begin{minipage}{.19\linewidth}
                \centering
                \includegraphics[width=\linewidth]{PerfFigures/AtkPeroformanceEpilepsy.png}
            \end{minipage}%
        \begin{minipage}{.19\linewidth}
                \centering
                \includegraphics[width=\linewidth]{PerfFigures/AtkPeroformanceERing.png}
            \end{minipage}
        \begin{minipage}{.19\linewidth}
                \centering
                \includegraphics[width=\linewidth]{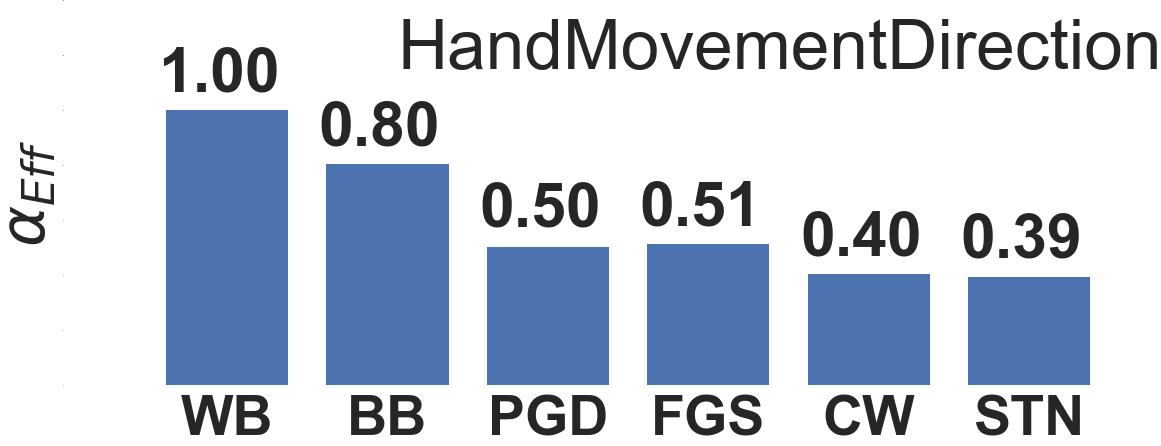}
            \end{minipage}%
        \begin{minipage}{.19\linewidth}
                \centering
                \includegraphics[width=\linewidth]{PerfFigures/AtkPeroformanceHeartbeat.png}
            \end{minipage}%
        \begin{minipage}{.19\linewidth}
                \centering
                \includegraphics[width=\linewidth]{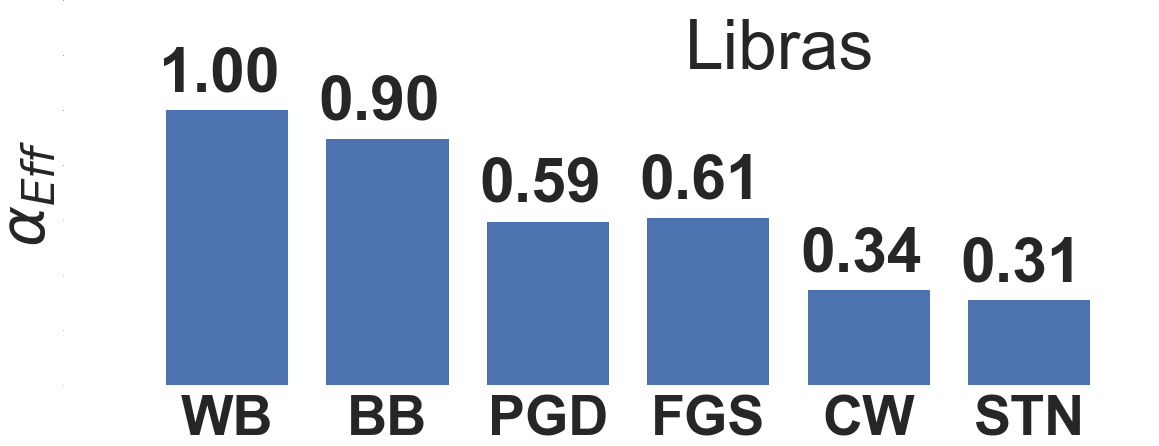}
            \end{minipage}%
        \begin{minipage}{.19\linewidth}
                \centering
                \includegraphics[width=\linewidth]{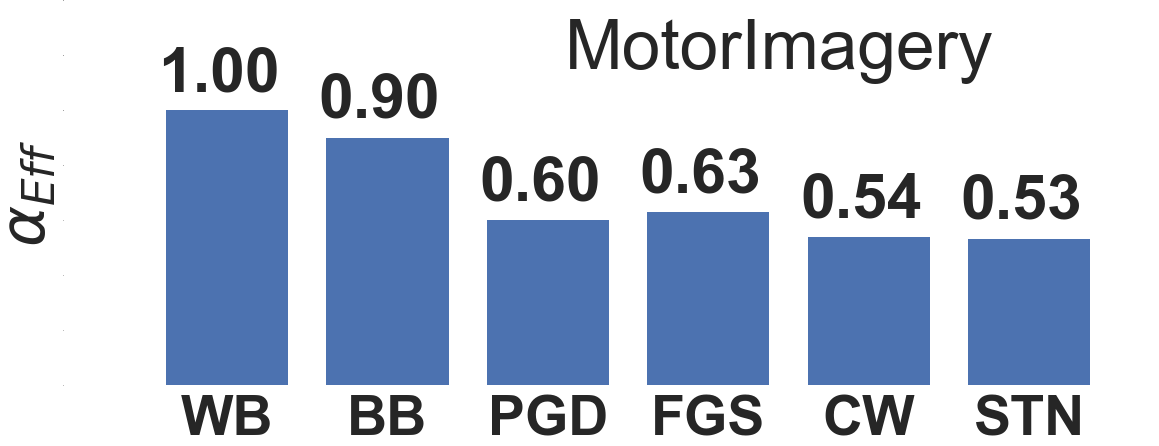}
            \end{minipage}%
        \begin{minipage}{.19\linewidth}
                \centering
                \includegraphics[width=\linewidth]{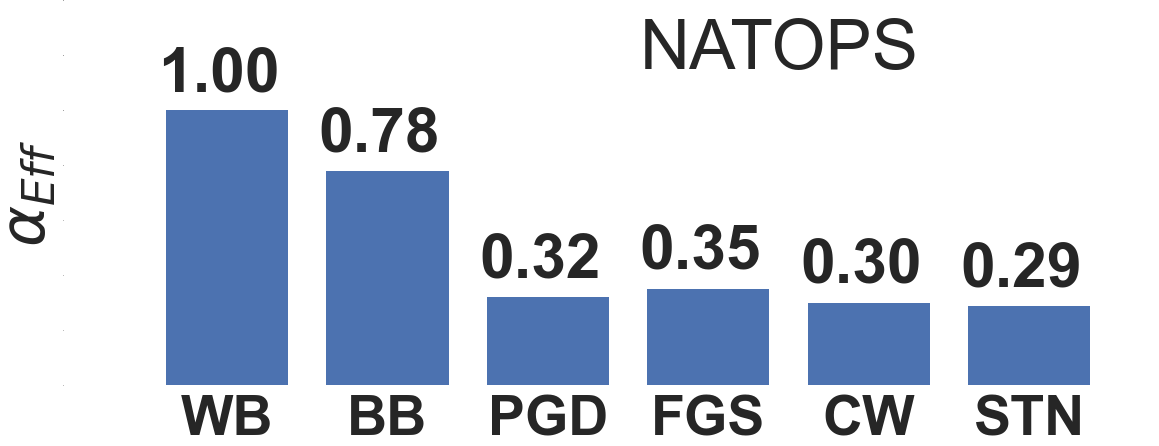}
            \end{minipage}
        \begin{minipage}{.19\linewidth}
                \centering
                \includegraphics[width=\linewidth]{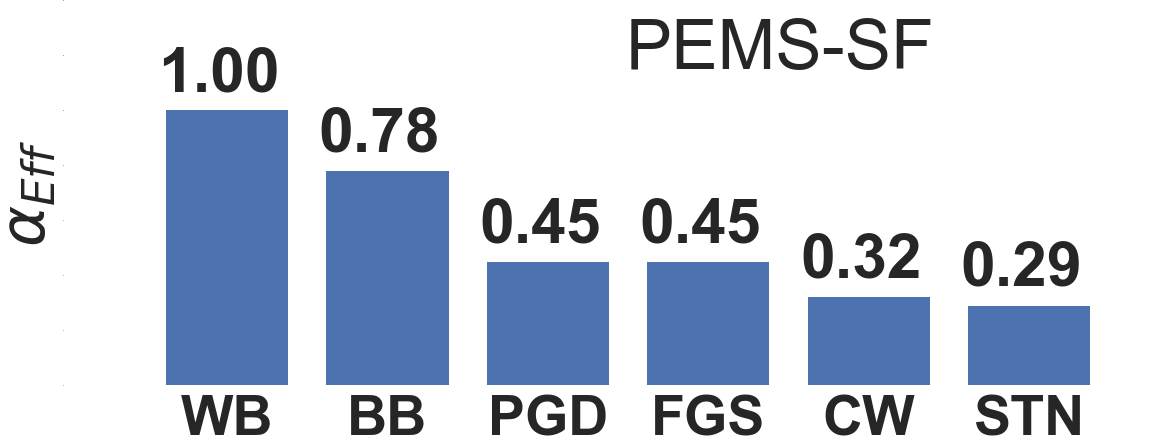}
            \end{minipage}%
        \begin{minipage}{.19\linewidth}
                \centering
                \includegraphics[width=\linewidth]{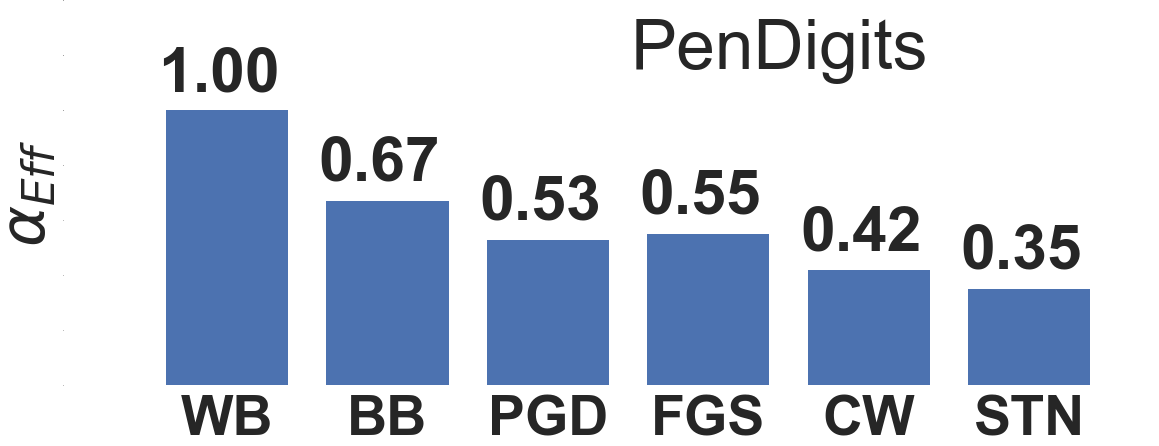}
            \end{minipage}%
        \begin{minipage}{.19\linewidth}
                \centering
                \includegraphics[width=\linewidth]{PerfFigures/AtkPeroformanceRacketSports.png}
            \end{minipage}%
        \begin{minipage}{.19\linewidth}
                \centering
                \includegraphics[width=\linewidth]{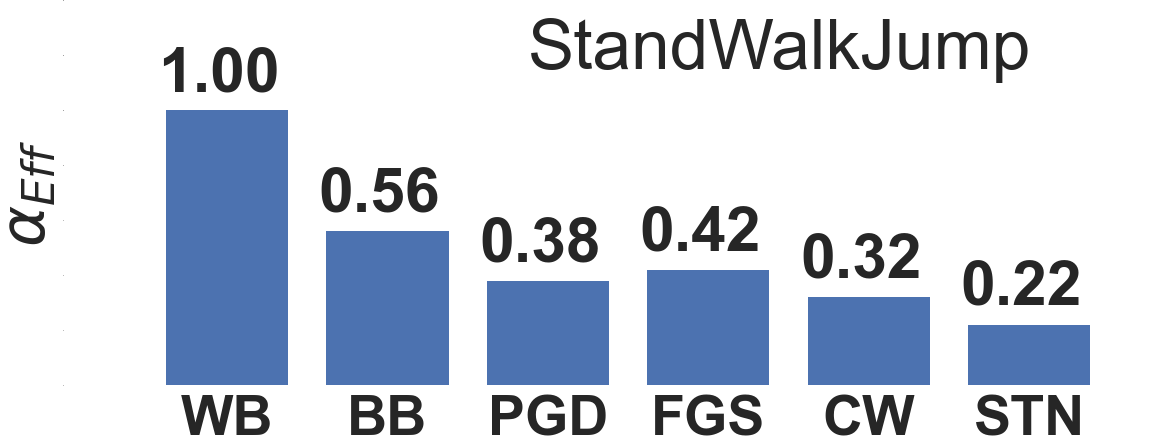}
            \end{minipage}%
        \begin{minipage}{.19\linewidth}
                \centering
                \includegraphics[width=\linewidth]{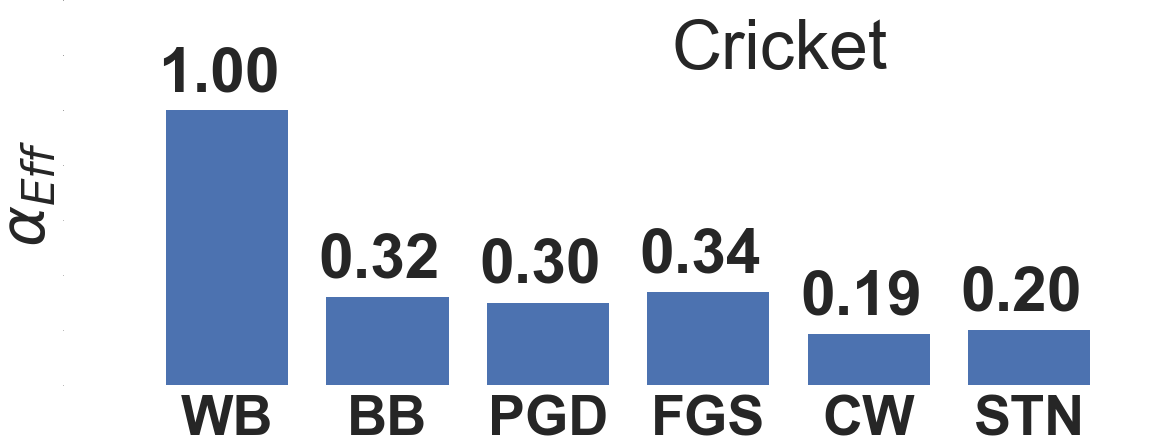}
            \end{minipage}
    \end{minipage}
    \vspace{-2ex}
\caption{Results for the effectiveness of adversarial examples from DTW-AR on different DNNs under white-box (WB) and black-box (BB) settings, and using adversarial training baselines (PGD, FGS, CW and STN) on all the UCR multivariate datasets} 
\label{fig:appendadvatk}
\end{figure*}

\begin{figure*}[t]
    \centering
        \begin{minipage}{\linewidth}
        \begin{minipage}{.19\linewidth}
                \centering
                \includegraphics[width=\linewidth]{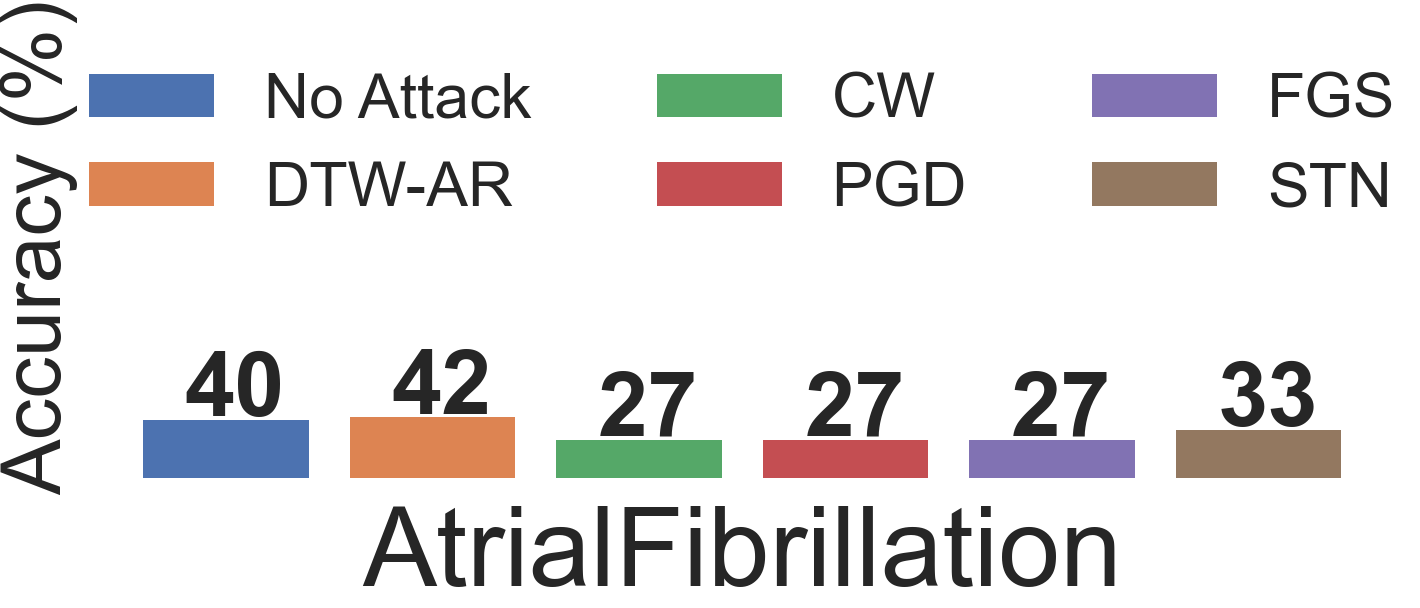}
            \end{minipage}%
        \begin{minipage}{.19\linewidth}
                \centering
                \includegraphics[width=\linewidth]{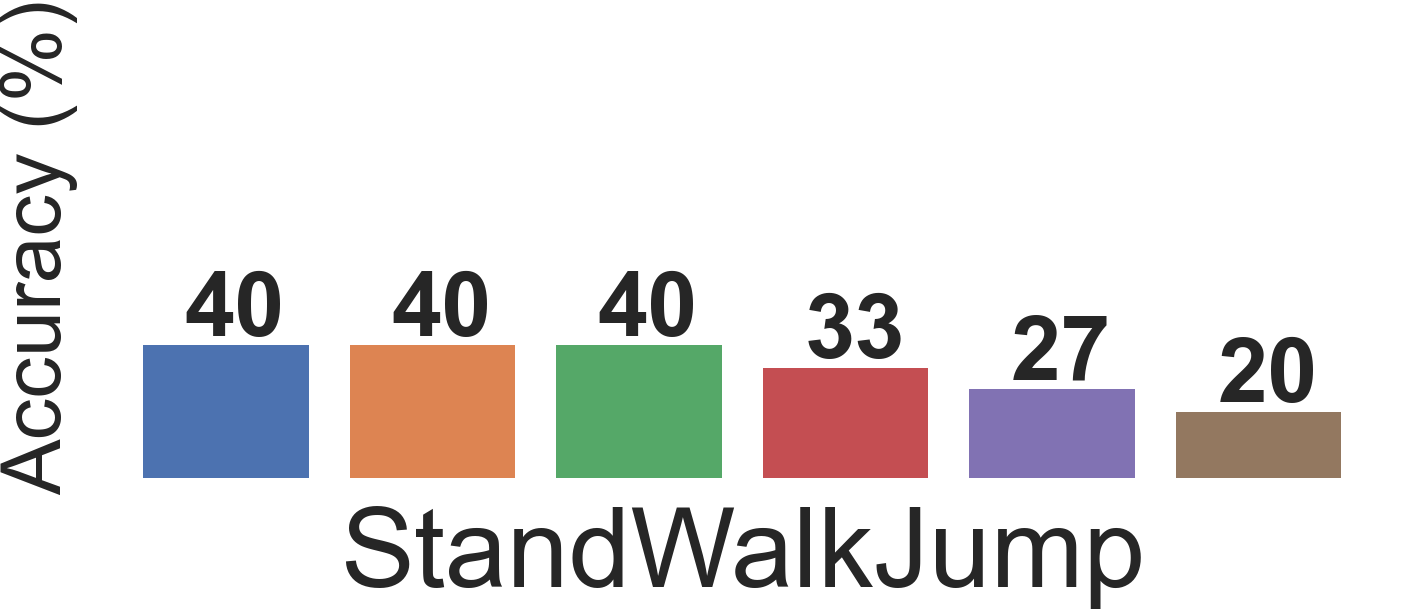}
            \end{minipage}%
        \begin{minipage}{.19\linewidth}
                \centering
                \includegraphics[width=\linewidth]{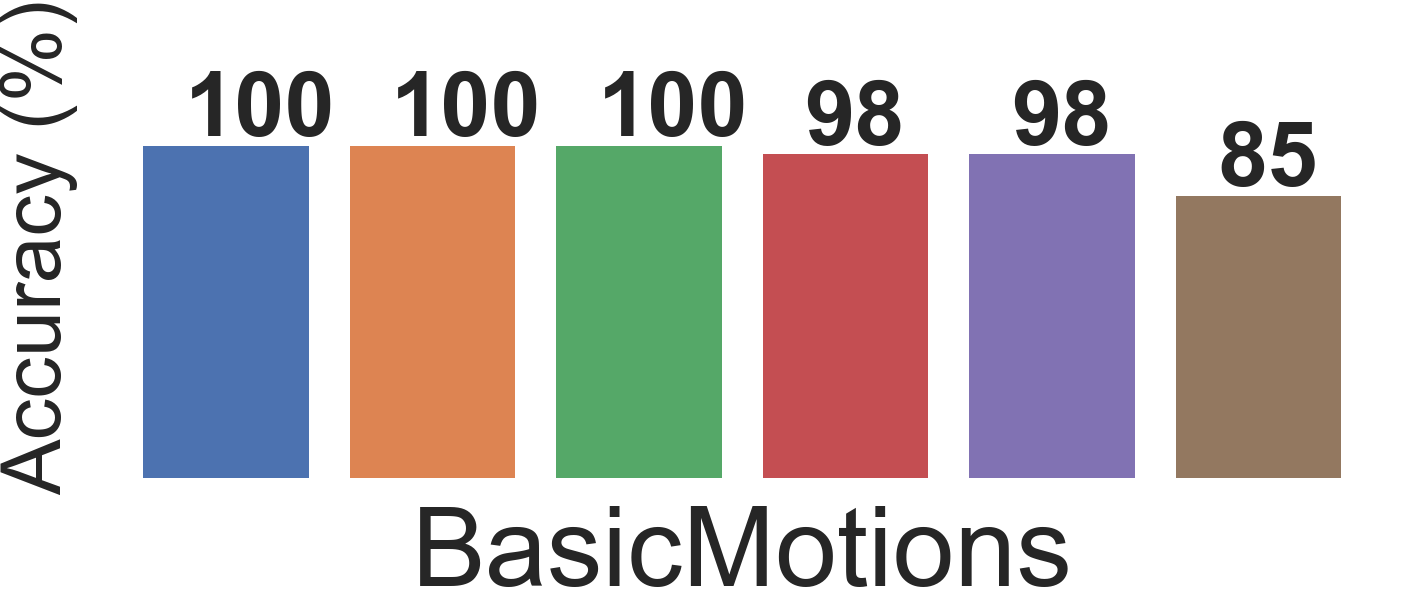}
            \end{minipage}%
        \begin{minipage}{.19\linewidth}
                \centering
                \includegraphics[width=\linewidth]{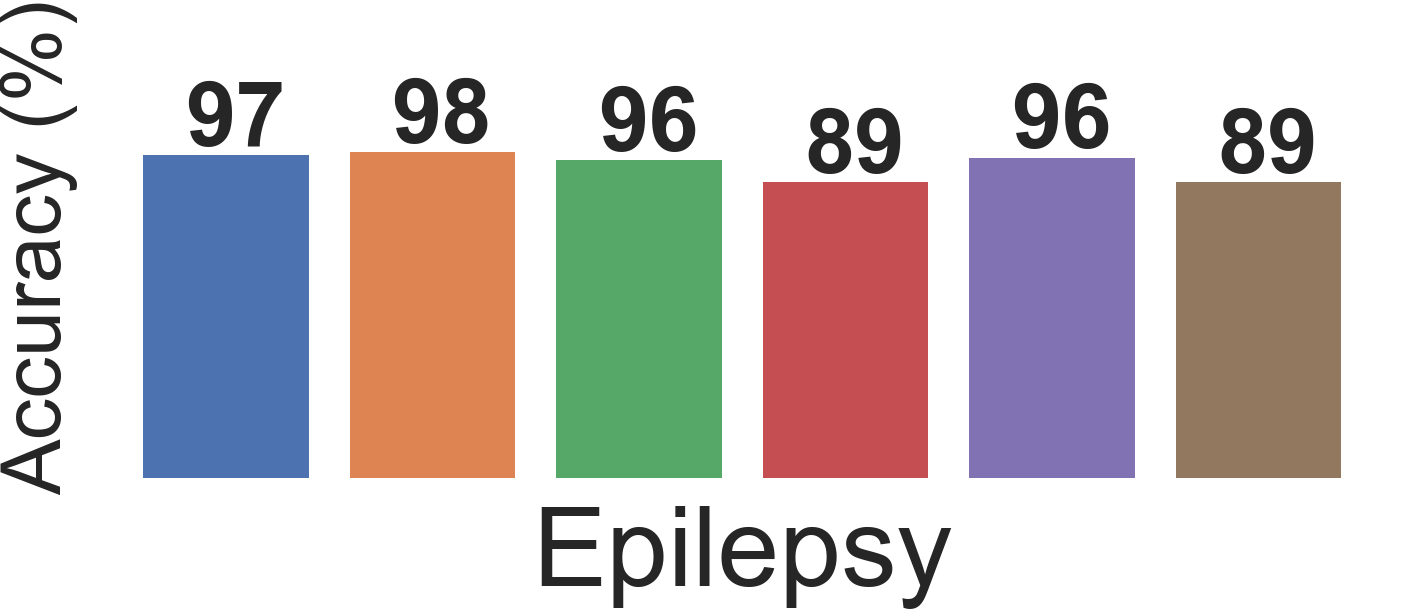}
            \end{minipage}%
        \begin{minipage}{.19\linewidth}
                \centering
                \includegraphics[width=\linewidth]{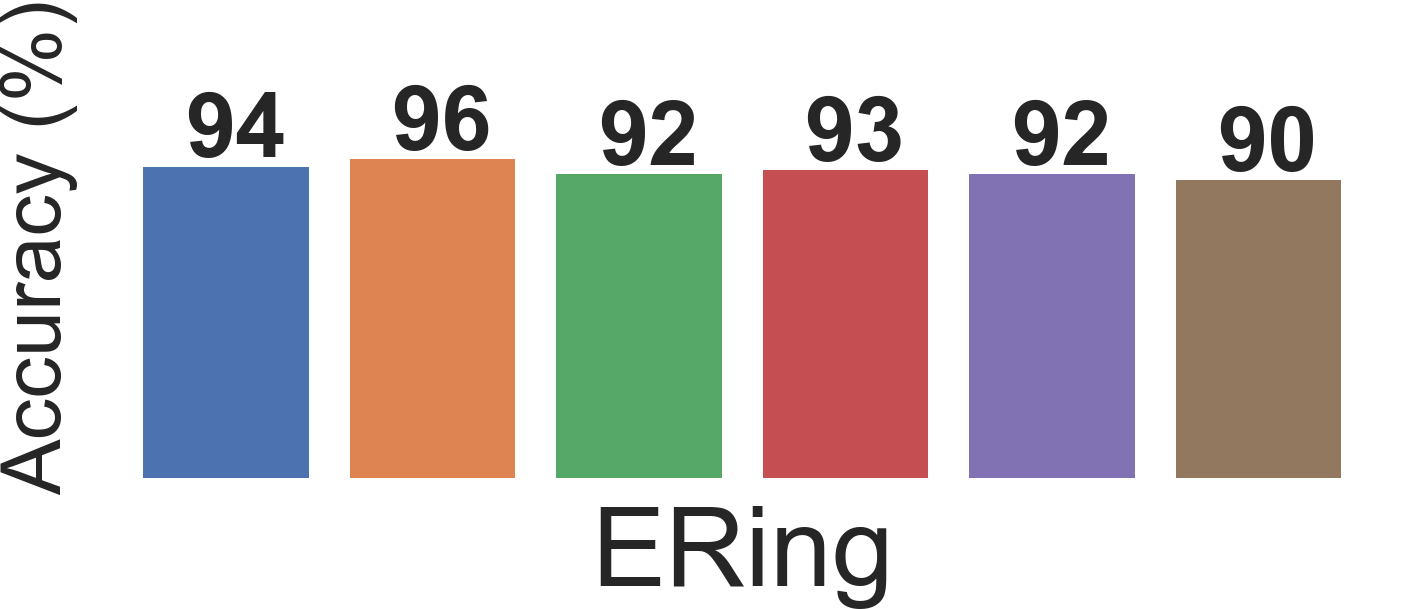}
            \end{minipage}
        \begin{minipage}{.19\linewidth}
                \centering
                \includegraphics[width=\linewidth]{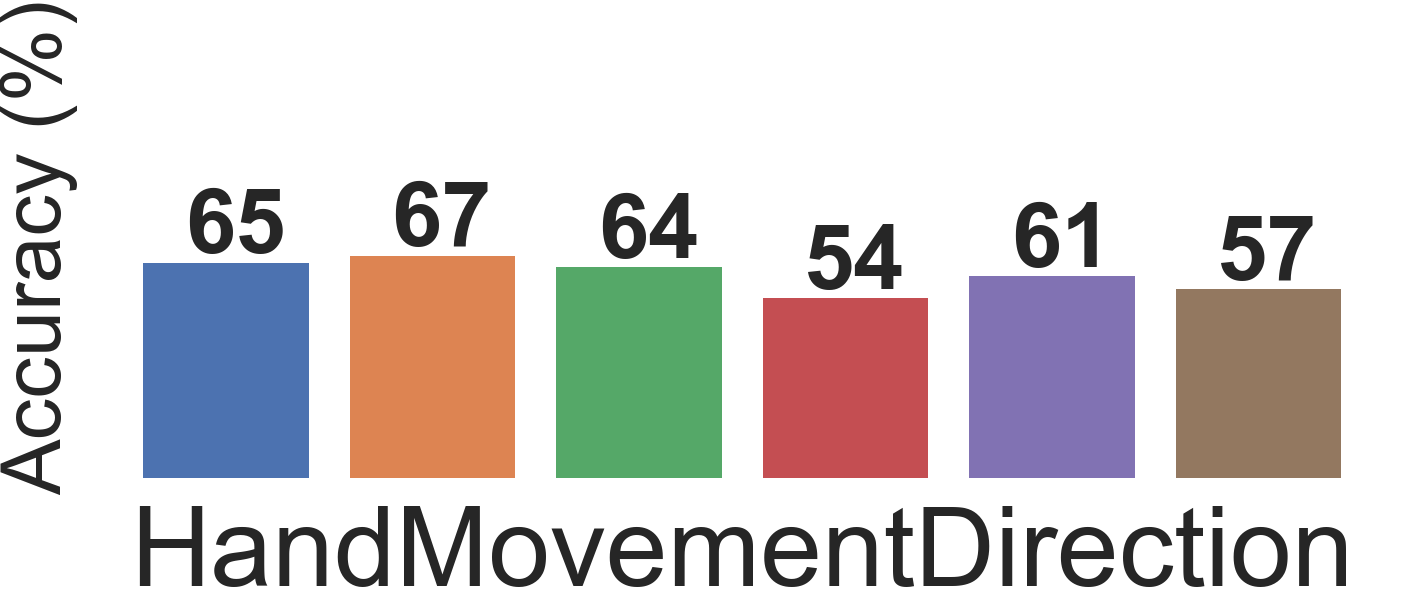}
            \end{minipage}%
        \begin{minipage}{.19\linewidth}
                \centering
                \includegraphics[width=\linewidth]{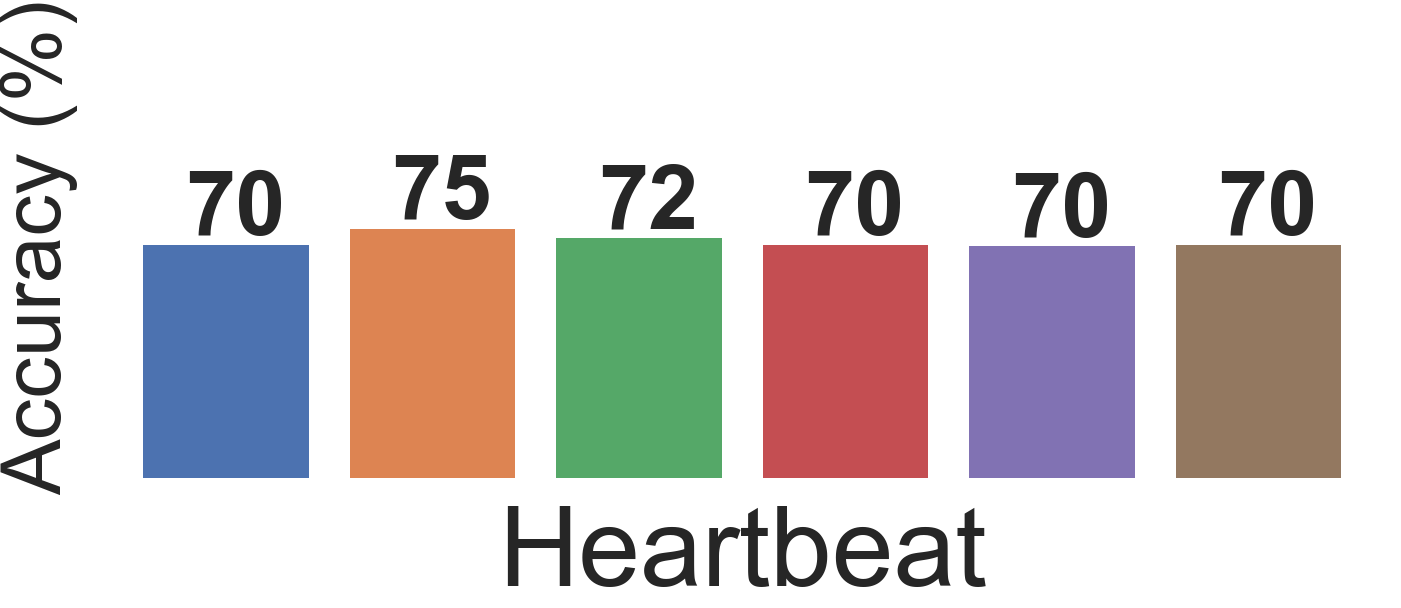}
            \end{minipage}%
        \begin{minipage}{.19\linewidth}
                \centering
                \includegraphics[width=\linewidth]{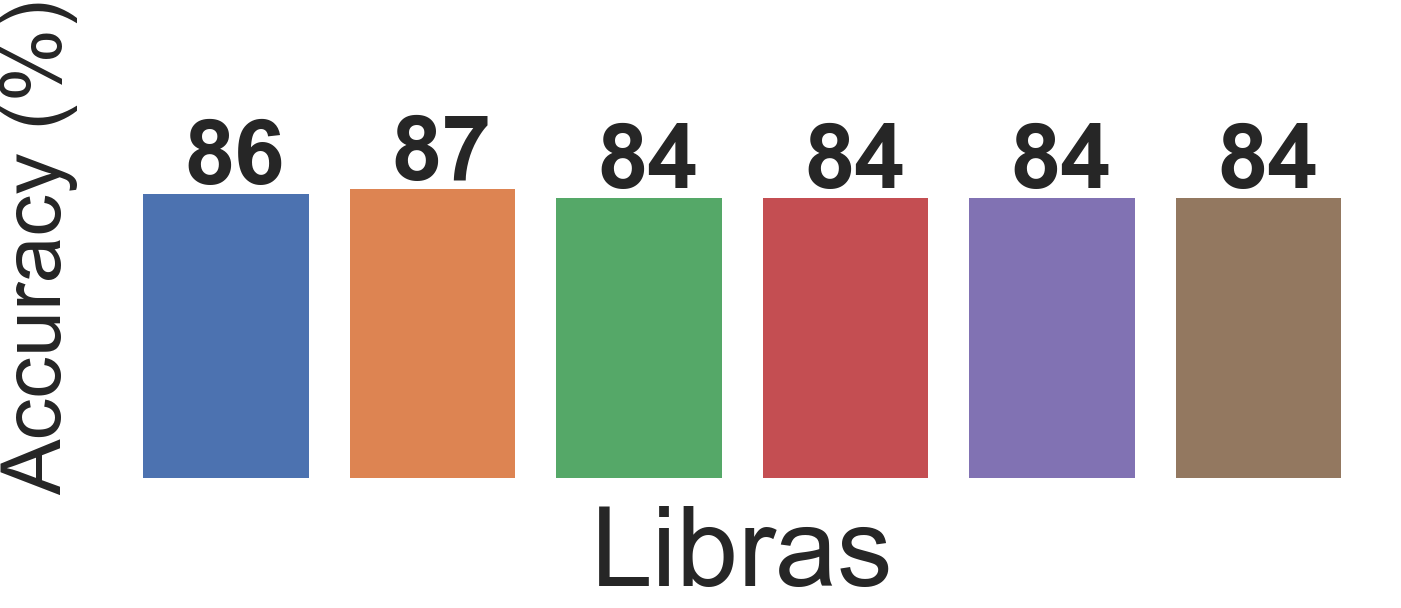}
            \end{minipage}%
        \begin{minipage}{.19\linewidth}
                \centering
                \includegraphics[width=\linewidth]{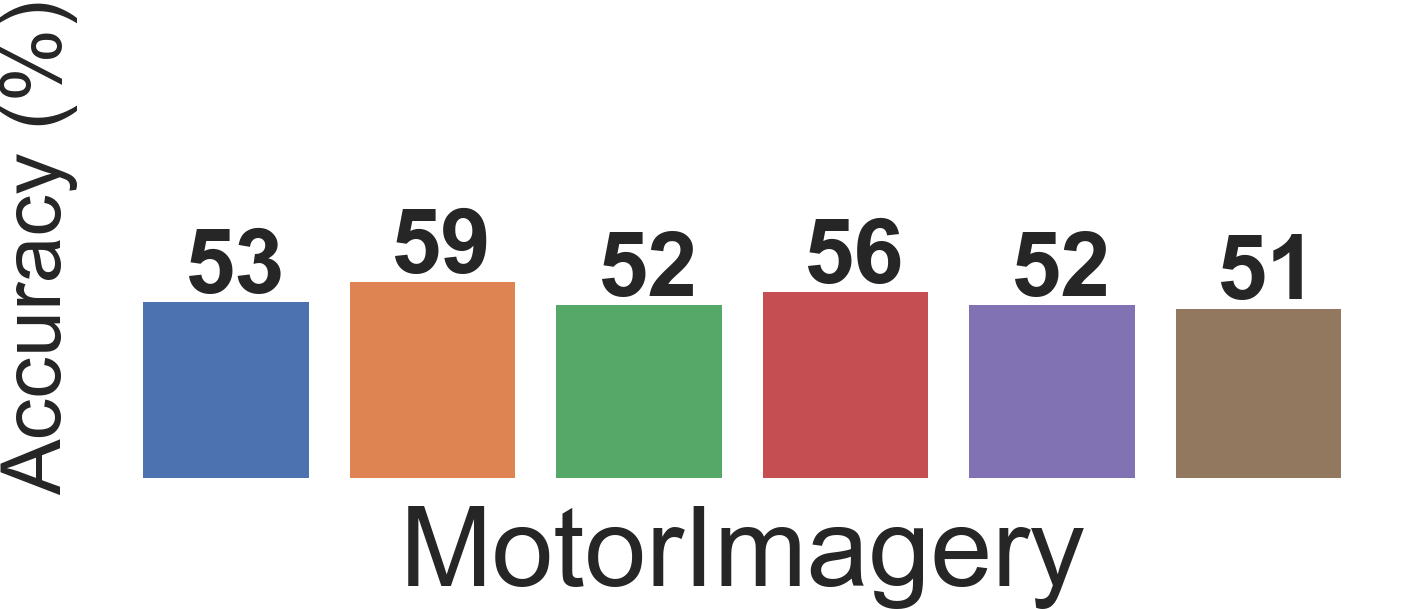}
            \end{minipage}%
        \begin{minipage}{.19\linewidth}
                \centering
                \includegraphics[width=\linewidth]{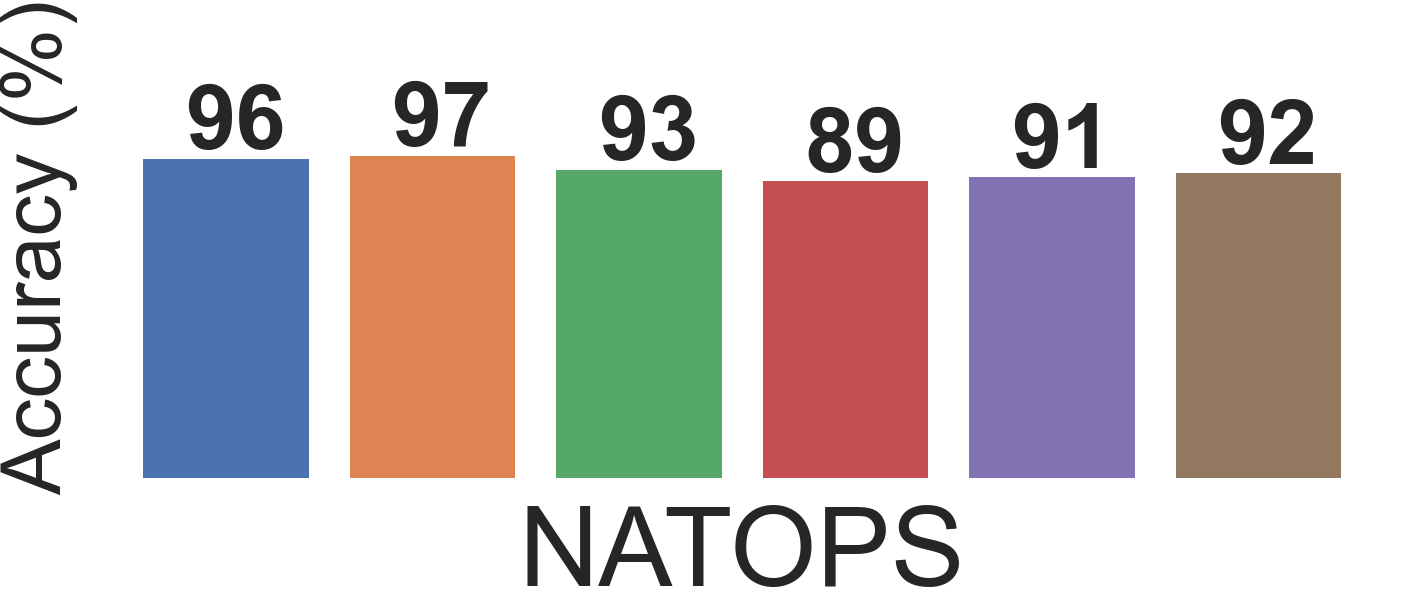}
            \end{minipage}
        \begin{minipage}{.19\linewidth}
                \centering
                \includegraphics[width=\linewidth]{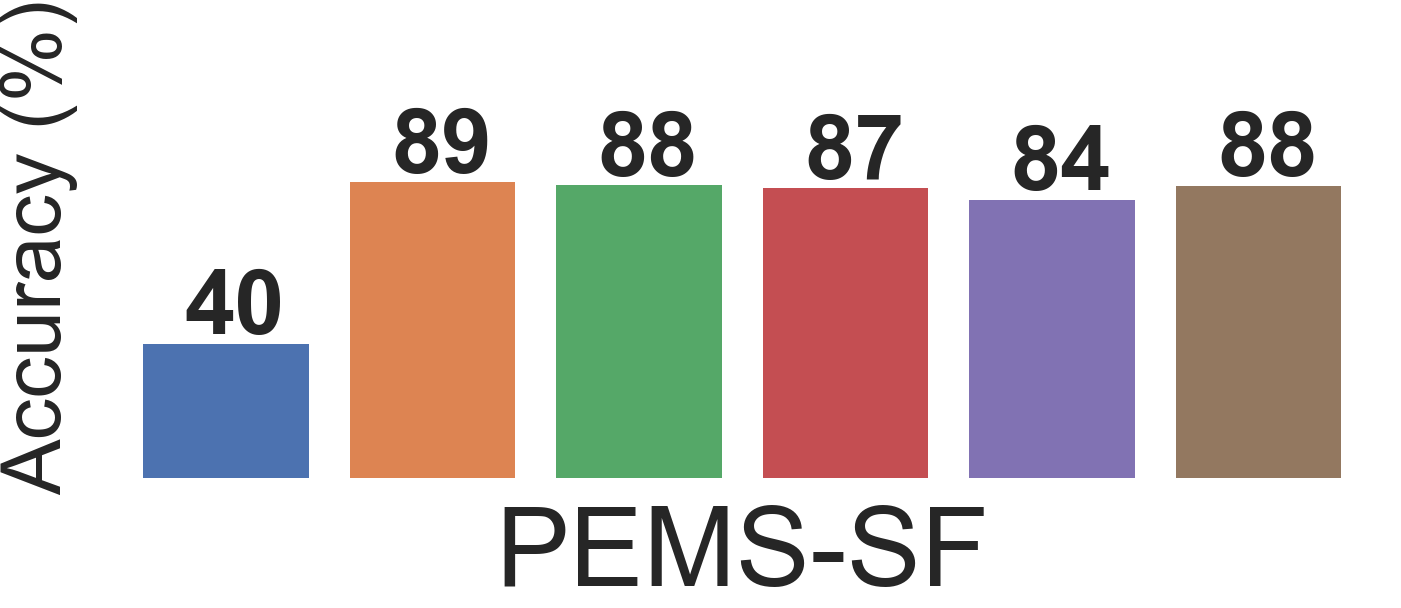}
            \end{minipage}%
        \begin{minipage}{.19\linewidth}
                \centering
                \includegraphics[width=\linewidth]{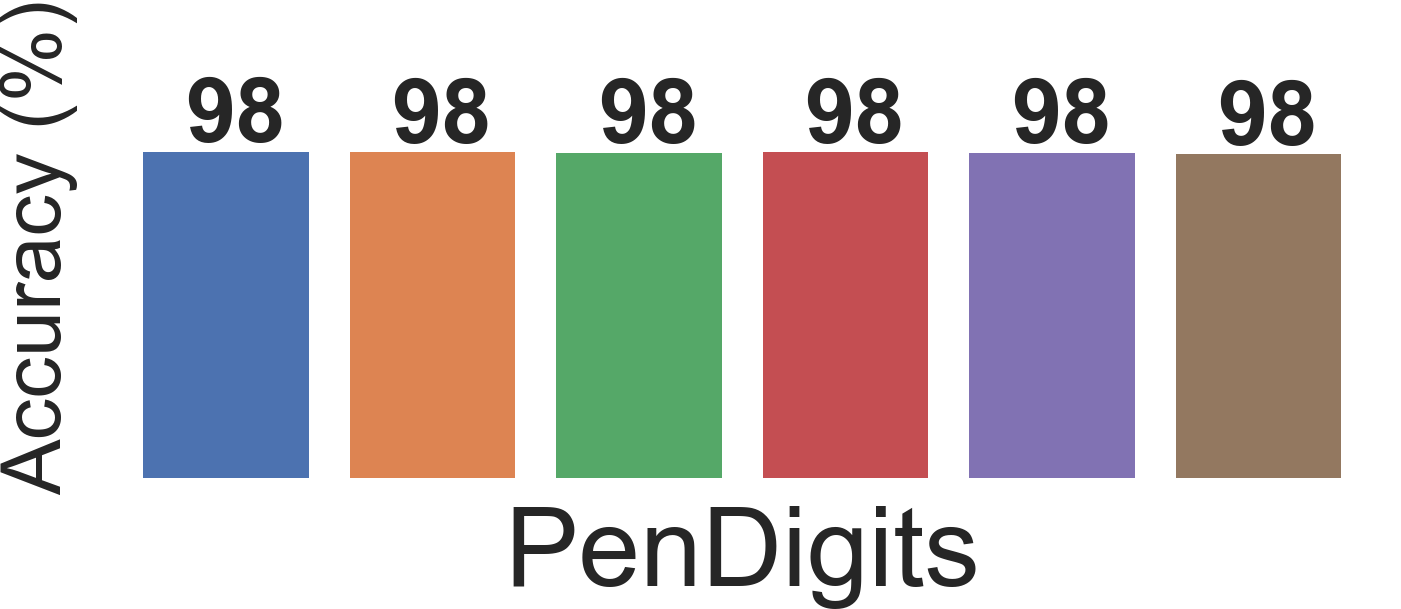}
            \end{minipage}%
        \begin{minipage}{.19\linewidth}
                \centering
                \includegraphics[width=\linewidth]{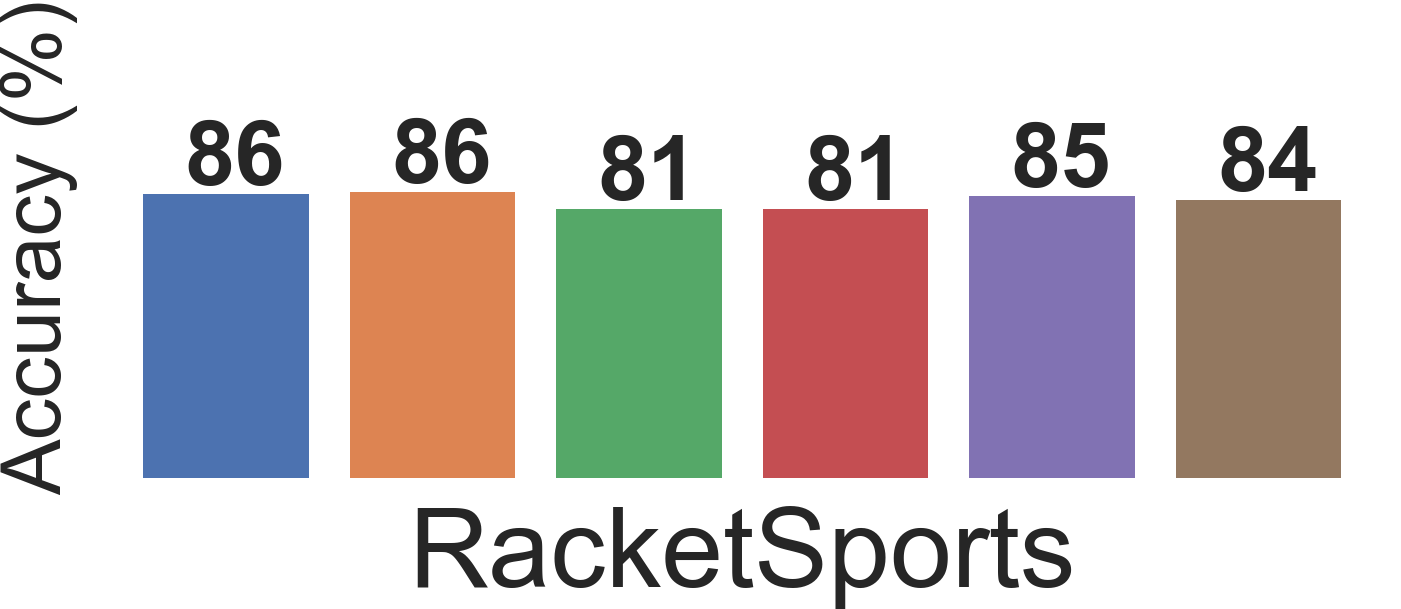}
            \end{minipage}%
        \begin{minipage}{.19\linewidth}
                \centering
                \includegraphics[width=\linewidth]{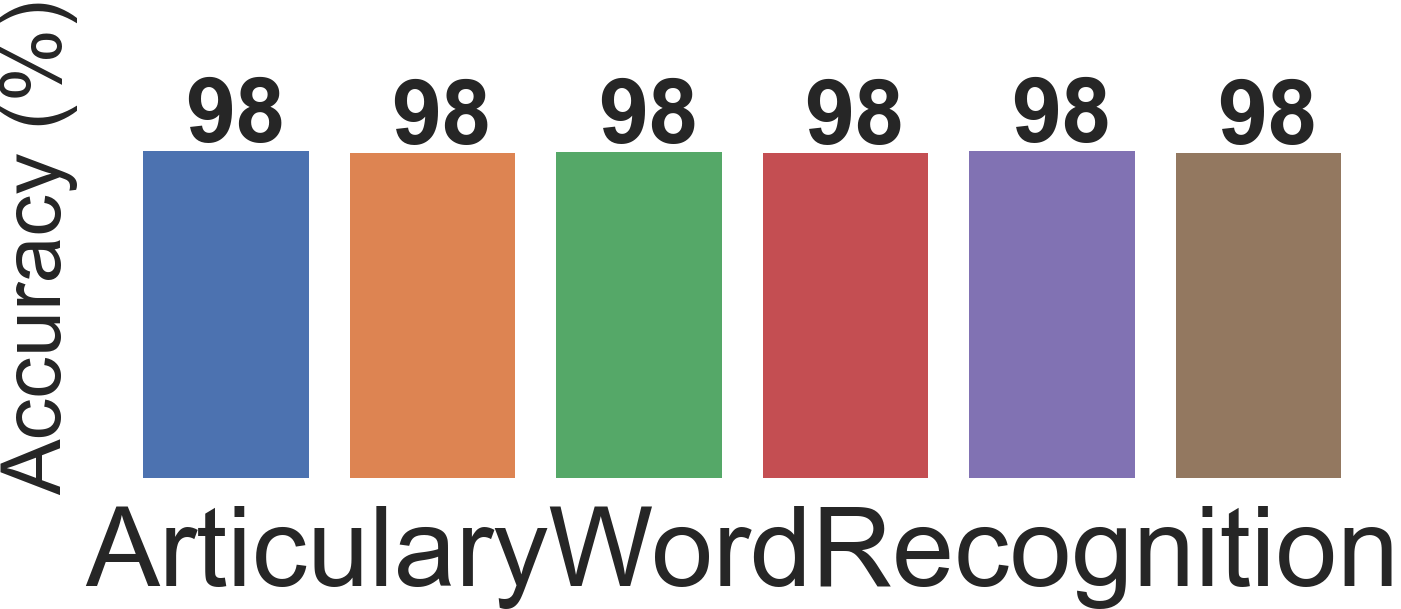}
            \end{minipage}%
        \begin{minipage}{.19\linewidth}
                \centering
                \includegraphics[width=\linewidth]{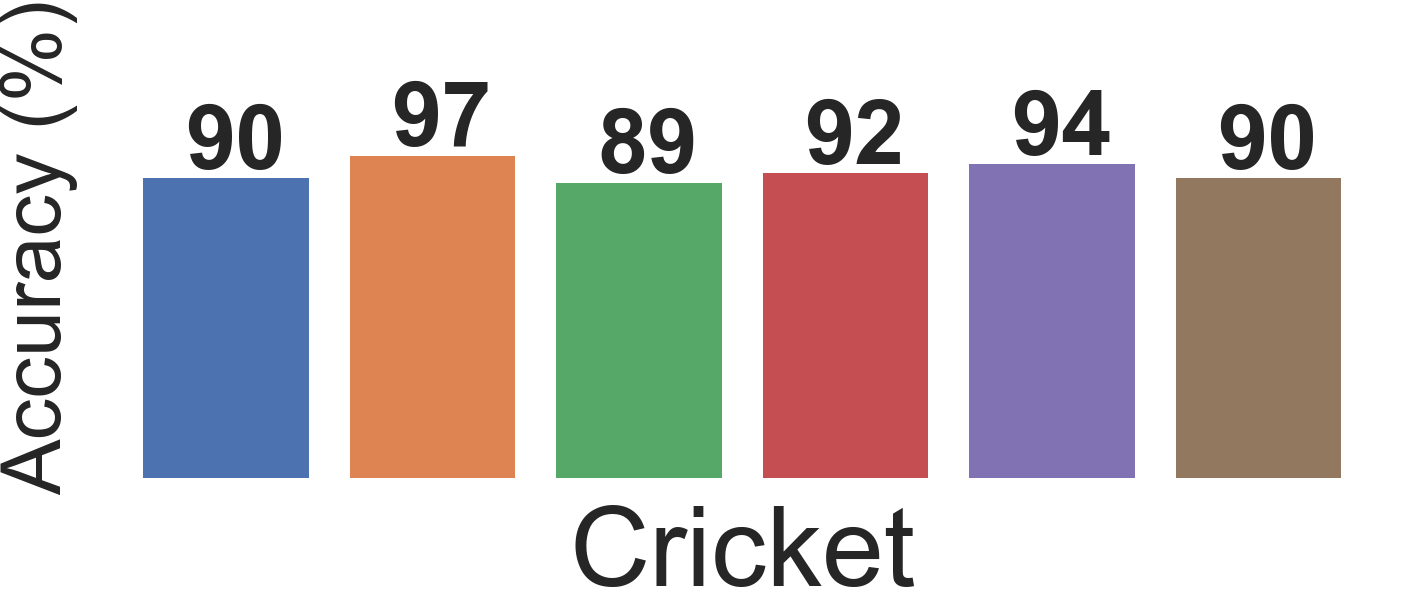}
            \end{minipage}
    \end{minipage}
\caption{Results of adversarial training using baseline attacks and DTW-AR on all the UCR multivariate datasets, and comparison with standard training without adversarial examples (No Attack) to classify clean data.}
\label{fig:appendcleanadvdef}
\vspace{-2ex}
\end{figure*}

\begin{figure*}[!h]
    \centering
        \begin{minipage}{\linewidth}
        \begin{minipage}{.19\linewidth}
                \centering
                \includegraphics[width=\linewidth]{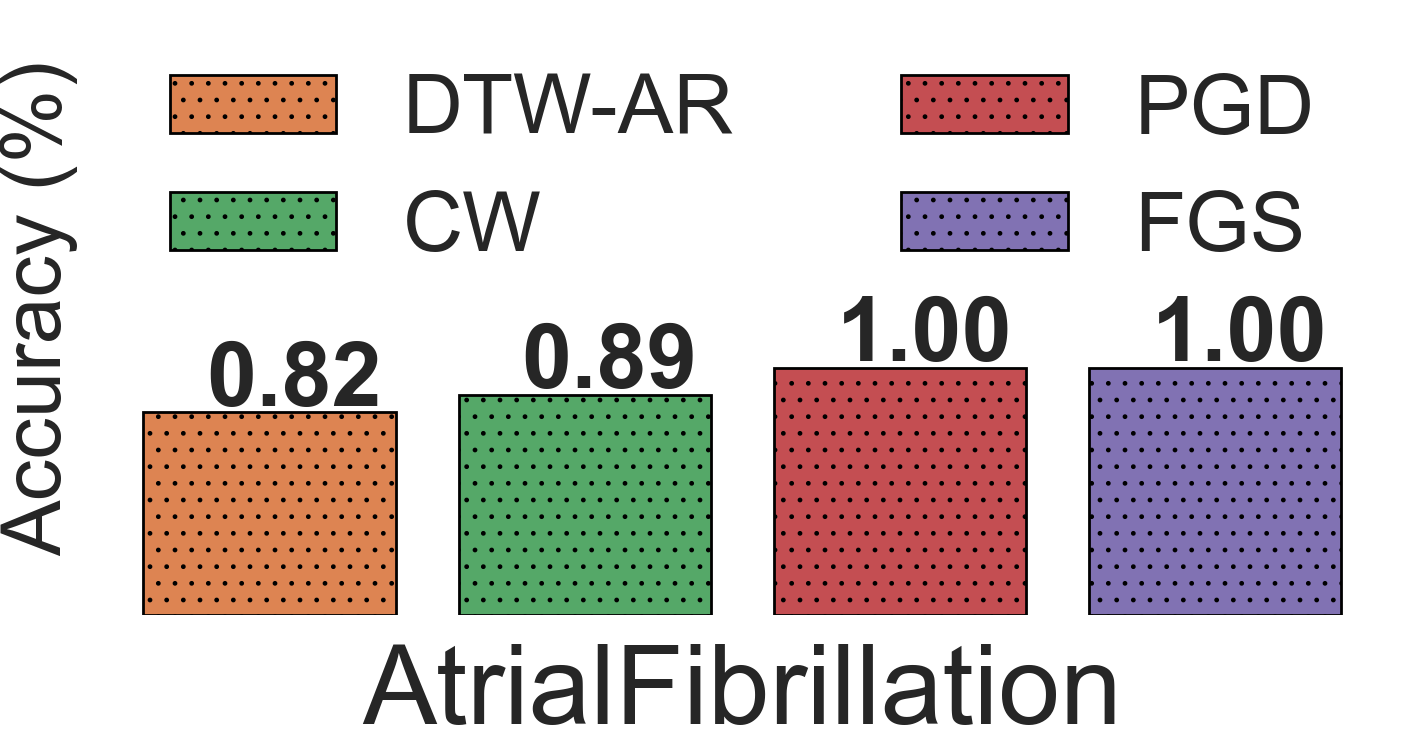}
            \end{minipage}%
        \begin{minipage}{.19\linewidth}
                \centering
                \includegraphics[width=\linewidth]{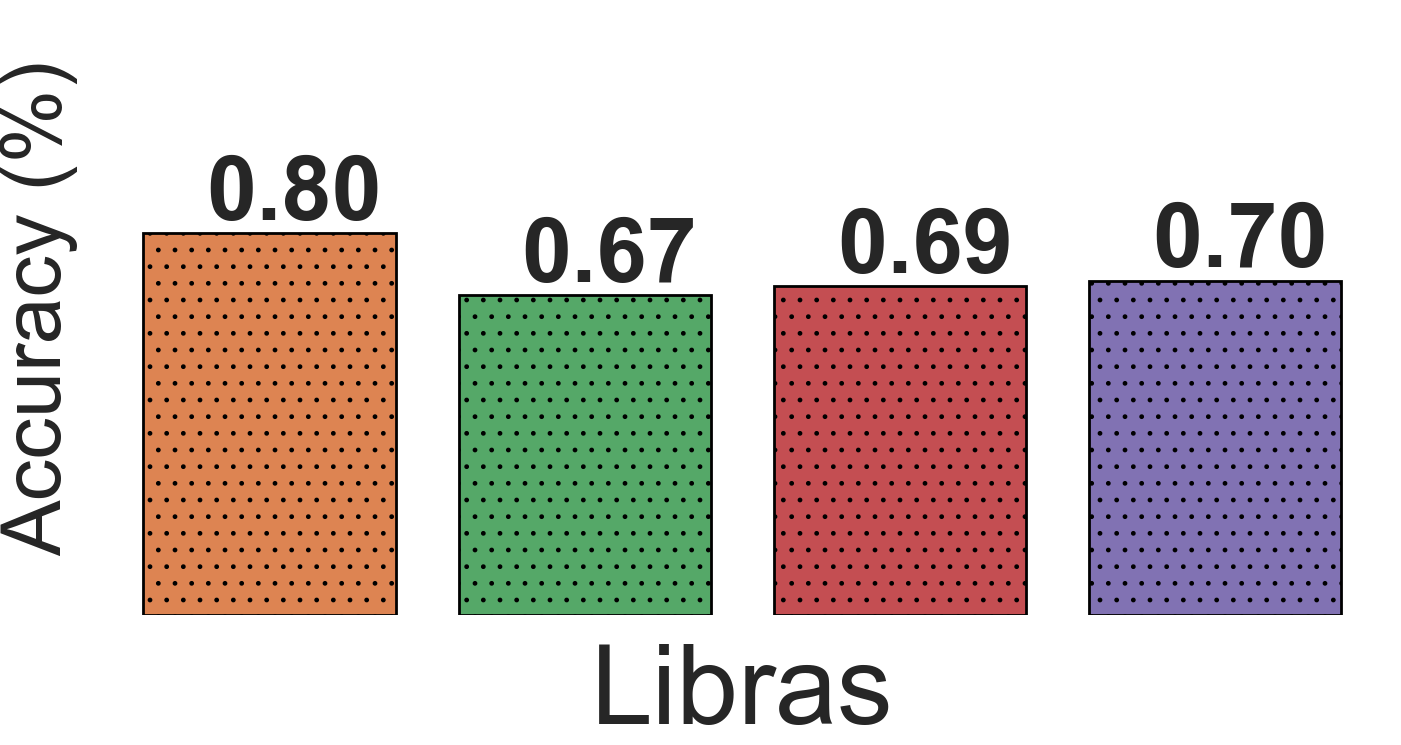}
            \end{minipage}%
        \begin{minipage}{.19\linewidth}
                \centering
                \includegraphics[width=\linewidth]{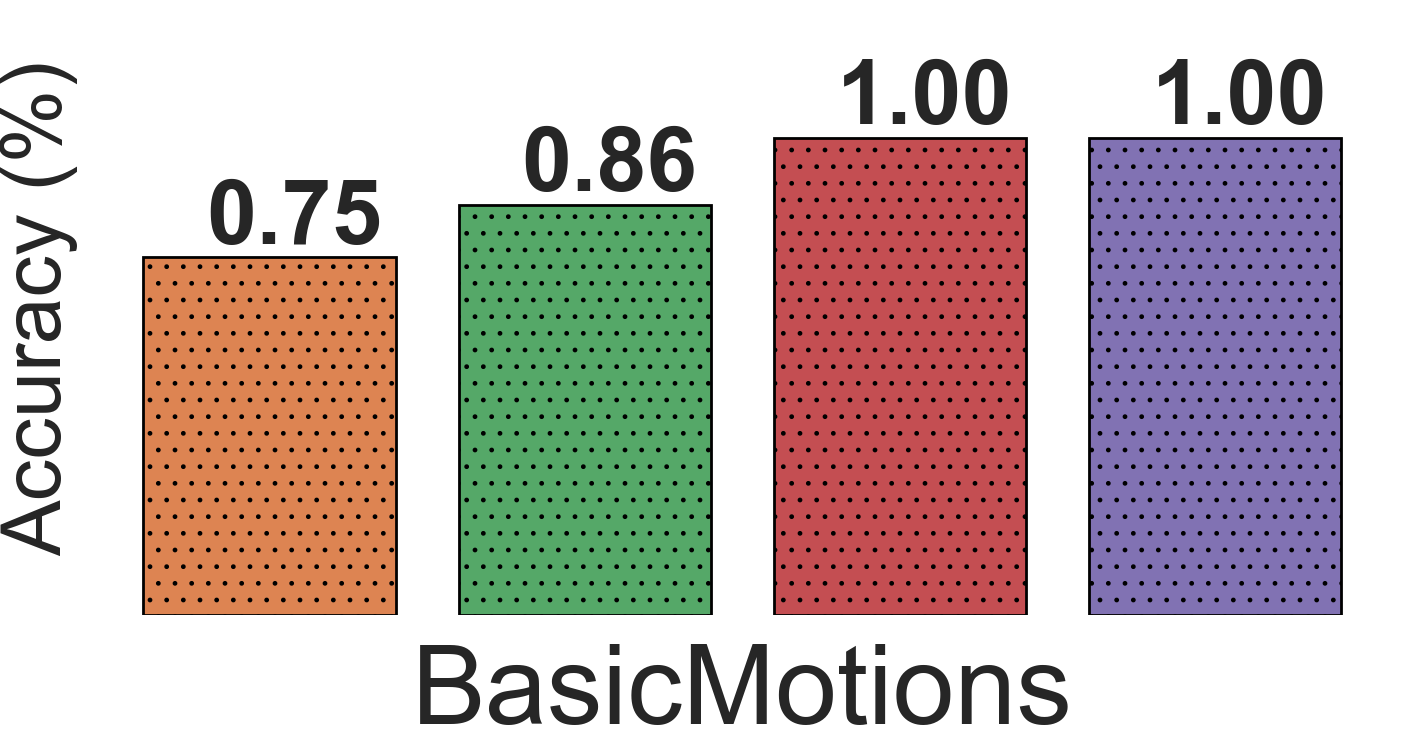}
            \end{minipage}%
        \begin{minipage}{.19\linewidth}
                \centering
                \includegraphics[width=\linewidth]{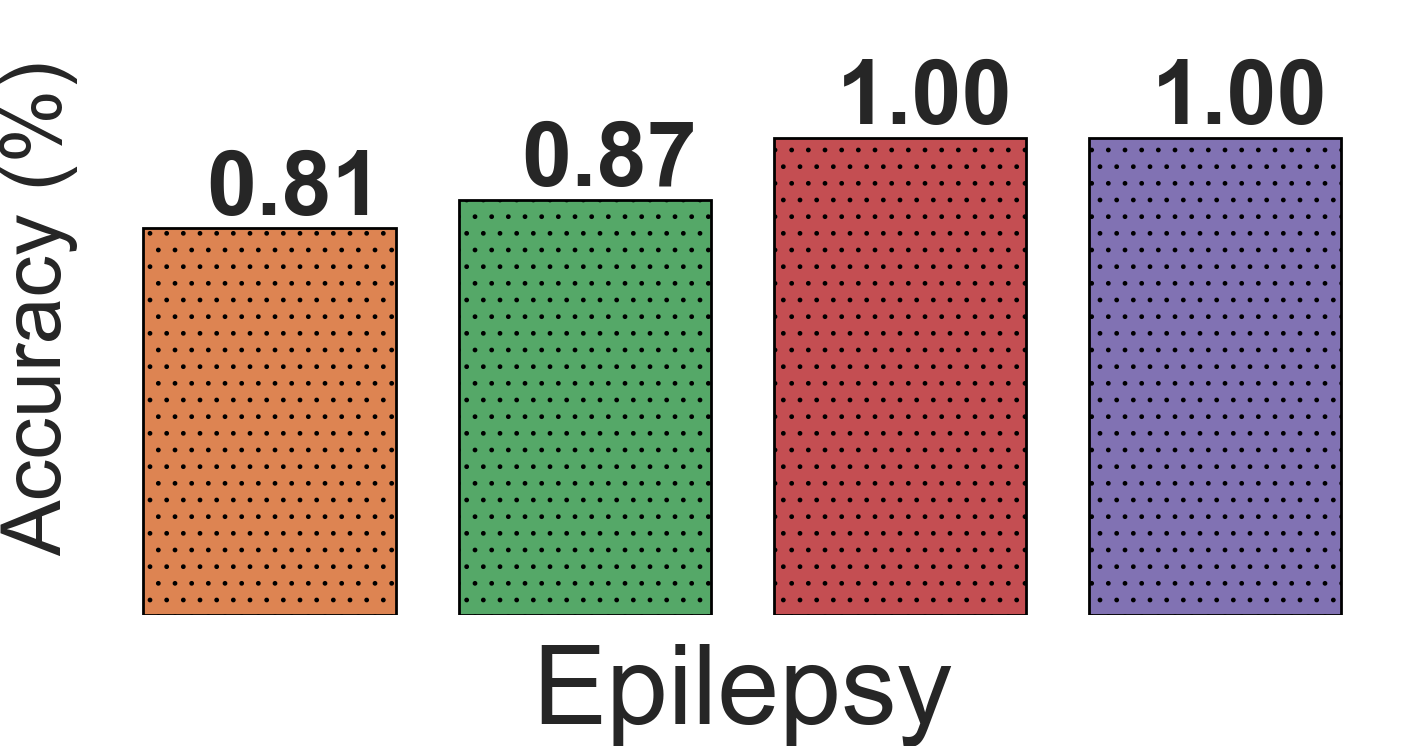}
            \end{minipage}%
        \begin{minipage}{.19\linewidth}
                \centering
                \includegraphics[width=\linewidth]{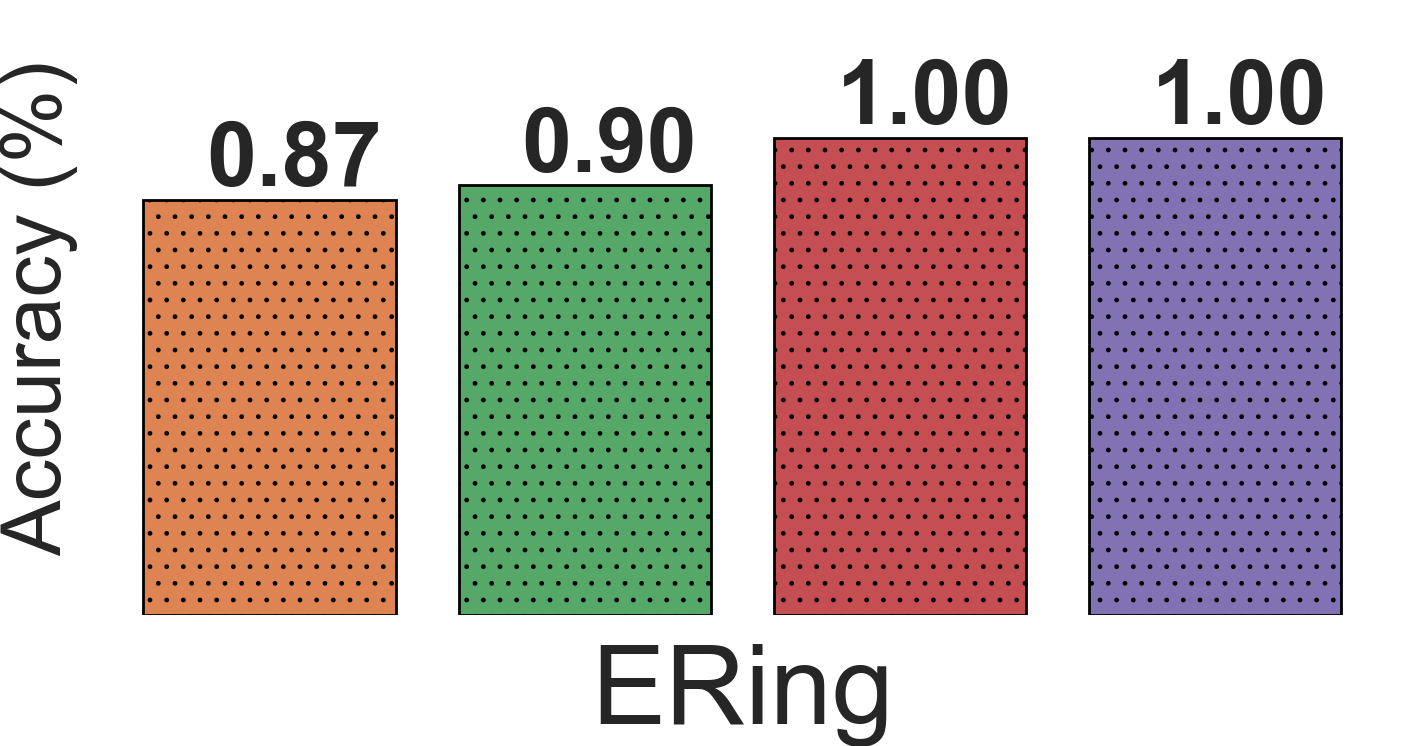}
            \end{minipage}
        \begin{minipage}{.19\linewidth}
                \centering
                \includegraphics[width=\linewidth]{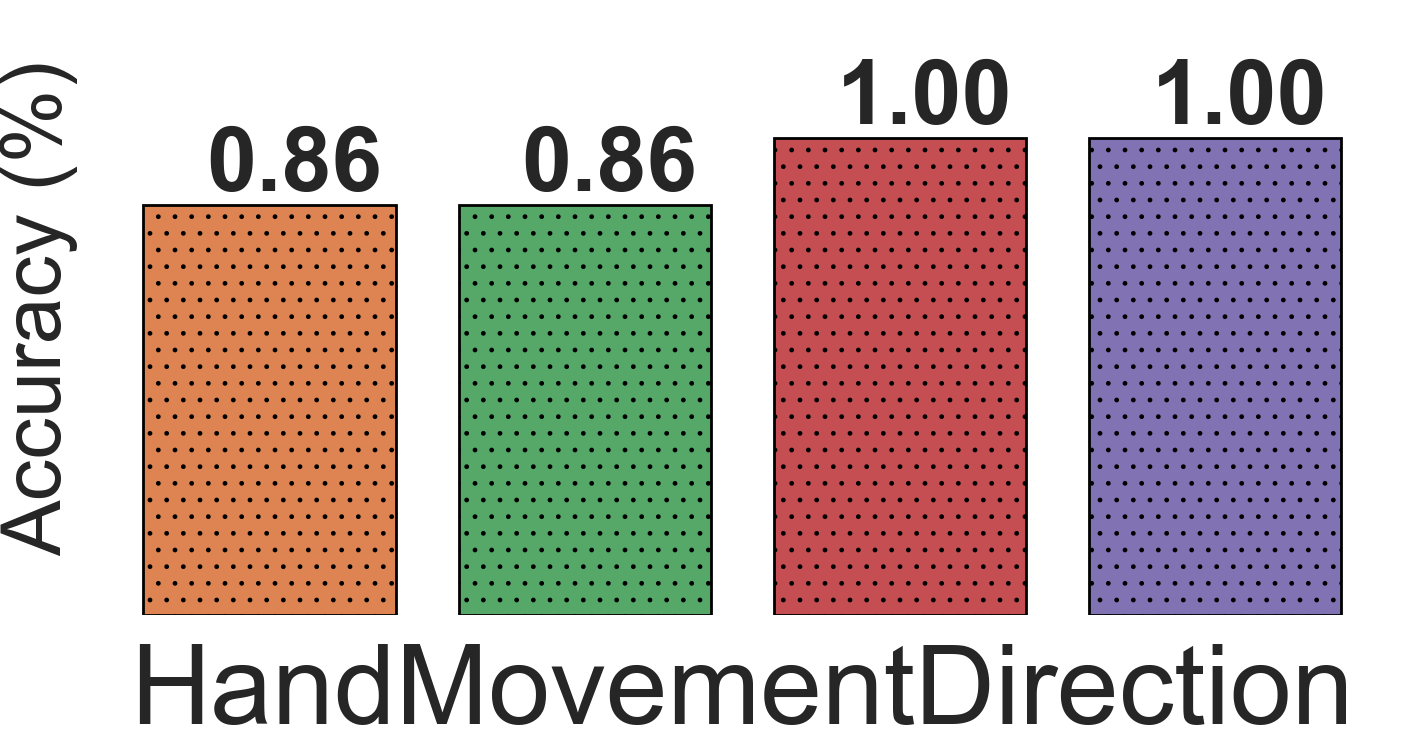}
            \end{minipage}%
        \begin{minipage}{.19\linewidth}
                \centering
                \includegraphics[width=\linewidth]{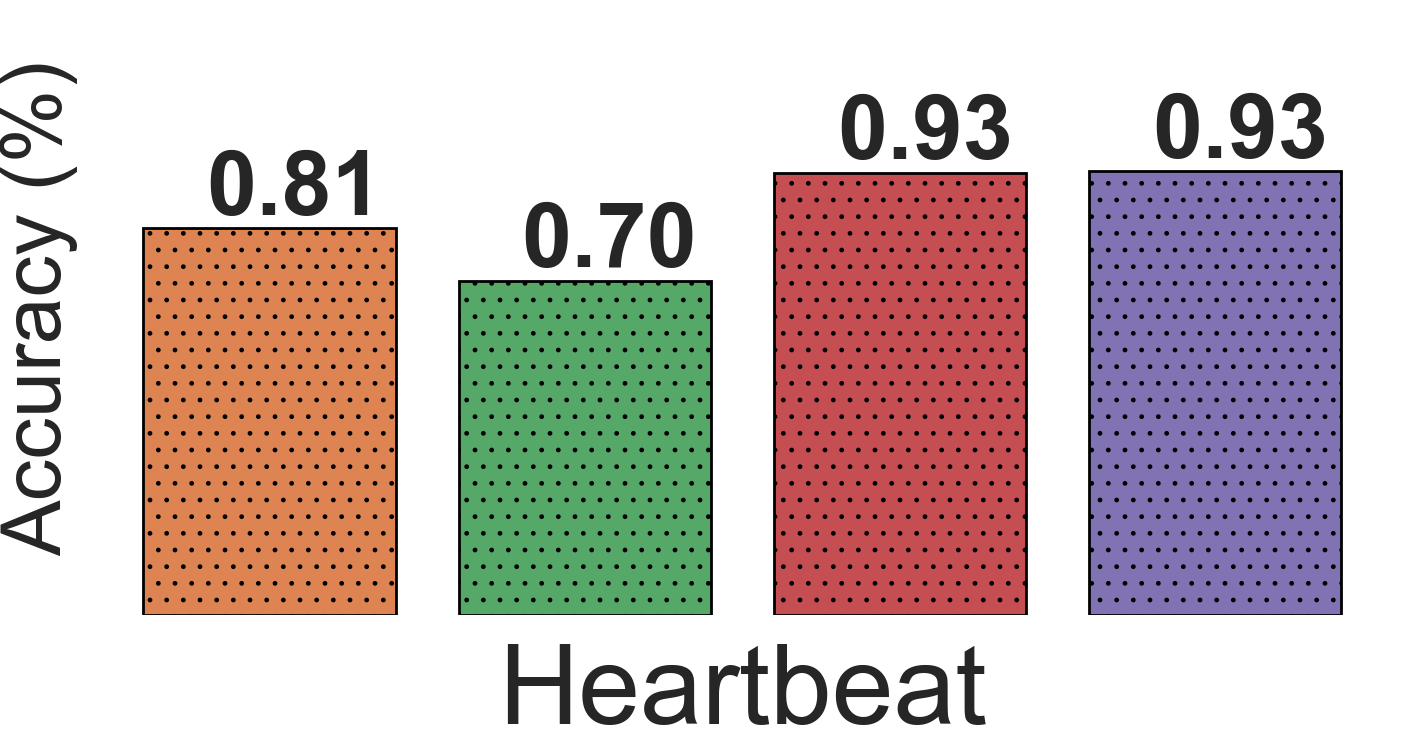}
            \end{minipage}%
        \begin{minipage}{.19\linewidth}
                \centering
                \includegraphics[width=\linewidth]{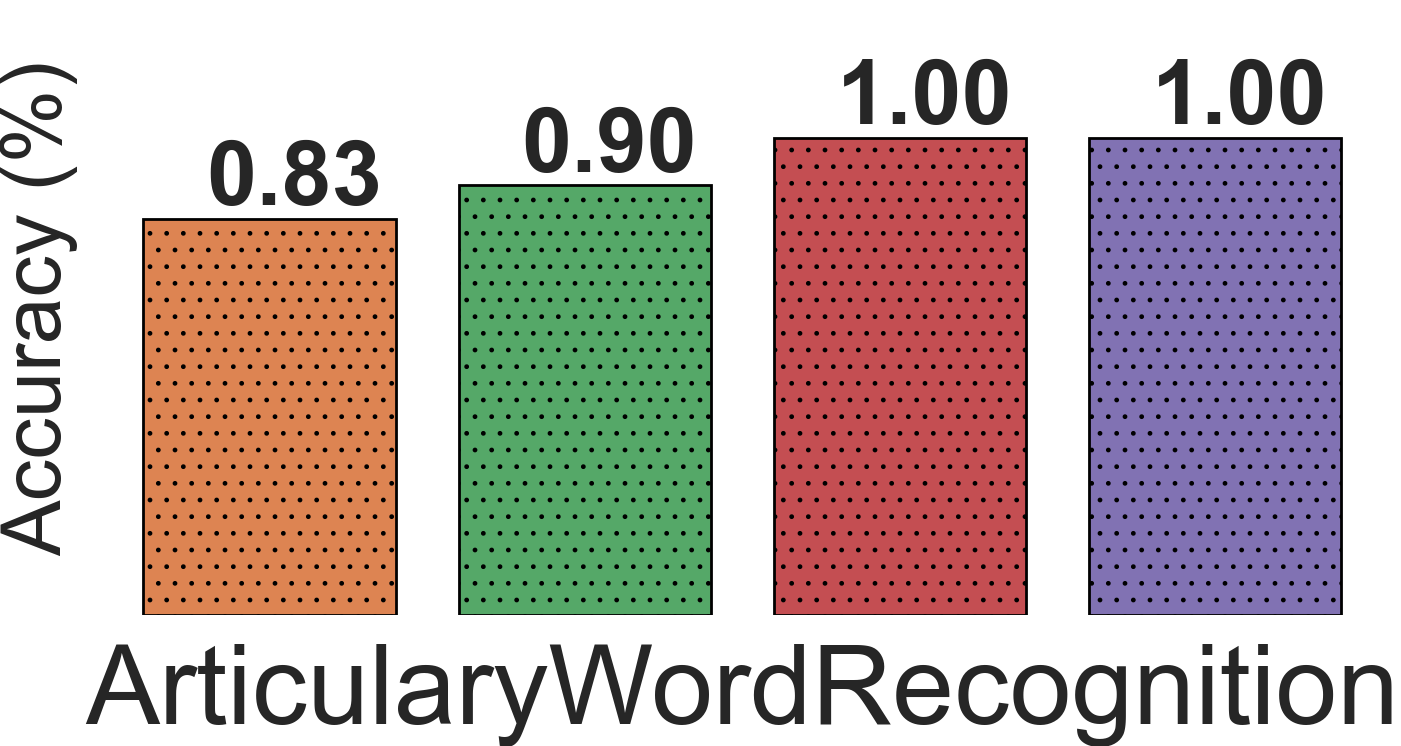}
            \end{minipage}%
        \begin{minipage}{.19\linewidth}
                \centering
                \includegraphics[width=\linewidth]{AppendPerf/ApAugPeroformanceBasicMotionsagainstAtk.png}
            \end{minipage}%
        \begin{minipage}{.19\linewidth}
                \centering
                \includegraphics[width=\linewidth]{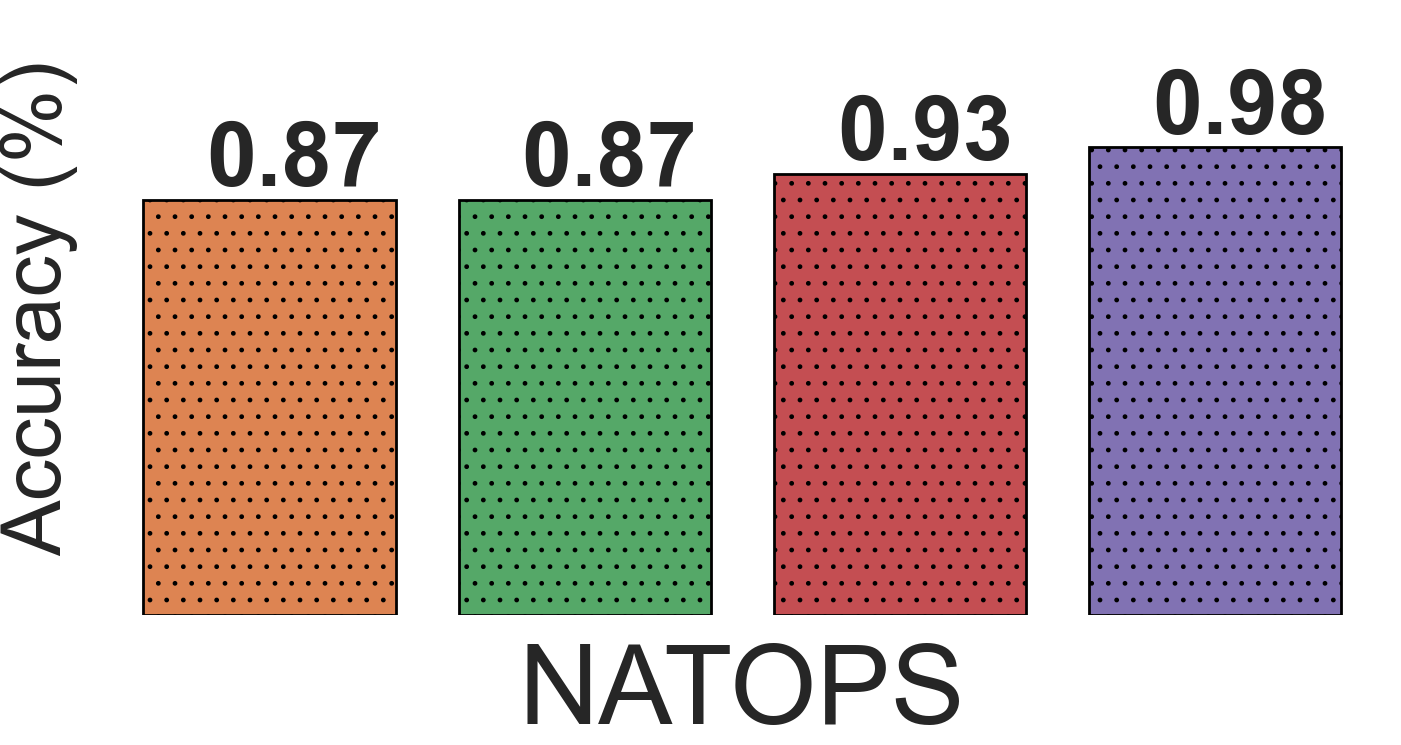}
            \end{minipage}
        \begin{minipage}{.19\linewidth}
                \centering
                \includegraphics[width=\linewidth]{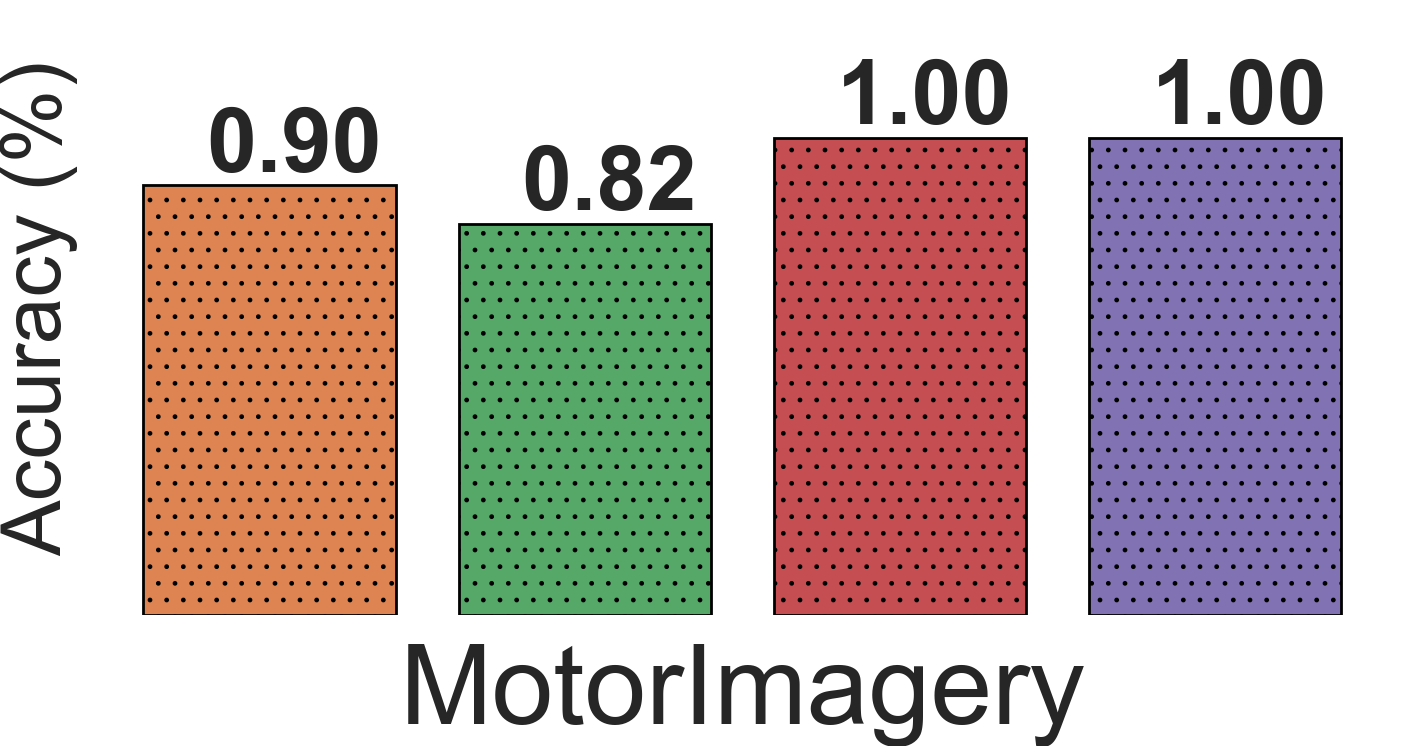}
            \end{minipage}%
        \begin{minipage}{.19\linewidth}
                \centering
                \includegraphics[width=\linewidth]{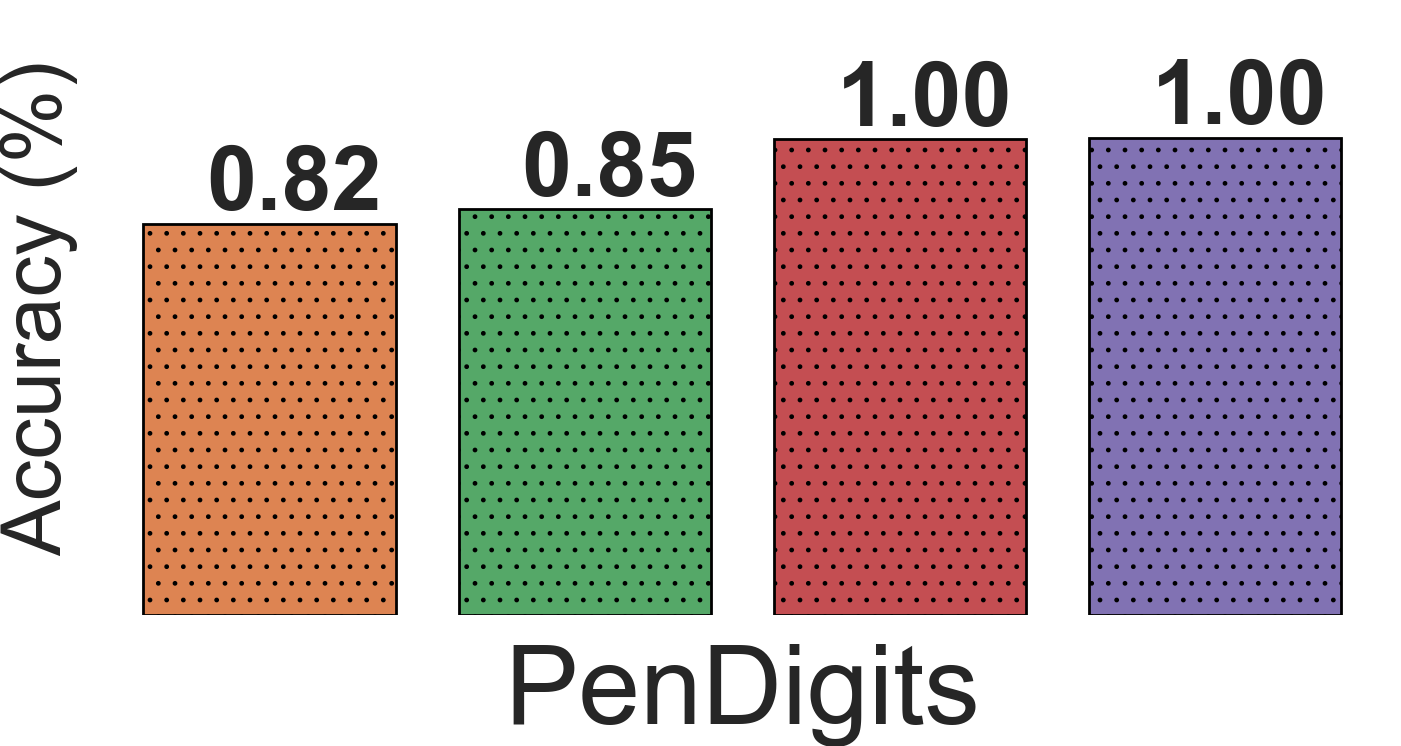}
            \end{minipage}%
        \begin{minipage}{.19\linewidth}
                \centering
                \includegraphics[width=\linewidth]{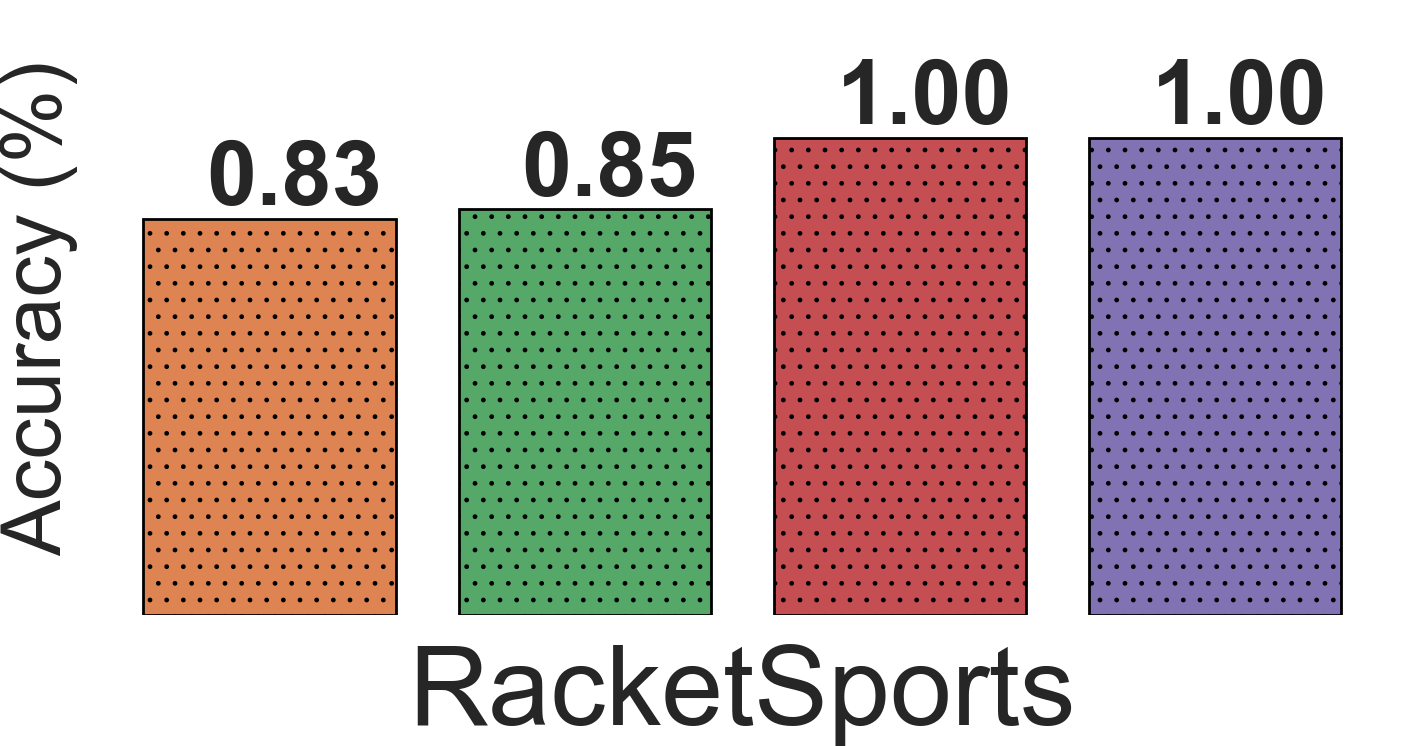}
            \end{minipage}%
        \begin{minipage}{.19\linewidth}
                \centering
                \includegraphics[width=\linewidth]{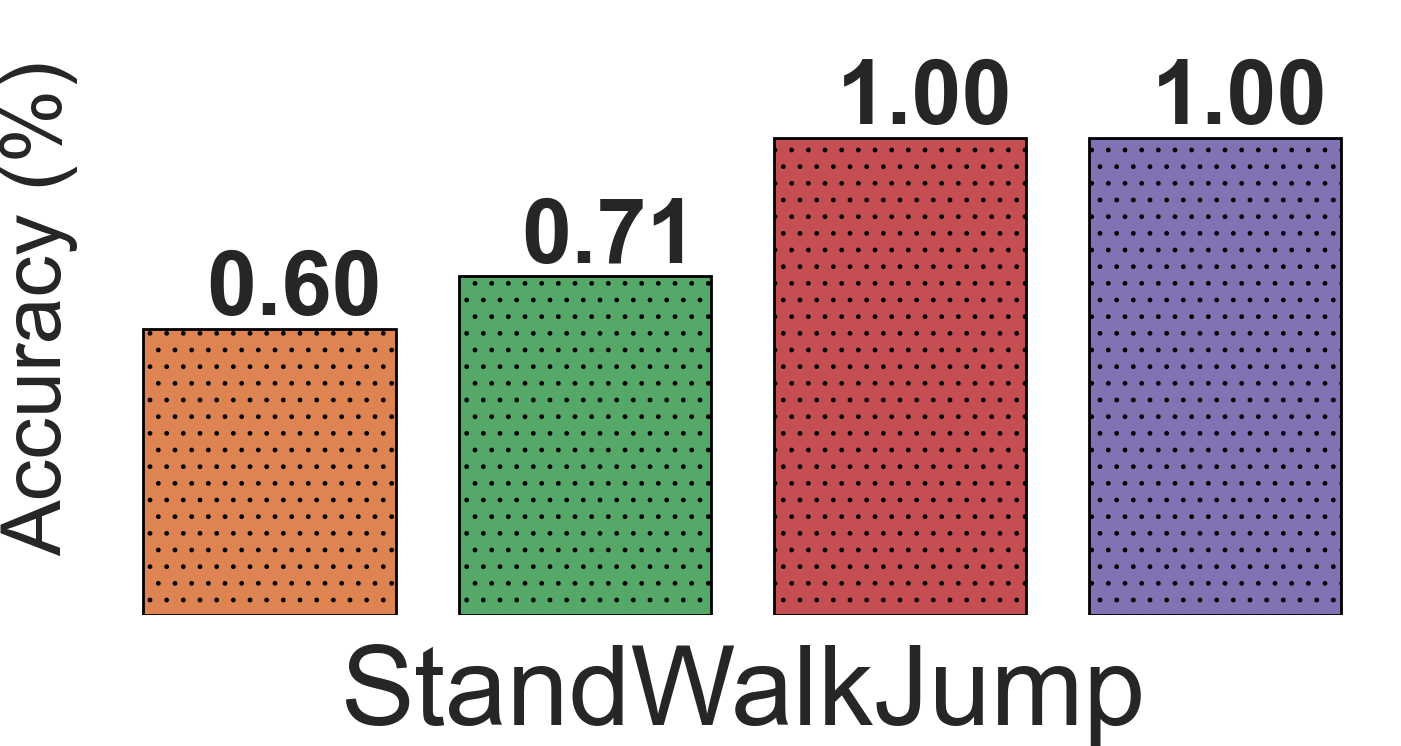}
            \end{minipage}%
        \begin{minipage}{.19\linewidth}
                \centering
                \includegraphics[width=\linewidth]{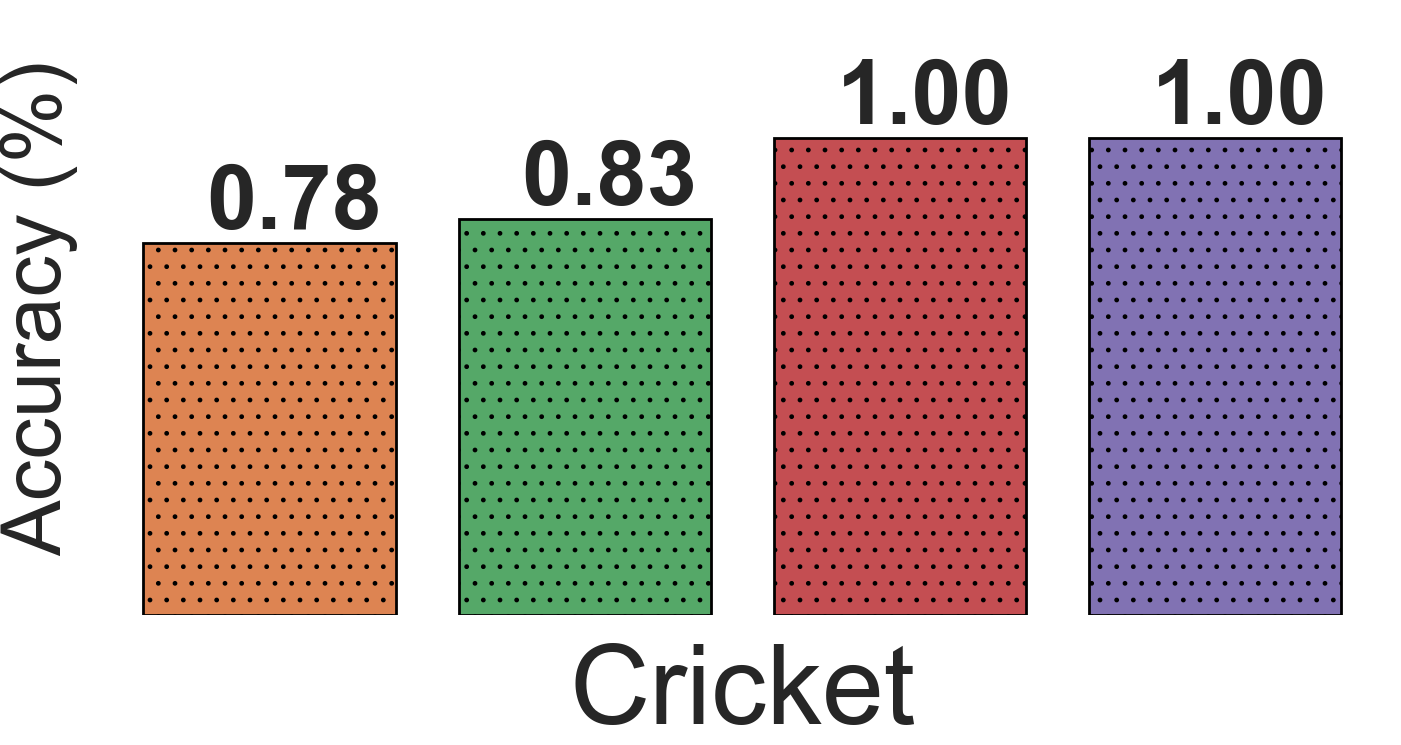}
            \end{minipage}
    \end{minipage}
    \vspace{-2ex}
\caption{Results of DTW-AR based adversarial training to predict the true labels of adversarial examples generated by DTW-AR and the baseline attack methods on all the UCR multivariate datasets. The adversarial examples considered are those that successfully fooled DNNs that do not use adversarial training.}
\label{fig:appendadvdef}
\end{figure*}

\noindent \textbf{Results on the full UCR multivariate dataset.} First, we provide in Figures \ref{fig:appendadvatk}, \ref{fig:appendcleanadvdef} and \ref{fig:appendadvdef}  the experiments conducted in the main paper on the \textbf{Effectiveness of adversarial attacks} and the \textbf{DTW-AR based adversarial training} on all the UCR multivariate dataset to our DTW-AR framework is general and highly-effective for all datasets.

\begin{figure*}[!h]
    \centering
        \begin{minipage}{\linewidth}
        \begin{minipage}{.19\linewidth}
                \centering
                \includegraphics[width=\linewidth]{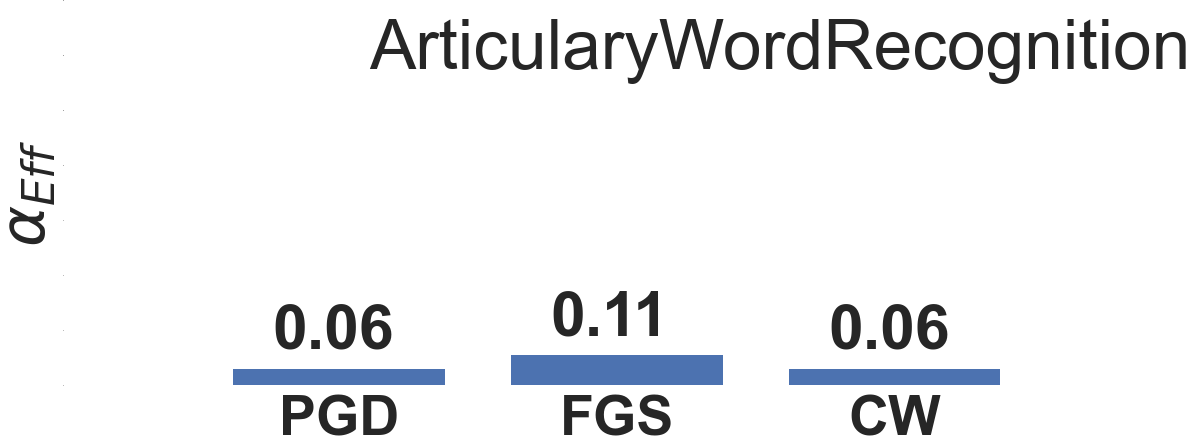}
            \end{minipage}%
        \begin{minipage}{.19\linewidth}
                \centering
                \includegraphics[width=\linewidth]{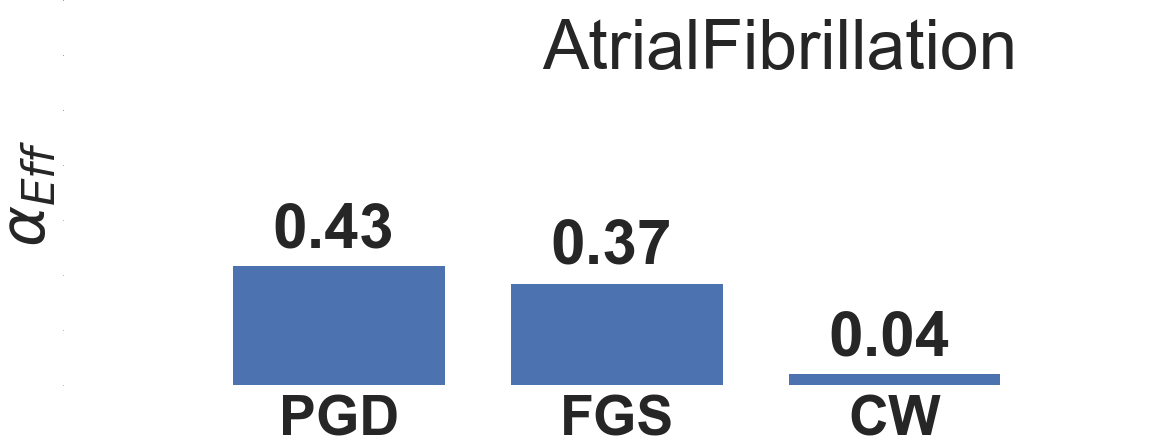}
            \end{minipage}%
        \begin{minipage}{.19\linewidth}
                \centering
                \includegraphics[width=\linewidth]{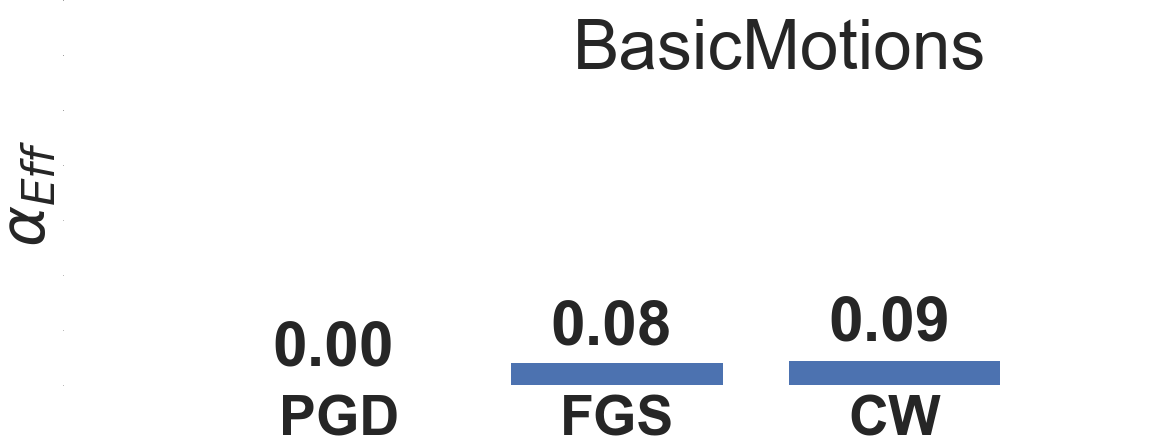}
            \end{minipage}%
        \begin{minipage}{.19\linewidth}
                \centering
                \includegraphics[width=\linewidth]{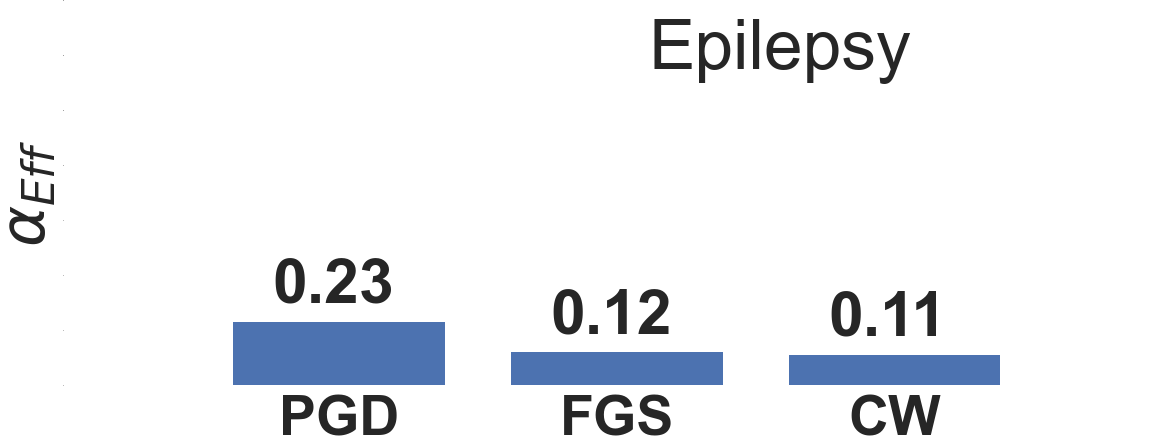}
            \end{minipage}%
        \begin{minipage}{.19\linewidth}
                \centering
                \includegraphics[width=\linewidth]{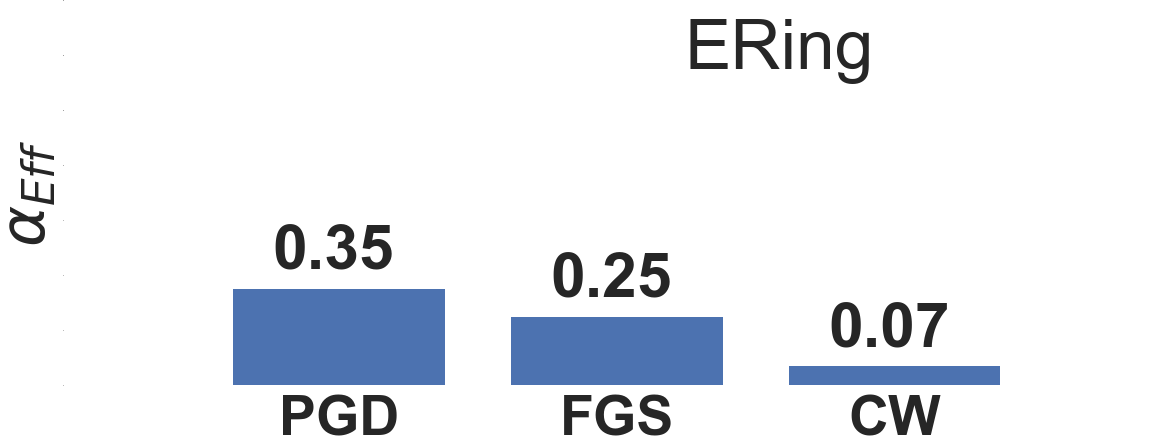}
            \end{minipage}
        \begin{minipage}{.19\linewidth}
                \centering
                \includegraphics[width=\linewidth]{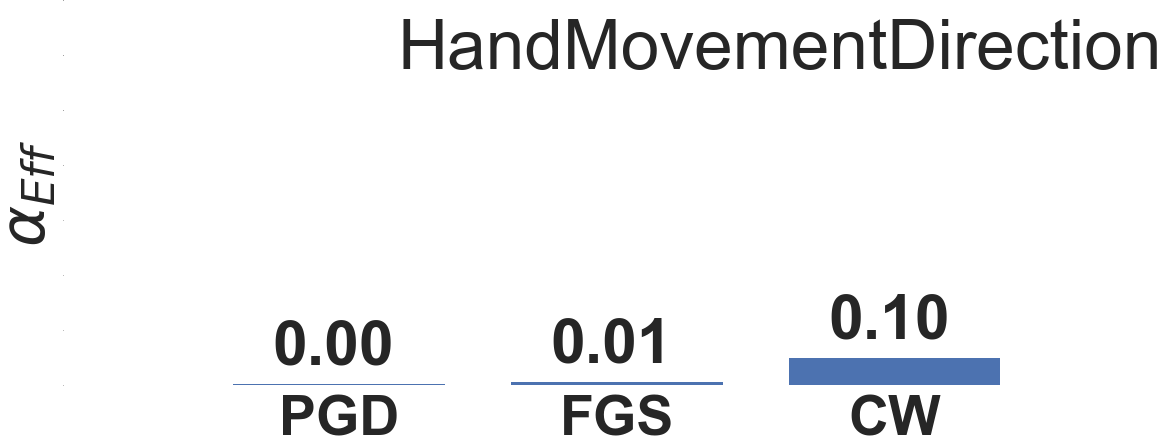}
            \end{minipage}%
        \begin{minipage}{.19\linewidth}
                \centering
                \includegraphics[width=\linewidth]{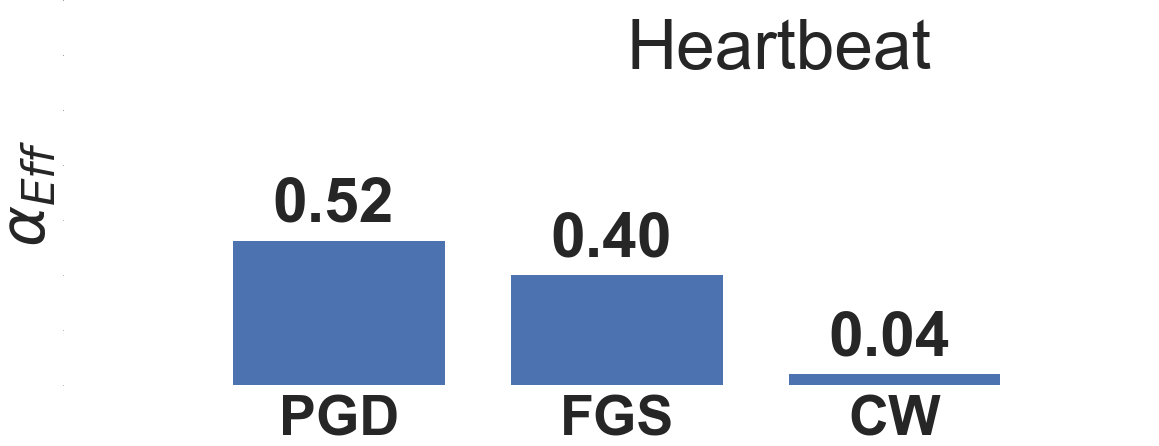}
            \end{minipage}%
        \begin{minipage}{.19\linewidth}
                \centering
                \includegraphics[width=\linewidth]{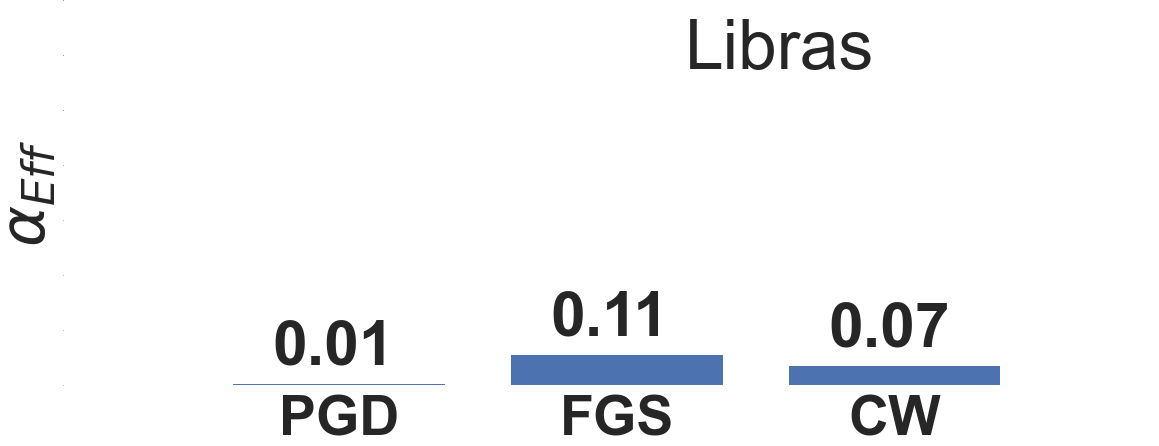}
            \end{minipage}%
        \begin{minipage}{.19\linewidth}
                \centering
                \includegraphics[width=\linewidth]{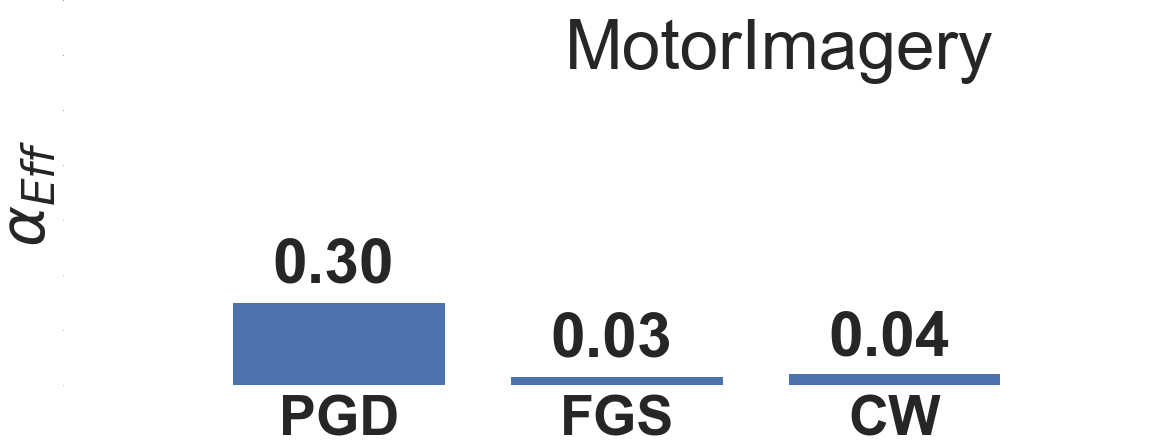}
            \end{minipage}%
        \begin{minipage}{.19\linewidth}
                \centering
                \includegraphics[width=\linewidth]{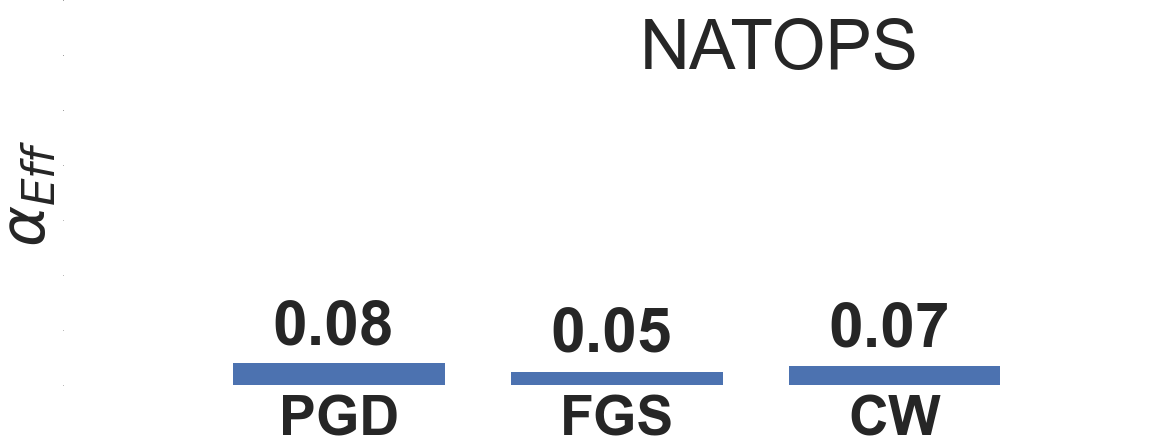}
            \end{minipage}
        \begin{minipage}{.19\linewidth}
                \centering
                \includegraphics[width=\linewidth]{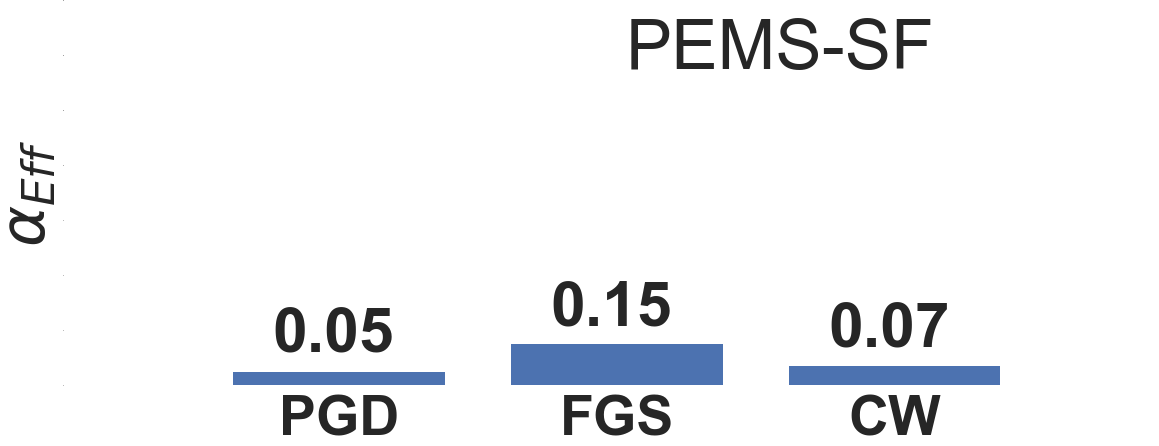}
            \end{minipage}%
        \begin{minipage}{.19\linewidth}
                \centering
                \includegraphics[width=\linewidth]{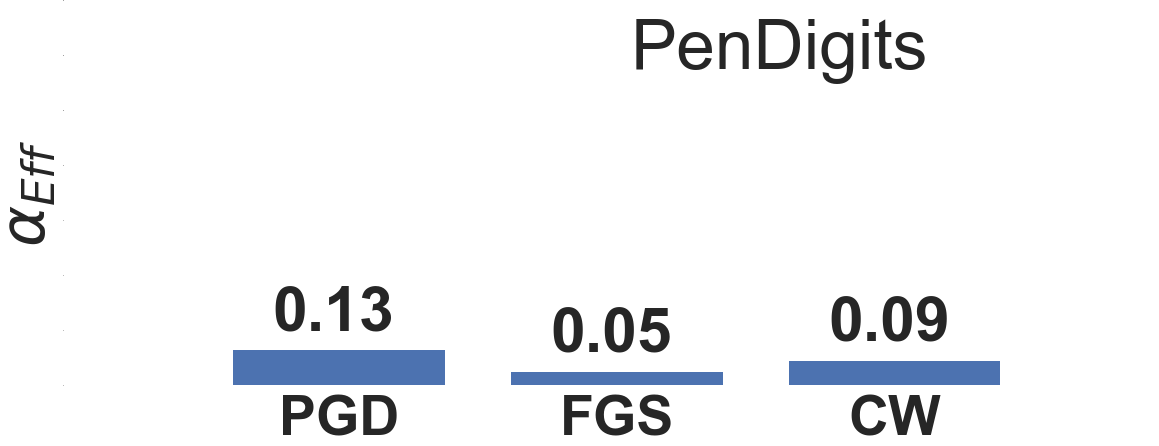}
            \end{minipage}%
        \begin{minipage}{.19\linewidth}
                \centering
                \includegraphics[width=\linewidth]{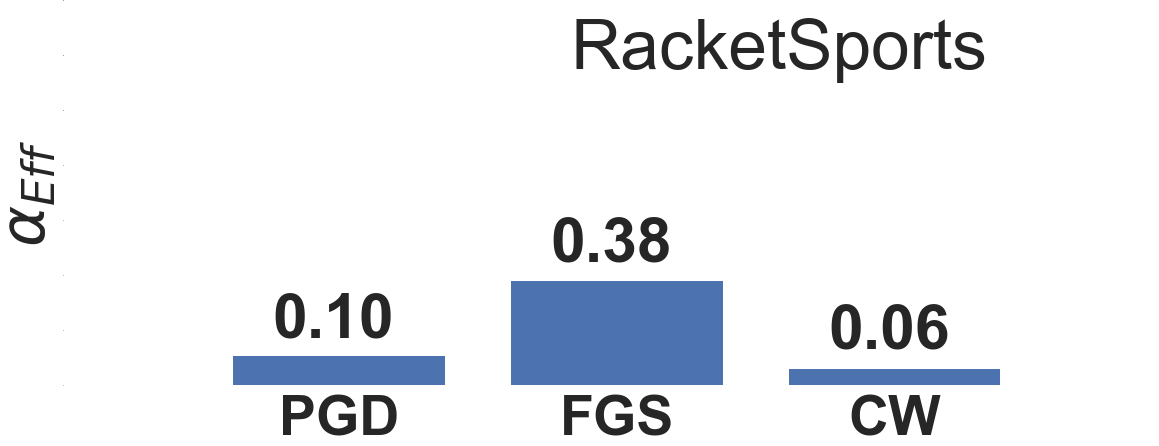}
            \end{minipage}%
        \begin{minipage}{.19\linewidth}
                \centering
                \includegraphics[width=\linewidth]{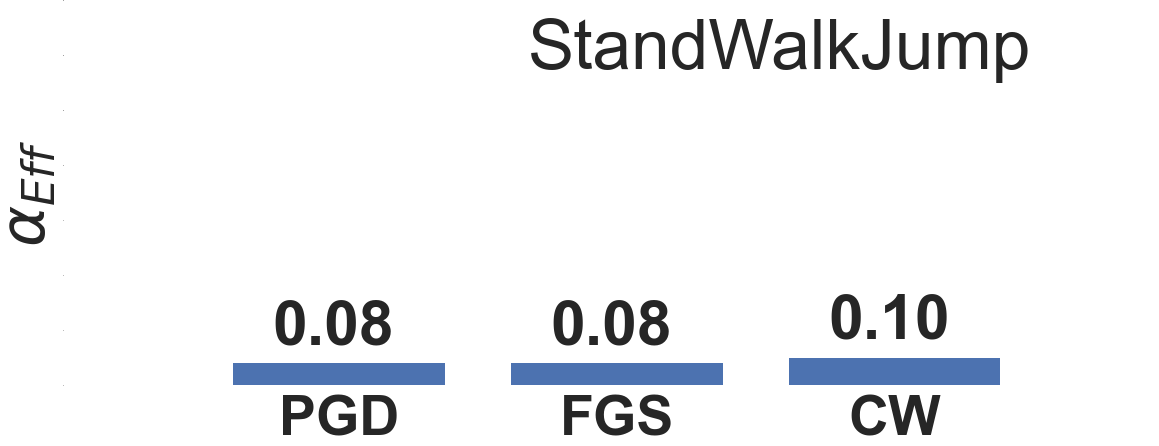}
            \end{minipage}%
        \begin{minipage}{.19\linewidth}
                \centering
                \includegraphics[width=\linewidth]{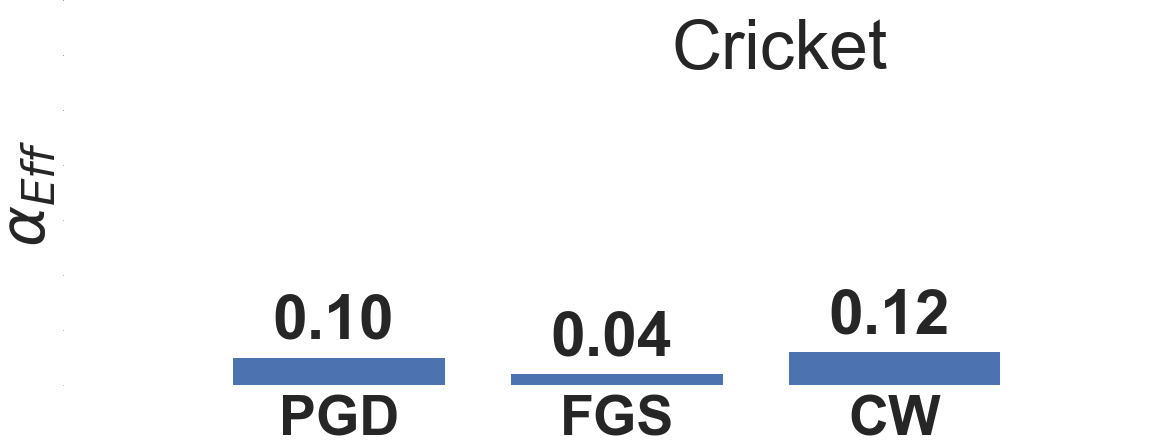}
            \end{minipage}
    \end{minipage}
\caption{Results for the effectiveness of adversarial examples from CW on different deep models using adversarial training baselines (PGD, FGS, CW).}
\label{fig:cwadvatk}
\end{figure*}

\begin{figure*}[!h]
    \centering
        \begin{minipage}{\linewidth}
        \begin{minipage}{.19\linewidth}
                \centering
                \includegraphics[width=\linewidth]{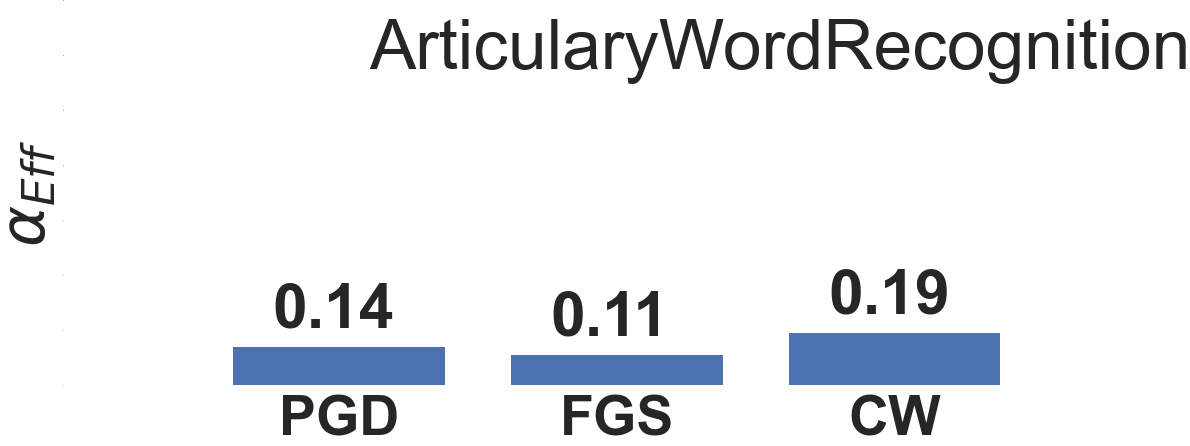}
            \end{minipage}%
        \begin{minipage}{.19\linewidth}
                \centering
                \includegraphics[width=\linewidth]{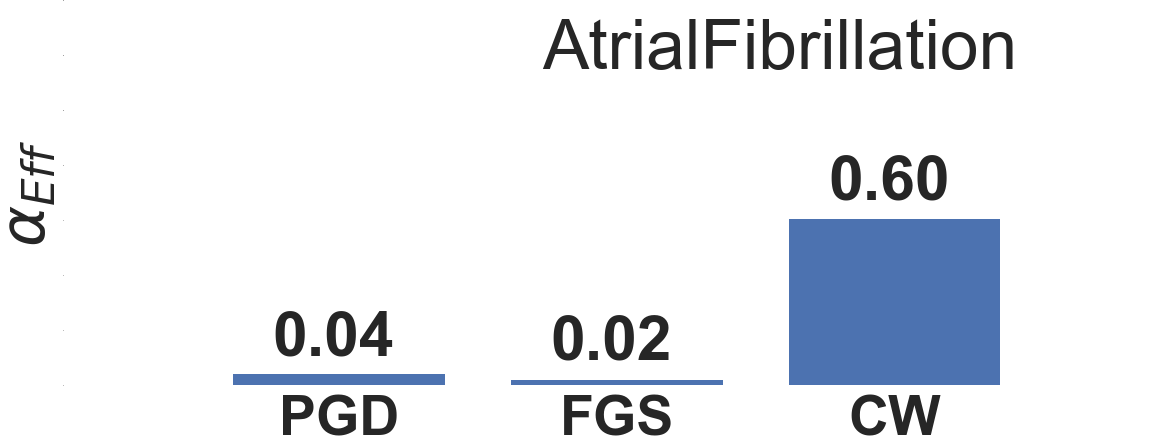}
            \end{minipage}%
        \begin{minipage}{.19\linewidth}
                \centering
                \includegraphics[width=\linewidth]{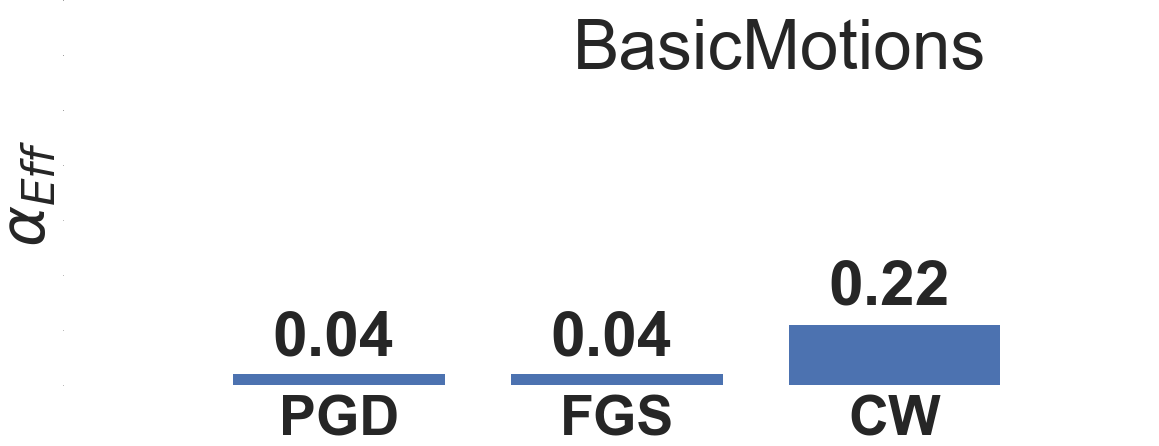}
            \end{minipage}%
        \begin{minipage}{.19\linewidth}
                \centering
                \includegraphics[width=\linewidth]{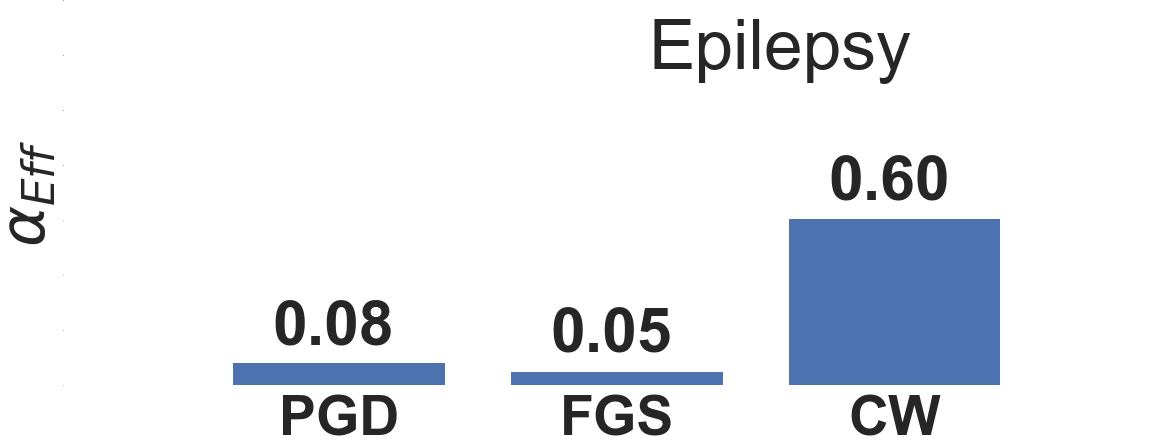}
            \end{minipage}%
        \begin{minipage}{.19\linewidth}
                \centering
                \includegraphics[width=\linewidth]{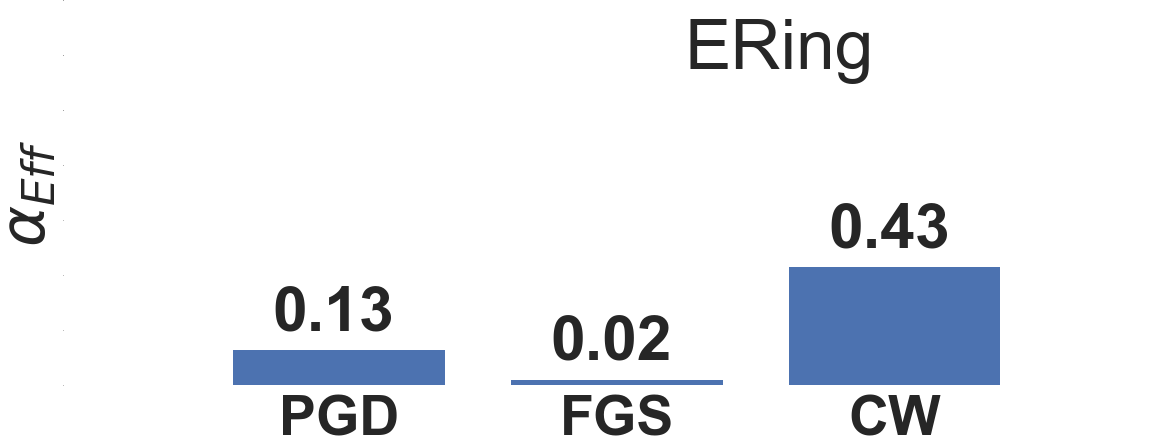}
            \end{minipage}
        \begin{minipage}{.19\linewidth}
                \centering
                \includegraphics[width=\linewidth]{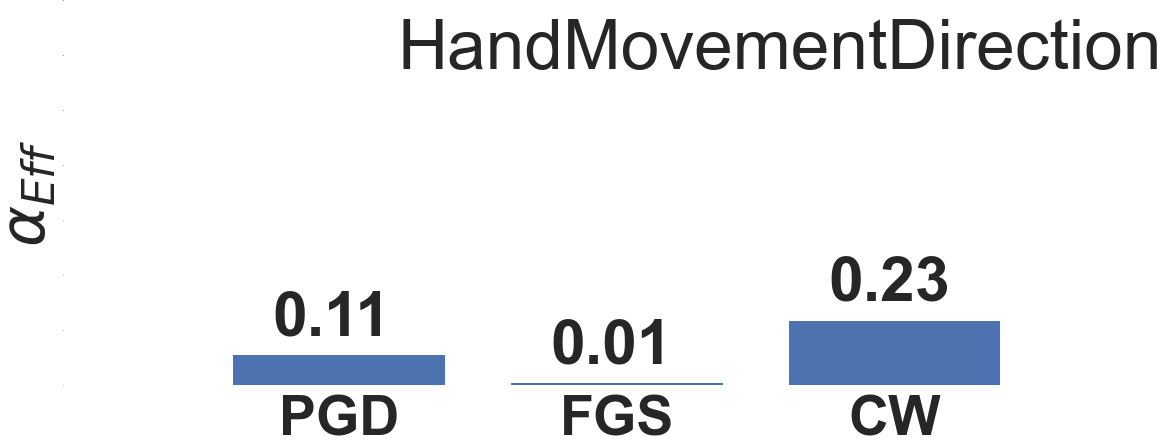}
            \end{minipage}%
        \begin{minipage}{.19\linewidth}
                \centering
                \includegraphics[width=\linewidth]{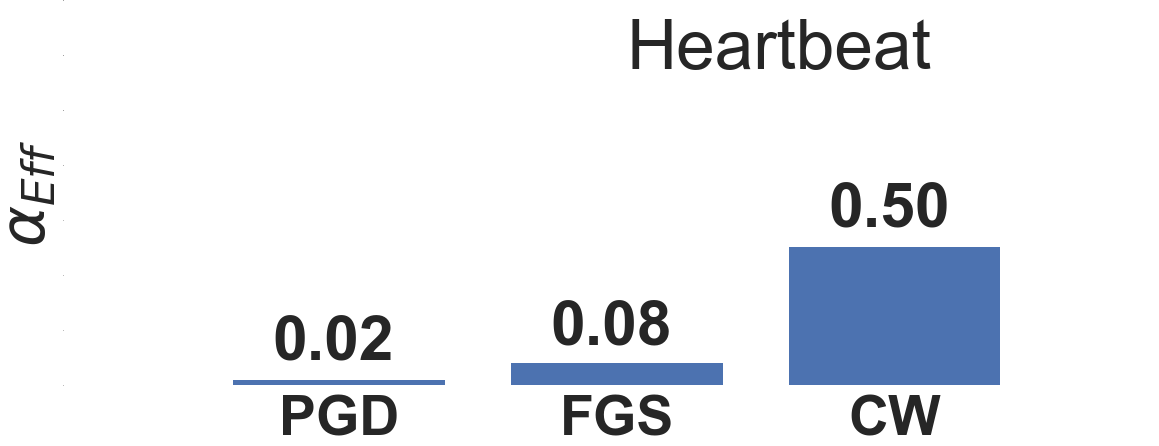}
            \end{minipage}%
        \begin{minipage}{.19\linewidth}
                \centering
                \includegraphics[width=\linewidth]{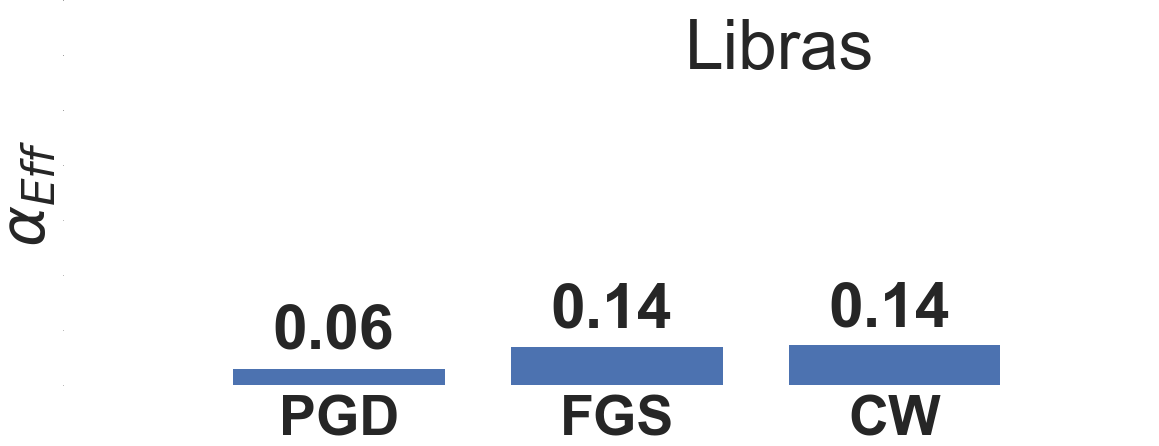}
            \end{minipage}%
        \begin{minipage}{.19\linewidth}
                \centering
                \includegraphics[width=\linewidth]{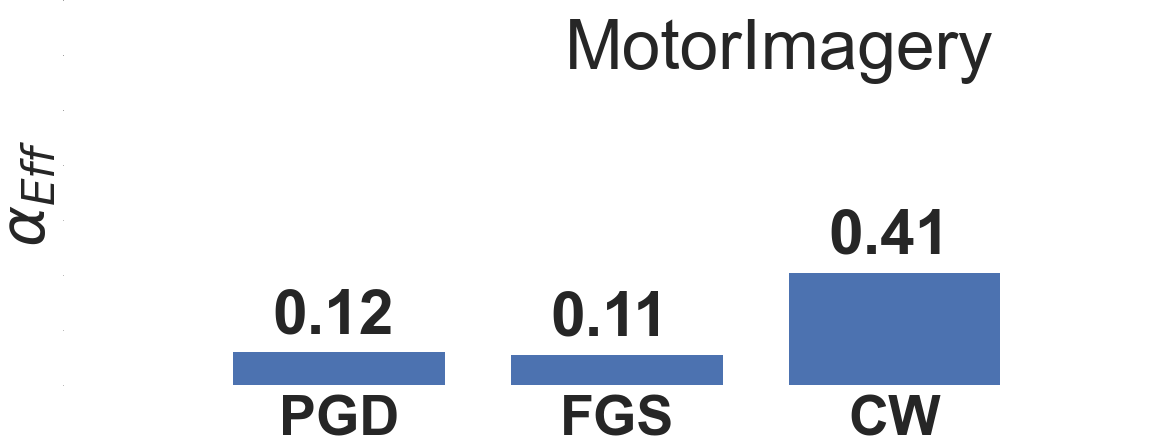}
            \end{minipage}%
        \begin{minipage}{.19\linewidth}
                \centering
                \includegraphics[width=\linewidth]{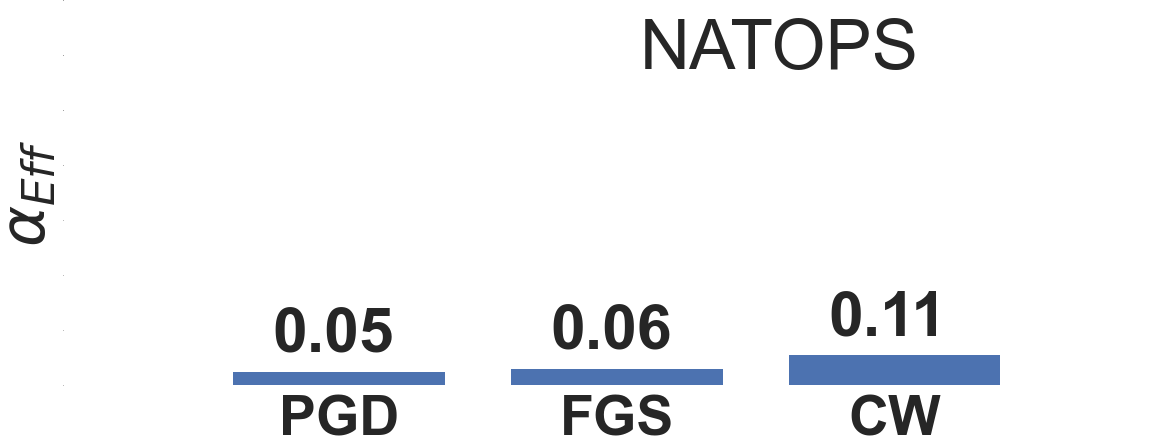}
            \end{minipage}
        \begin{minipage}{.19\linewidth}
                \centering
                \includegraphics[width=\linewidth]{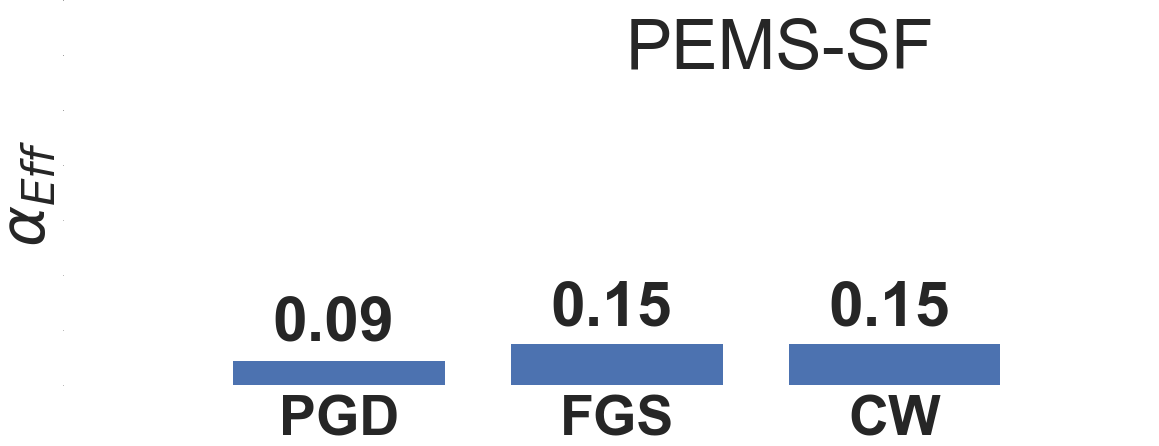}
            \end{minipage}%
        \begin{minipage}{.19\linewidth}
                \centering
                \includegraphics[width=\linewidth]{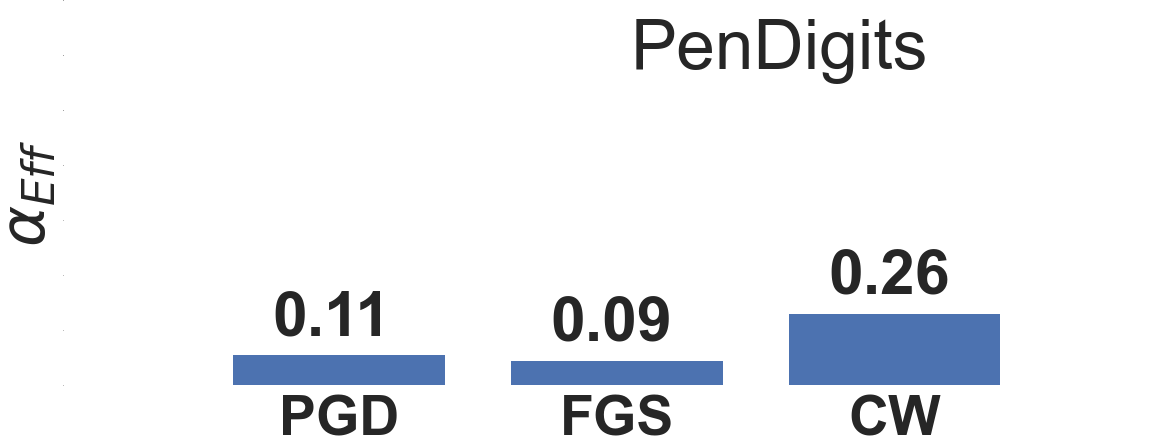}
            \end{minipage}%
        \begin{minipage}{.19\linewidth}
                \centering
                \includegraphics[width=\linewidth]{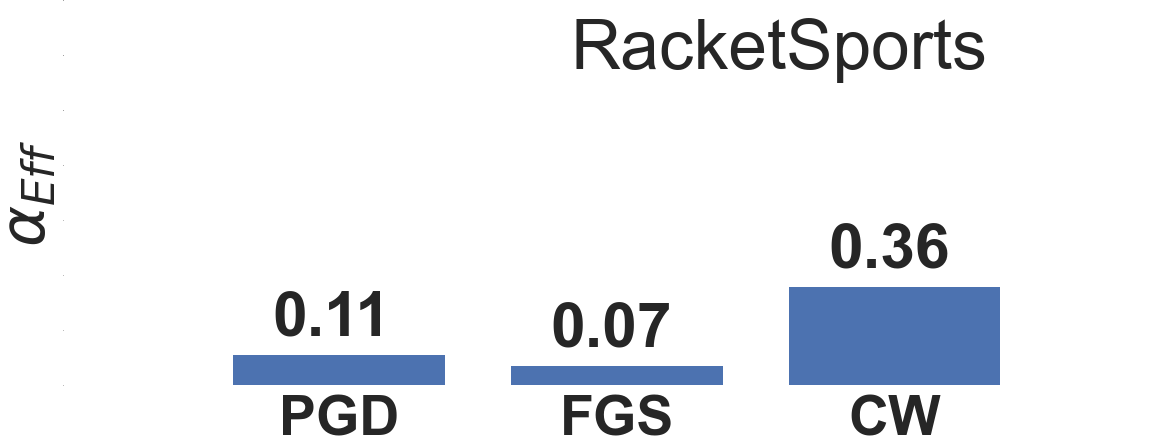}
            \end{minipage}%
        \begin{minipage}{.19\linewidth}
                \centering
                \includegraphics[width=\linewidth]{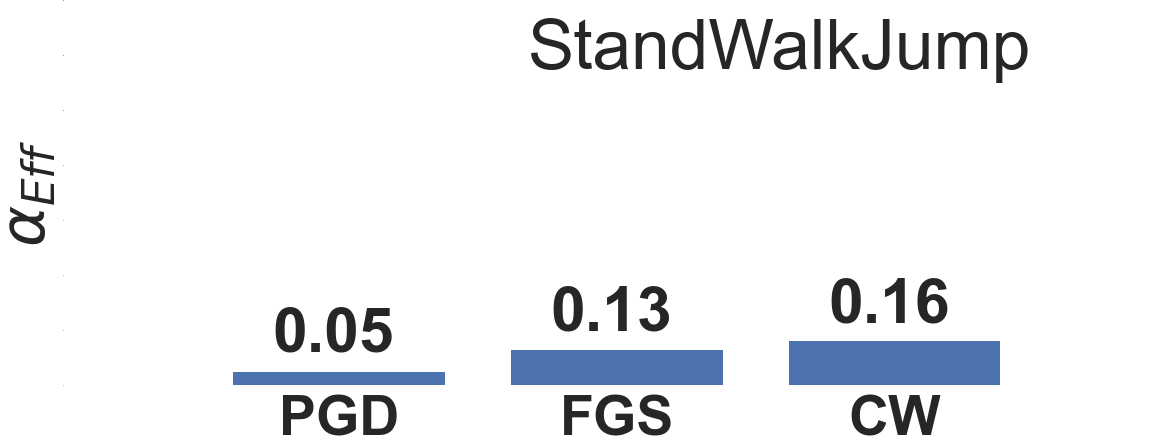}
            \end{minipage}%
        \begin{minipage}{.19\linewidth}
                \centering
                \includegraphics[width=\linewidth]{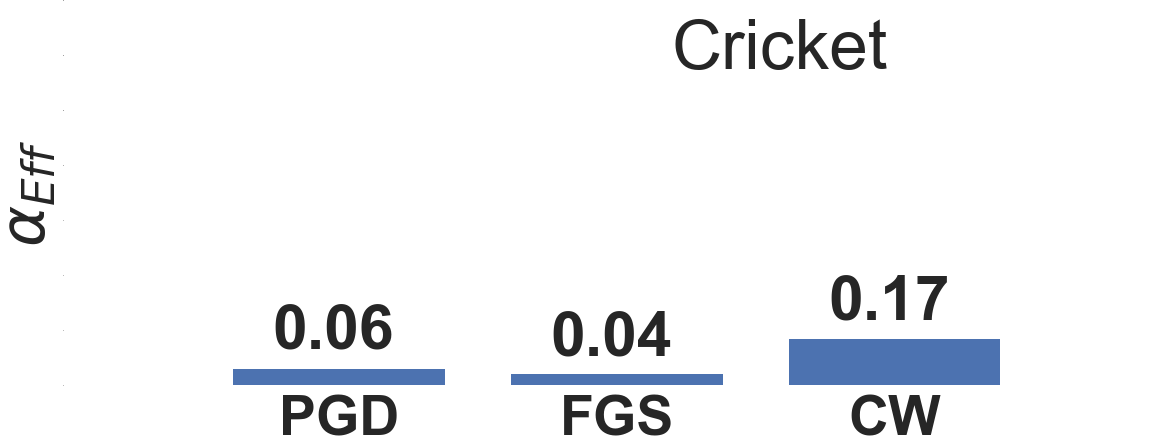}
            \end{minipage}
    \end{minipage}
\caption{Results for the effectiveness of adversarial examples from PGD and FGS on different deep models using adversarial training baselines (PGD, FGS, CW).}
\label{fig:pgdadvatk}
\end{figure*}

In Figure \ref{fig:appendadvatk}, we can observe that DTW-AR performs lower ($\alpha_{Eff} \le 0.5$) for some cases. We explain below how the other baseline attacks fail to outperform the proposed DTW-AR method on the same datasets. In Figure \ref{fig:cwadvatk} and \ref{fig:pgdadvatk}, we show results to evaluate the effectiveness of  baseline attacks against the models shown in Figure \ref{fig:appendadvatk}. These results show that DTW-AR is more effective in fooling DNNs created using baseline attacks-based adversarial training. For datasets where DTW-AR did not succeed in fooling the deep models with a high score, Figure \ref{fig:cwadvatk} and \ref{fig:pgdadvatk} show that baselines fail to outperform our proposed DTW-AR attack.
We also demonstrate in Figure \ref{fig:appendadvdef} that the baselines are not suitable for time-series domain since DTW-AR based adversarial training is able to defend against these attacks.

\vspace{1.0ex}

\begin{figure}[!h]
    \centering
\begin{minipage}{\linewidth}  
    \begin{minipage}{.47\linewidth}
            \centering
            \includegraphics[width=\linewidth]{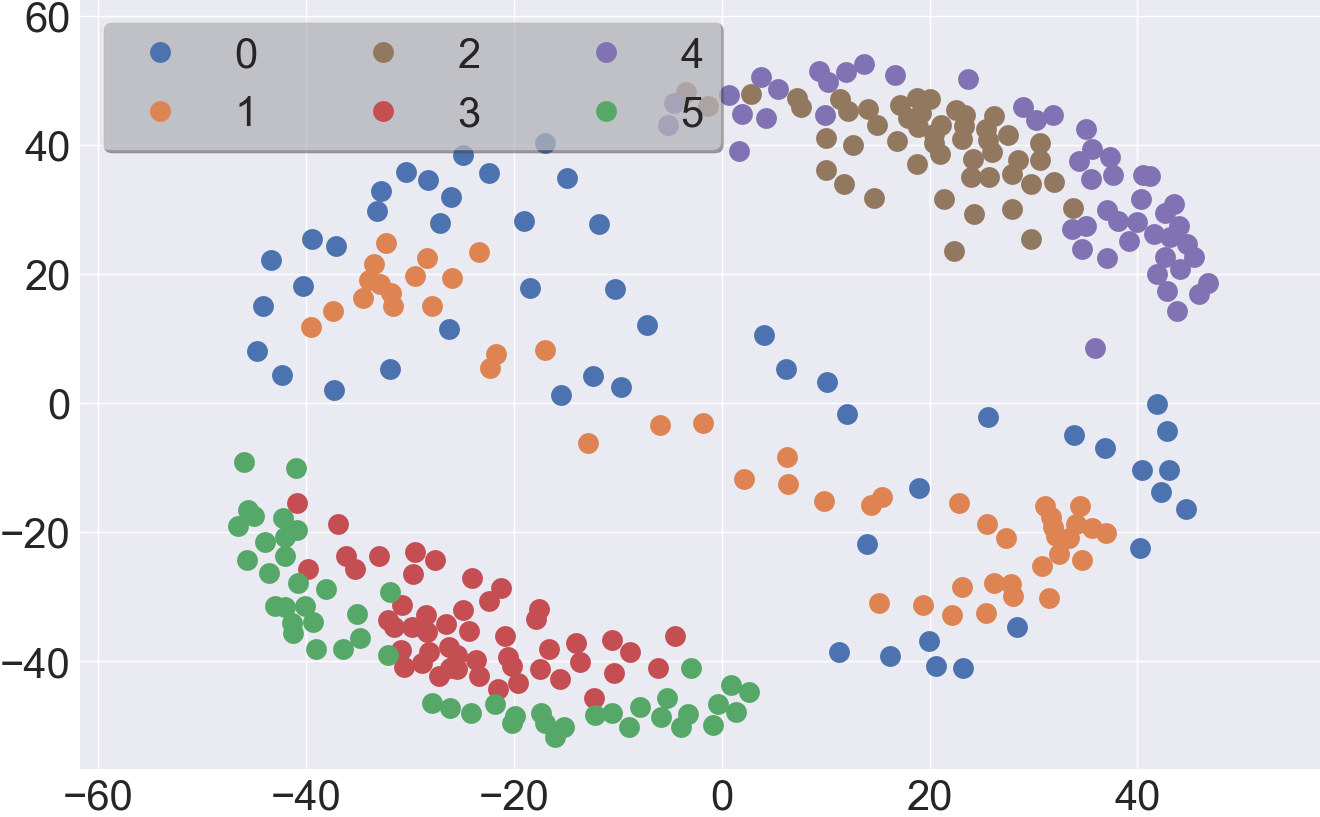}
        \end{minipage}%
    \begin{minipage}{.47\linewidth}
            \centering
            \includegraphics[width=\linewidth]{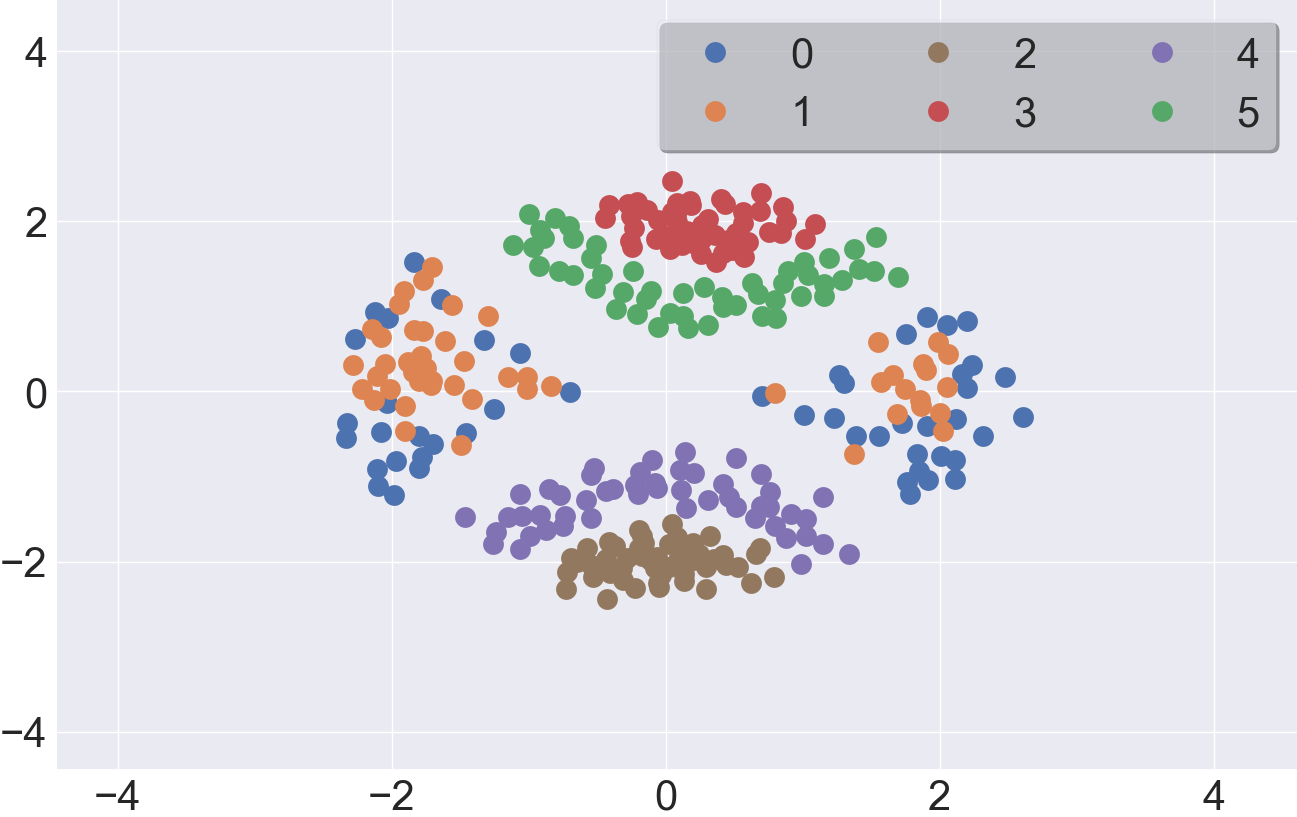}
        \end{minipage}
    \begin{minipage}{.47\linewidth}
            \centering
            \includegraphics[width=\linewidth]{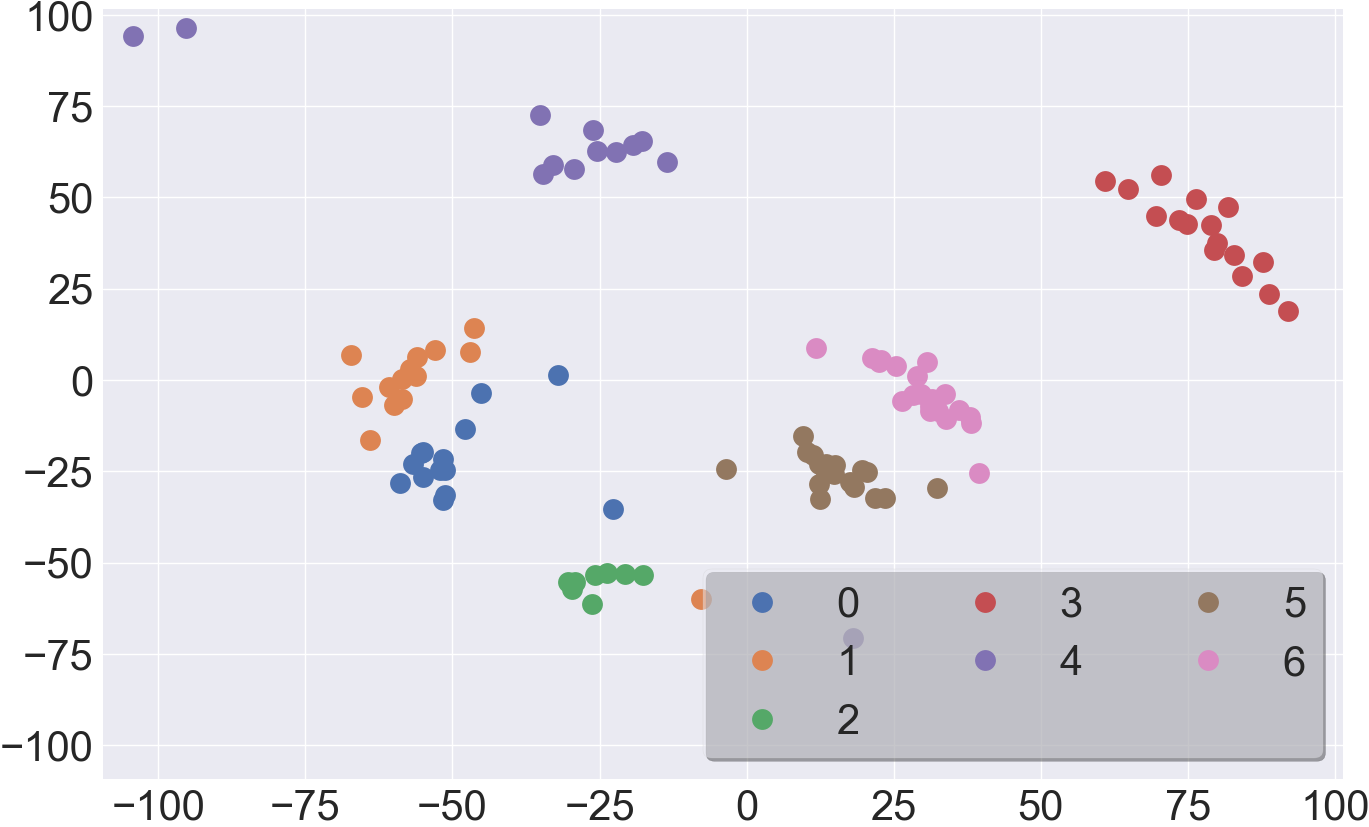}
        \end{minipage}%
    \begin{minipage}{.47\linewidth}
            \centering
            \includegraphics[width=\linewidth]{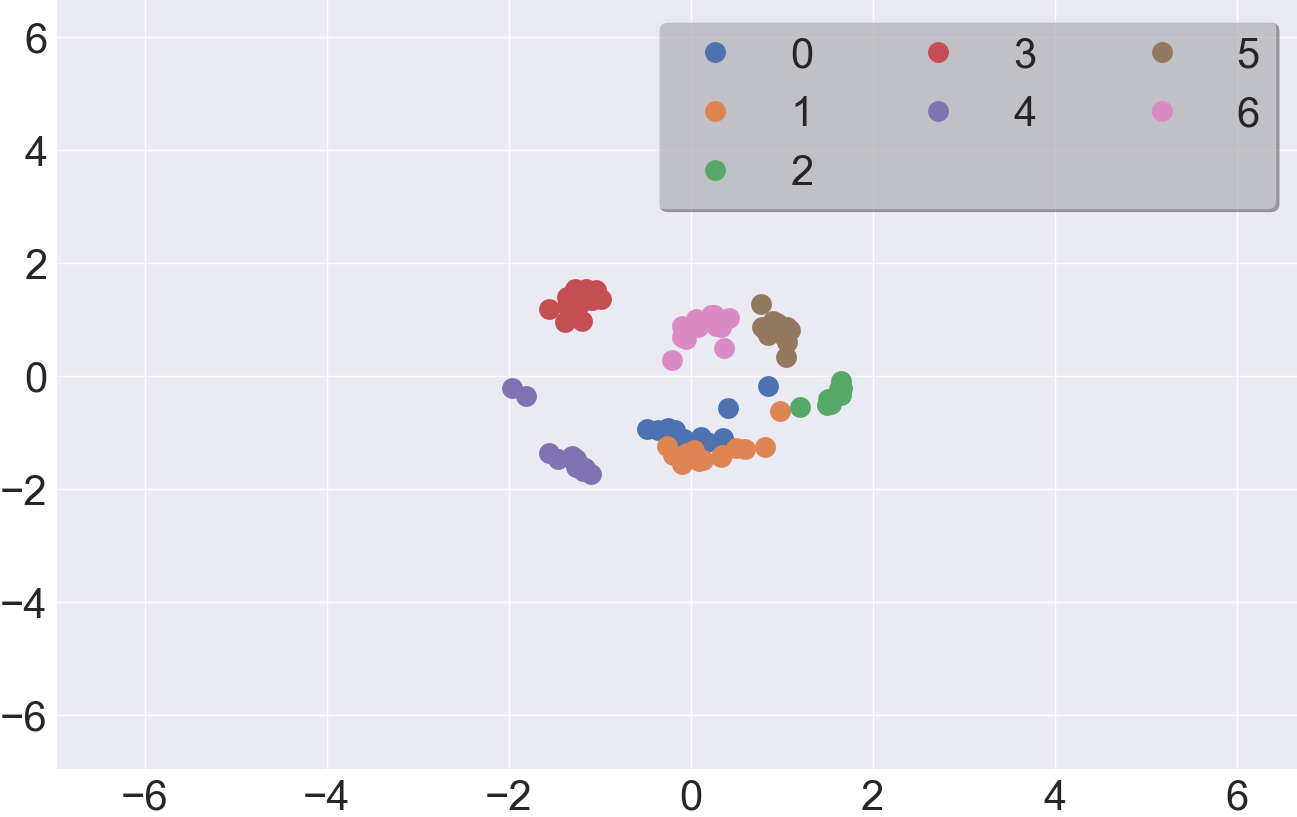}
        \end{minipage}
    \begin{minipage}{.47\linewidth}
    \vspace{1em}
            \centering
            $l_1$ Space
        \end{minipage}%
    \begin{minipage}{.47\linewidth}
    \vspace{1em}
            \centering
            $l_{\infty}$ Space
        \end{minipage}
\end{minipage}
\caption{Multi-dimensional scaling results showing the labeled data distribution in spaces using $l_1$ as a similarity measure (left column) and $l_{\infty}$ (right column) for two datasets: SC (top row) and Plane (bottom row).}
\label{fig:spacel1}
\end{figure}

\noindent \textbf{Results and Discussion on $l_1$ and $l_{\infty}$.} Figure \ref{fig:spacel1} shows the MDS results of \texttt{SC} and \texttt{Plane} in the space using $l_1$ as a similarity measure (left) and in the space using $l_{\infty}$ as a similarity measure (right).  We can observe that similar to the Euclidean space, there is a substantial entanglement between different classes. Thus, using $l_1$ and $l_{\infty}$ comes with similar drawbacks to using the Euclidean space for adversarial studies.

\begin{figure*}[!h]
    \centering
        \begin{minipage}{\linewidth}
        \begin{minipage}{.19\linewidth}
                \centering
                \includegraphics[width=\linewidth]{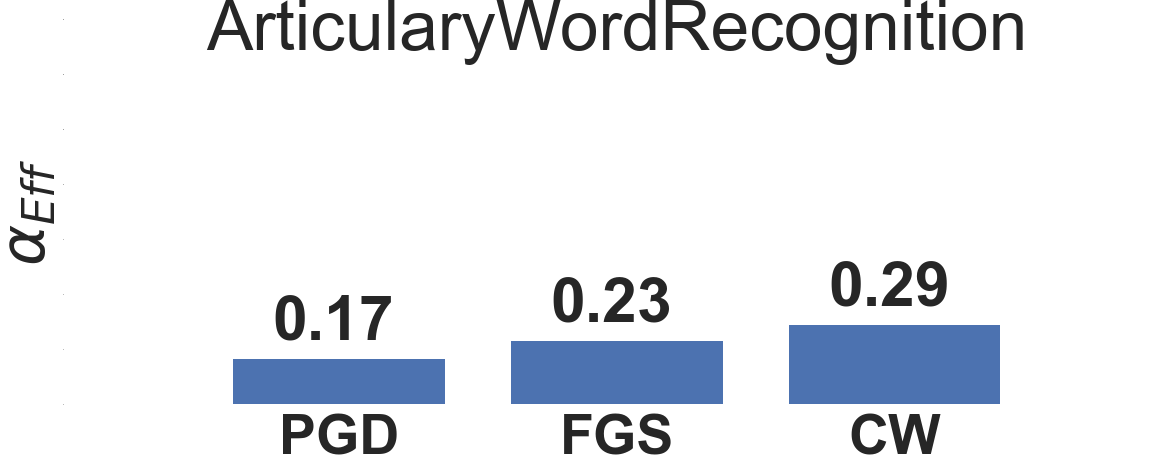}
            \end{minipage}%
        \begin{minipage}{.19\linewidth}
                \centering
                \includegraphics[width=\linewidth]{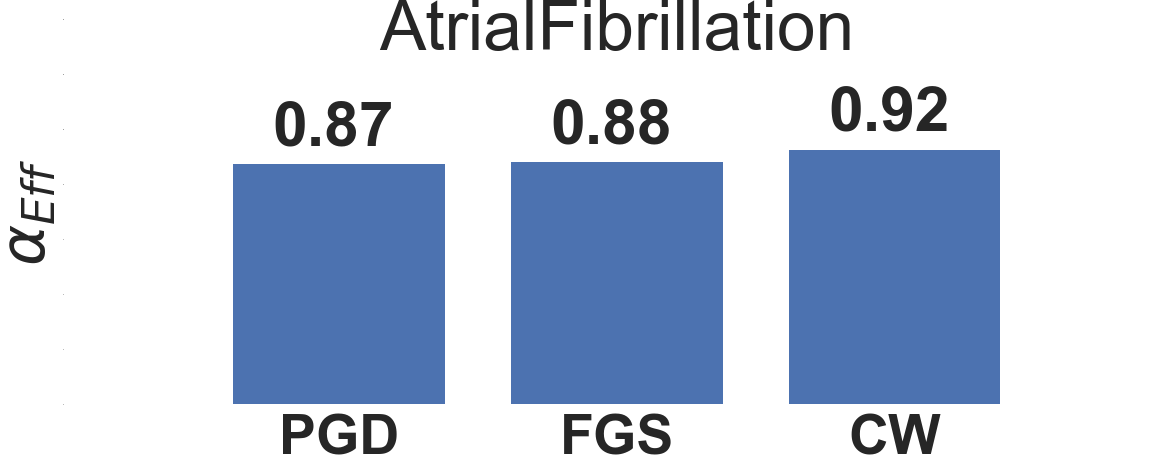}
            \end{minipage}%
        \begin{minipage}{.19\linewidth}
                \centering
                \includegraphics[width=\linewidth]{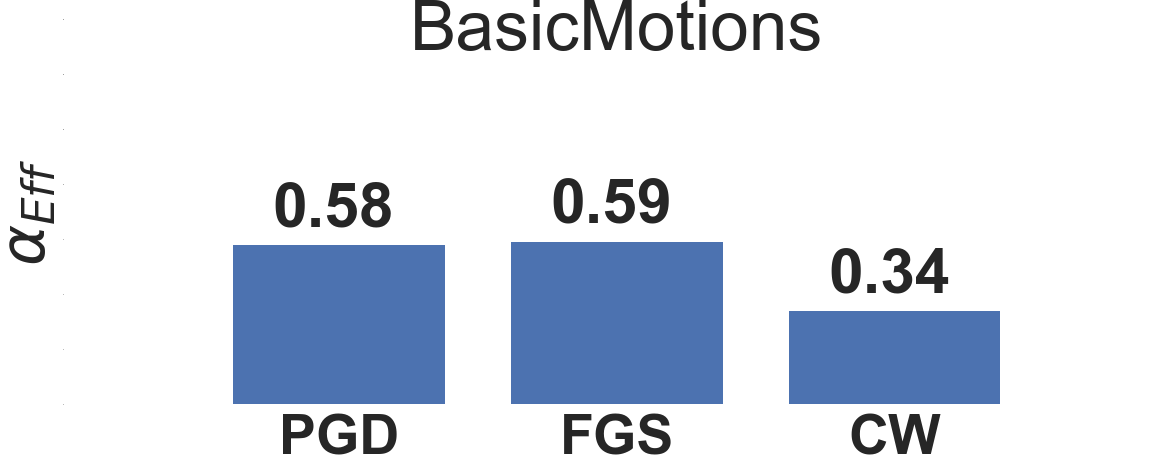}
            \end{minipage}%
        \begin{minipage}{.19\linewidth}
                \centering
                \includegraphics[width=\linewidth]{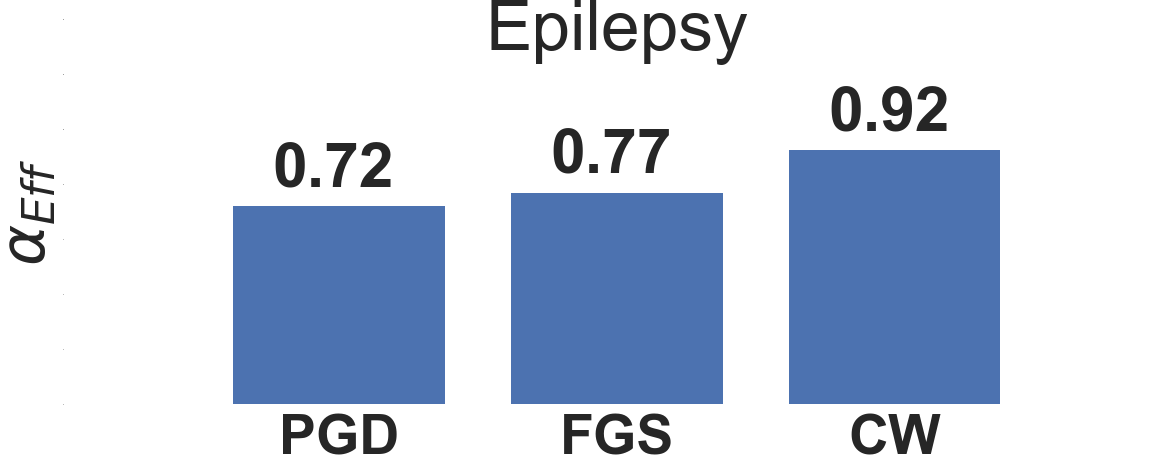}
            \end{minipage}%
        \begin{minipage}{.19\linewidth}
                \centering
                \includegraphics[width=\linewidth]{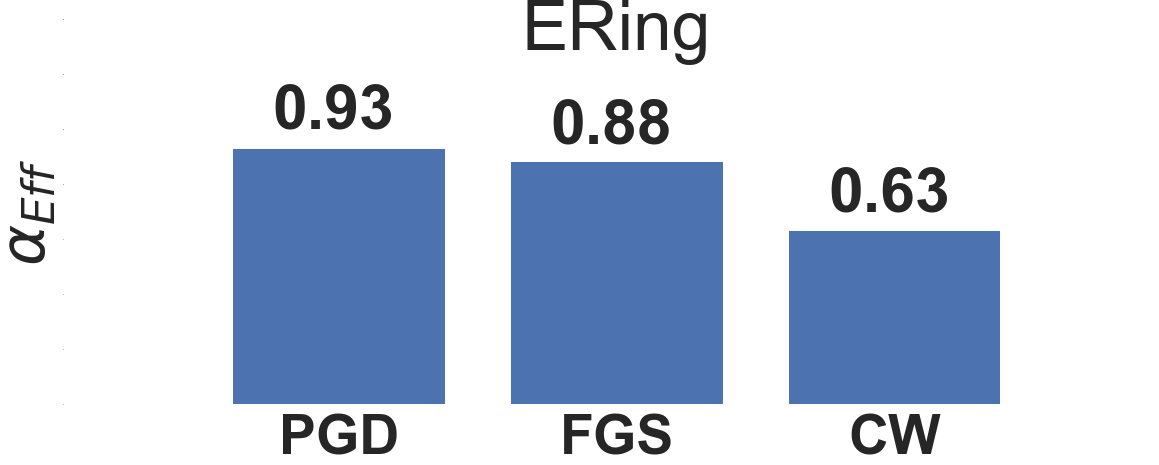}
            \end{minipage}
        \begin{minipage}{.19\linewidth}
                \centering
                \includegraphics[width=\linewidth]{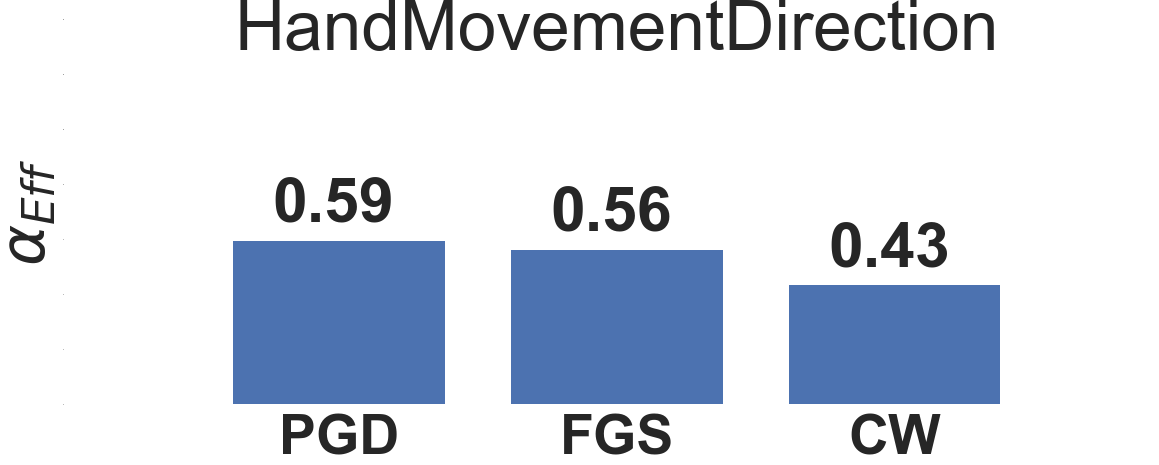}
            \end{minipage}%
        \begin{minipage}{.19\linewidth}
                \centering
                \includegraphics[width=\linewidth]{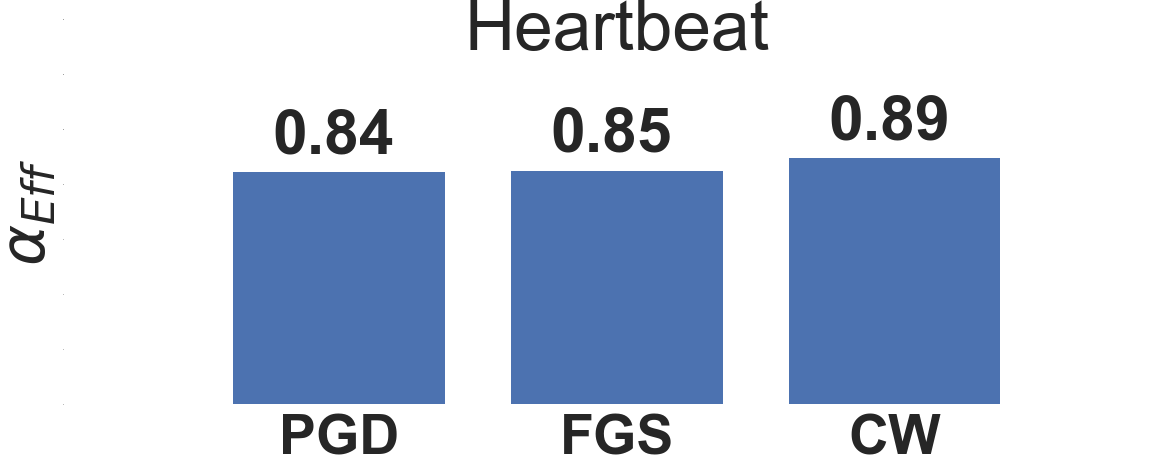}
            \end{minipage}%
        \begin{minipage}{.19\linewidth}
                \centering
                \includegraphics[width=\linewidth]{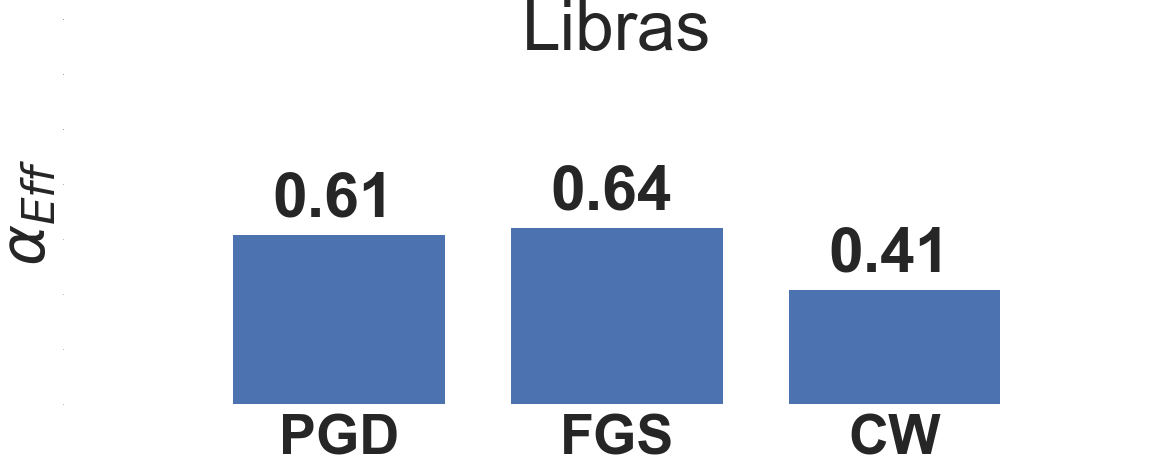}
            \end{minipage}%
        \begin{minipage}{.19\linewidth}
                \centering
                \includegraphics[width=\linewidth]{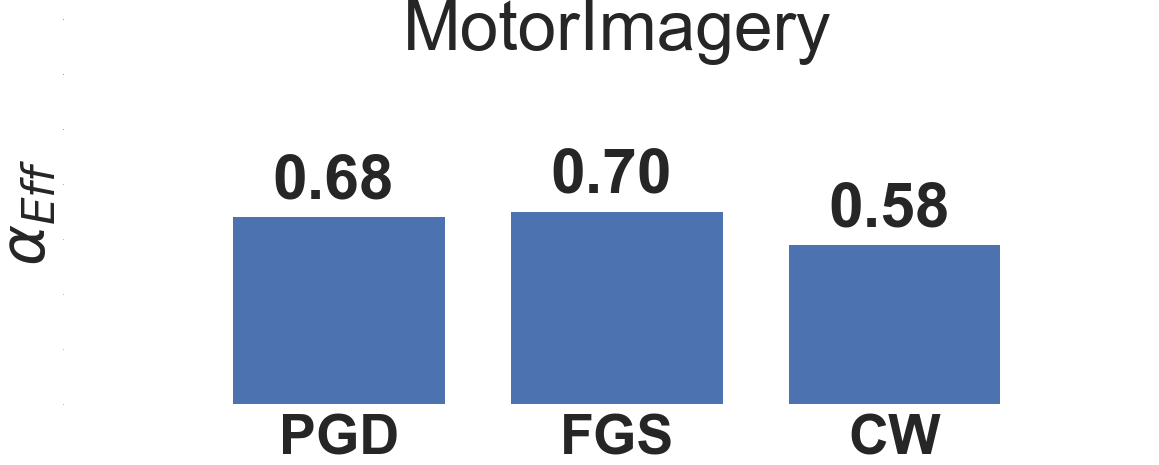}
            \end{minipage}%
        \begin{minipage}{.19\linewidth}
                \centering
                \includegraphics[width=\linewidth]{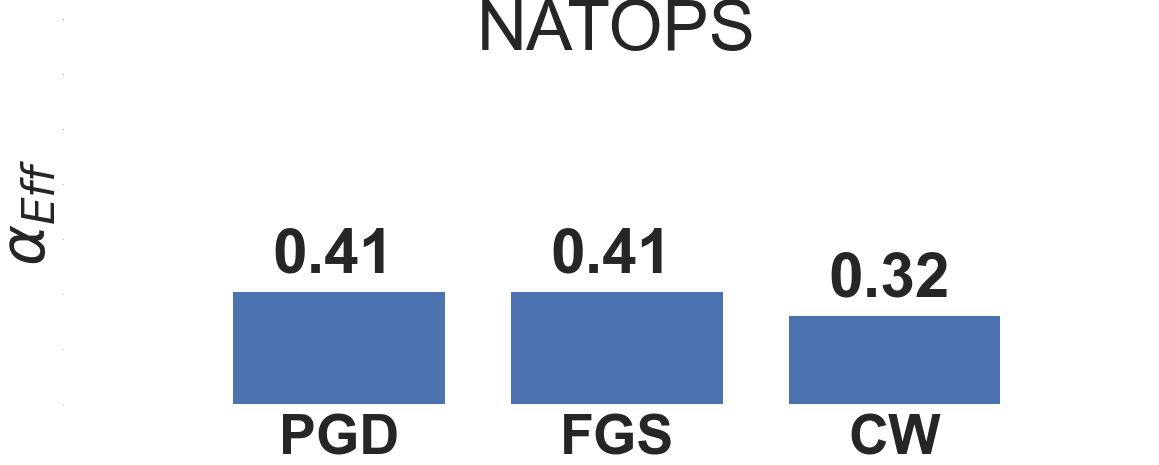}
            \end{minipage}
        \begin{minipage}{.19\linewidth}
                \centering
                \includegraphics[width=\linewidth]{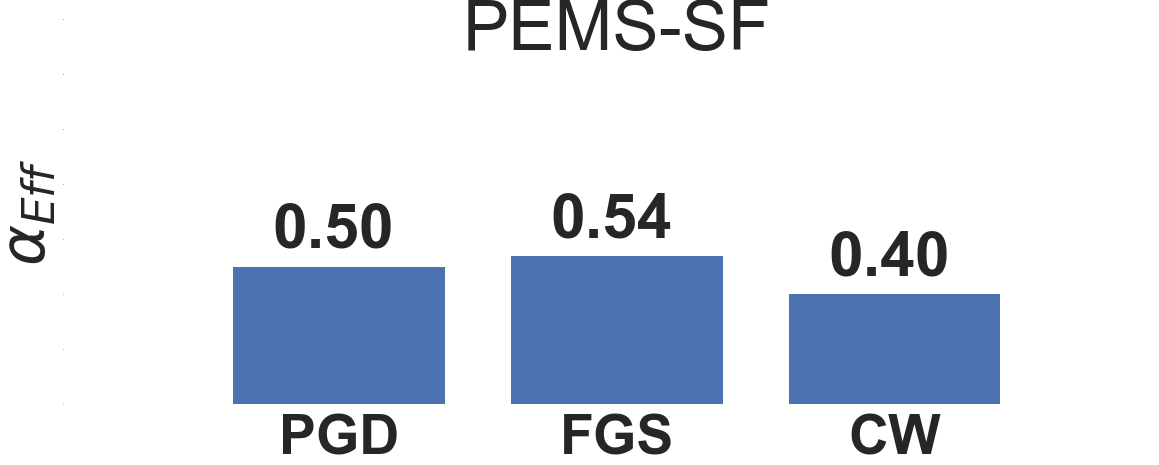}
            \end{minipage}%
        \begin{minipage}{.19\linewidth}
                \centering
                \includegraphics[width=\linewidth]{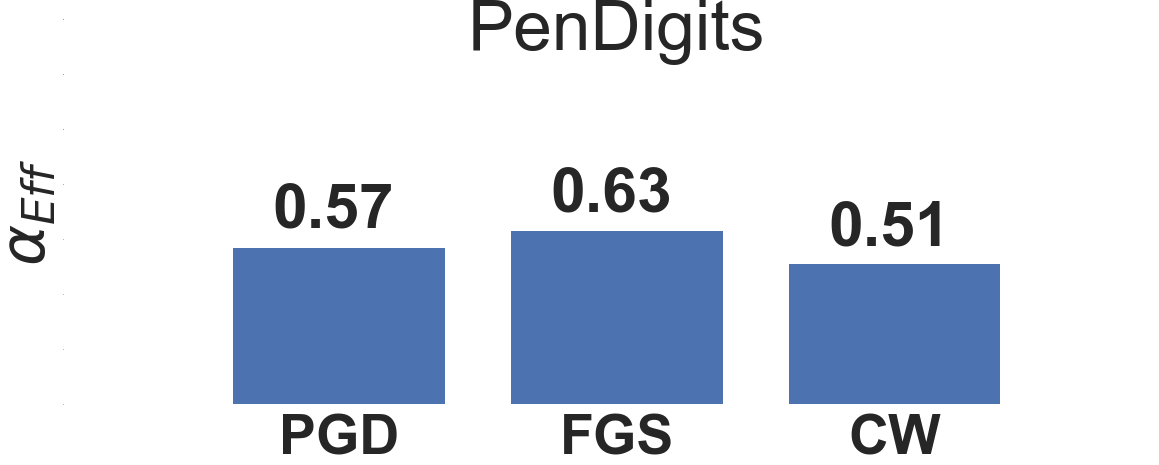}
            \end{minipage}%
        \begin{minipage}{.19\linewidth}
                \centering
                \includegraphics[width=\linewidth]{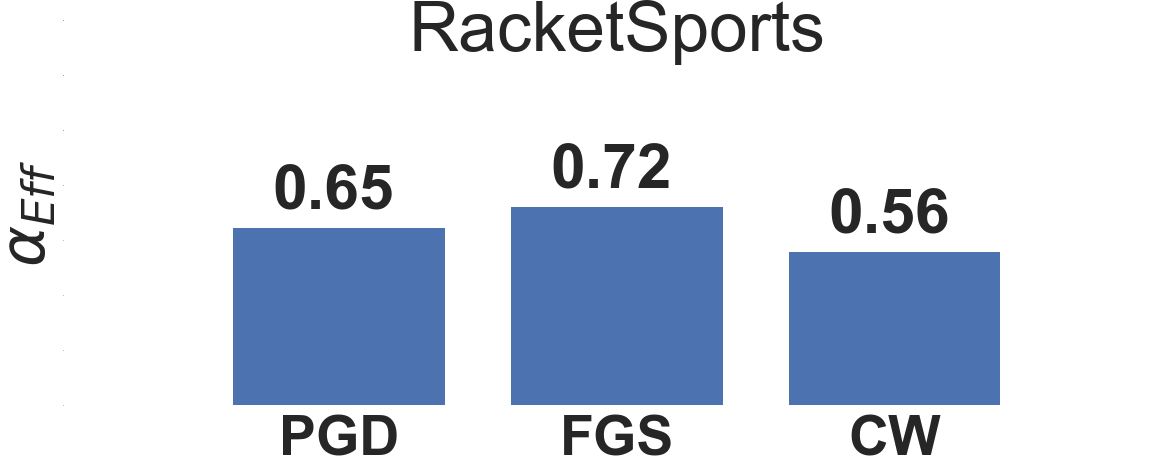}
            \end{minipage}%
        \begin{minipage}{.19\linewidth}
                \centering
                \includegraphics[width=\linewidth]{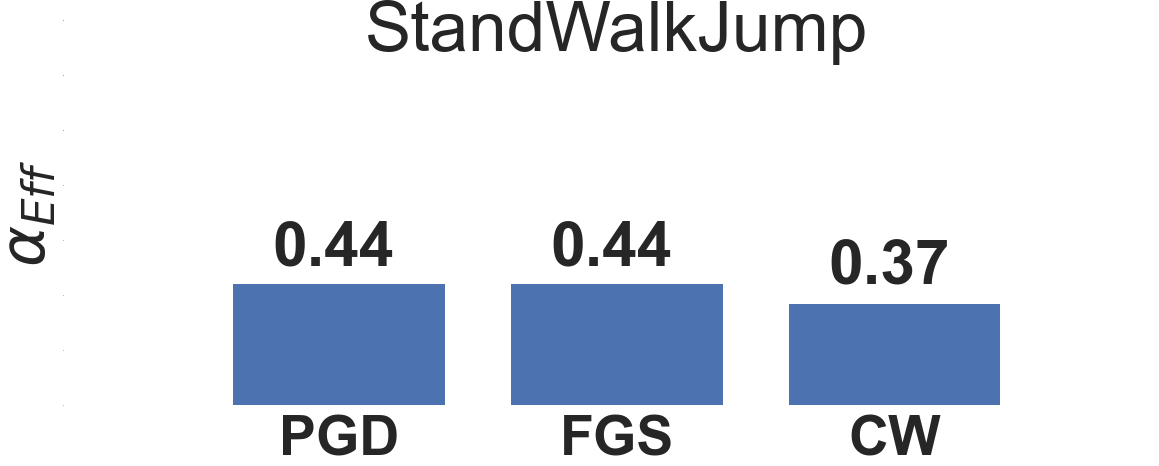}
            \end{minipage}%
        \begin{minipage}{.19\linewidth}
                \centering
                \includegraphics[width=\linewidth]{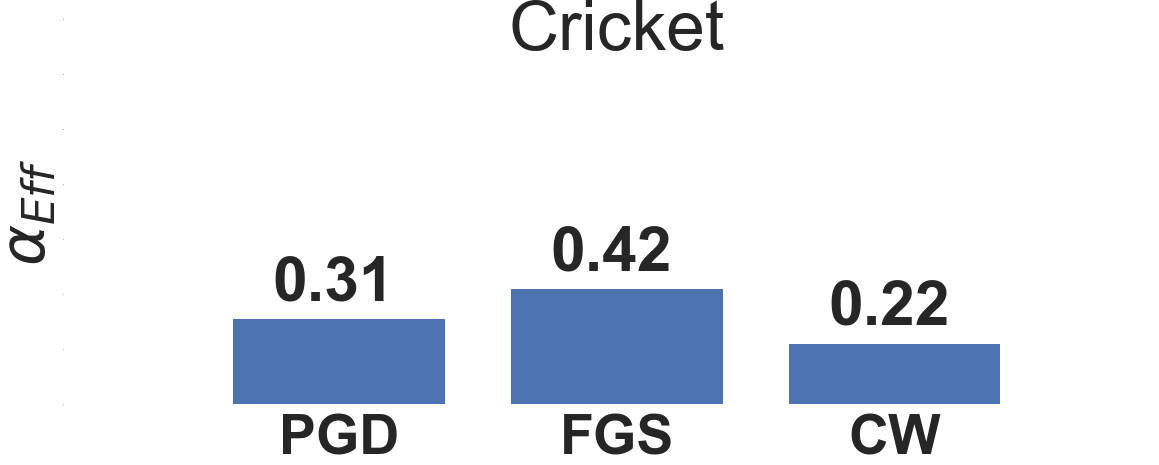}
            \end{minipage}
    \end{minipage}
\caption{Results for the effectiveness of adversarial examples from DTW-AR on different deep models using adversarial training baselines (PGD, FGS, CW) with $l_1$-norm.}
\label{fig:advl1}
\end{figure*}

\begin{figure*}[!h]
    \centering
        \begin{minipage}{\linewidth}
        \begin{minipage}{.19\linewidth}
                \centering
                \includegraphics[width=\linewidth]{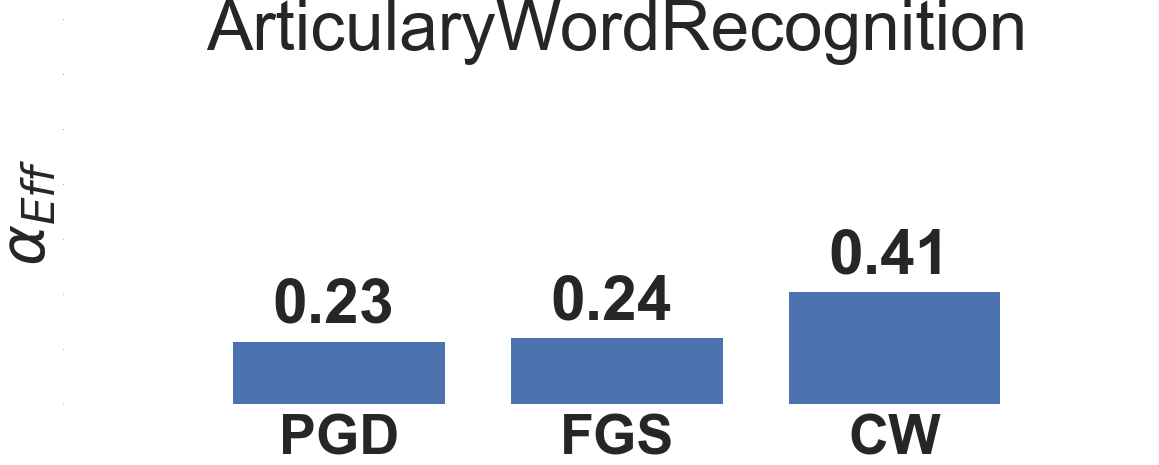}
            \end{minipage}%
        \begin{minipage}{.19\linewidth}
                \centering
                \includegraphics[width=\linewidth]{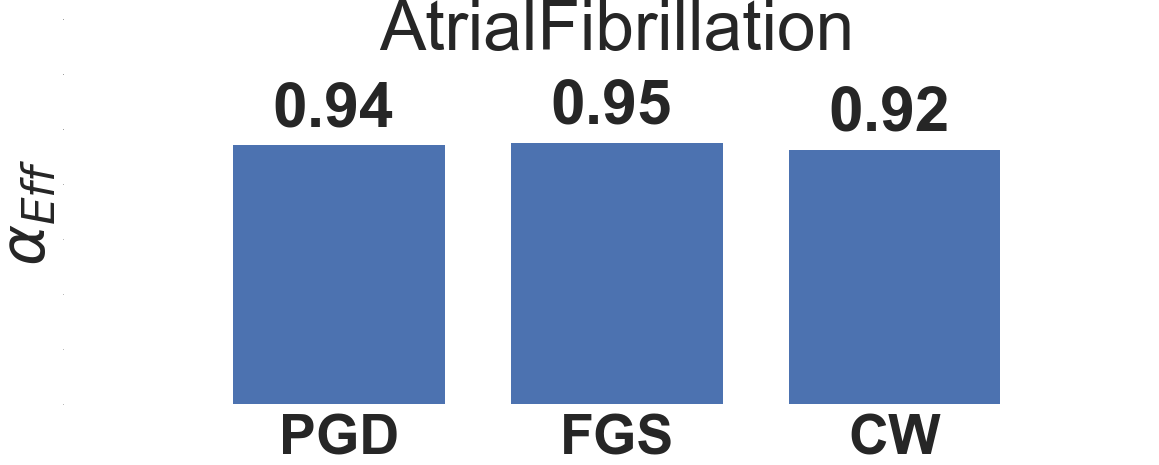}
            \end{minipage}%
        \begin{minipage}{.19\linewidth}
                \centering
                \includegraphics[width=\linewidth]{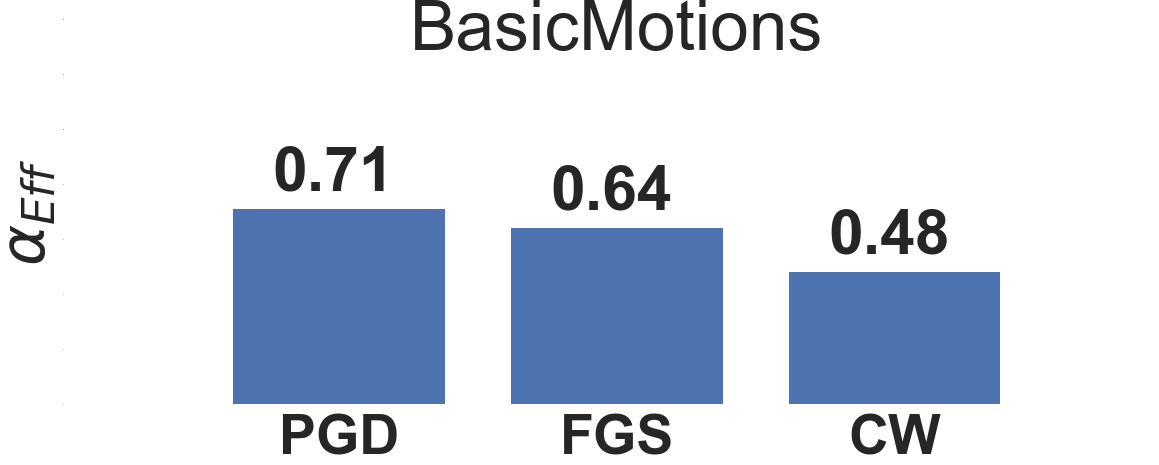}
            \end{minipage}%
        \begin{minipage}{.19\linewidth}
                \centering
                \includegraphics[width=\linewidth]{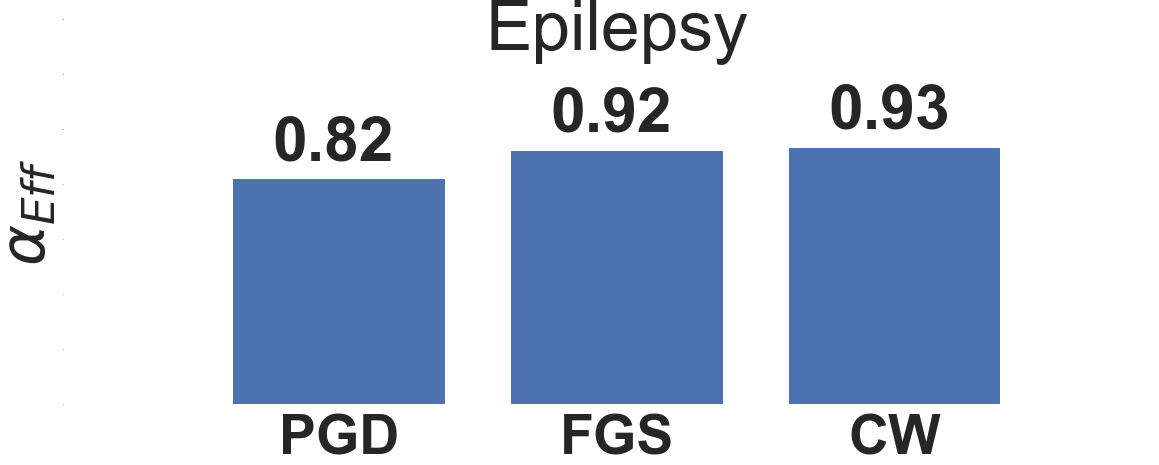}
            \end{minipage}%
        \begin{minipage}{.19\linewidth}
                \centering
                \includegraphics[width=\linewidth]{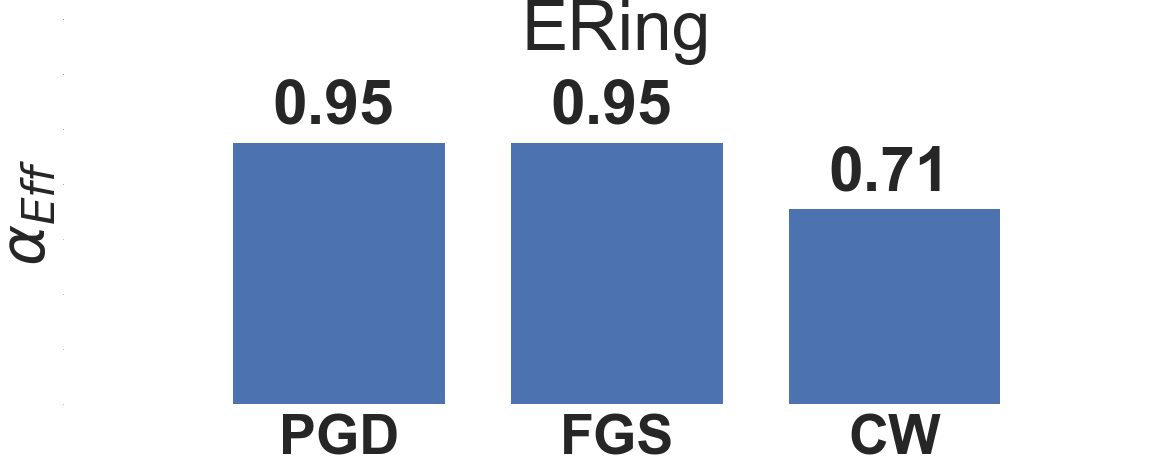}
            \end{minipage}
        \begin{minipage}{.19\linewidth}
                \centering
                \includegraphics[width=\linewidth]{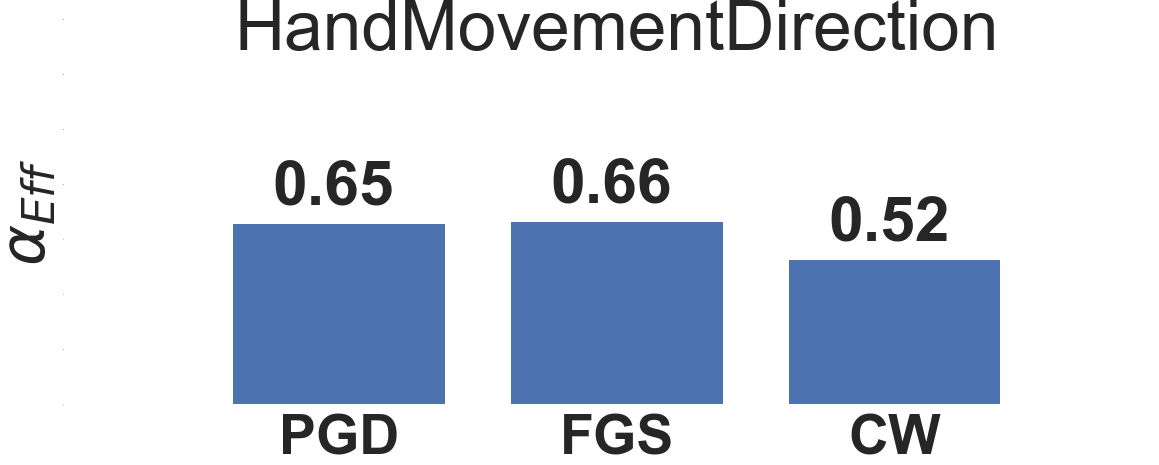}
            \end{minipage}%
        \begin{minipage}{.19\linewidth}
                \centering
                \includegraphics[width=\linewidth]{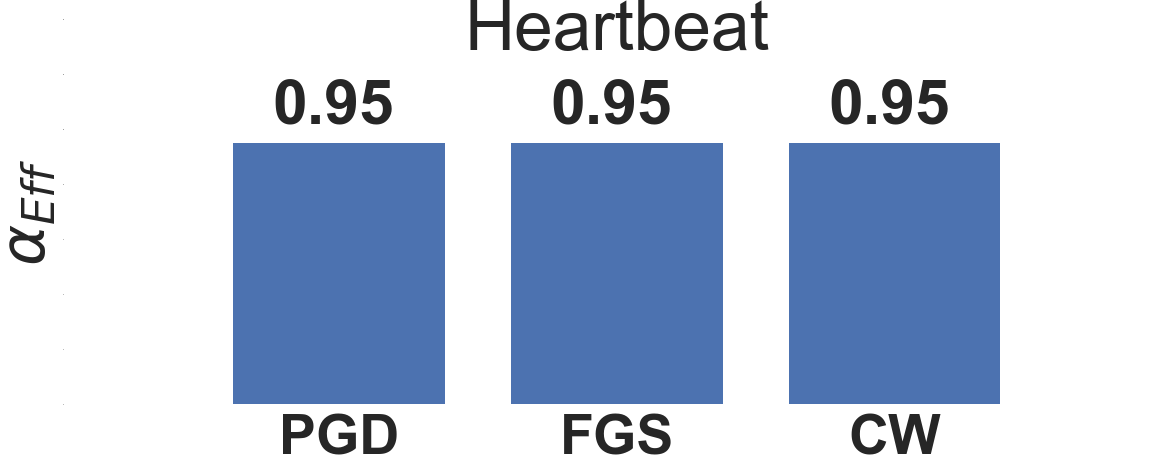}
            \end{minipage}%
        \begin{minipage}{.19\linewidth}
                \centering
                \includegraphics[width=\linewidth]{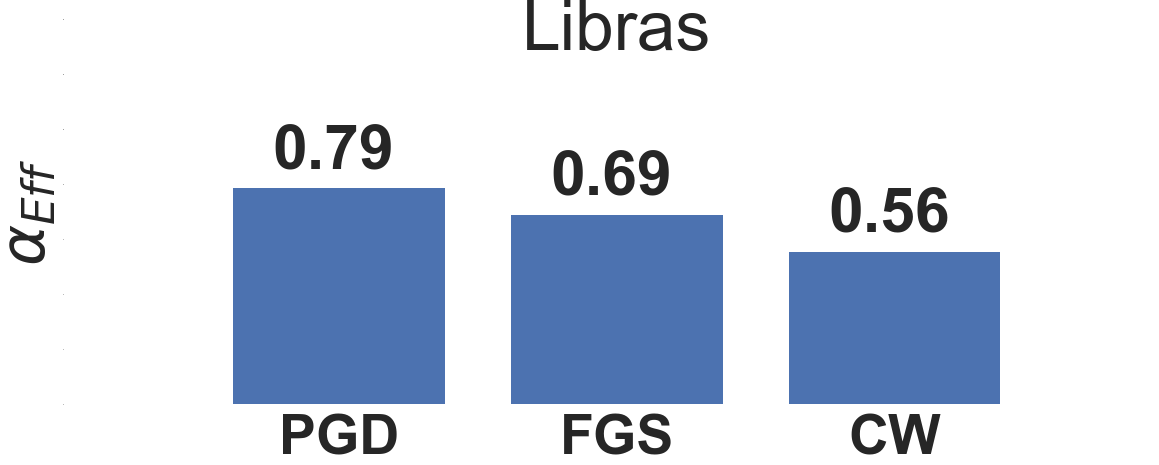}
            \end{minipage}%
        \begin{minipage}{.19\linewidth}
                \centering
                \includegraphics[width=\linewidth]{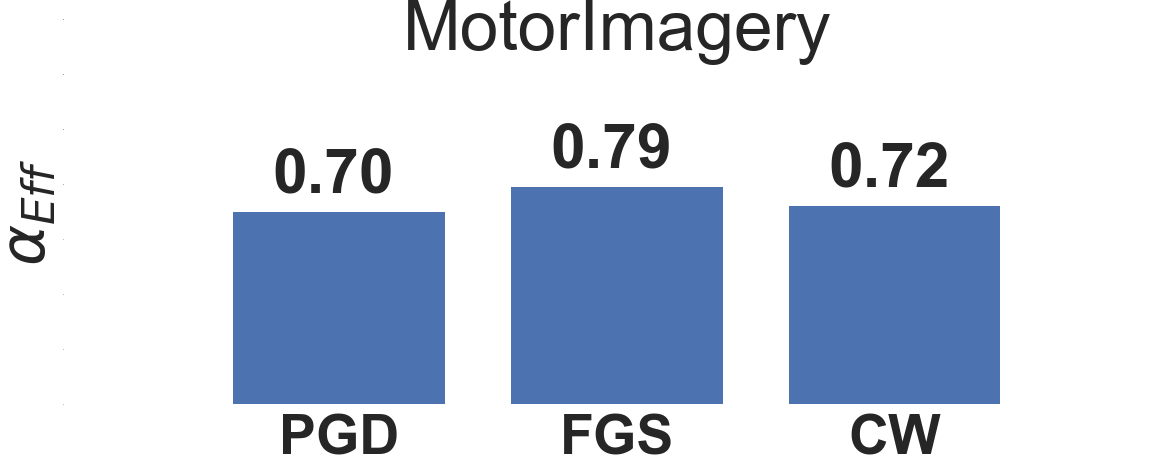}
            \end{minipage}%
        \begin{minipage}{.19\linewidth}
                \centering
                \includegraphics[width=\linewidth]{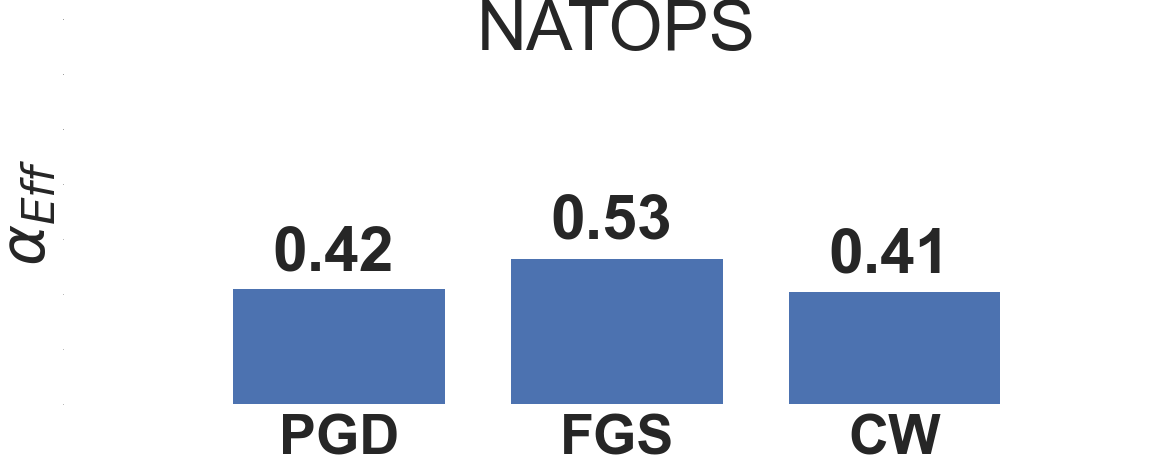}
            \end{minipage}
        \begin{minipage}{.19\linewidth}
                \centering
                \includegraphics[width=\linewidth]{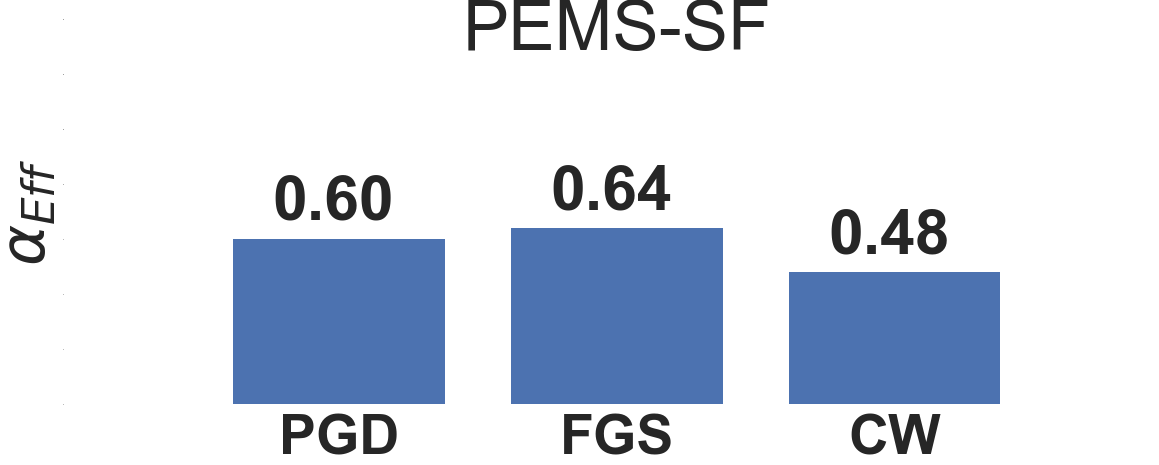}
            \end{minipage}%
        \begin{minipage}{.19\linewidth}
                \centering
                \includegraphics[width=\linewidth]{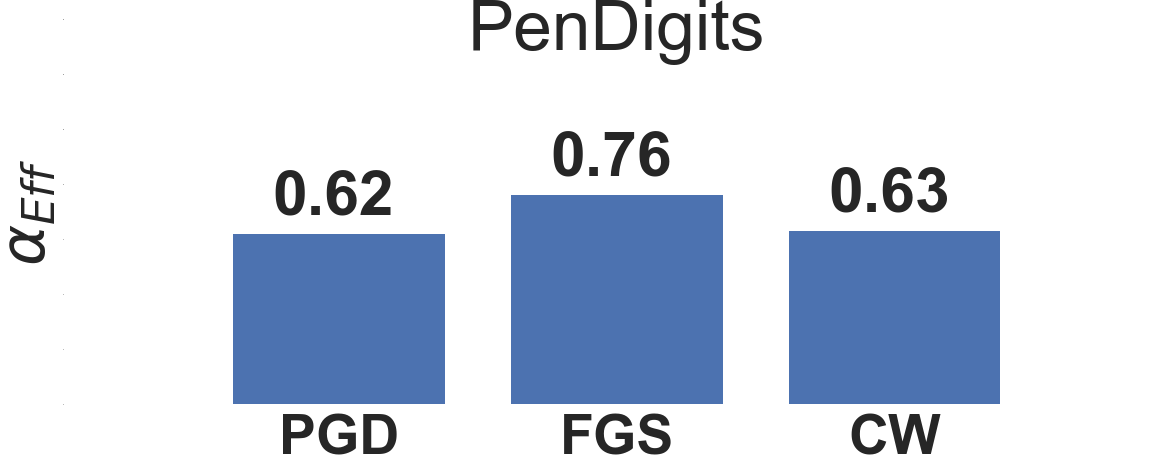}
            \end{minipage}%
        \begin{minipage}{.19\linewidth}
                \centering
                \includegraphics[width=\linewidth]{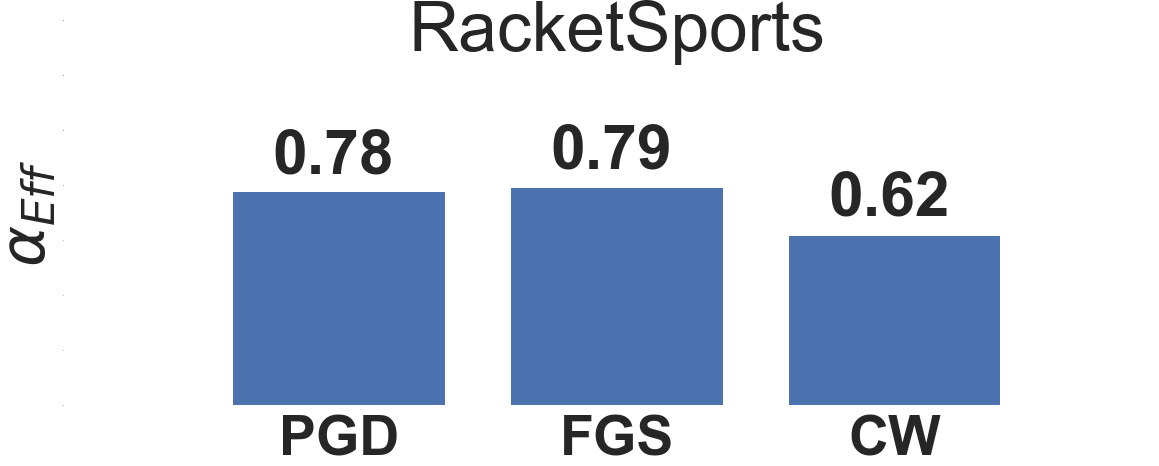}
            \end{minipage}%
        \begin{minipage}{.19\linewidth}
                \centering
                \includegraphics[width=\linewidth]{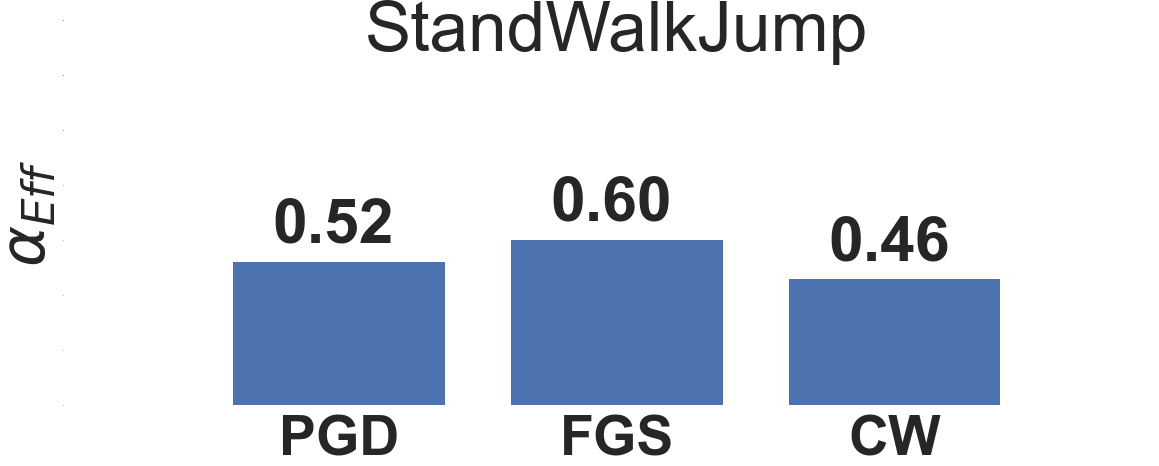}
            \end{minipage}%
        \begin{minipage}{.19\linewidth}
                \centering
                \includegraphics[width=\linewidth]{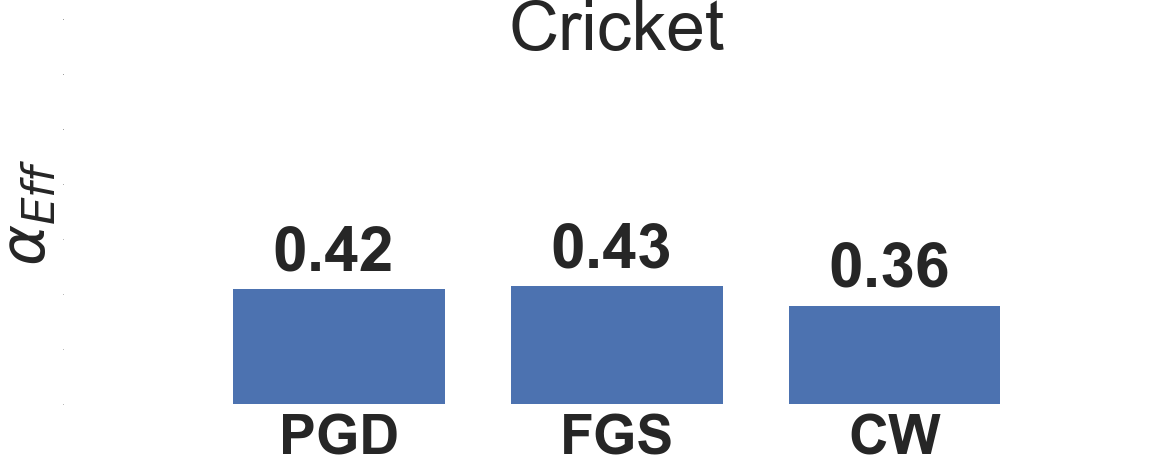}
            \end{minipage}
    \end{minipage}
\caption{Results for the effectiveness of adversarial examples from DTW-AR on different deep models using adversarial training baselines (PGD, FGS, CW) with $l_{\infty}$-norm.}
\label{fig:advlinf}
\end{figure*}

Figure \ref{fig:advl1} and \ref{fig:advlinf} show the results of DTW-AR based adversarial training against adversarial attacks generated using $l_1$ and $l_{\infty}$ as a metric.

We conclude that DTW-AR is able to generalize against attacks in other spaces than the Euclidean one. Since the Manhattan distance is similar to the Euclidean distance in the point-to-point matching, and $\infty$-norm describes a signal by solely its maximum value, DTW measure is still considered a better similarity measure. The empirical success of the DTW-AR suggests that the framework can be further analyzed theoretically and empirically for future research into adversarially robust classification when compared to different alternative similarity measures.

\begin{figure}[!h]
    \centering
    \includegraphics[width=\linewidth]{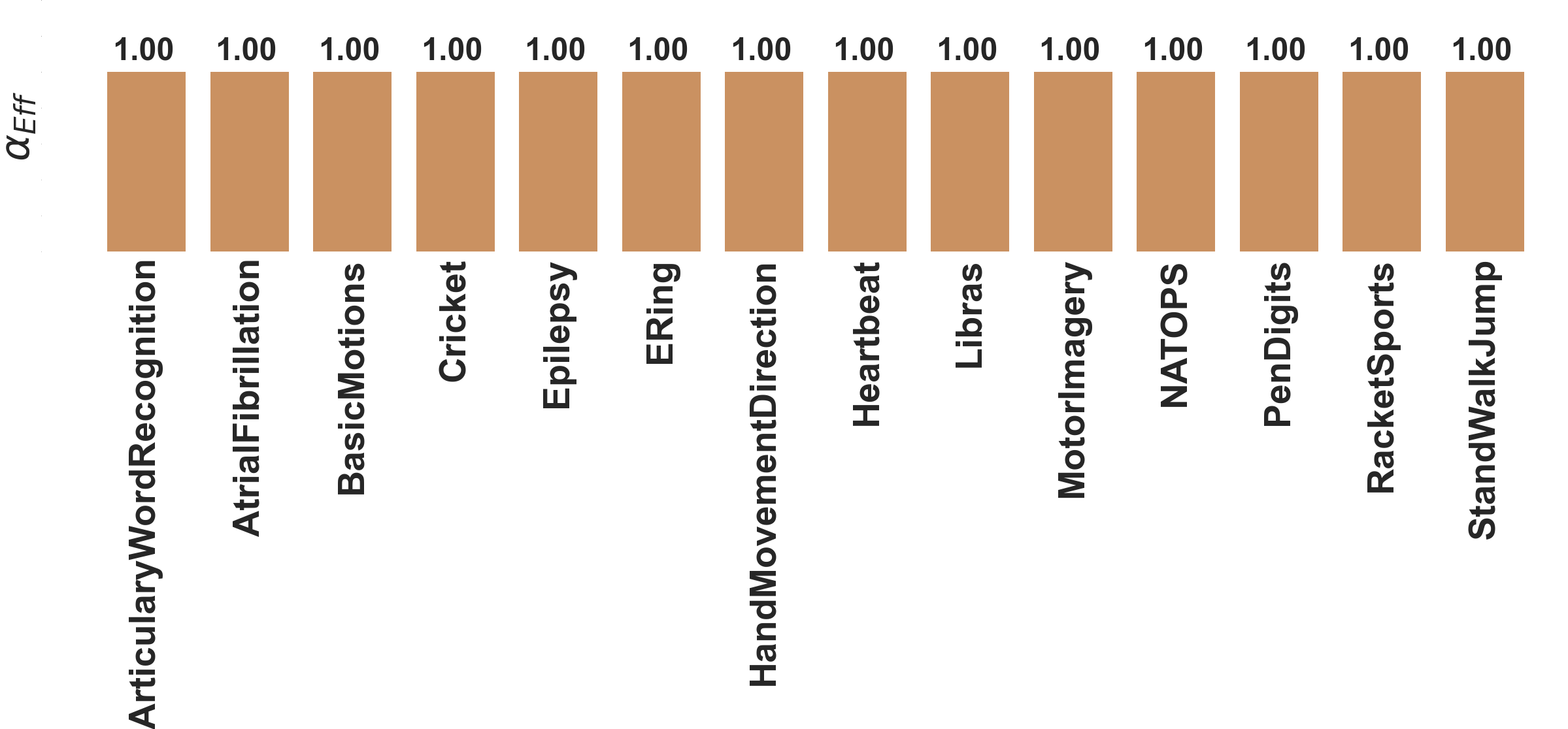}
    \caption{Results of the success rate of DTW-AR adversarial trained model to predict the true label of adversarial attack generated from \cite{karim2020adversarial}.}
    \label{fig:karim}
\end{figure}
\vspace{1.0ex}
\noindent \textbf{Comparison with \cite{karim2020adversarial} on the full MV UCR dataset.} Figure \ref{fig:karim} shows that the observations made within the main paper are still valid over the different datasets.

\vspace{1.0ex}
\noindent \textbf{DTW-AR on non-CNN models} We want to clarify that the proposed DTW based adversarial framework is model-agnostic. The adversarial attack algorithm is based on Equations (5) and (7). Equation (7) relies on the output of the pre-softmax layer of the deep model and is independent of the model's core architecture. If the main layers of the model are based on convolutional layers, recurrent layers or attention layers, our proposed algorithm (Algorithm 1 in the main paper) is the same. As both LSTM and Transformer based models are typically used for forecasting applications [1,2], we focus on 1D-CNNs in this work. In Tables \ref{tab:addexp} and \ref{tab:addexp2}, we provide additional results on non-CNN models. We clearly observe that DTW-AR remains effective on non-CNN models using the efficiency metric $\alpha_{Eff} \in [0,1]$ over the created adversarial examples. $\alpha_{Eff} = \frac{\text{\# Adv. examples s.t.}F(X)==y_{target}}{\text{\# Adv. examples}}$ (higher means better attacks) was introduced in the paper to measure the capability of adversarial examples to fool a given DNN $F_{\theta}$ to output the true class-label. Therefore, DTW-AR generalizes over any given deep model. Furthermore, we note that all our assumptions and claims made in the paper are not based specifically on CNN models, but are general to any DNN classifier $F_{\theta}$. Finally, we note that LSTM-based models are slower in execution than CNN models and Transformer-based models are too complex for classification tasks whereas CNNs perform similarly with less computational resources.

\begin{table}[!h]
    \centering
    \caption{Results for the effectiveness $\alpha_{Eff}$ of adversarial examples from DTW-AR using LSTM-based deep neural network in a black-box (BB) setting.}
    \resizebox{\linewidth}{!}{%
    \begin{tabular}{|l|c|c|c|c|} 
    \hline
     & \multicolumn{1}{c|}{Clean Test Accuracy} & \multicolumn{3}{c|}{$\alpha_{Eff}$} \\ \hline
    Dataset& Standard & Standard & PGD Adv. Trn. & FGS Adv. Trn. \\ \hline 
     Atrial Fibrillation & 0.26 & 0.97 & 0.93 & 0.95\\ \hline
     Epilepsy & 0.61 & 0.89 & 0.75 & 0.75 \\ \hline
     ERing & 0.74 & 0.98 & 0.91 & 0.91 \\ \hline
     Heartbeat & 0.73 & 0.65 & 0.55 & 0.54 \\ \hline
     RacketSports & 0.86 & 0.91 & 0.85 & 0.85 \\ \hline
    \end{tabular}
    }
    \label{tab:addexp}
\end{table}

\begin{table}[!h]
    \centering
    \caption{Results for the effectiveness $\alpha_{Eff}$ of adversarial examples from DTW-AR using Transformer-based deep neural network in a black-box (BB) setting.}
    \resizebox{\linewidth}{!}{%
    \begin{tabular}{|l|c|c|c|c|} 
    \hline
     & \multicolumn{1}{c|}{Clean Test Accuracy} & \multicolumn{3}{c|}{$\alpha_{Eff}$} \\ \hline
    Dataset& Standard & Standard & PGD Adv. Trn. & FGS Adv. Trn. \\ \hline 
     Atrial Fibrillation & 0.22 & 0.95 & 0.95 & 0.95 \\ \hline
     Epilepsy & 0.61 & 0.85 & 0.65 & 0.64 \\ \hline
     ERing & 0.38 & 0.73 & 0.69 & 0.65 \\ \hline
     Heartbeat & 0.71 & 0.91 & 0.85 & 0.85 \\ \hline
     RacketSports& 0.66 & 0.73 & 0.66 & 0.65  \\ \hline
    \end{tabular}
    }
    \label{tab:addexp2}
\end{table}

\vspace{1.0ex}

\begin{figure}[t]
    \centering
    \begin{minipage}{\linewidth}
        \begin{minipage}{.46\linewidth}
                \centering
                \includegraphics[width=\linewidth]{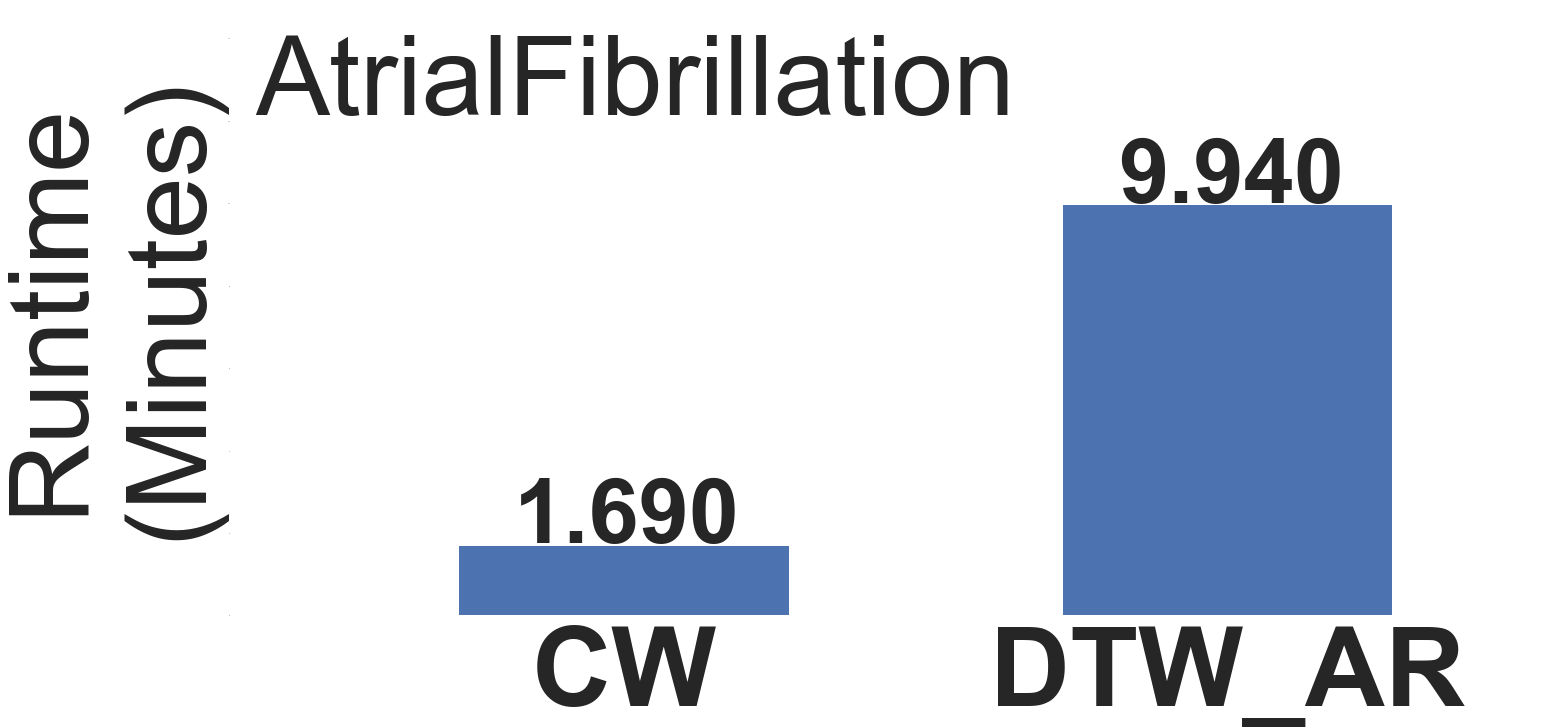}
            \end{minipage}%
            \hfill
        \begin{minipage}{.46\linewidth}
                \centering
                \includegraphics[width=\linewidth]{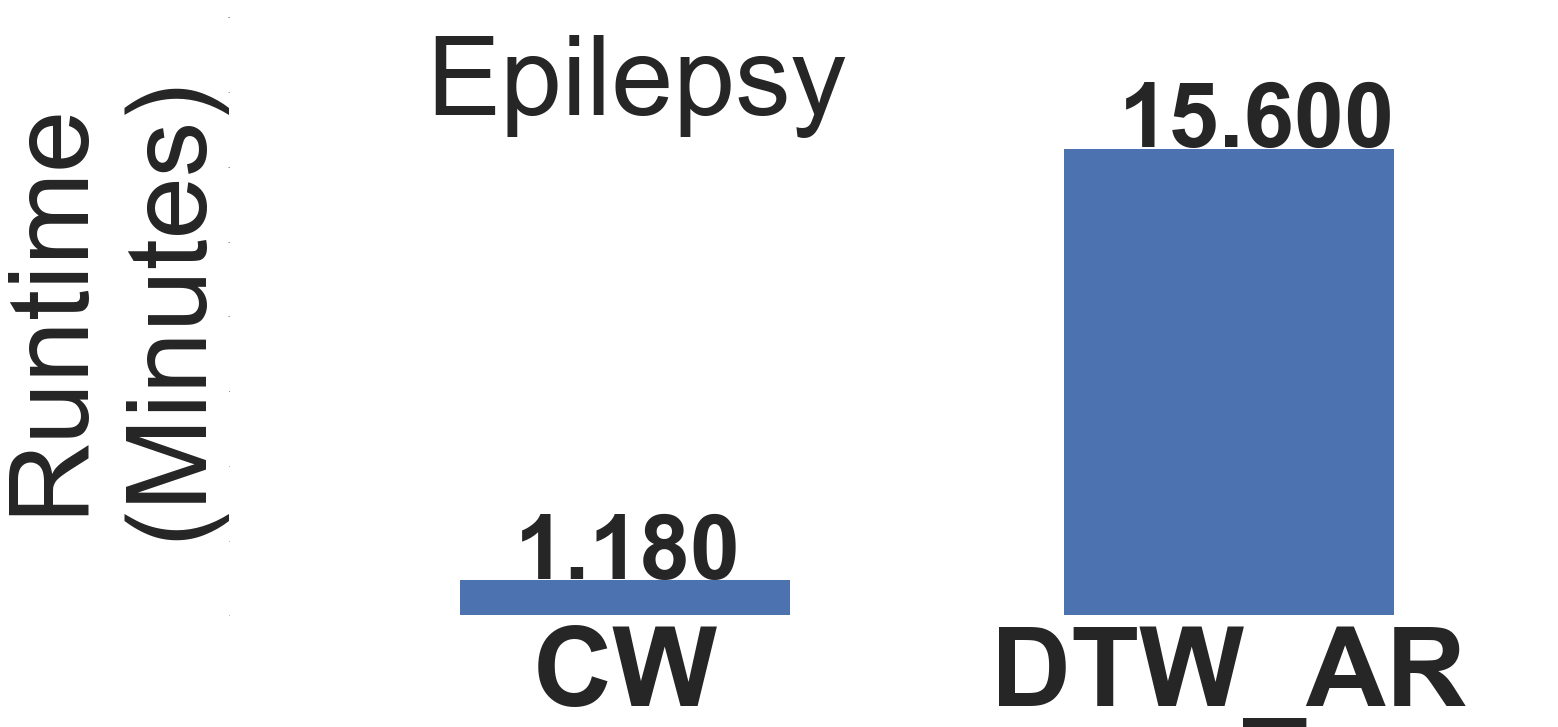}
            \end{minipage}
        \begin{minipage}{.46\linewidth}
                \centering
                \includegraphics[width=\linewidth]{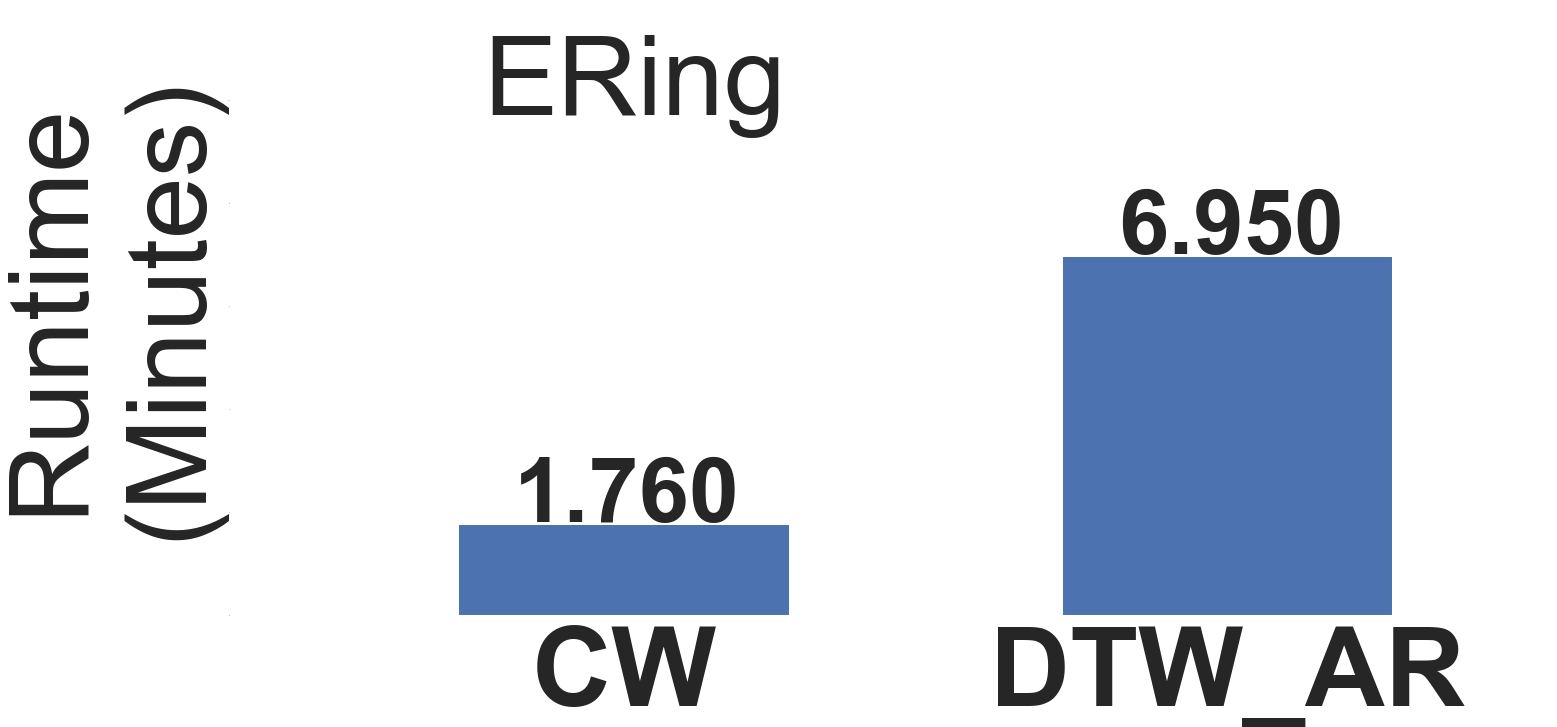}
            \end{minipage}%
            \hfill
        \begin{minipage}{.46\linewidth}
                \centering
                \includegraphics[width=\linewidth]{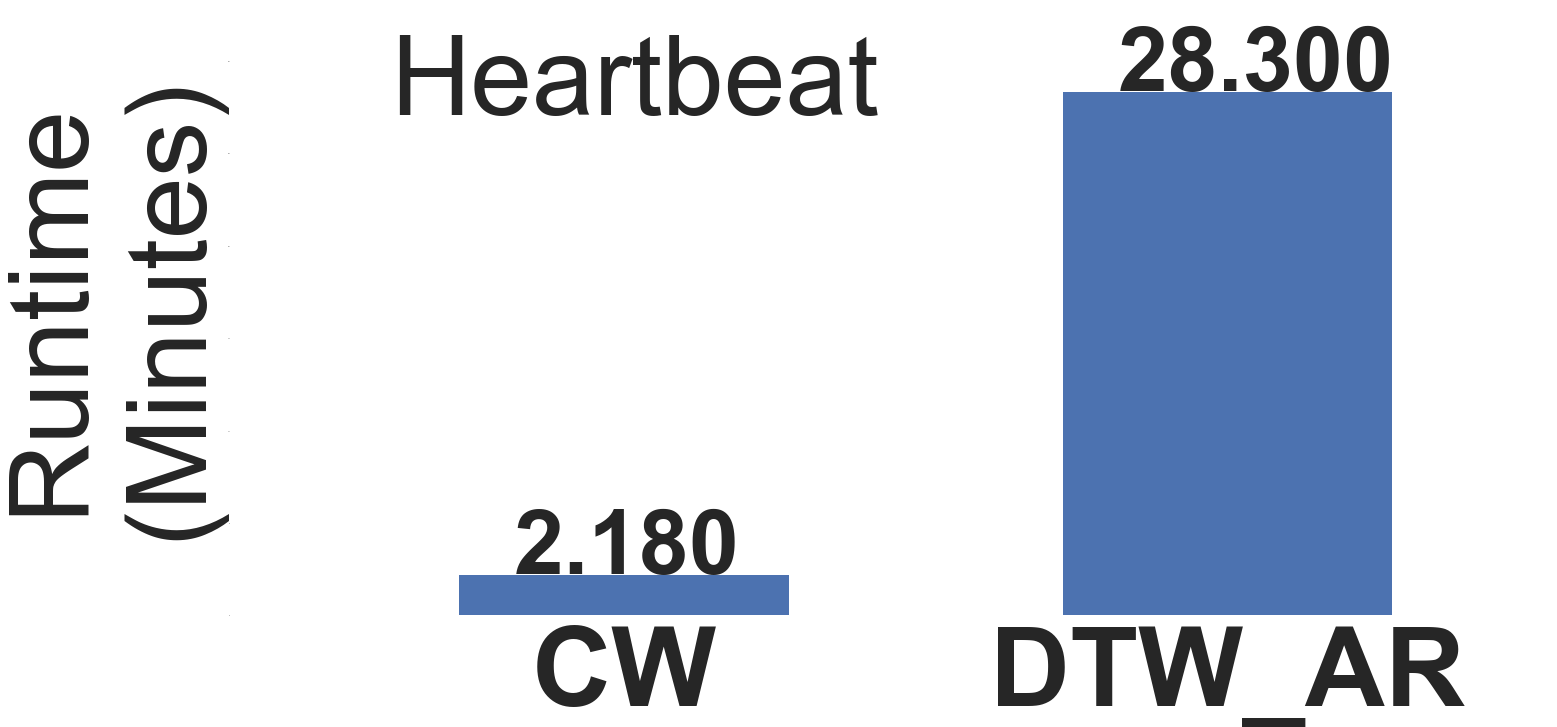}
            \end{minipage}
            \centering
        \begin{minipage}{.46\linewidth}
                \centering
                \includegraphics[width=\linewidth]{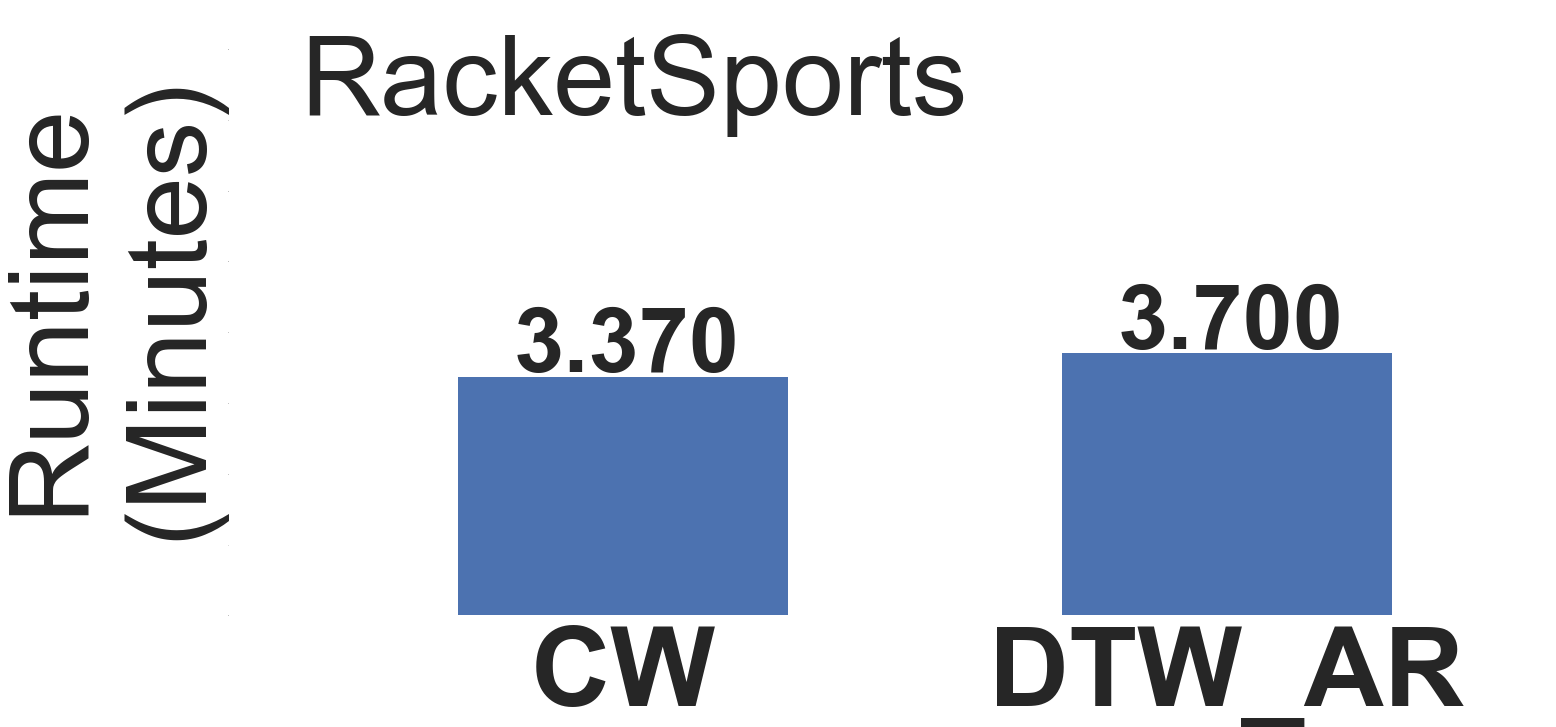}
            \end{minipage}
    \end{minipage}
    \caption{Average runtime for CW and DTW-AR to create one targeted adversarial example (run on NVIDIA Titan Xp GPU).}
    \label{fig:cwruntime}
\end{figure}

\noindent \textbf{Runtime comparison of DTW-AR vs. Carlini \& Wagner.} As explained in Section 3, one main advantage of DTW-AR is reducing the time complexity of using DTW to create adversarial examples. We provide in Figure \ref{fig:cwruntime} a comparison of the average runtime per iteration to create one targeted adversarial example by iterative baseline methods. We note that we only compare to CW because FGSM and PGD are not considered targeted attacks, and Karim et al. \cite{karim2020adversarial}, fails to create adversarial examples for every input. While we observe that CW is faster, we note that we have already demonstrated empirically (Figure \ref{fig:appendadvatk} and \ref{fig:cwadvatk}) that DTW-AR always outperforms CW in both effectiveness of adversarial examples and adversarial training. We also observe differences in the DTW-AR's runtime across datasets. DTW-AR is relatively quick for small-size data such as \textit{RacketSports} ($30\times6$) and slower for large-size data such as \textit{HeartBeat} ($405\times61$). The additional runtime cost is explained by the proposed loss function in Equation \ref{eq:dtwloss} that guarantees a highly-similar adversarial example. For future work, we aim to optimize the implementation of DTW-AR to further reduce the runtime on large time-series datasets.

\begin{figure*}[t]
    \begin{minipage}{\linewidth}
    \centering
        \begin{minipage}{0.19\linewidth}
            \centering
            \includegraphics[width=\linewidth]{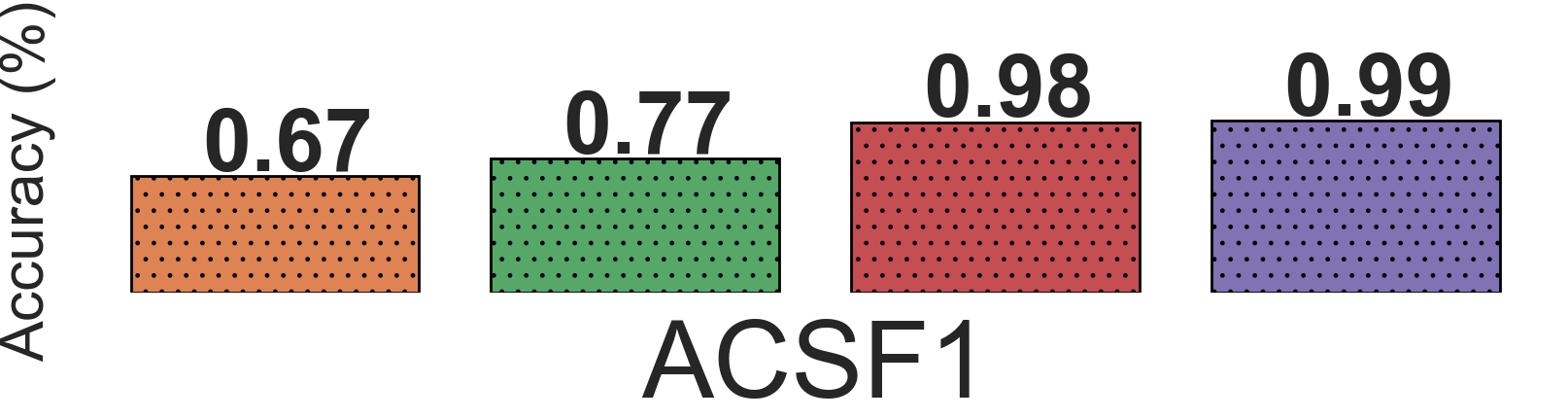}
        \end{minipage}%
\hfill 
        \begin{minipage}{0.19\linewidth}
            \centering
            \includegraphics[width=\linewidth]{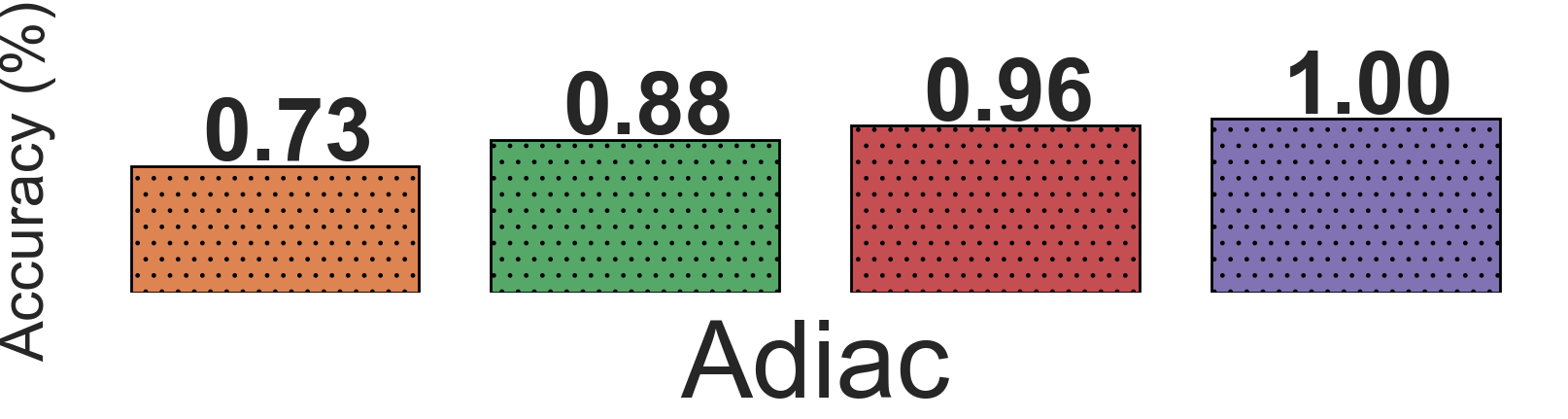}
        \end{minipage}%
\hfill 
        \begin{minipage}{0.19\linewidth}
            \centering
            \includegraphics[width=\linewidth]{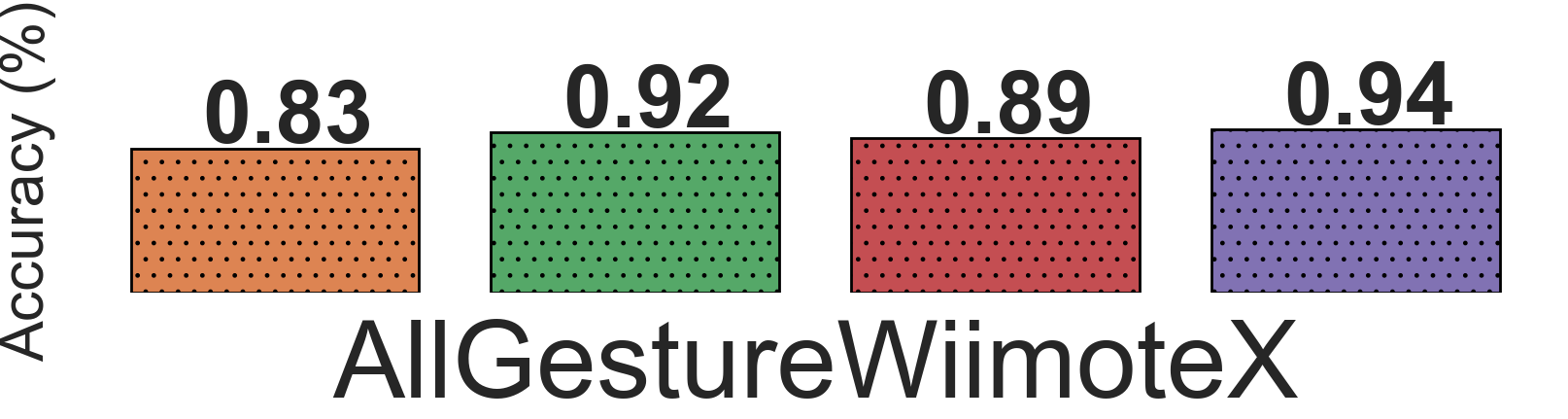}
        \end{minipage}%
\hfill 
        \begin{minipage}{0.19\linewidth}
            \centering
            \includegraphics[width=\linewidth]{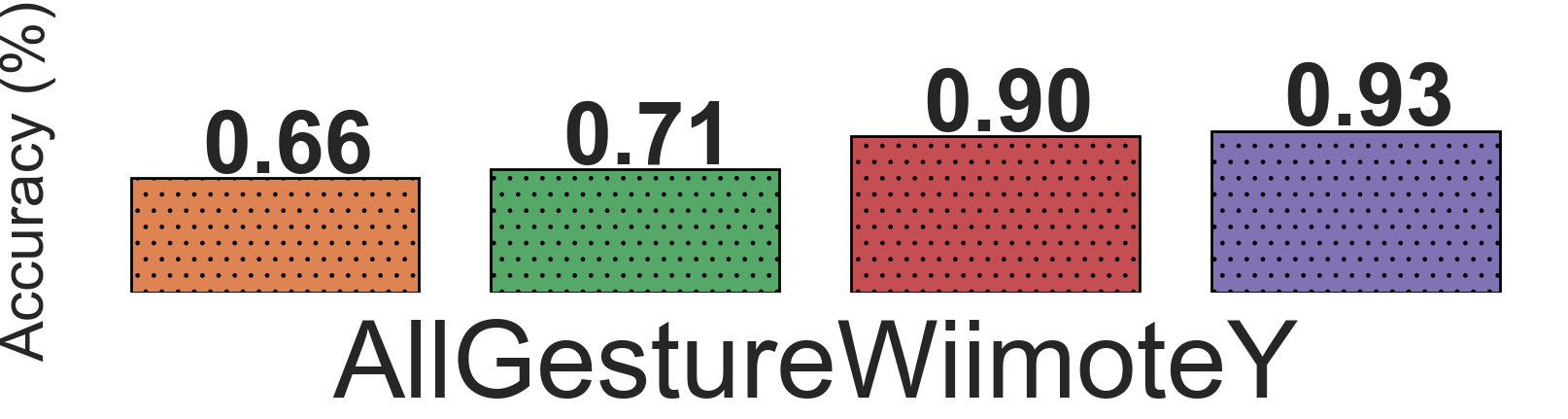}
        \end{minipage}%
\hfill 
        \begin{minipage}{0.19\linewidth}
            \centering
            \includegraphics[width=\linewidth]{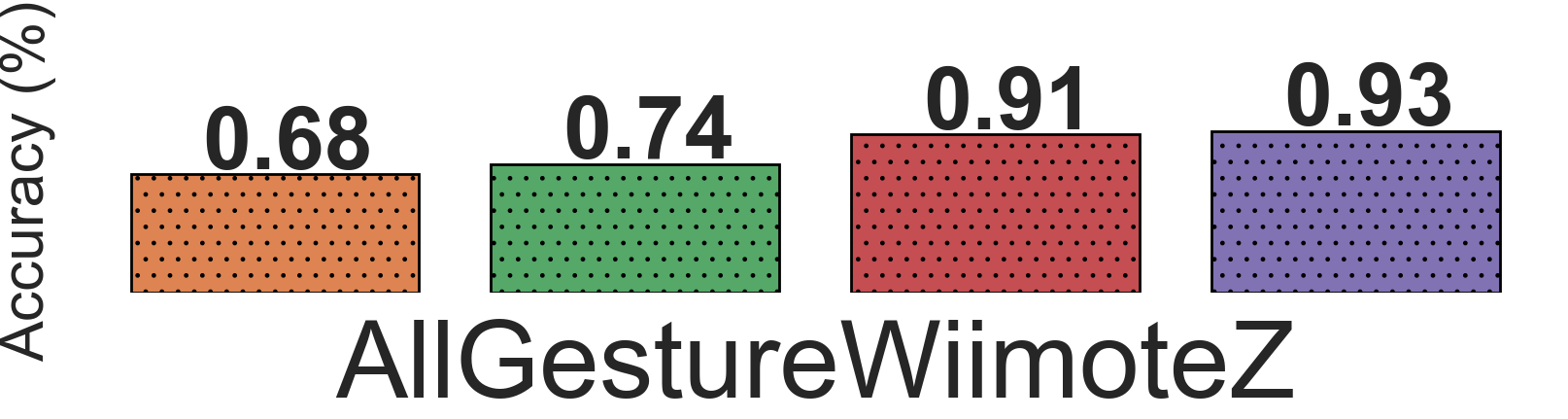}
        \end{minipage}
        \begin{minipage}{0.19\linewidth}
            \centering
            \includegraphics[width=\linewidth]{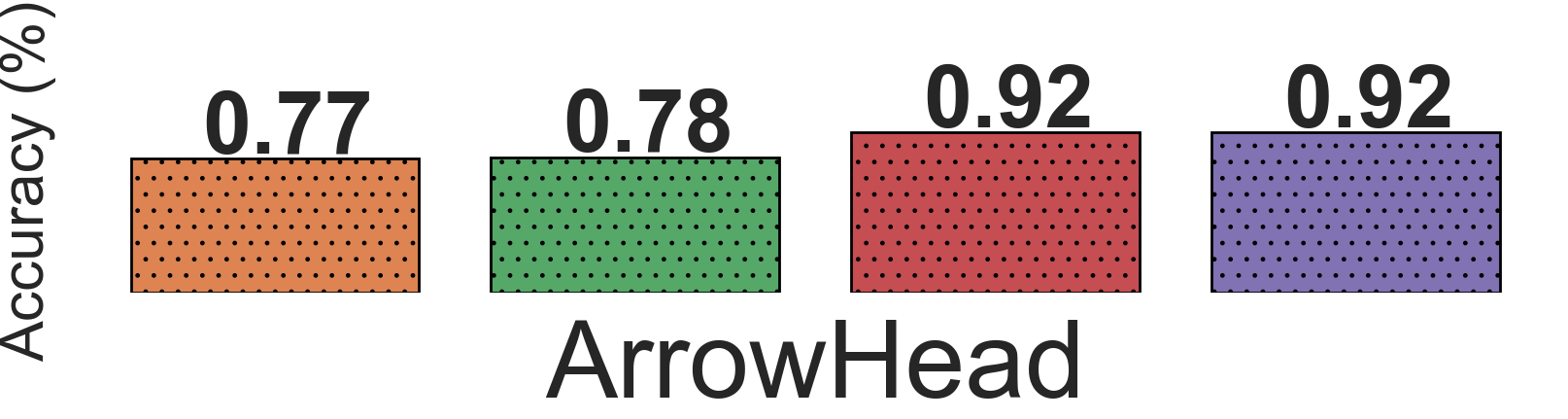}
        \end{minipage}%
\hfill 
        \begin{minipage}{0.19\linewidth}
            \centering
            \includegraphics[width=\linewidth]{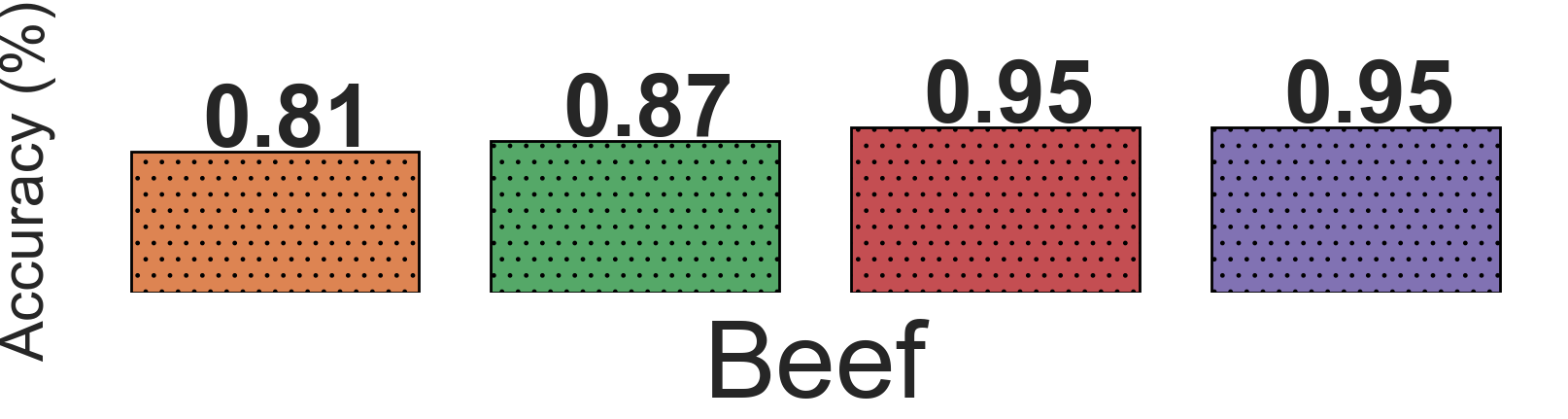}
        \end{minipage}%
\hfill 
        \begin{minipage}{0.19\linewidth}
            \centering
            \includegraphics[width=\linewidth]{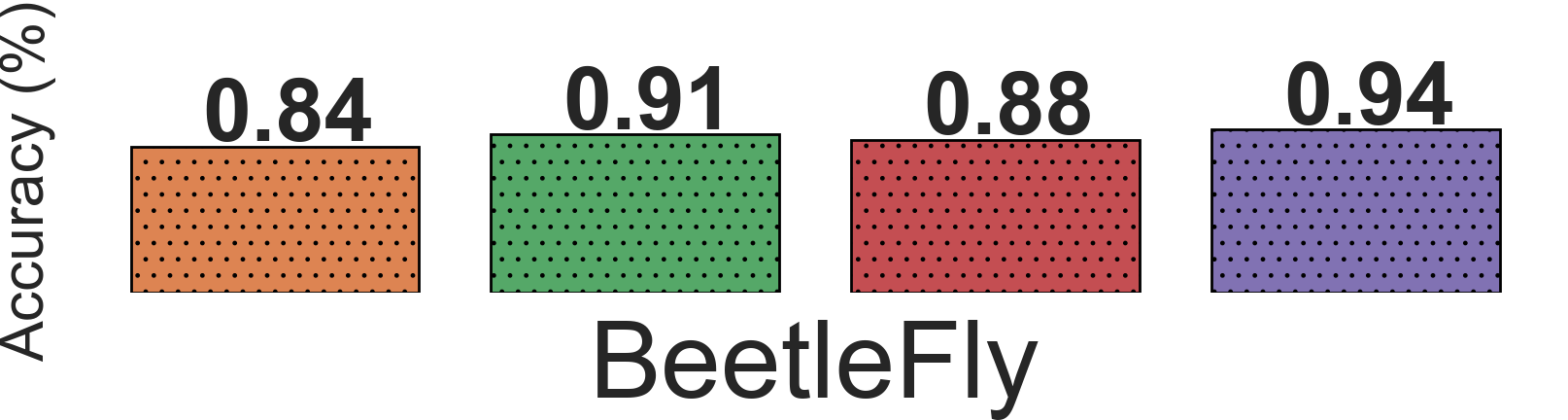}
        \end{minipage}%
\hfill 
        \begin{minipage}{0.19\linewidth}
            \centering
            \includegraphics[width=\linewidth]{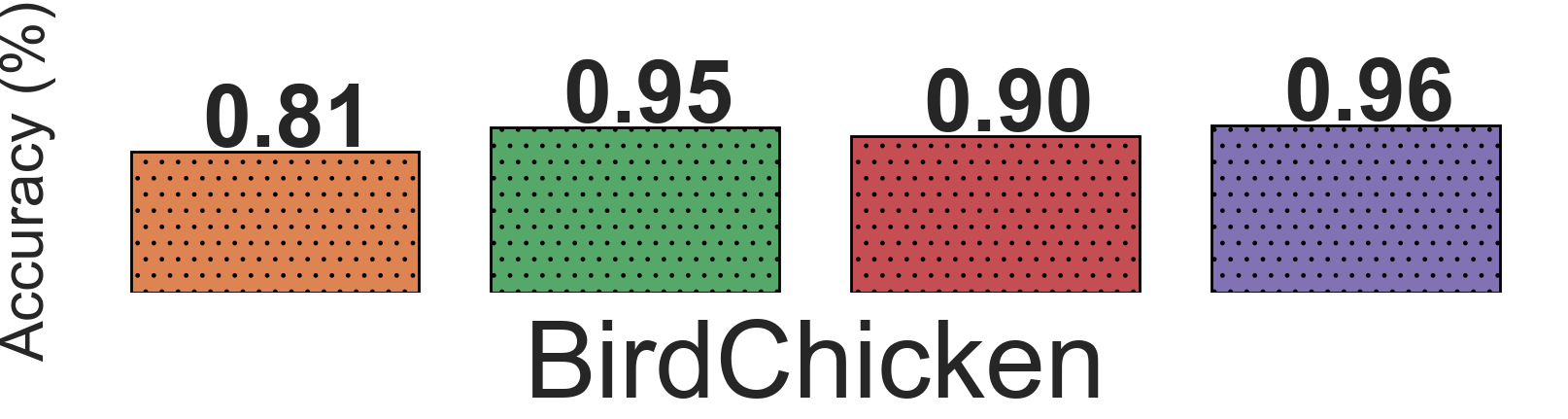}
        \end{minipage}%
\hfill 
        \begin{minipage}{0.19\linewidth}
            \centering
            \includegraphics[width=\linewidth]{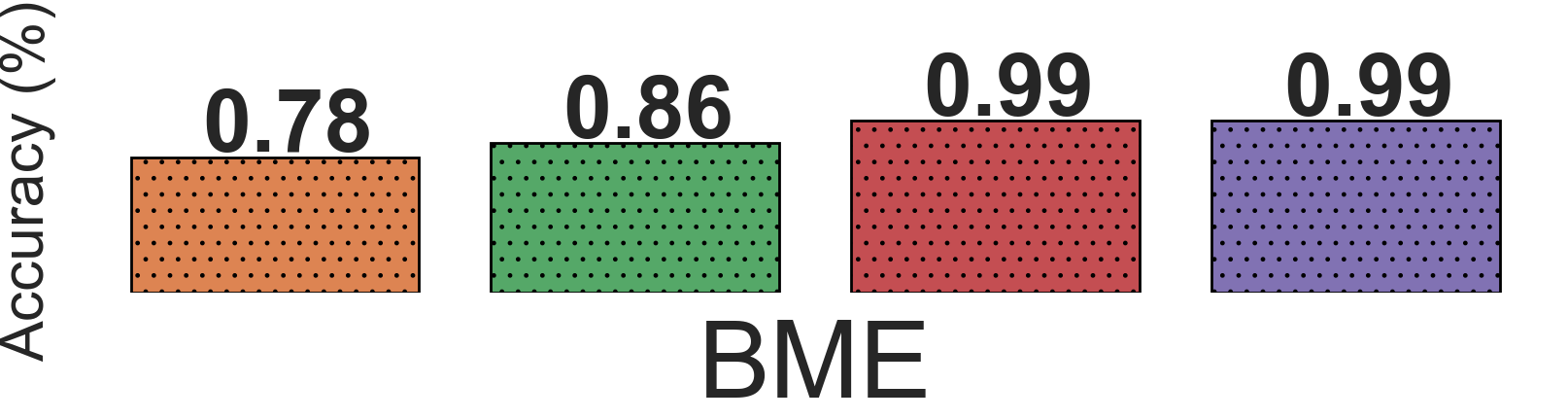}
        \end{minipage}
        \begin{minipage}{0.19\linewidth}
            \centering
            \includegraphics[width=\linewidth]{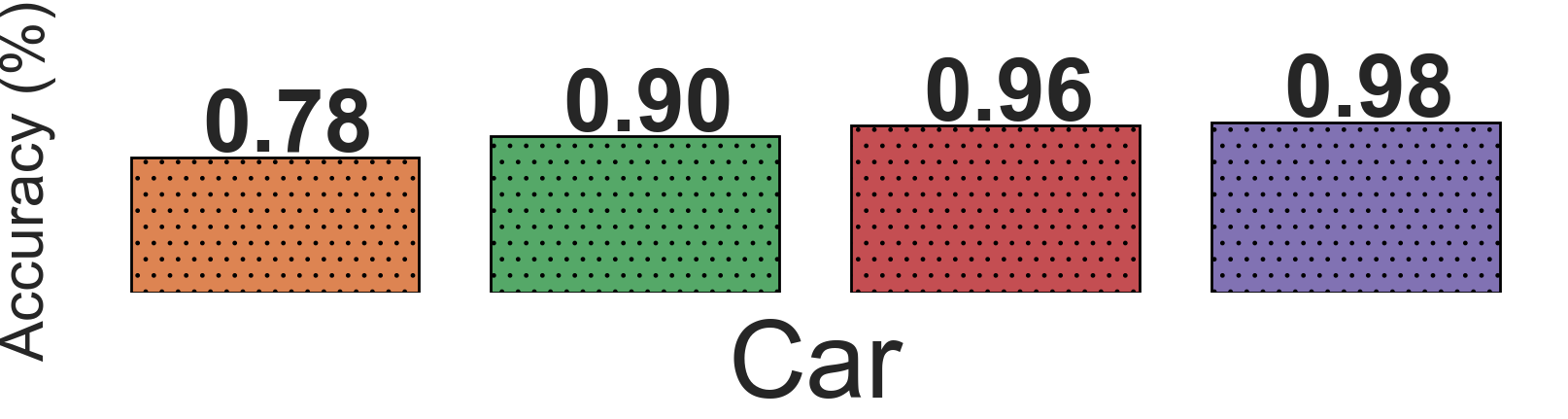}
        \end{minipage}%
\hfill 
        \begin{minipage}{0.19\linewidth}
            \centering
            \includegraphics[width=\linewidth]{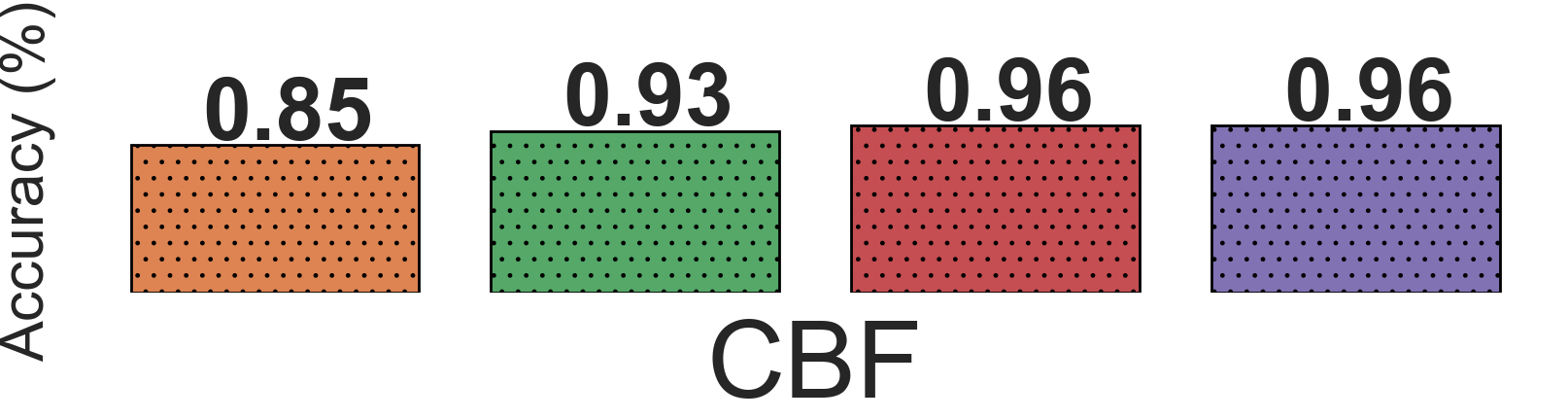}
        \end{minipage}%
\hfill 
        \begin{minipage}{0.19\linewidth}
            \centering
            \includegraphics[width=\linewidth]{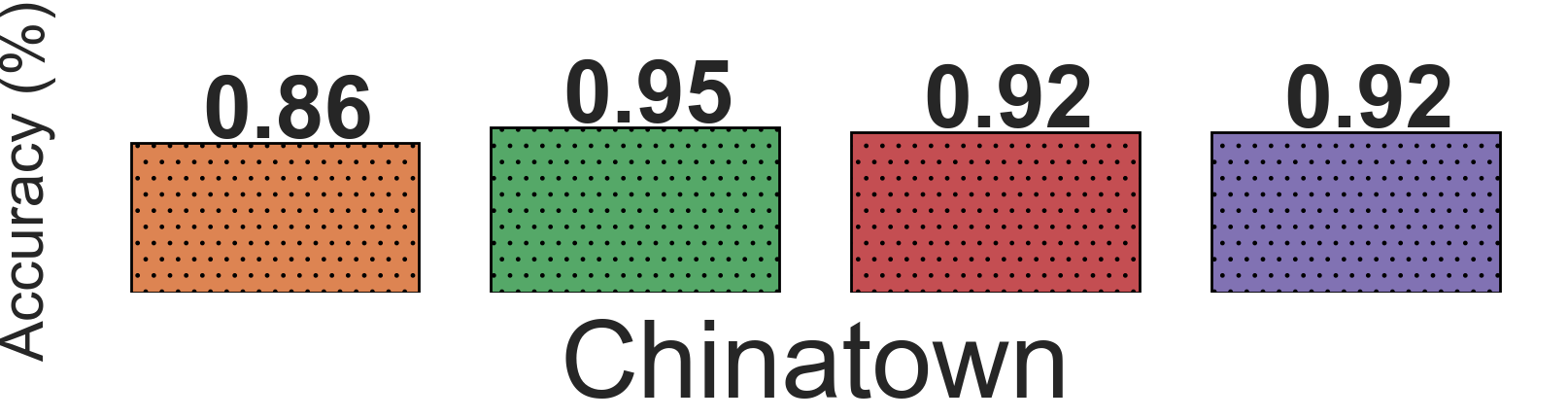}
        \end{minipage}%
\hfill 
        \begin{minipage}{0.19\linewidth}
            \centering
            \includegraphics[width=\linewidth]{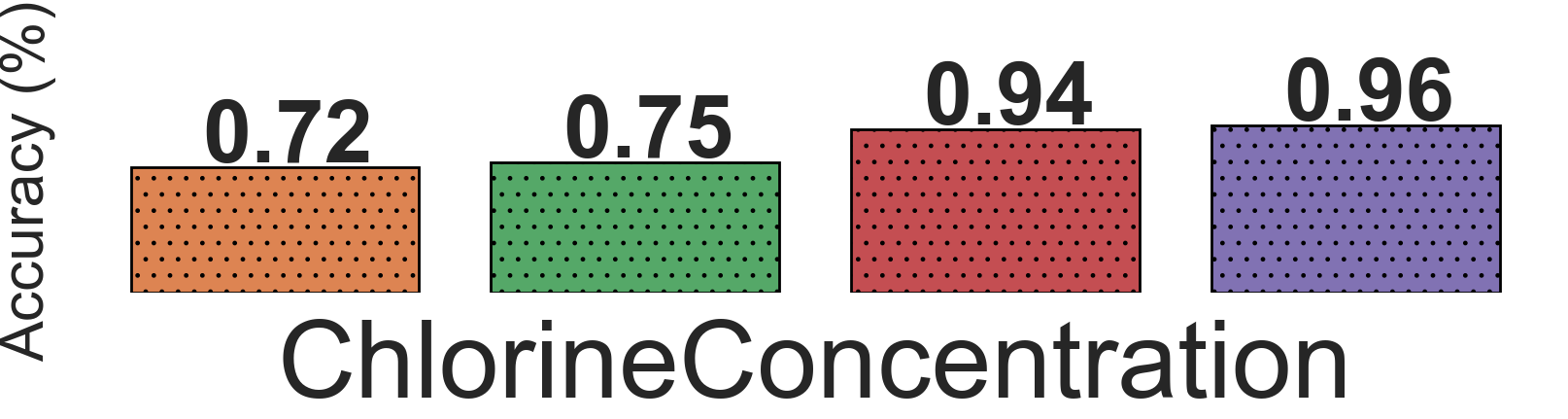}
        \end{minipage}%
\hfill 
        \begin{minipage}{0.19\linewidth}
            \centering
            \includegraphics[width=\linewidth]{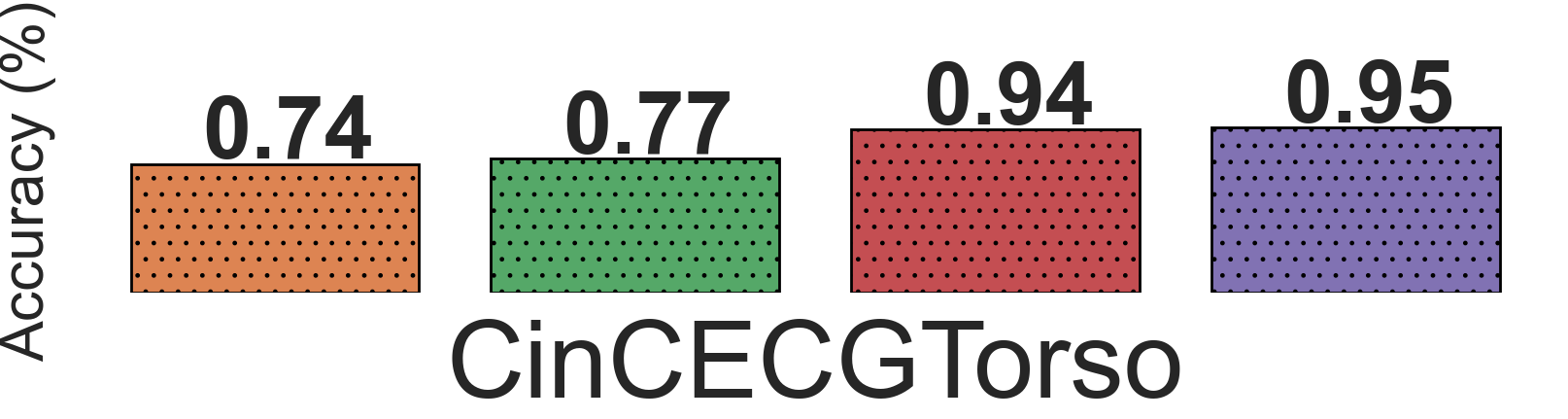}
        \end{minipage}
        \begin{minipage}{0.19\linewidth}
            \centering
            \includegraphics[width=\linewidth]{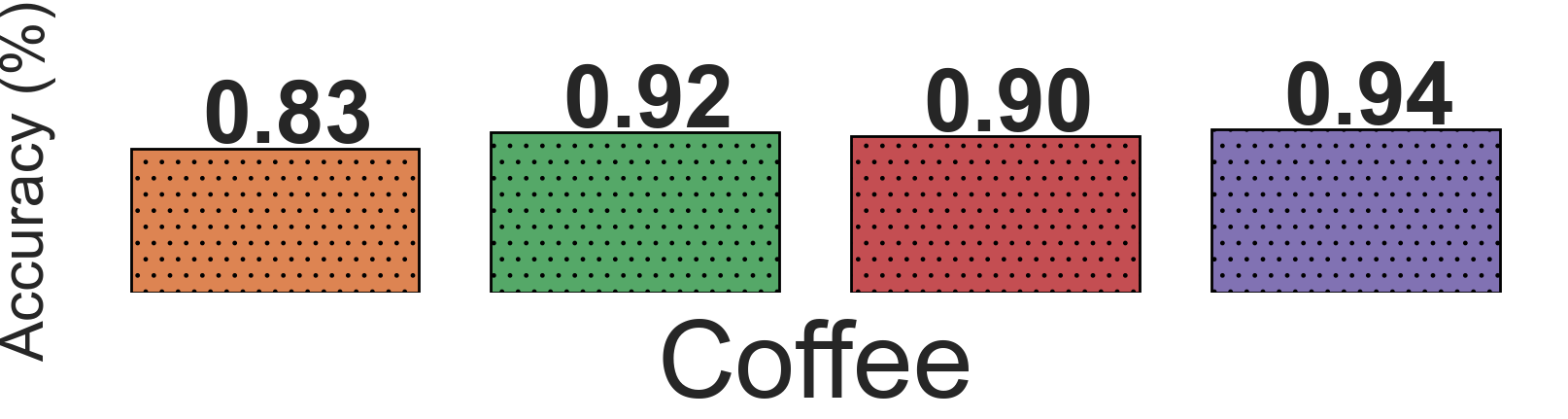}
        \end{minipage}%
\hfill 
        \begin{minipage}{0.19\linewidth}
            \centering
            \includegraphics[width=\linewidth]{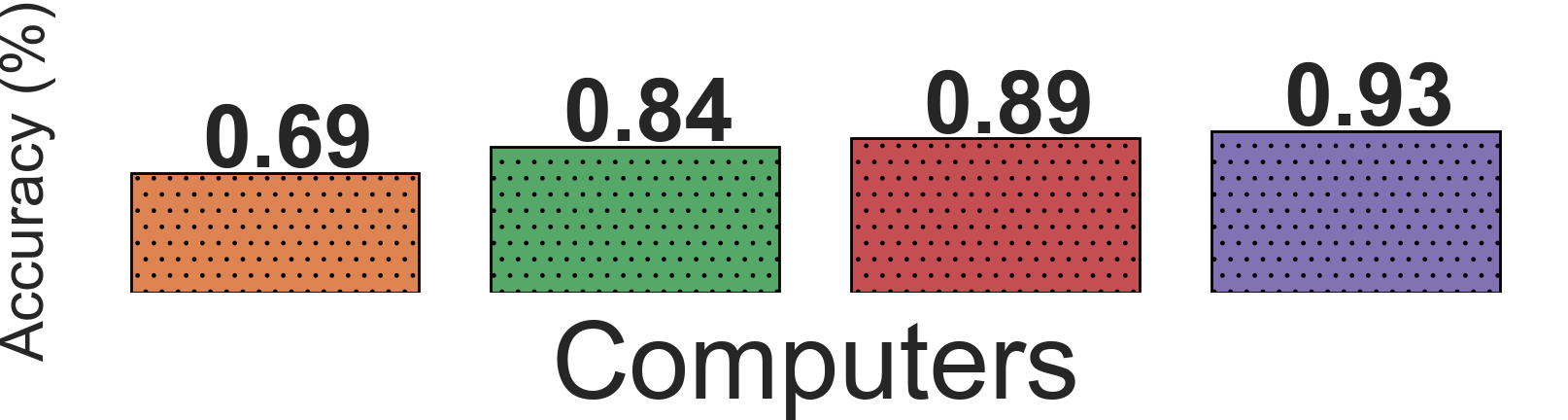}
        \end{minipage}%
\hfill 
        \begin{minipage}{0.19\linewidth}
            \centering
            \includegraphics[width=\linewidth]{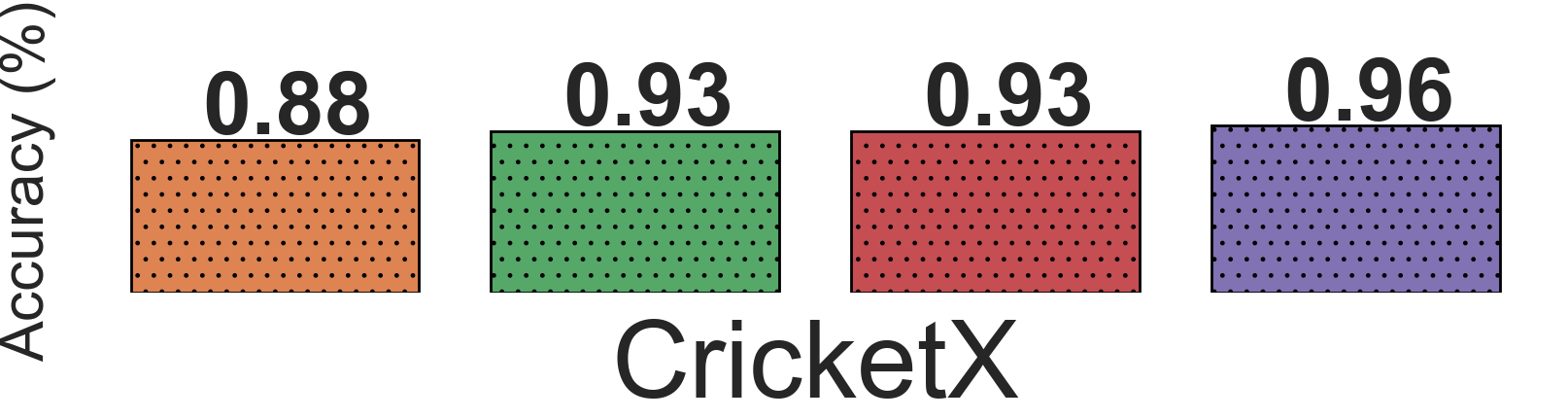}
        \end{minipage}%
\hfill 
        \begin{minipage}{0.19\linewidth}
            \centering
            \includegraphics[width=\linewidth]{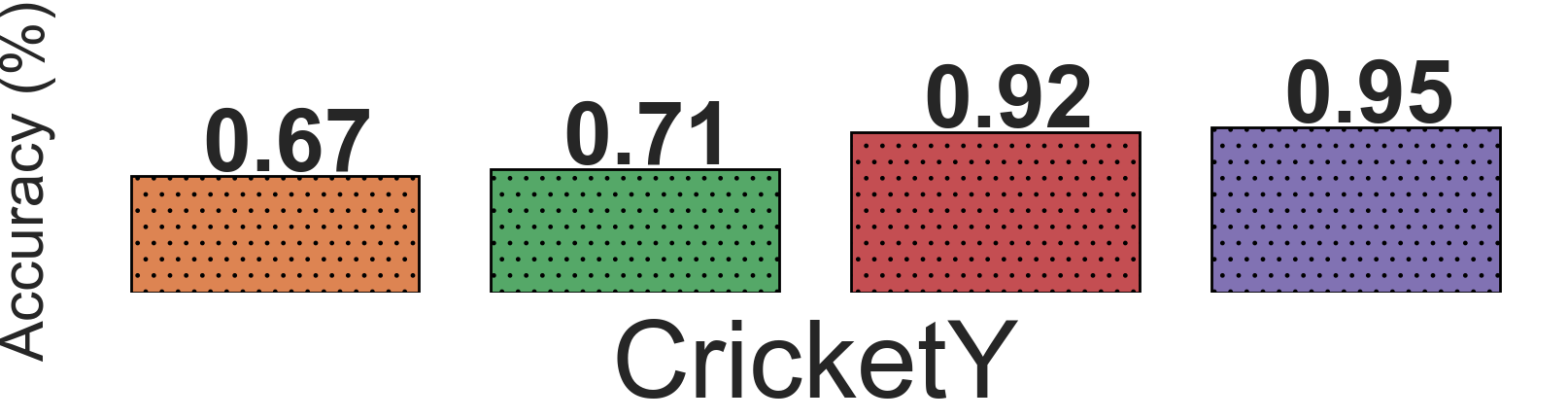}
        \end{minipage}%
\hfill 
        \begin{minipage}{0.19\linewidth}
            \centering
            \includegraphics[width=\linewidth]{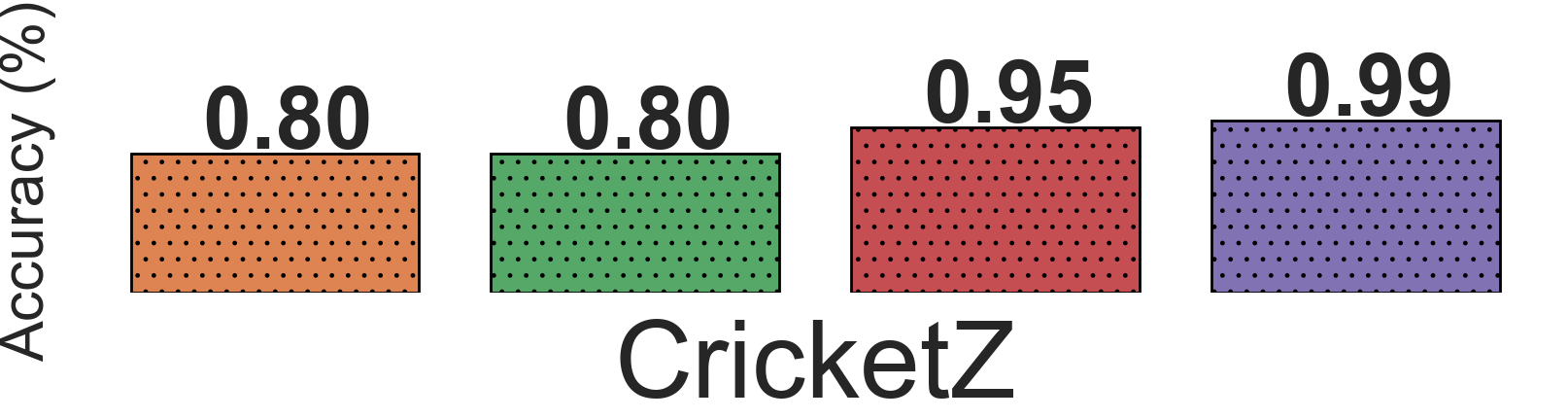}
        \end{minipage}
        \begin{minipage}{0.19\linewidth}
            \centering
            \includegraphics[width=\linewidth]{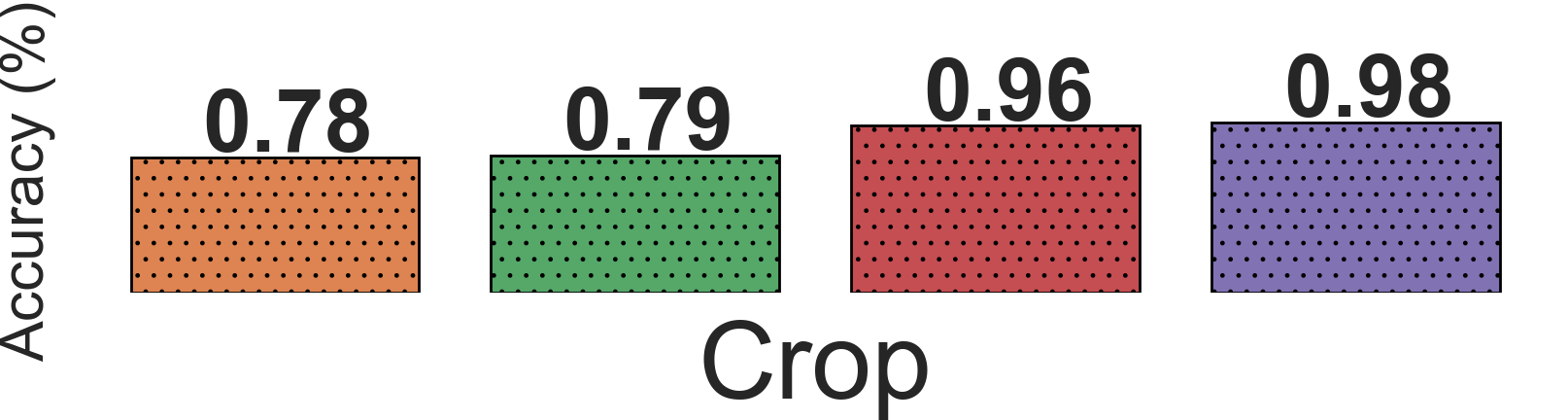}
        \end{minipage}%
\hfill 
        \begin{minipage}{0.19\linewidth}
            \centering
            \includegraphics[width=\linewidth]{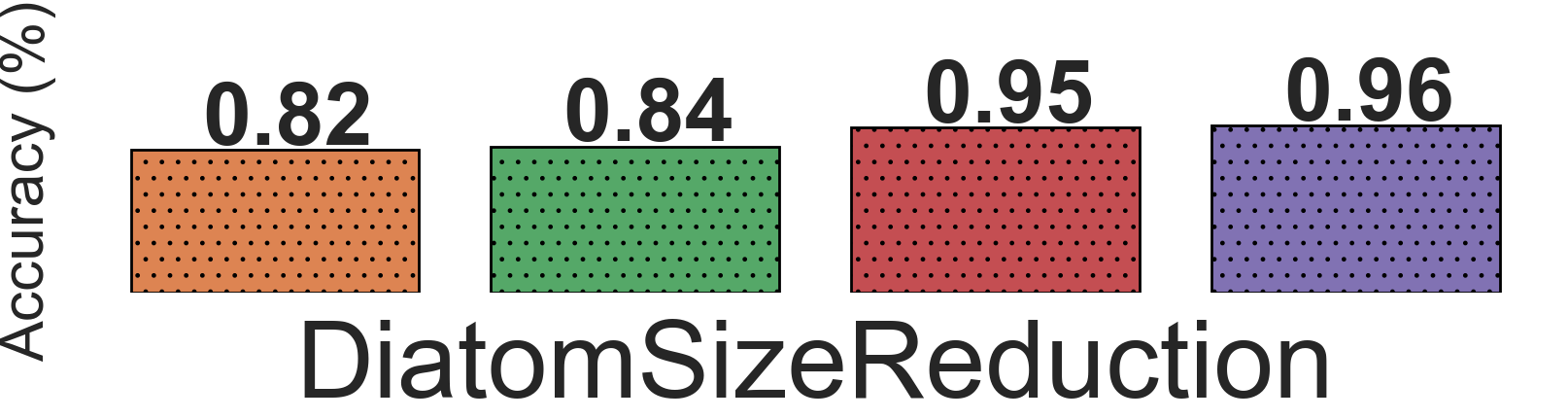}
        \end{minipage}%
\hfill 
        \begin{minipage}{0.19\linewidth}
            \centering
            \includegraphics[width=\linewidth]{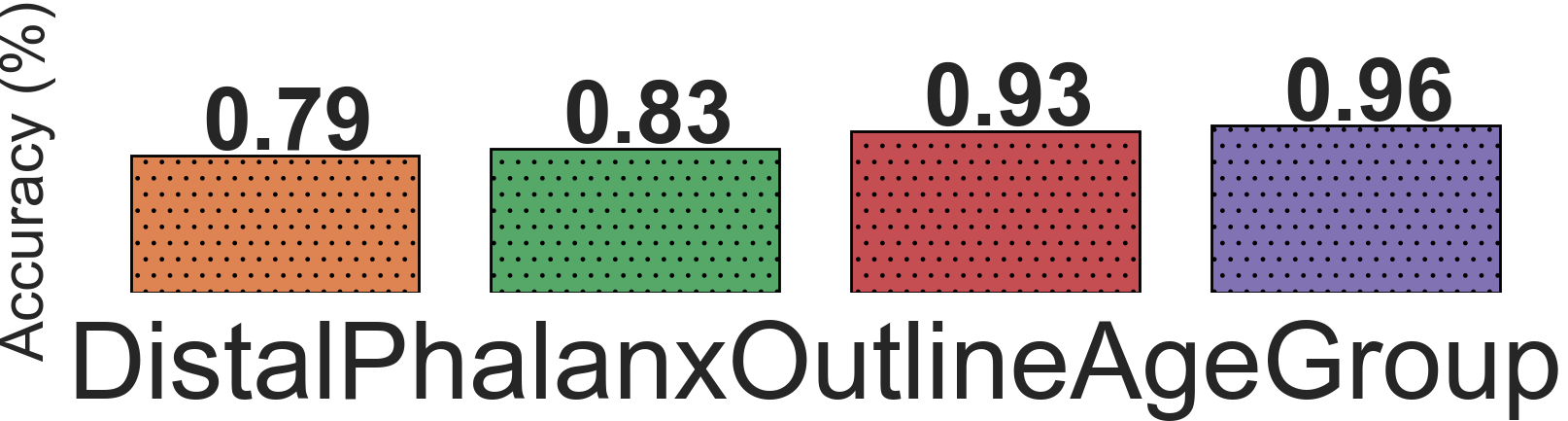}
        \end{minipage}%
\hfill 
        \begin{minipage}{0.19\linewidth}
            \centering
            \includegraphics[width=\linewidth]{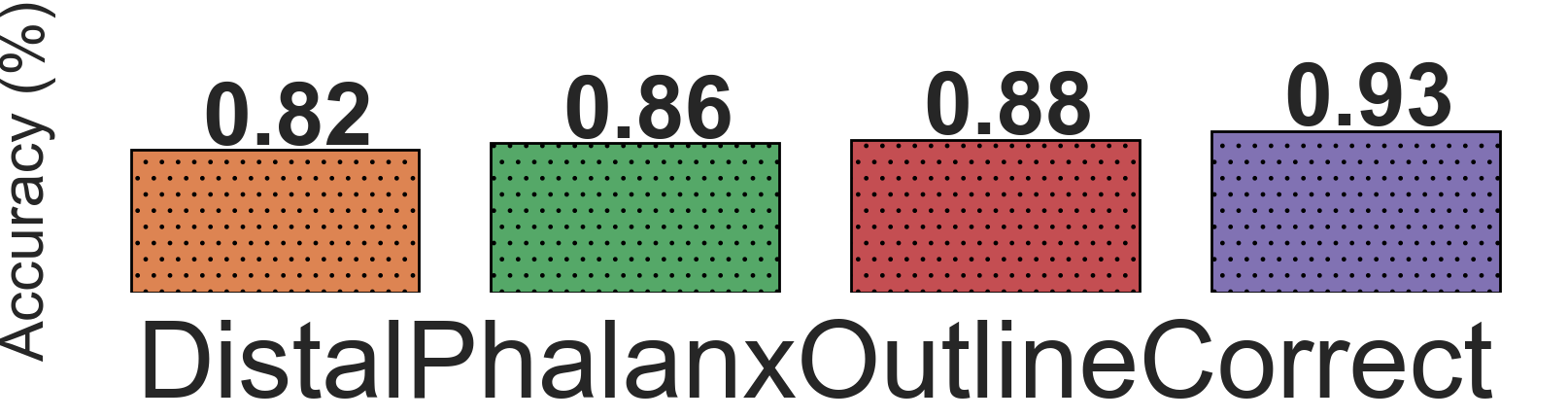}
        \end{minipage}%
\hfill 
        \begin{minipage}{0.19\linewidth}
            \centering
            \includegraphics[width=\linewidth]{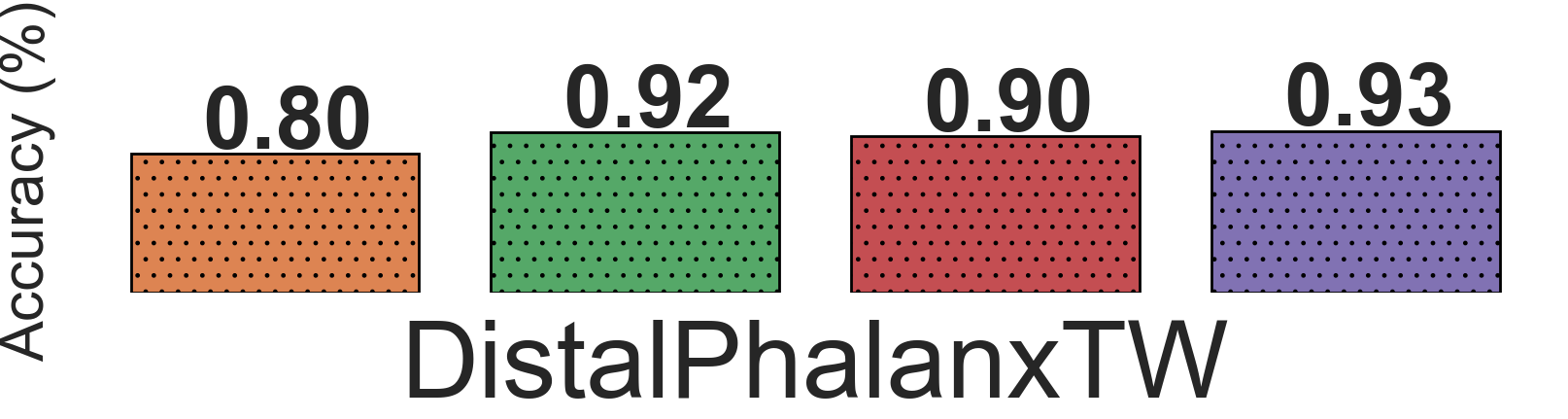}
        \end{minipage}
        \begin{minipage}{0.19\linewidth}
            \centering
            \includegraphics[width=\linewidth]{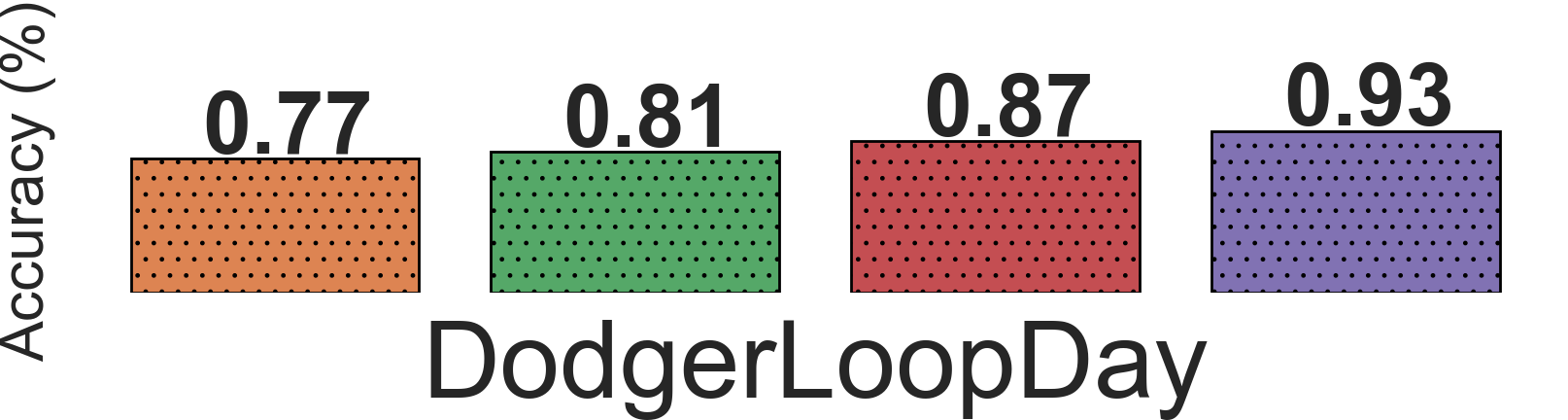}
        \end{minipage}%
\hfill 
        \begin{minipage}{0.19\linewidth}
            \centering
            \includegraphics[width=\linewidth]{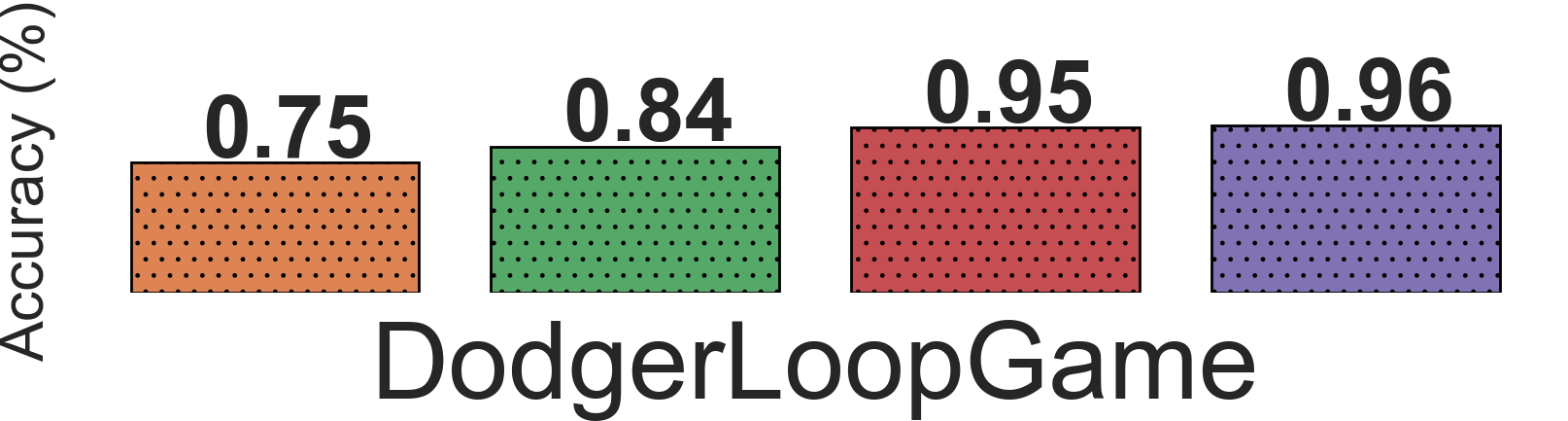}
        \end{minipage}%
\hfill 
        \begin{minipage}{0.19\linewidth}
            \centering
            \includegraphics[width=\linewidth]{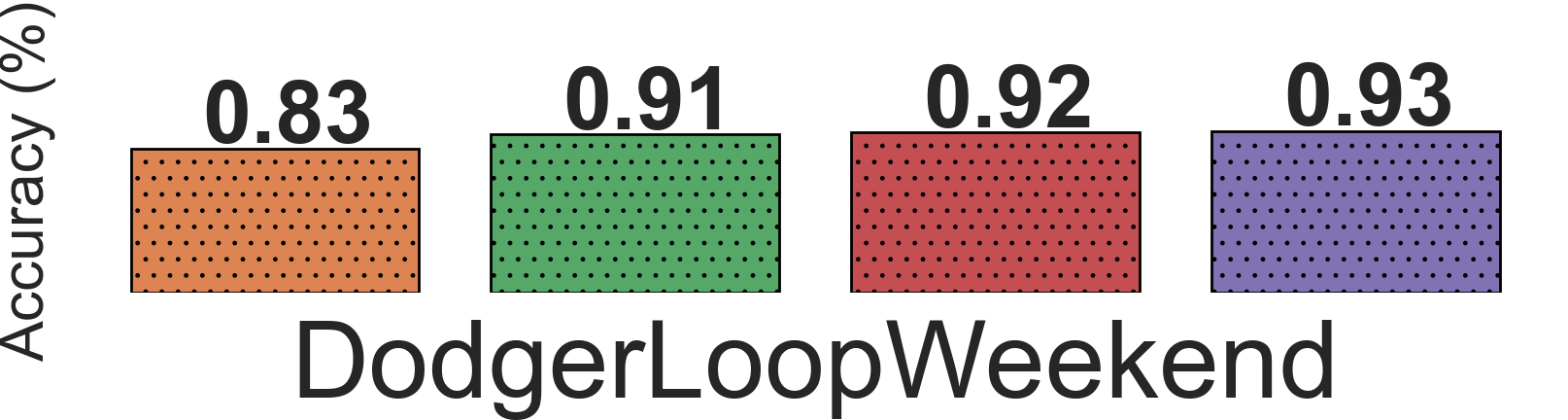}
        \end{minipage}%
\hfill 
        \begin{minipage}{0.19\linewidth}
            \centering
            \includegraphics[width=\linewidth]{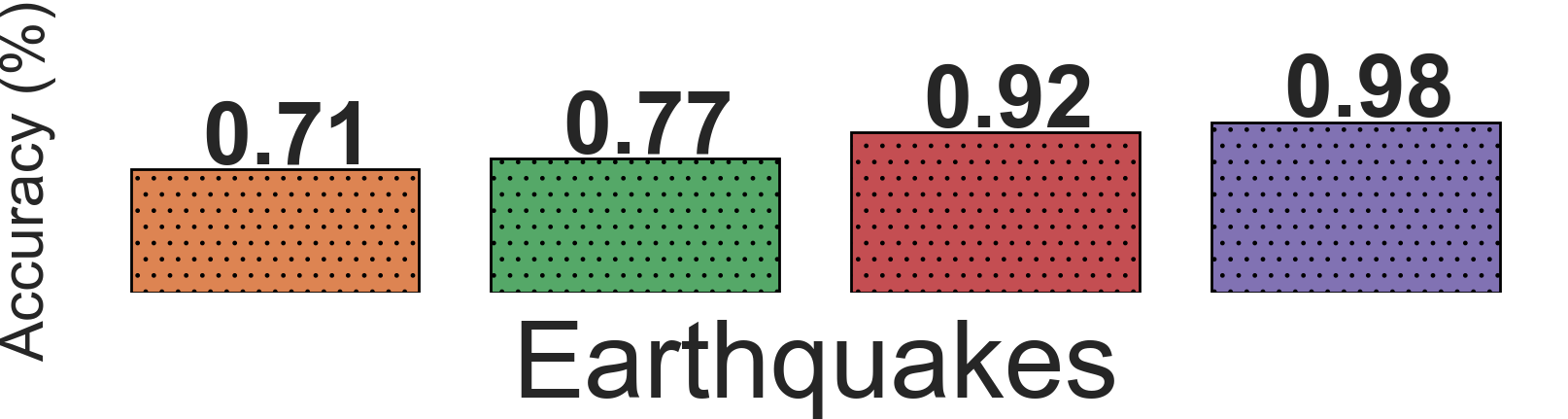}
        \end{minipage}%
\hfill 
        \begin{minipage}{0.19\linewidth}
            \centering
            \includegraphics[width=\linewidth]{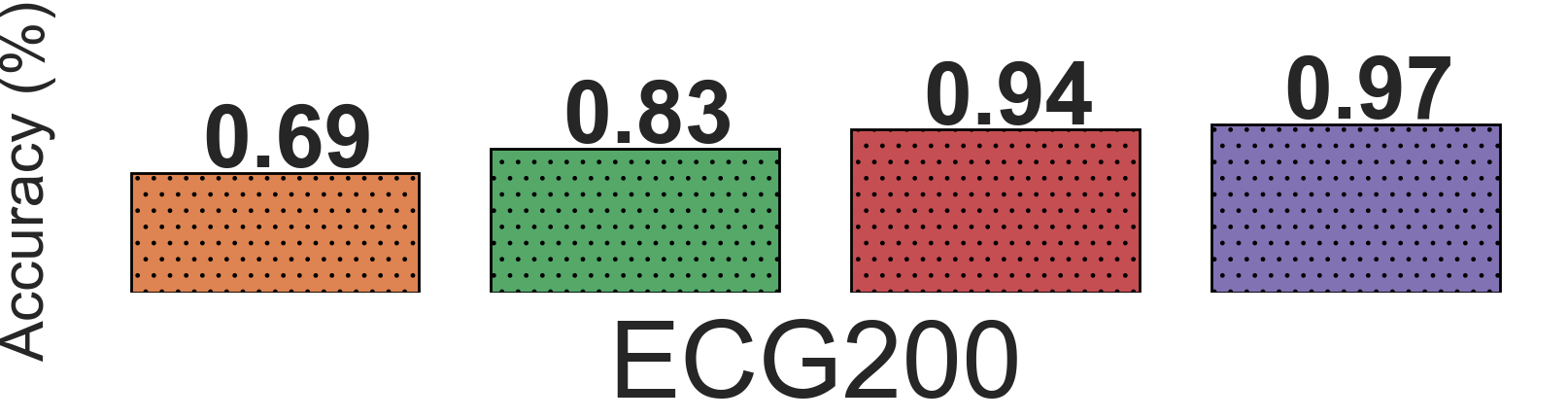}
        \end{minipage}
        \begin{minipage}{0.19\linewidth}
            \centering
            \includegraphics[width=\linewidth]{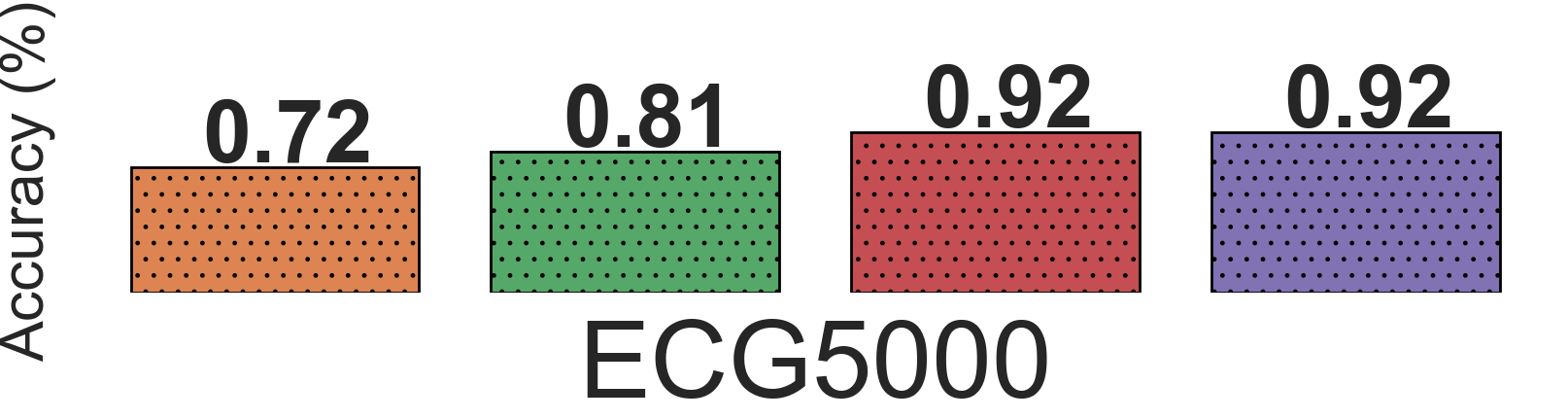}
        \end{minipage}%
\hfill 
        \begin{minipage}{0.19\linewidth}
            \centering
            \includegraphics[width=\linewidth]{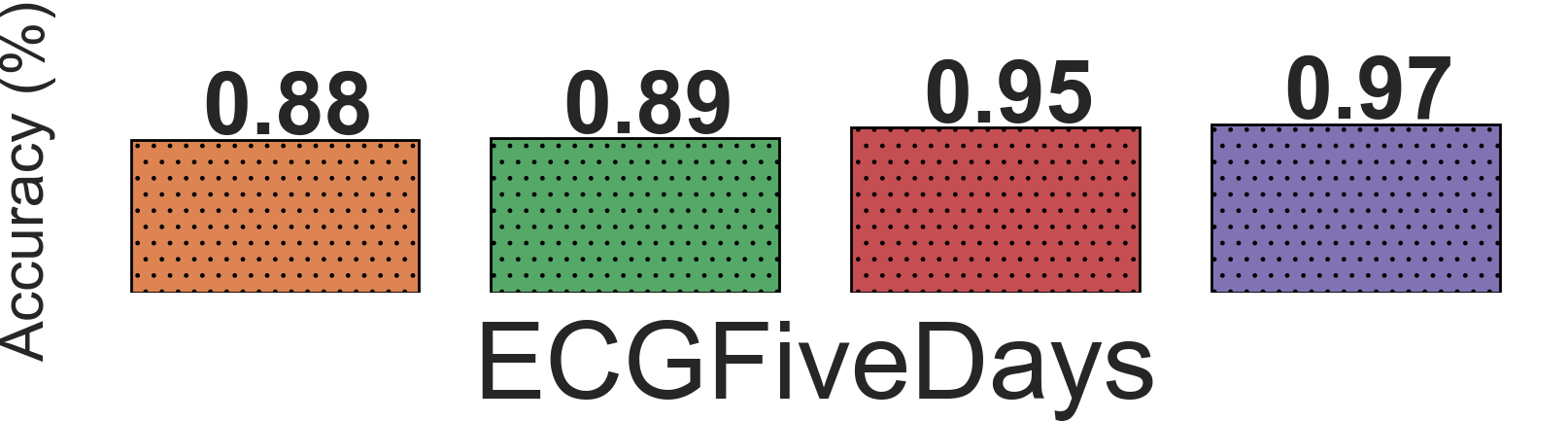}
        \end{minipage}%
\hfill 
        \begin{minipage}{0.19\linewidth}
            \centering
            \includegraphics[width=\linewidth]{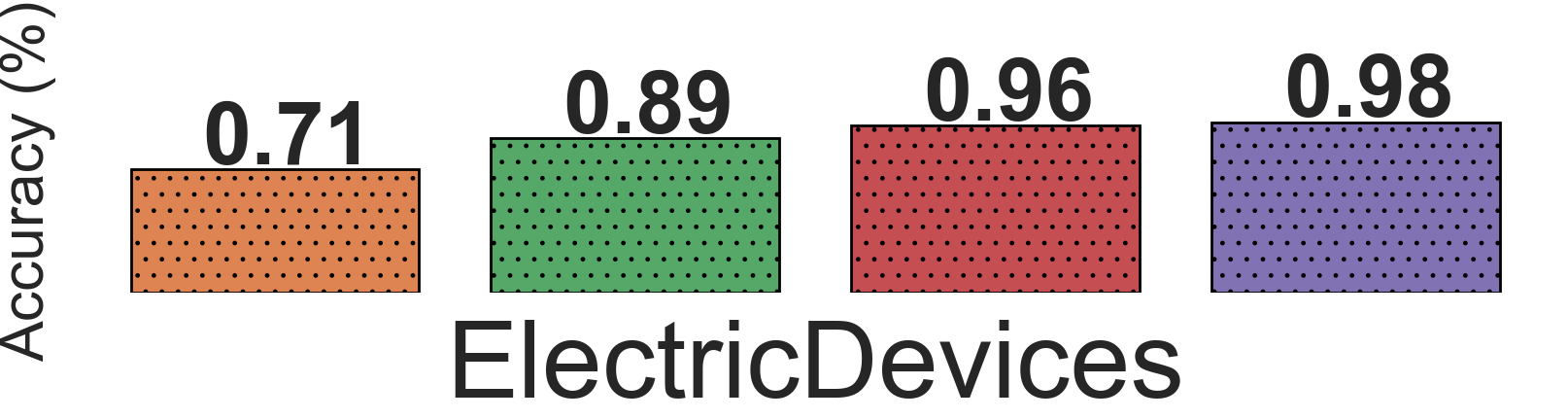}
        \end{minipage}%
\hfill 
        \begin{minipage}{0.19\linewidth}
            \centering
            \includegraphics[width=\linewidth]{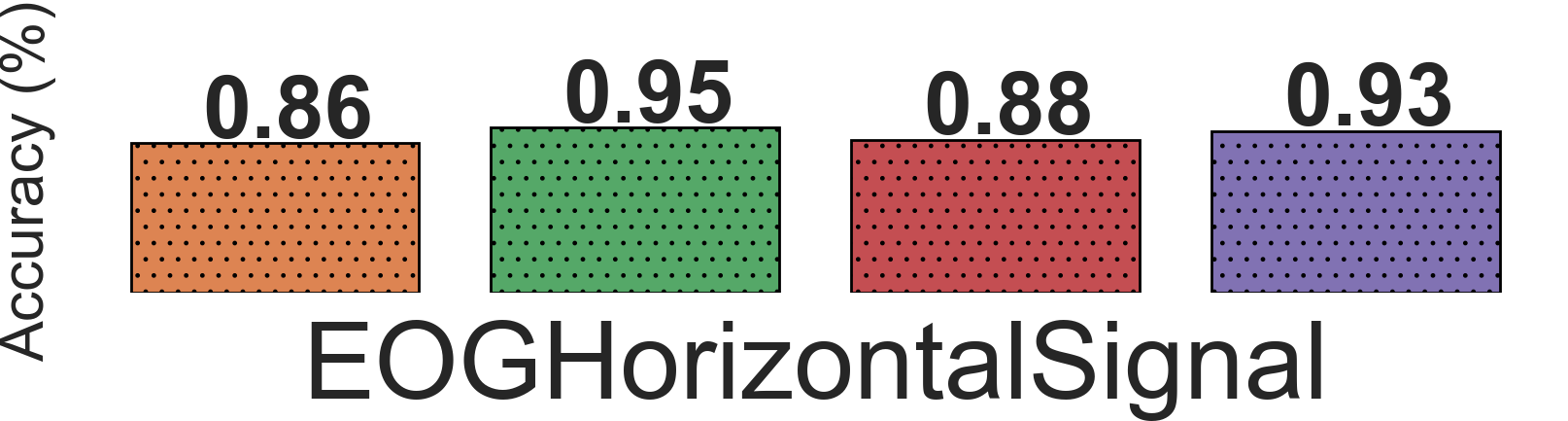}
        \end{minipage}%
\hfill 
        \begin{minipage}{0.19\linewidth}
            \centering
            \includegraphics[width=\linewidth]{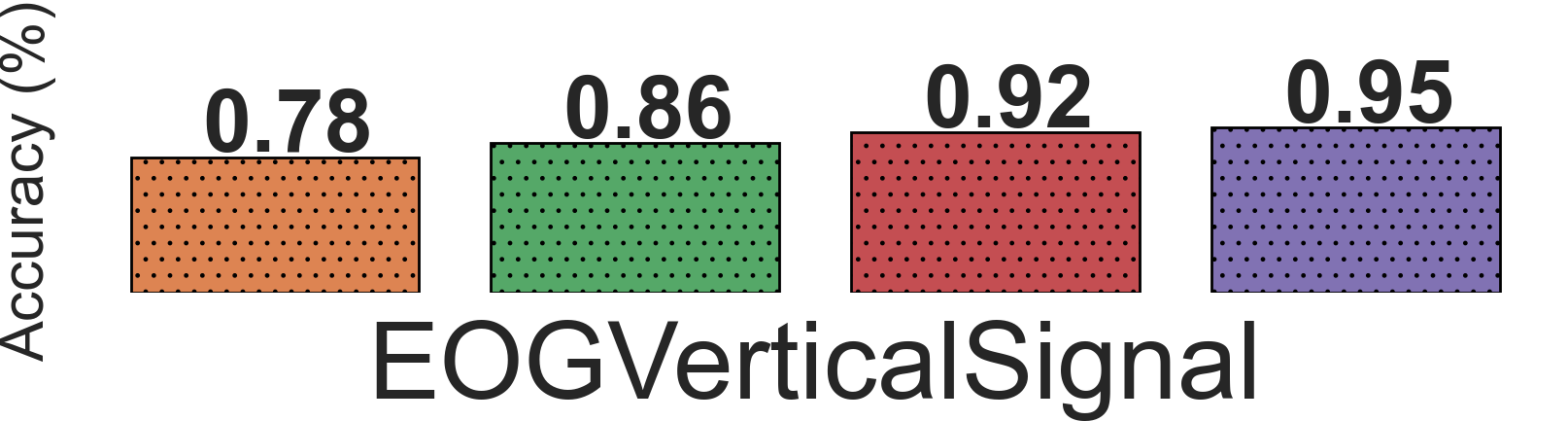}
        \end{minipage}
        \begin{minipage}{0.19\linewidth}
            \centering
            \includegraphics[width=\linewidth]{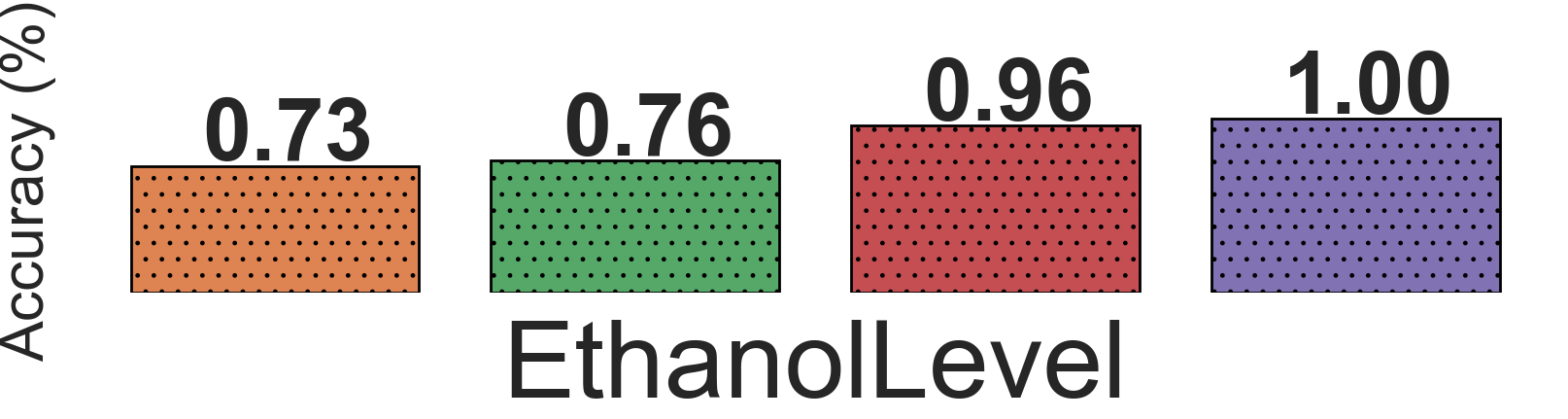}
        \end{minipage}%
\hfill 
        \begin{minipage}{0.19\linewidth}
            \centering
            \includegraphics[width=\linewidth]{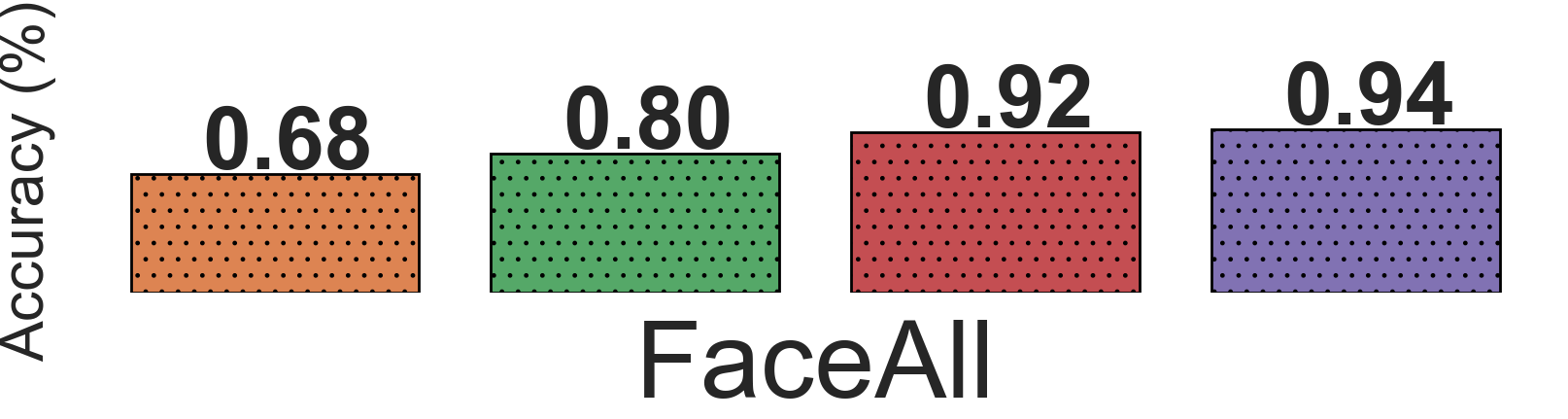}
        \end{minipage}%
\hfill 
        \begin{minipage}{0.19\linewidth}
            \centering
            \includegraphics[width=\linewidth]{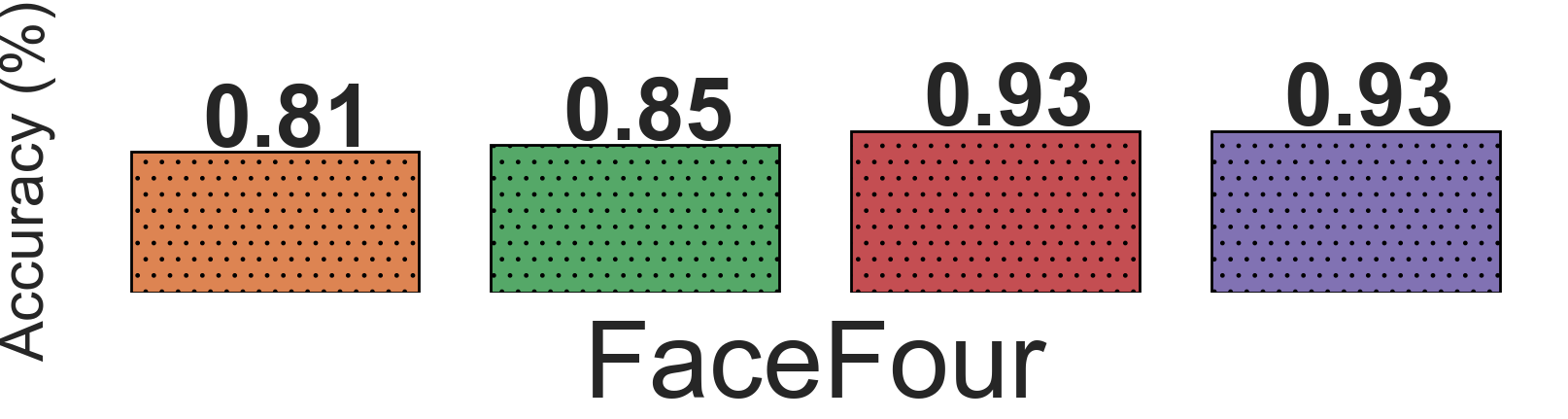}
        \end{minipage}%
\hfill 
        \begin{minipage}{0.19\linewidth}
            \centering
            \includegraphics[width=\linewidth]{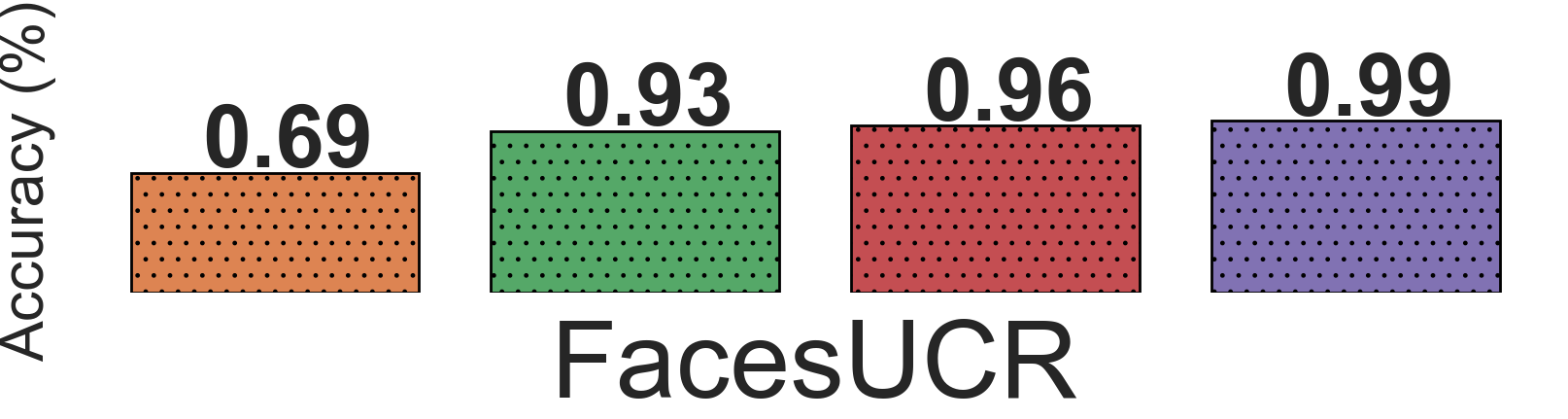}
        \end{minipage}%
\hfill 
        \begin{minipage}{0.19\linewidth}
            \centering
            \includegraphics[width=\linewidth]{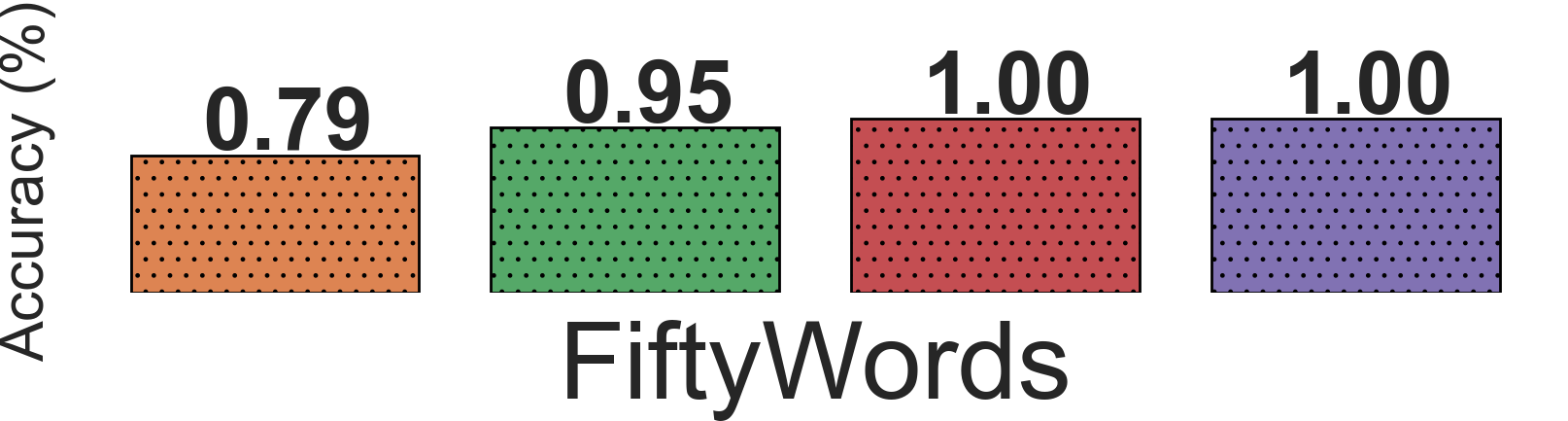}
        \end{minipage}
        \begin{minipage}{0.19\linewidth}
            \centering
            \includegraphics[width=\linewidth]{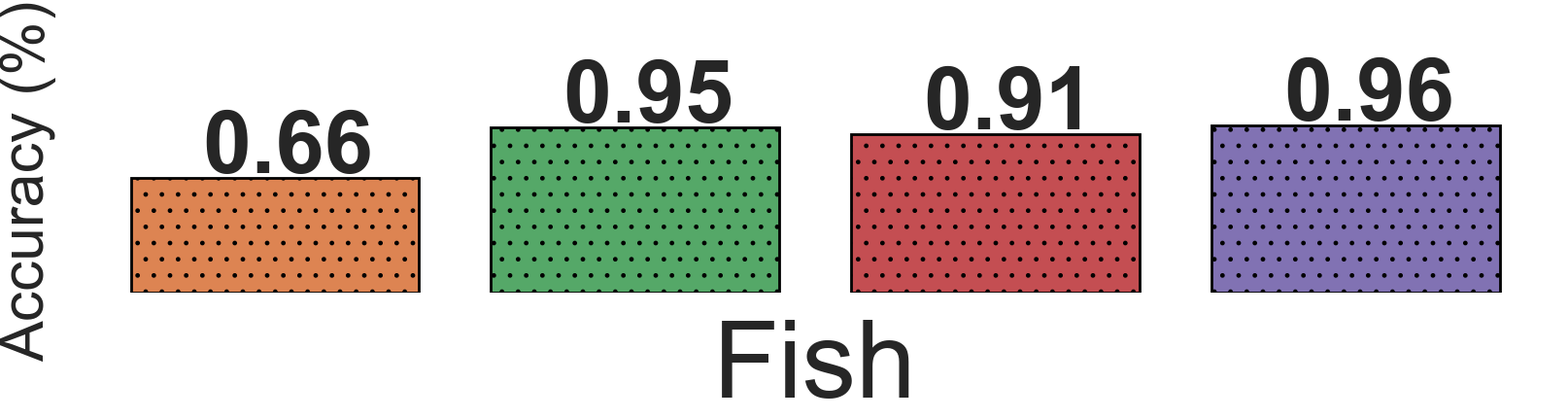}
        \end{minipage}%
\hfill 
        \begin{minipage}{0.19\linewidth}
            \centering
            \includegraphics[width=\linewidth]{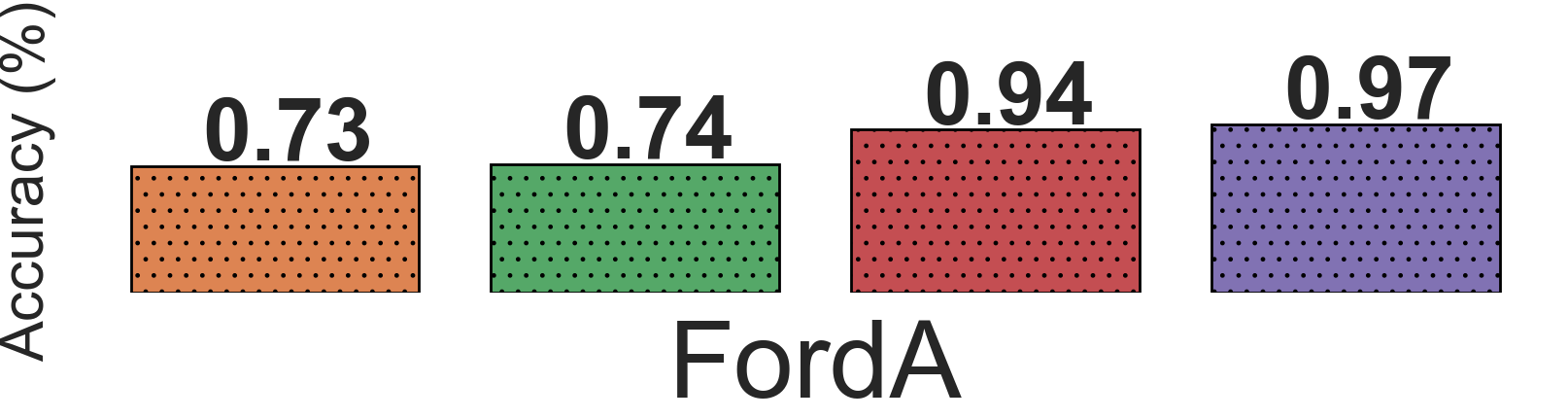}
        \end{minipage}%
\hfill 
        \begin{minipage}{0.19\linewidth}
            \centering
            \includegraphics[width=\linewidth]{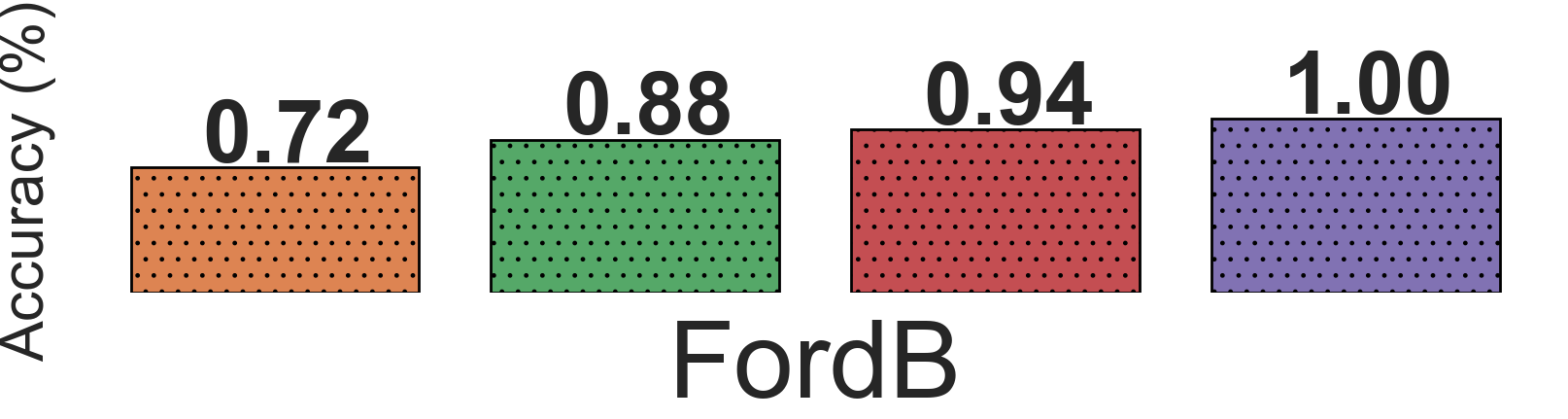}
        \end{minipage}%
\hfill 
        \begin{minipage}{0.19\linewidth}
            \centering
            \includegraphics[width=\linewidth]{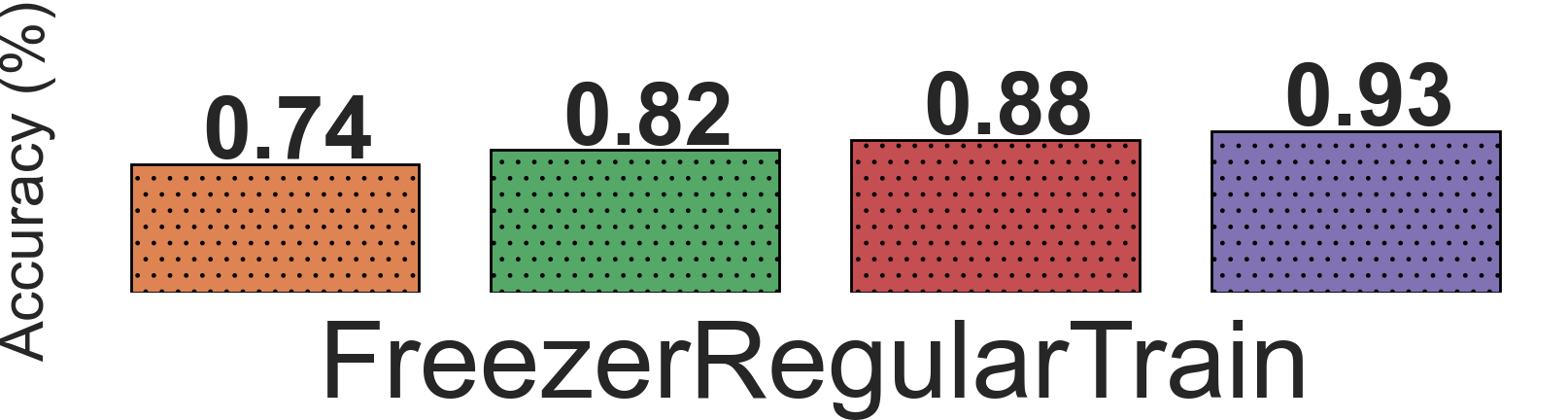}
        \end{minipage}%
\hfill 
        \begin{minipage}{0.19\linewidth}
            \centering
            \includegraphics[width=\linewidth]{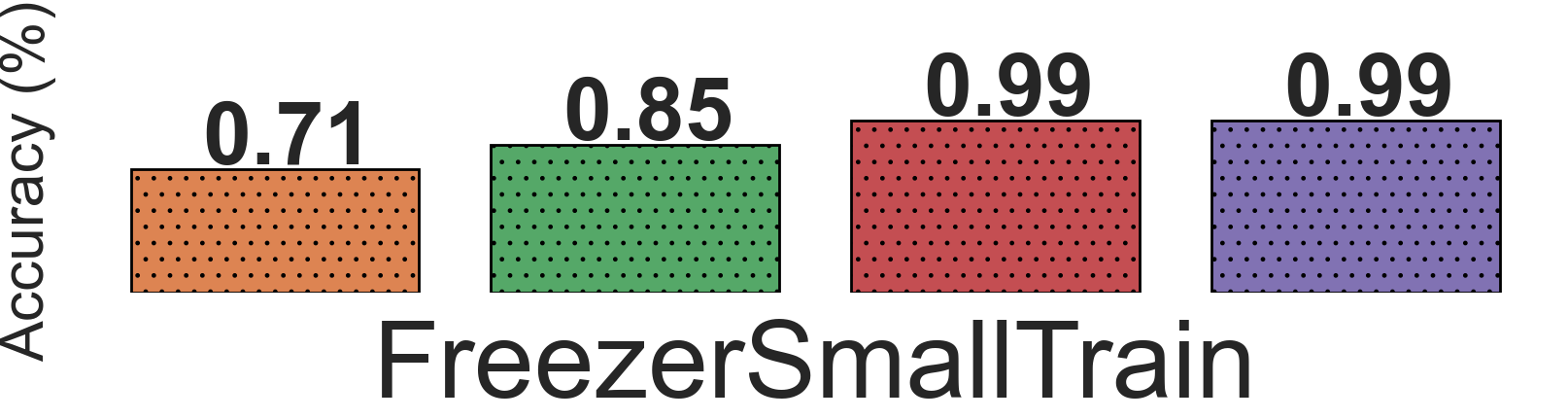}
        \end{minipage}
        \begin{minipage}{0.19\linewidth}
            \centering
            \includegraphics[width=\linewidth]{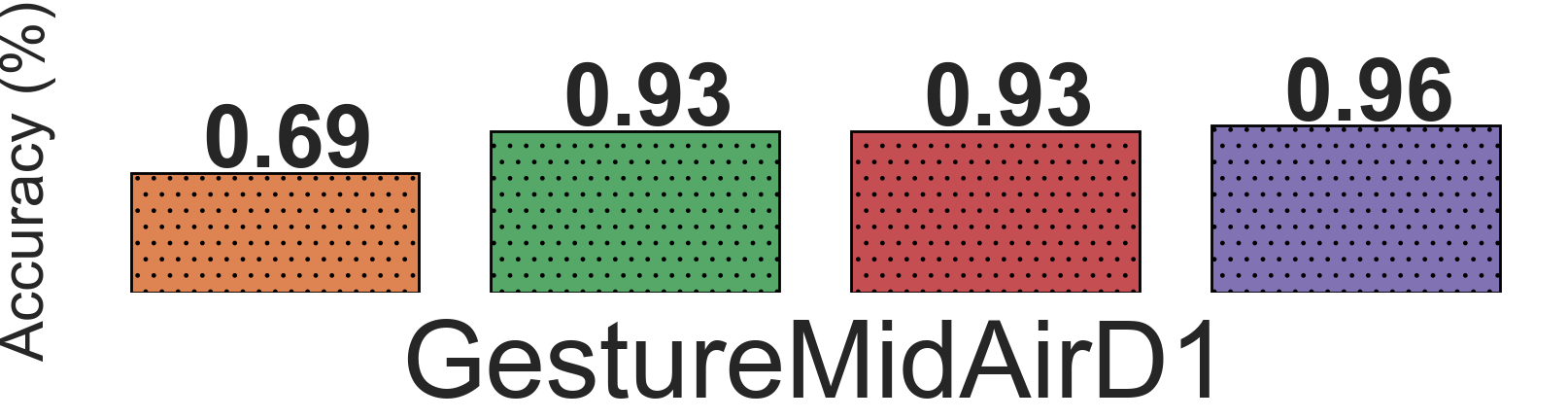}
        \end{minipage}%
\hfill 
        \begin{minipage}{0.19\linewidth}
            \centering
            \includegraphics[width=\linewidth]{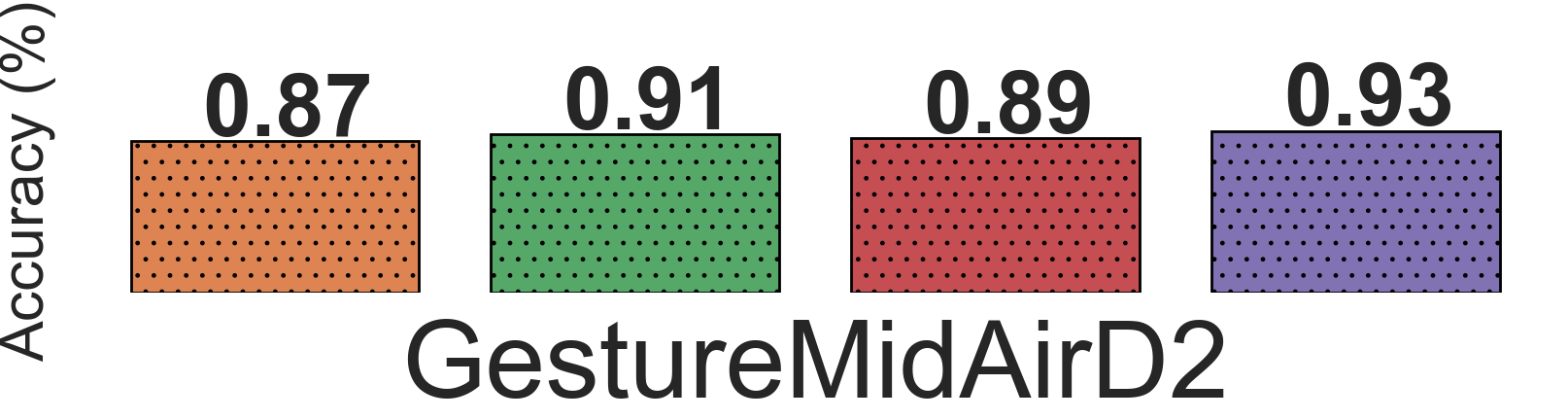}
        \end{minipage}%
\hfill 
        \begin{minipage}{0.19\linewidth}
            \centering
            \includegraphics[width=\linewidth]{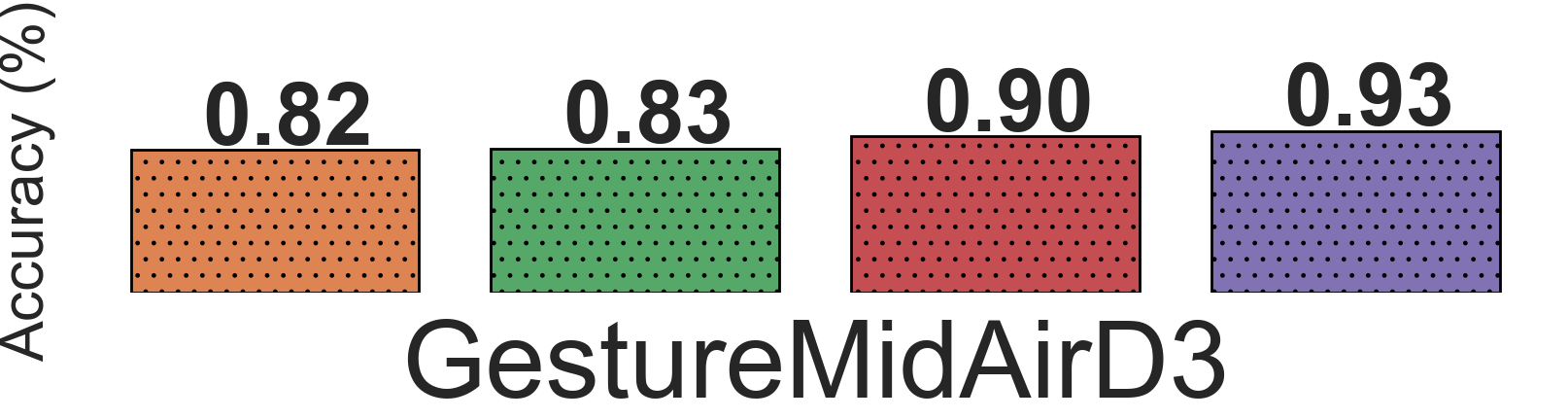}
        \end{minipage}%
\hfill 
        \begin{minipage}{0.19\linewidth}
            \centering
            \includegraphics[width=\linewidth]{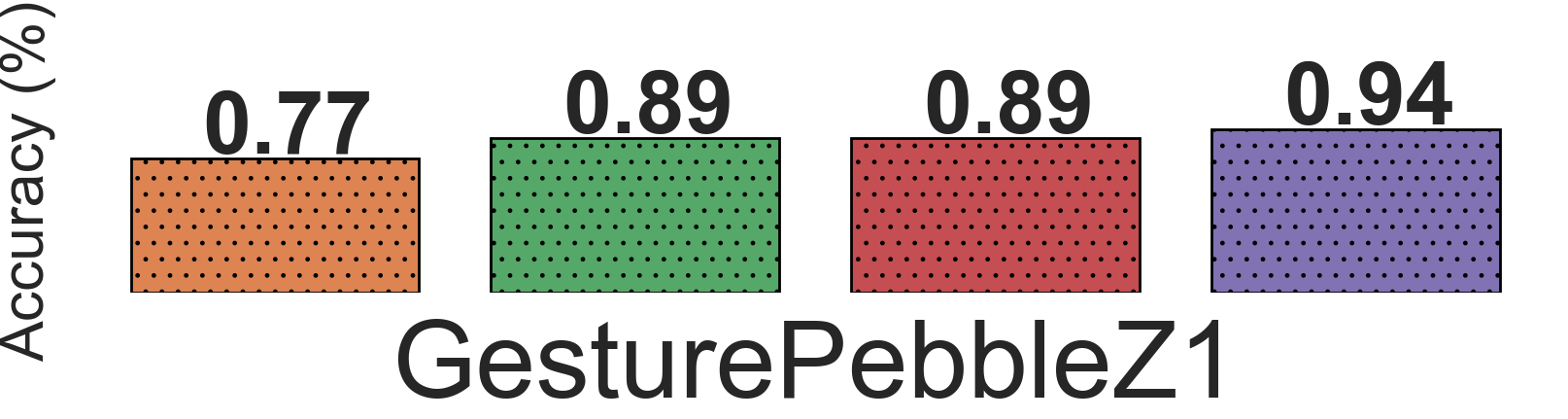}
        \end{minipage}%
\hfill 
        \begin{minipage}{0.19\linewidth}
            \centering
            \includegraphics[width=\linewidth]{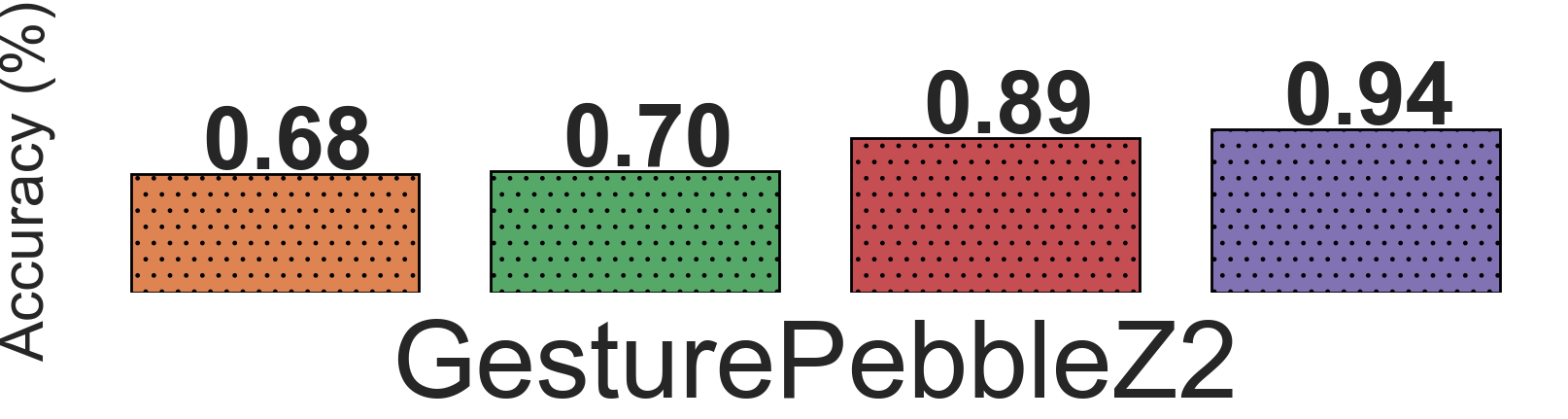}
        \end{minipage}
        \begin{minipage}{0.19\linewidth}
            \centering
            \includegraphics[width=\linewidth]{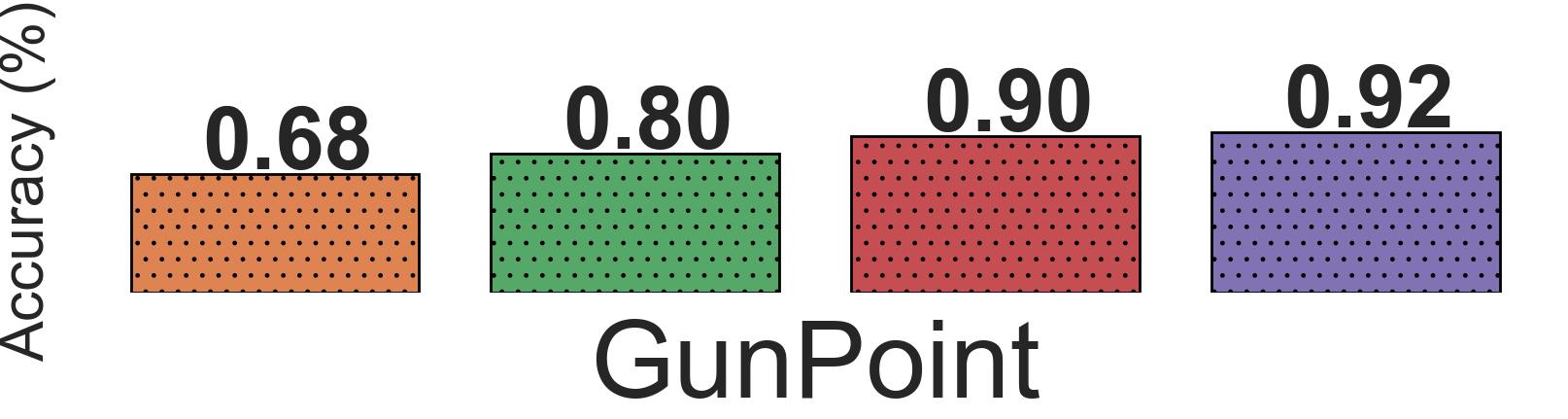}
        \end{minipage}%
\hfill 
        \begin{minipage}{0.19\linewidth}
            \centering
            \includegraphics[width=\linewidth]{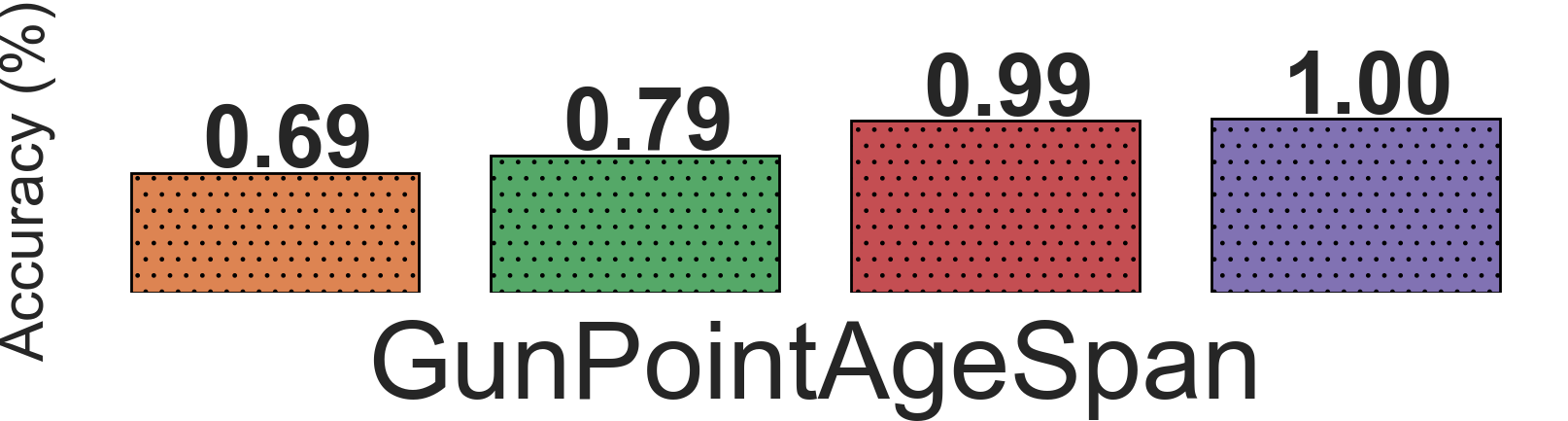}
        \end{minipage}%
\hfill 
        \begin{minipage}{0.19\linewidth}
            \centering
            \includegraphics[width=\linewidth]{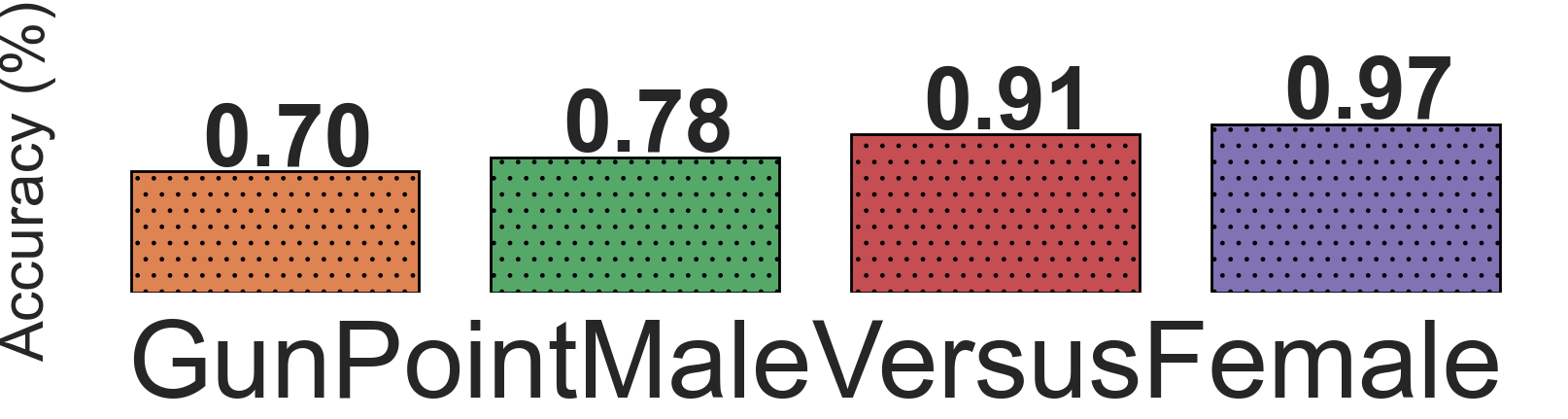}
        \end{minipage}%
\hfill 
        \begin{minipage}{0.19\linewidth}
            \centering
            \includegraphics[width=\linewidth]{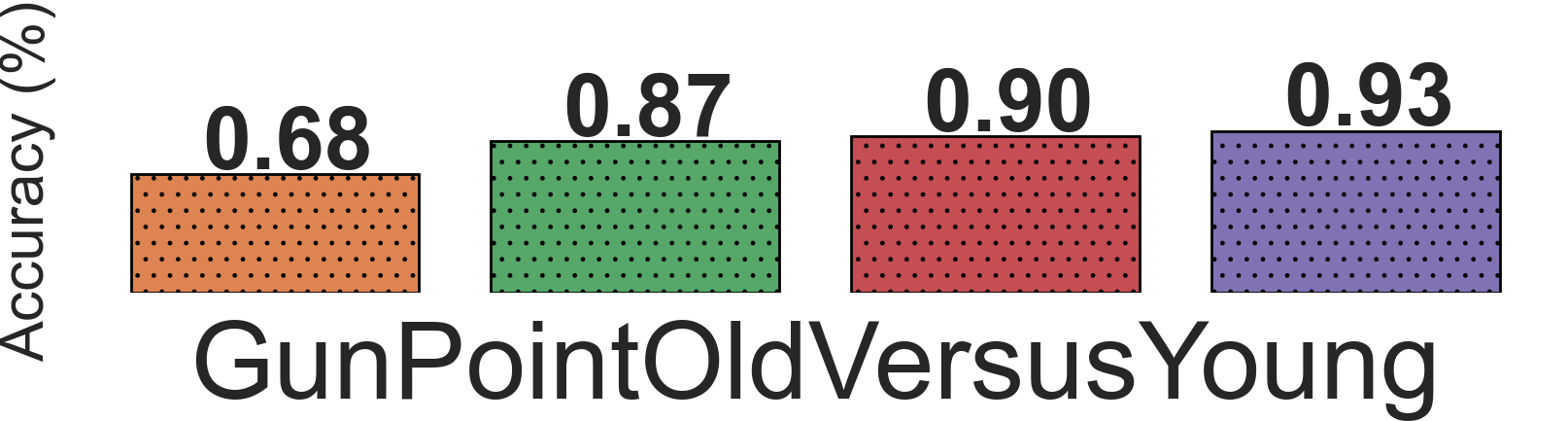}
        \end{minipage}%
\hfill 
        \begin{minipage}{0.19\linewidth}
            \centering
            \includegraphics[width=\linewidth]{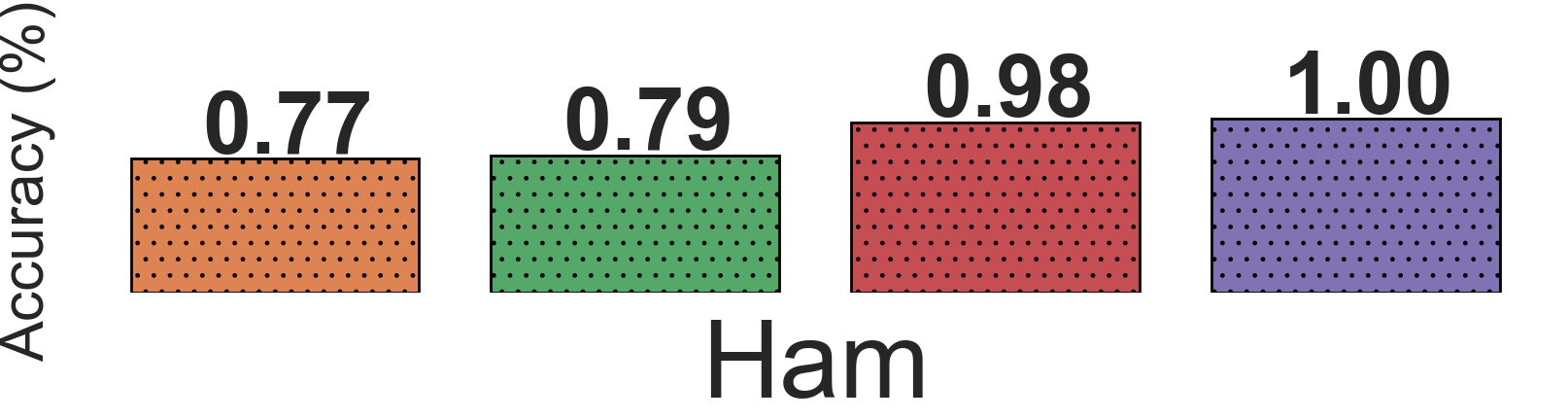}
        \end{minipage}
        \begin{minipage}{0.19\linewidth}
            \centering
            \includegraphics[width=\linewidth]{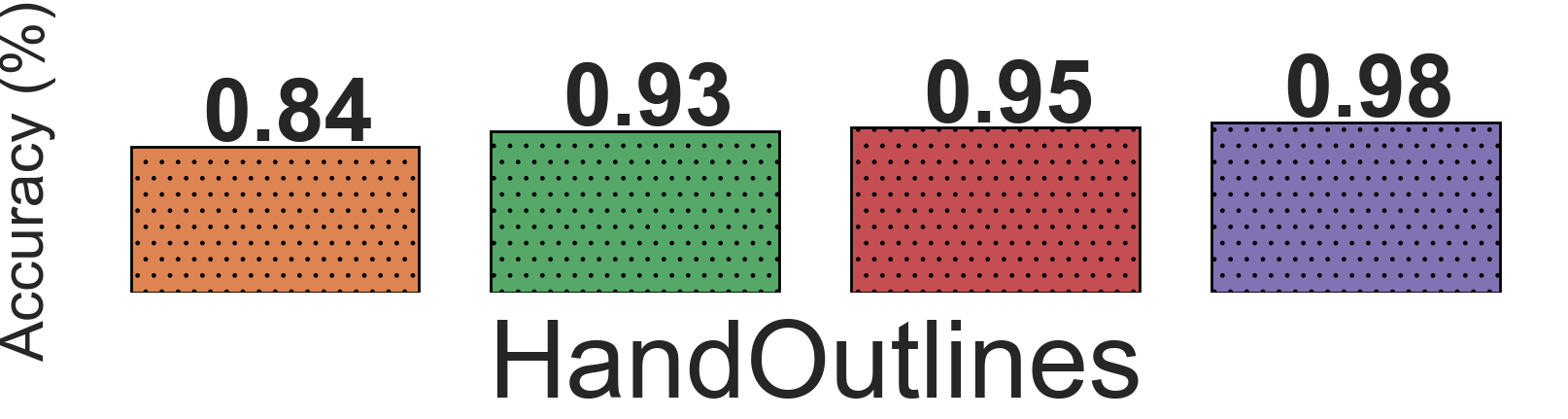}
        \end{minipage}%
\hfill 
        \begin{minipage}{0.19\linewidth}
            \centering
            \includegraphics[width=\linewidth]{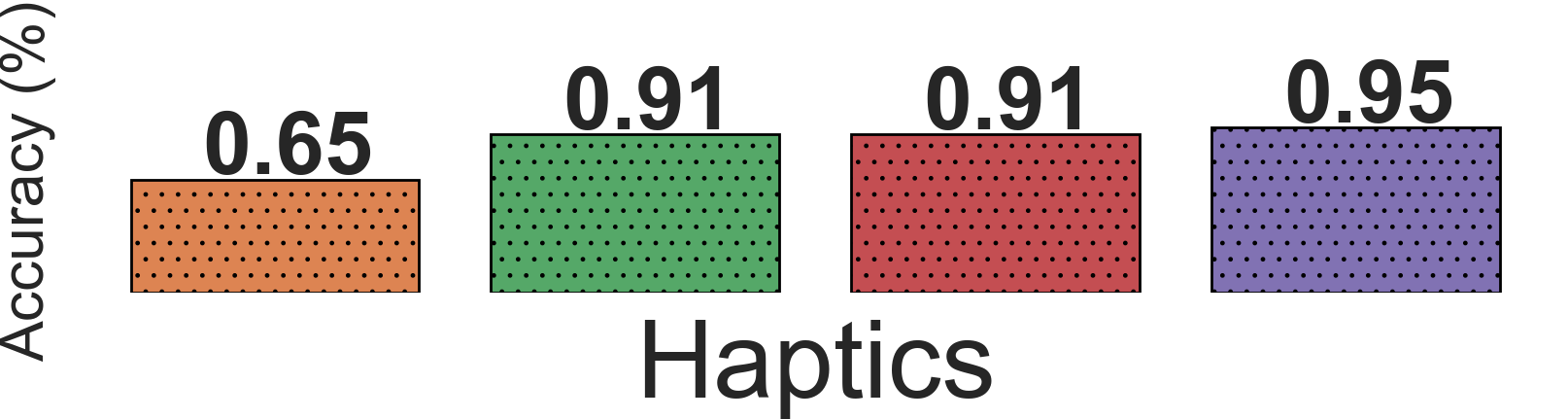}
        \end{minipage}%
\hfill 
        \begin{minipage}{0.19\linewidth}
            \centering
            \includegraphics[width=\linewidth]{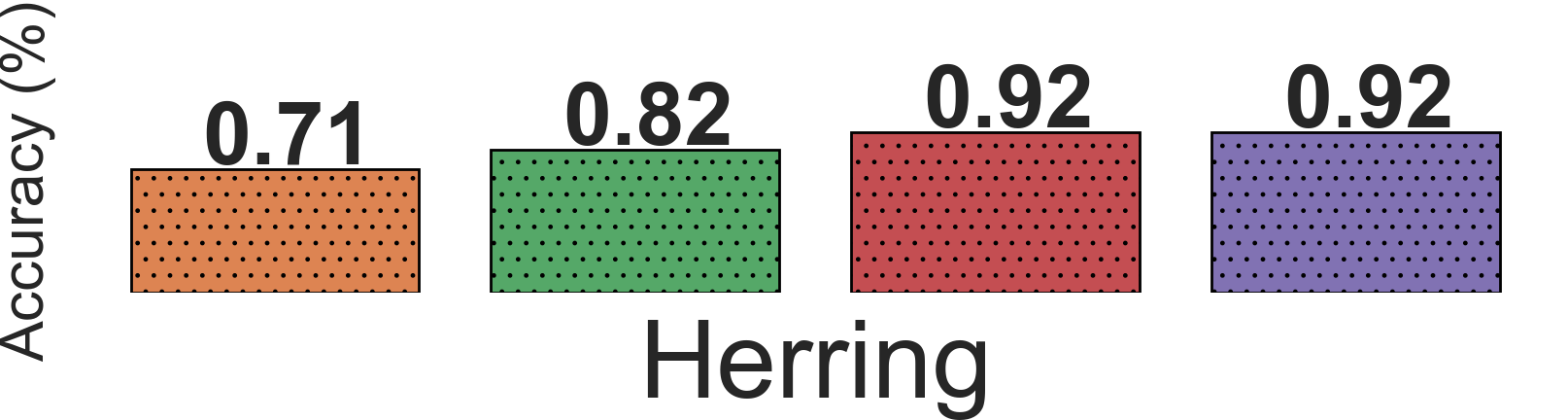}
        \end{minipage}%
\hfill 
        \begin{minipage}{0.19\linewidth}
            \centering
            \includegraphics[width=\linewidth]{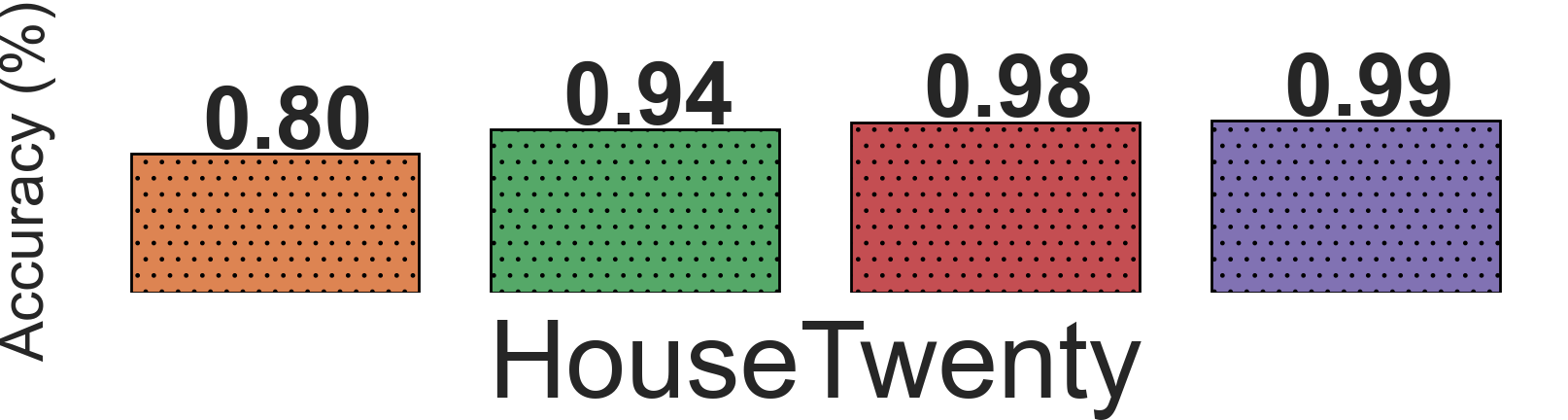}
        \end{minipage}%
\hfill 
        \begin{minipage}{0.19\linewidth}
            \centering
            \includegraphics[width=\linewidth]{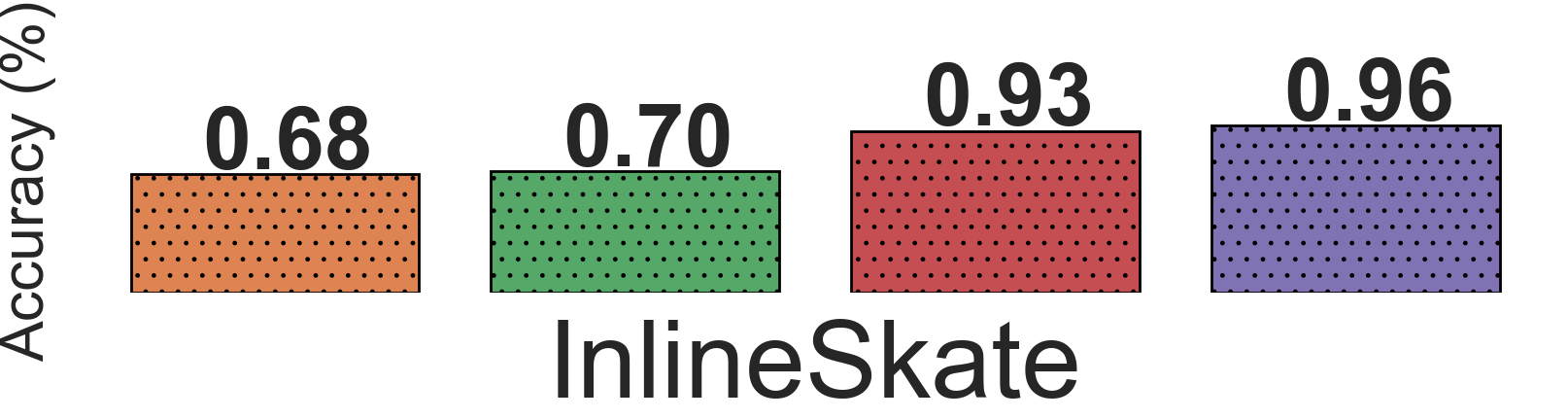}
        \end{minipage}
        \begin{minipage}{0.19\linewidth}
            \centering
            \includegraphics[width=\linewidth]{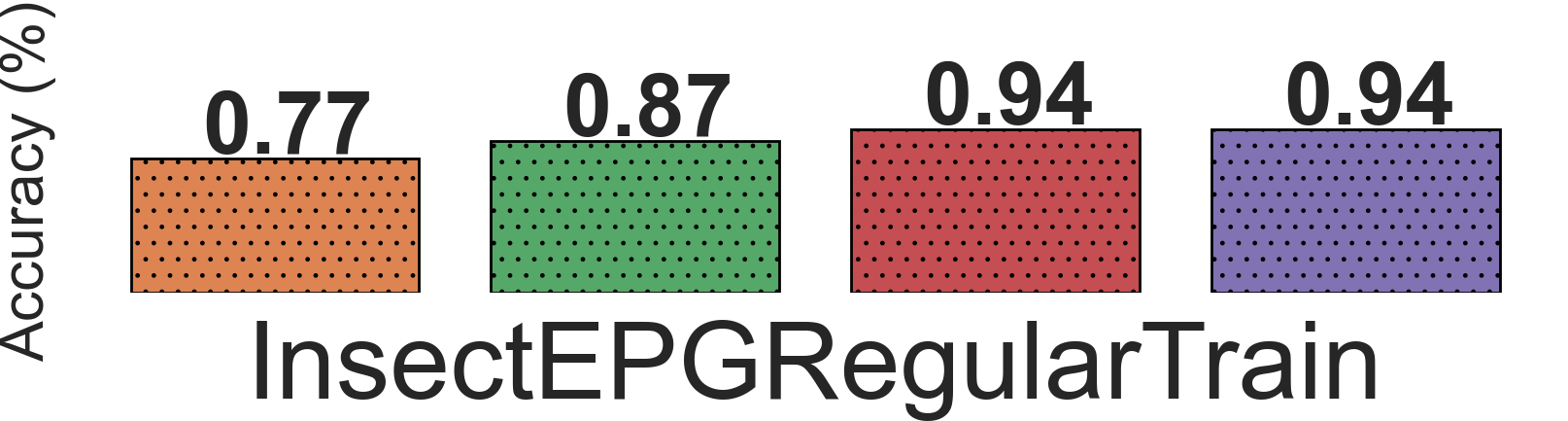}
        \end{minipage}%
\hfill 
        \begin{minipage}{0.19\linewidth}
            \centering
            \includegraphics[width=\linewidth]{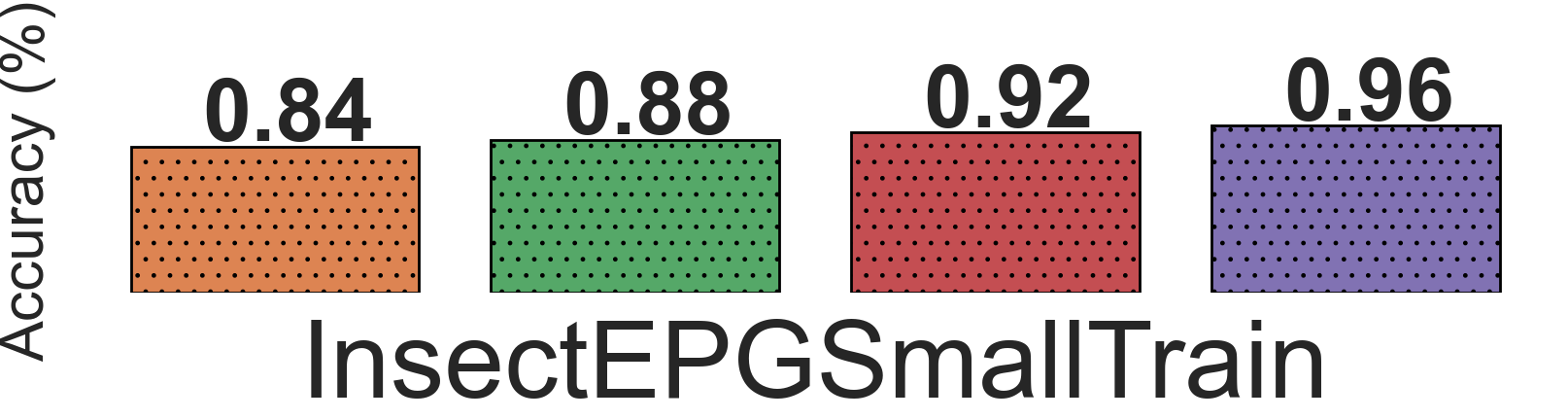}
        \end{minipage}%
\hfill 
        \begin{minipage}{0.19\linewidth}
            \centering
            \includegraphics[width=\linewidth]{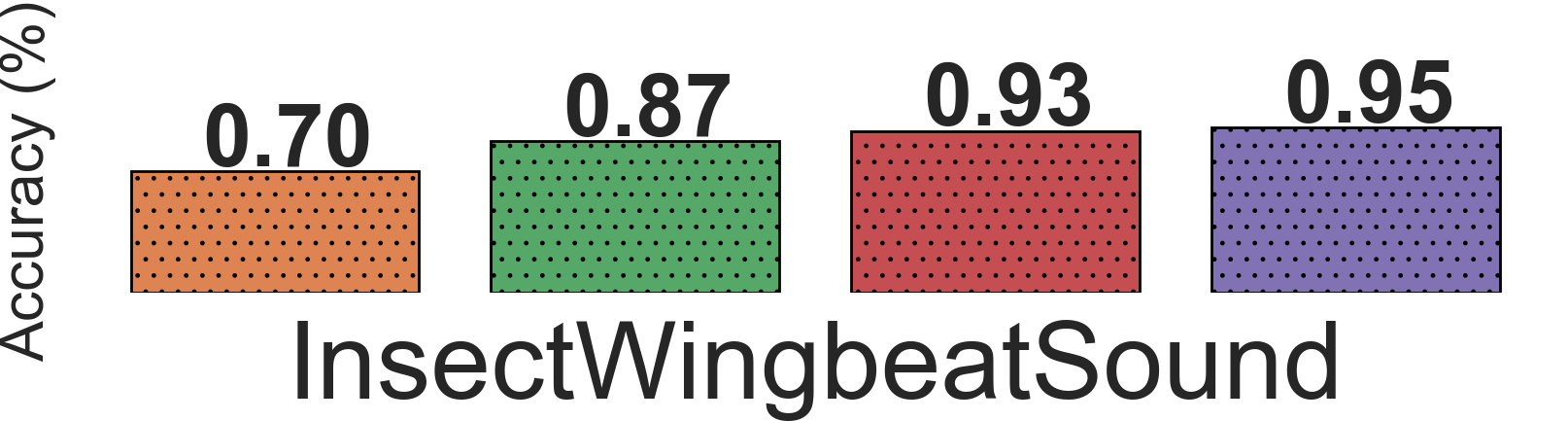}
        \end{minipage}%
\hfill 
        \begin{minipage}{0.19\linewidth}
            \centering
            \includegraphics[width=\linewidth]{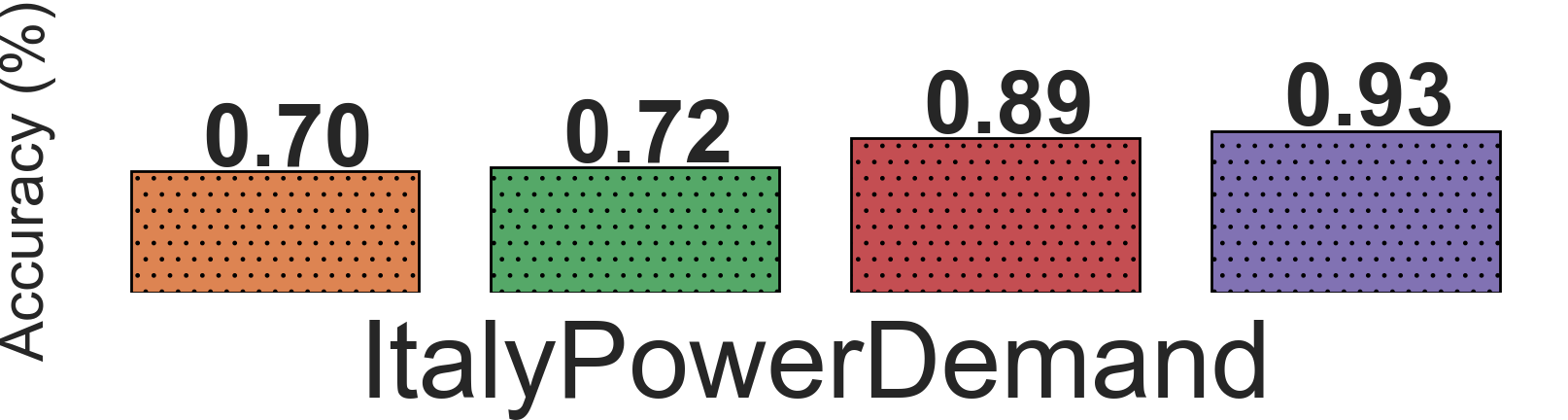}
        \end{minipage}%
\hfill 
        \begin{minipage}{0.19\linewidth}
            \centering
            \includegraphics[width=\linewidth]{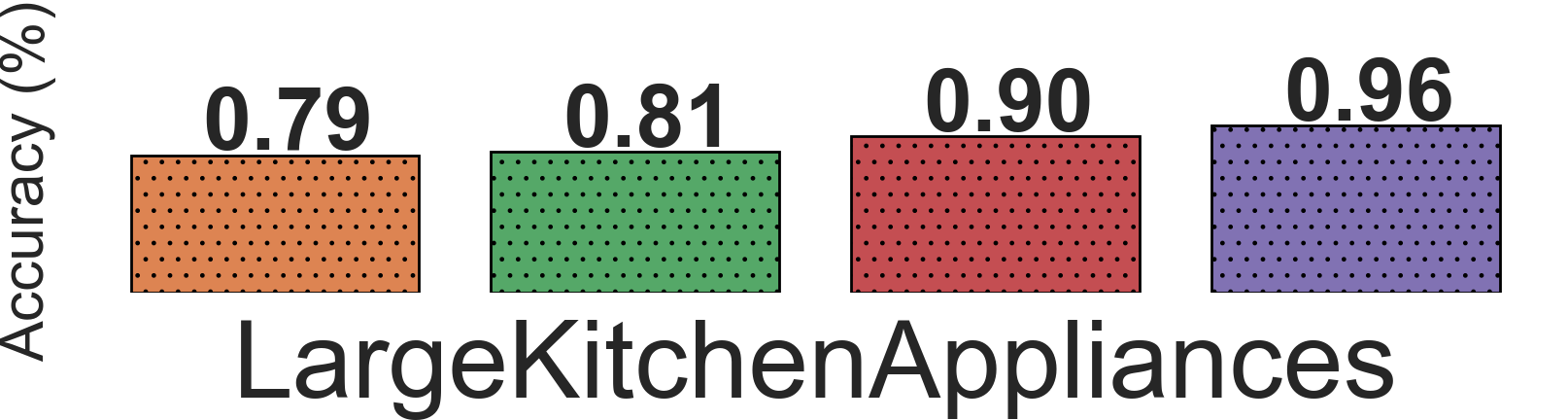}
        \end{minipage}
        \begin{minipage}{0.19\linewidth}
            \centering
            \includegraphics[width=\linewidth]{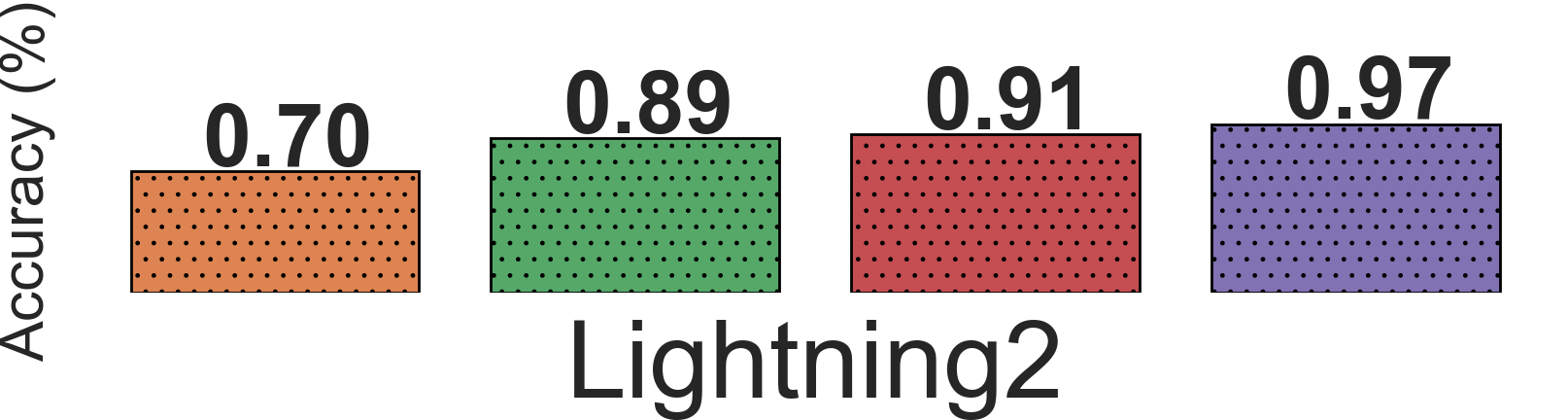}
        \end{minipage}%
\hfill 
        \begin{minipage}{0.19\linewidth}
            \centering
            \includegraphics[width=\linewidth]{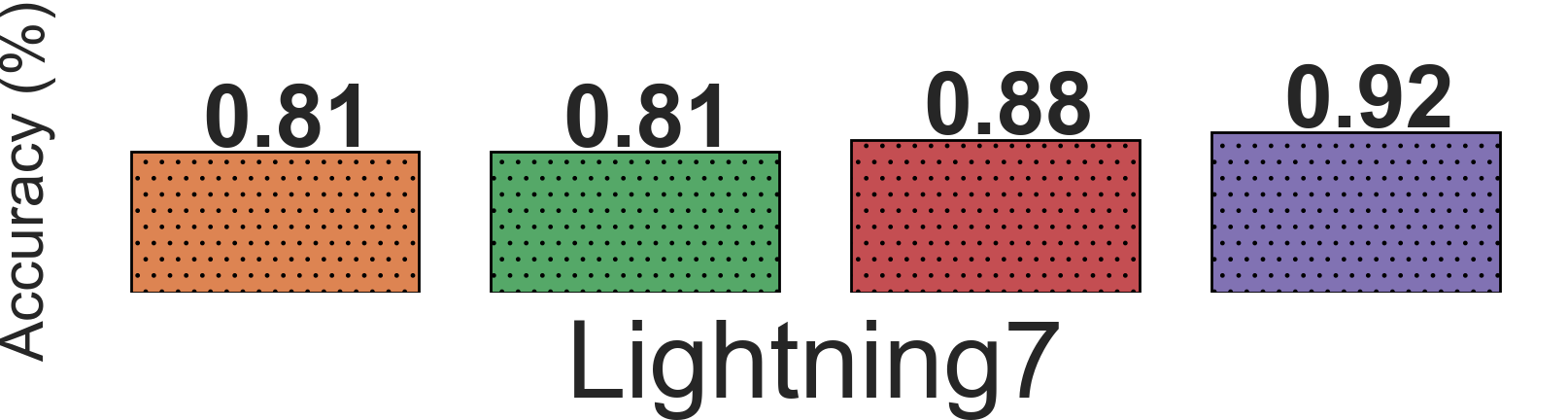}
        \end{minipage}%
\hfill 
        \begin{minipage}{0.19\linewidth}
            \centering
            \includegraphics[width=\linewidth]{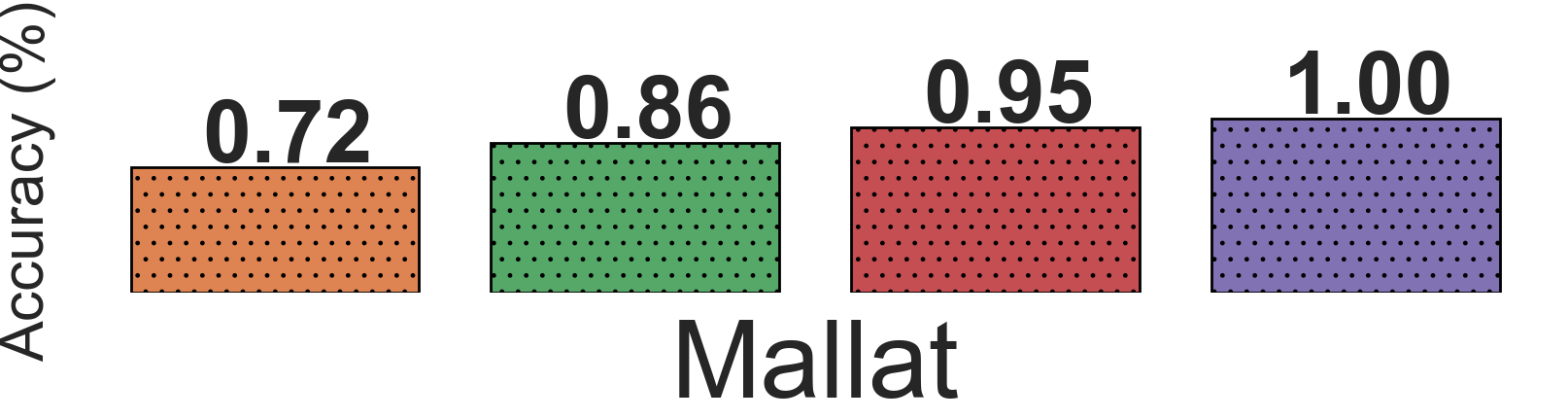}
        \end{minipage}%
\hfill 
        \begin{minipage}{0.19\linewidth}
            \centering
            \includegraphics[width=\linewidth]{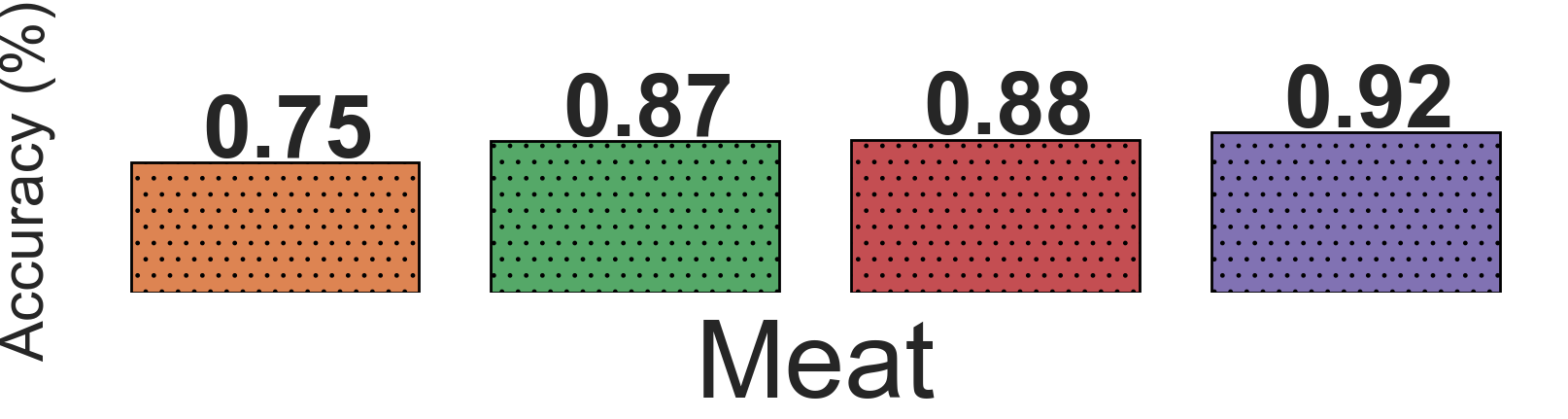}
        \end{minipage}%
\hfill 
        \begin{minipage}{0.19\linewidth}
            \centering
            \includegraphics[width=\linewidth]{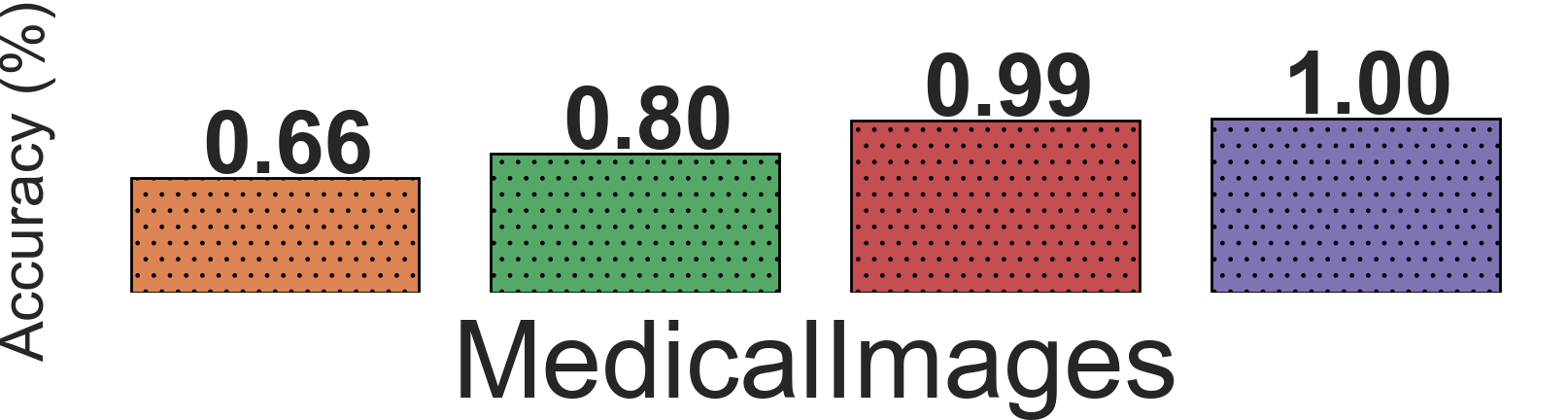}
        \end{minipage}
        \begin{minipage}{0.19\linewidth}
            \centering
            \includegraphics[width=\linewidth]{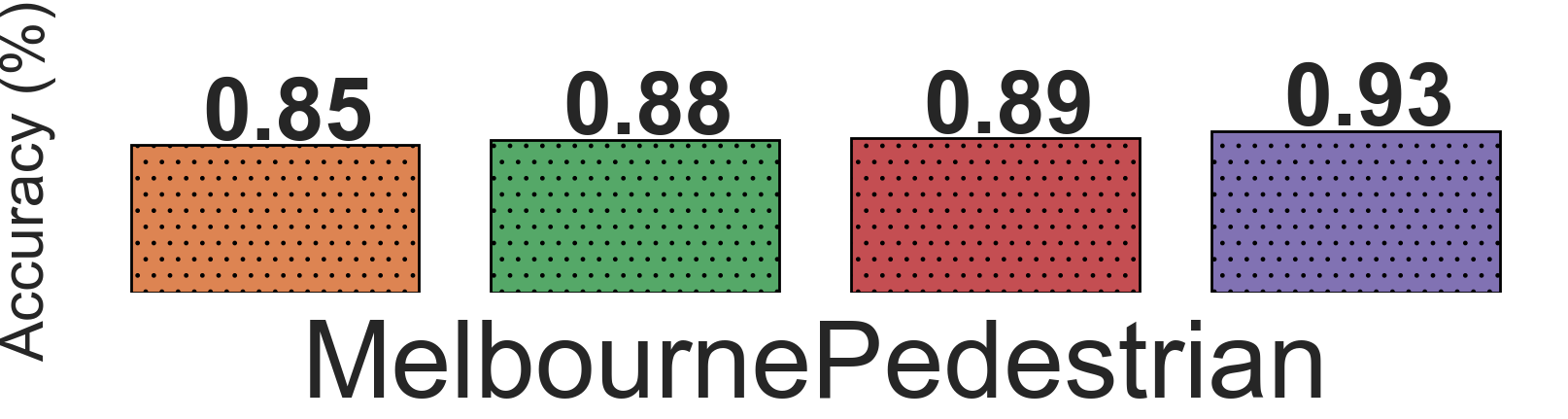}
        \end{minipage}%
\hfill 
        \begin{minipage}{0.19\linewidth}
            \centering
            \includegraphics[width=\linewidth]{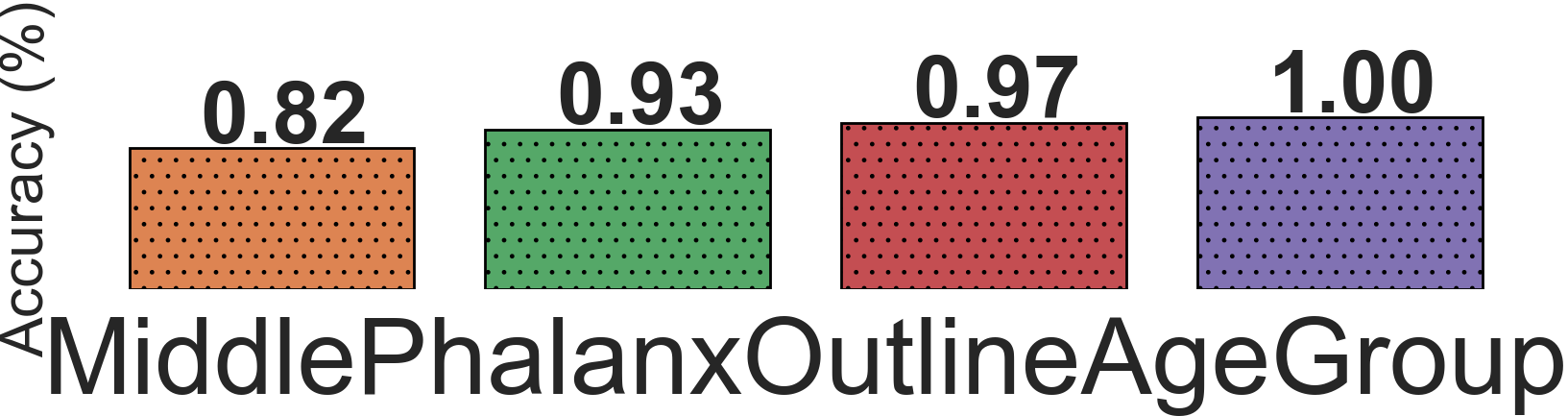}
        \end{minipage}%
\hfill 
        \begin{minipage}{0.19\linewidth}
            \centering
            \includegraphics[width=\linewidth]{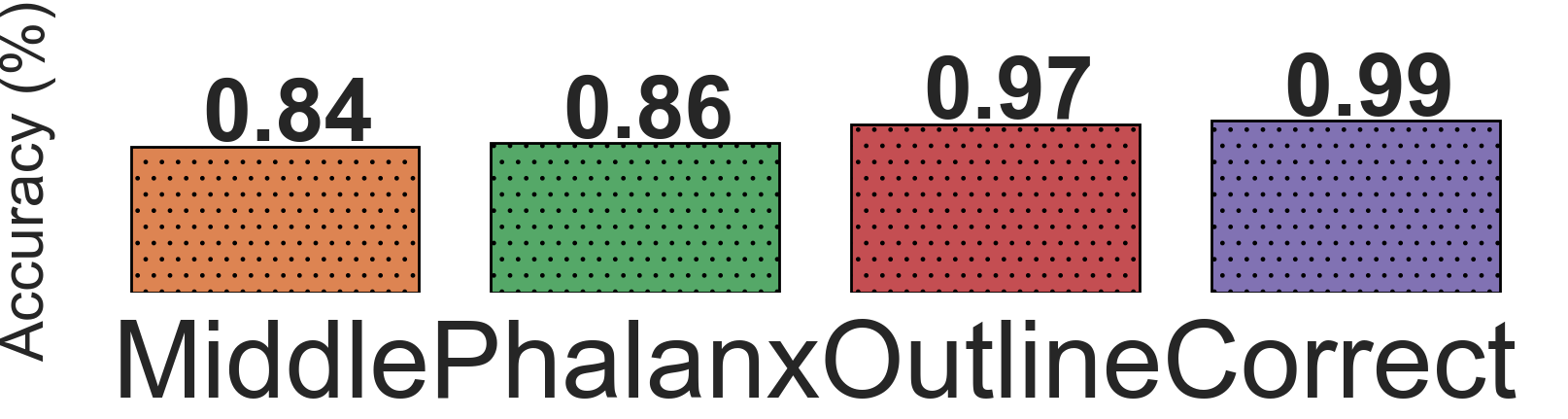}
        \end{minipage}%
\hfill 
        \begin{minipage}{0.19\linewidth}
            \centering
            \includegraphics[width=\linewidth]{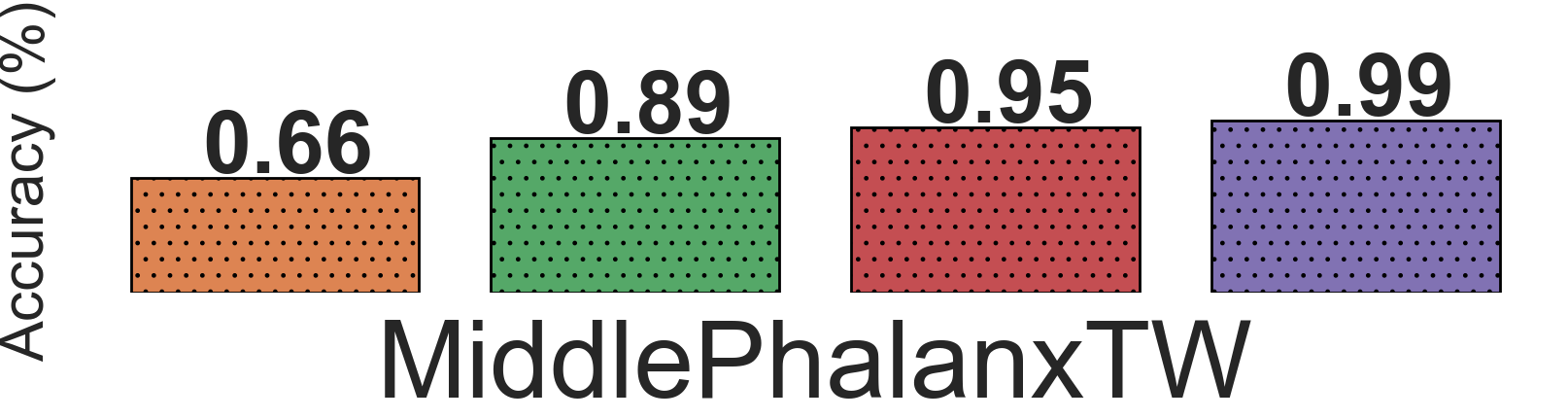}
        \end{minipage}%
\hfill 
        \begin{minipage}{0.19\linewidth}
            \centering
            \includegraphics[width=\linewidth]{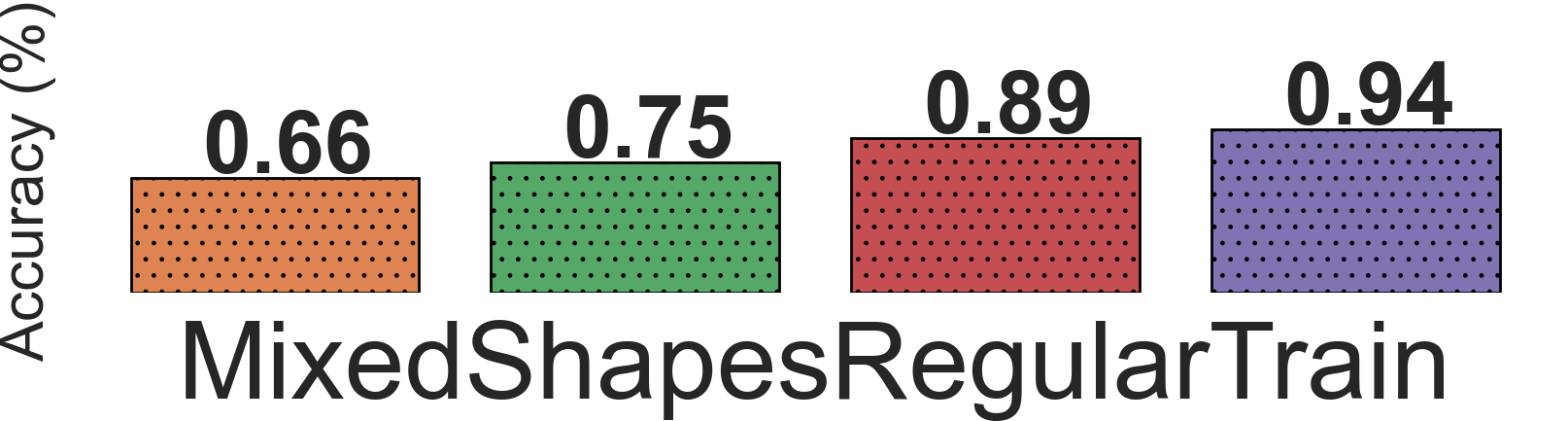}
        \end{minipage}
        \begin{minipage}{0.19\linewidth}
            \centering
            \includegraphics[width=\linewidth]{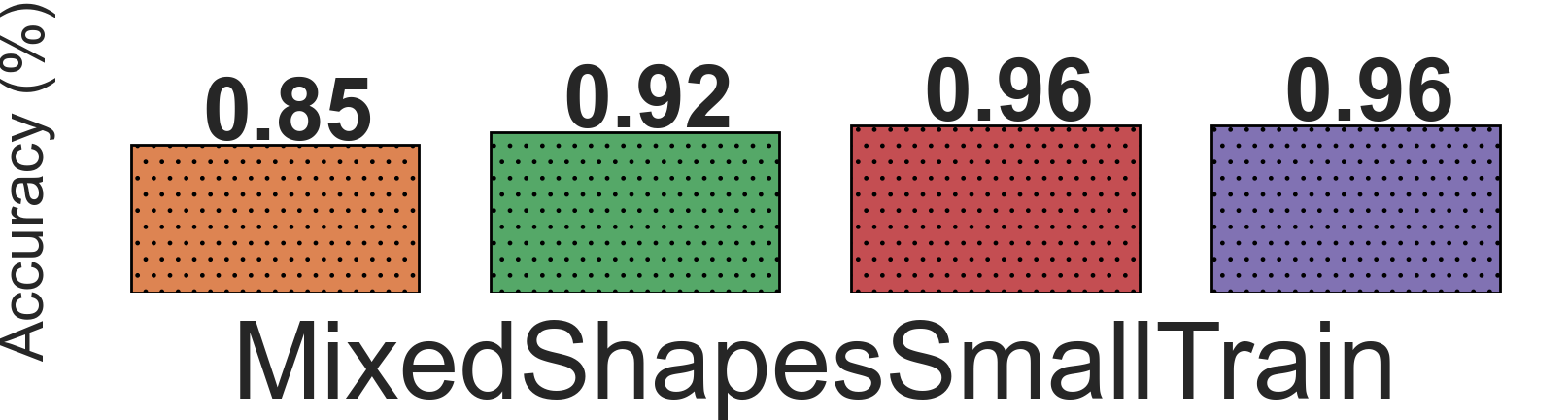}
        \end{minipage}%
\hfill 
        \begin{minipage}{0.19\linewidth}
            \centering
            \includegraphics[width=\linewidth]{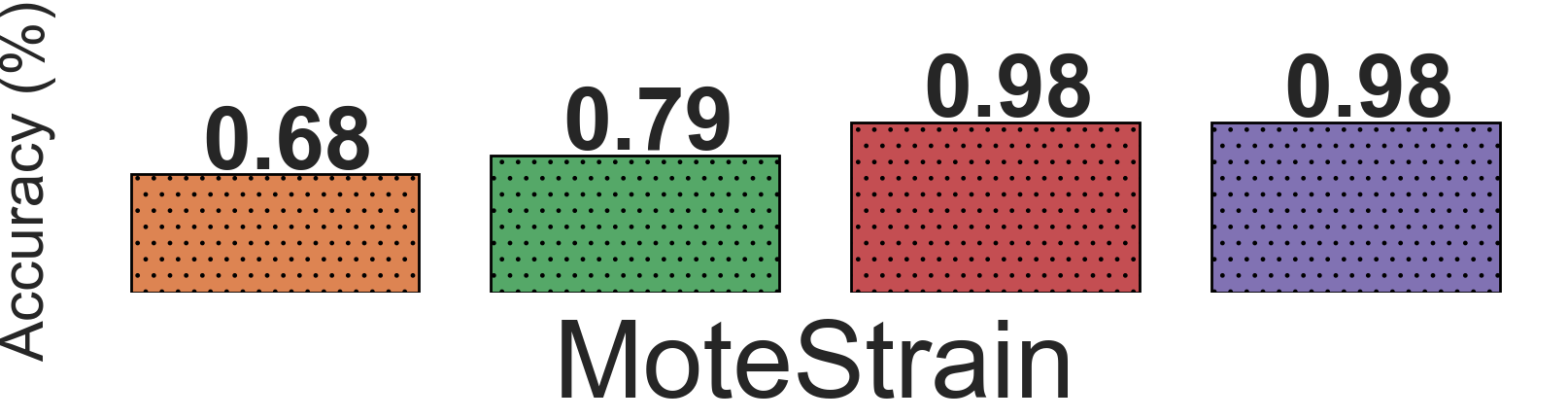}
        \end{minipage}%
\hfill 
        \begin{minipage}{0.19\linewidth}
            \centering
            \includegraphics[width=\linewidth]{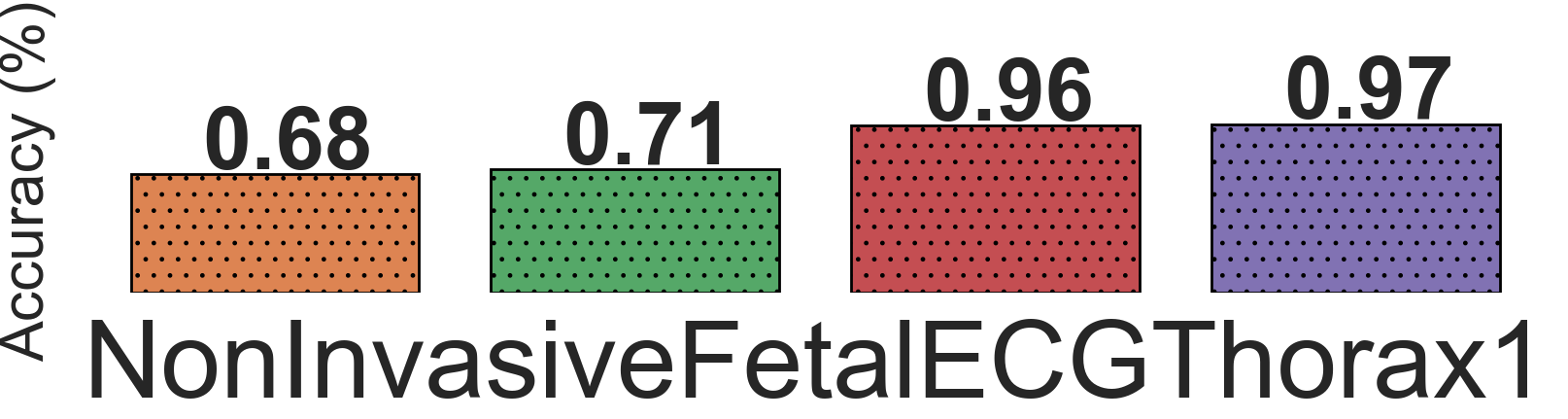}
        \end{minipage}%
\hfill 
        \begin{minipage}{0.19\linewidth}
            \centering
            \includegraphics[width=\linewidth]{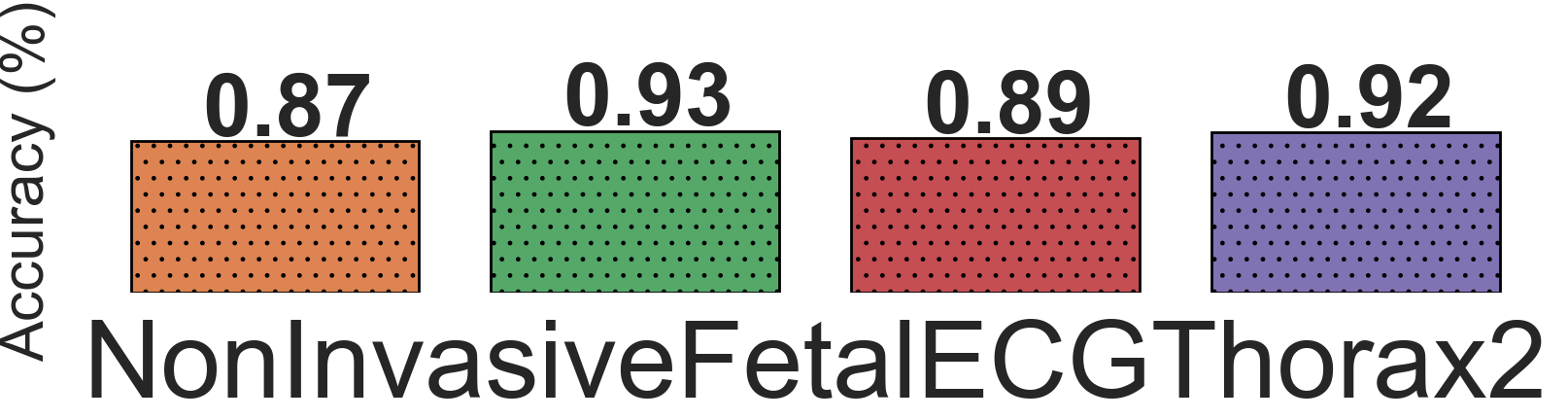}
        \end{minipage}%
\hfill 
        \begin{minipage}{0.19\linewidth}
            \centering
            \includegraphics[width=\linewidth]{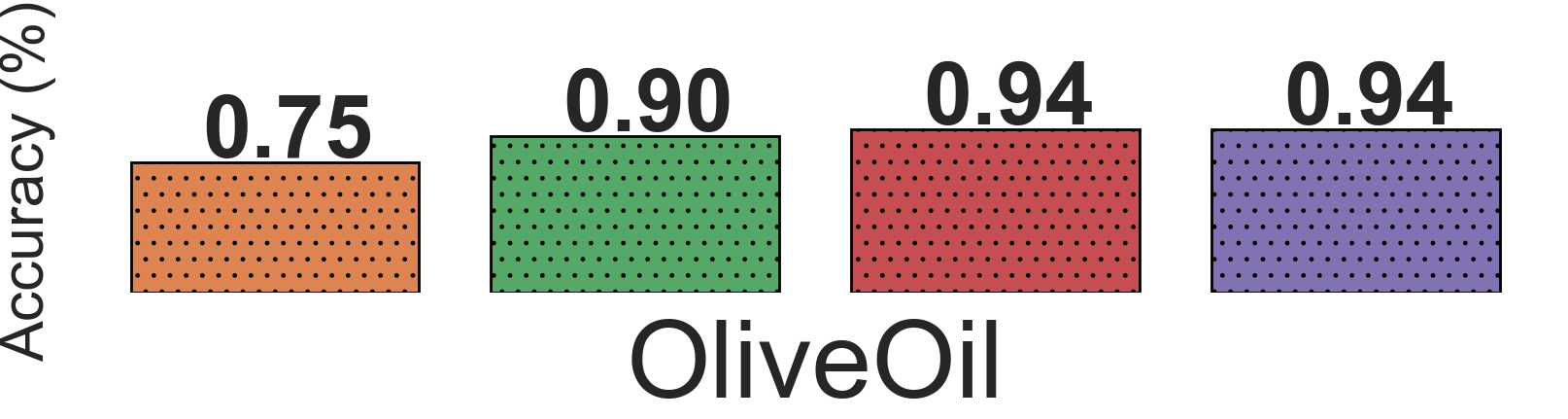}
        \end{minipage}
        \begin{minipage}{0.19\linewidth}
            \centering
            \includegraphics[width=\linewidth]{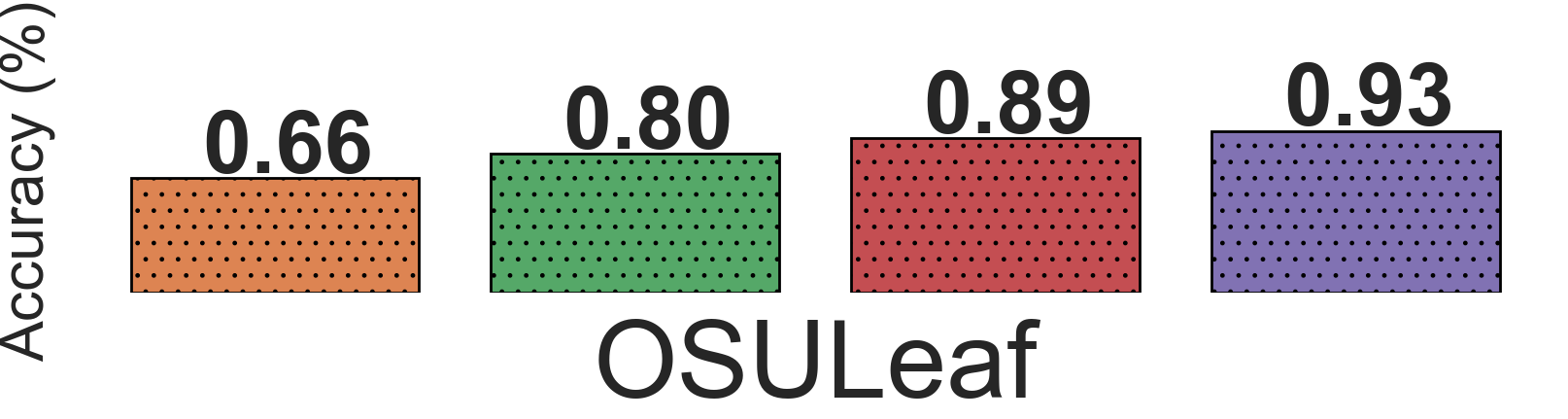}
        \end{minipage}%
\hfill 
        \begin{minipage}{0.19\linewidth}
            \centering
            \includegraphics[width=\linewidth]{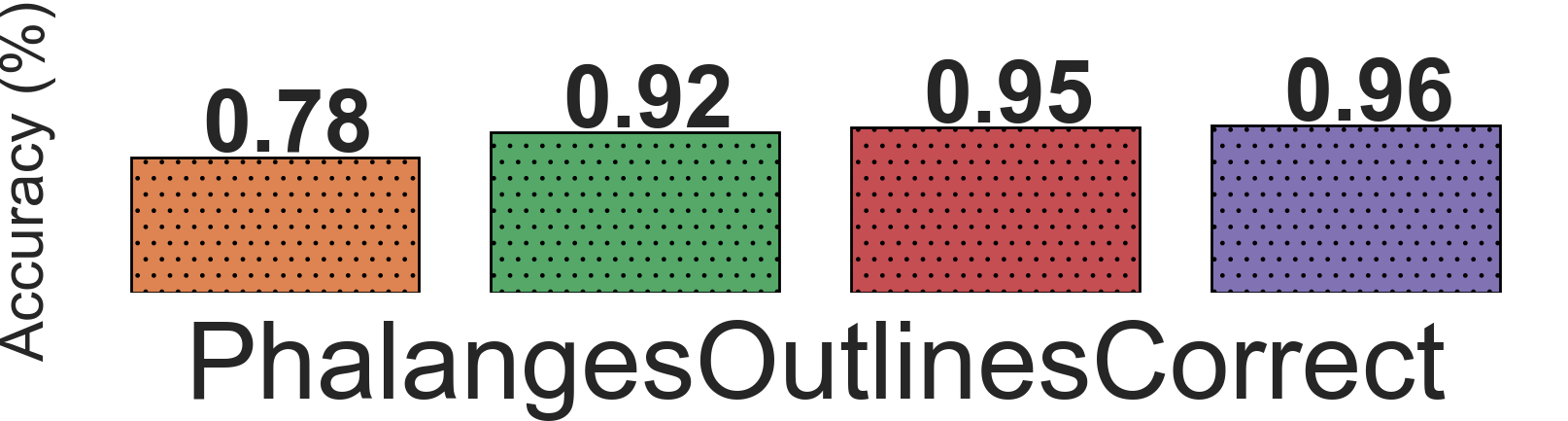}
        \end{minipage}%
\hfill 
        \begin{minipage}{0.19\linewidth}
            \centering
            \includegraphics[width=\linewidth]{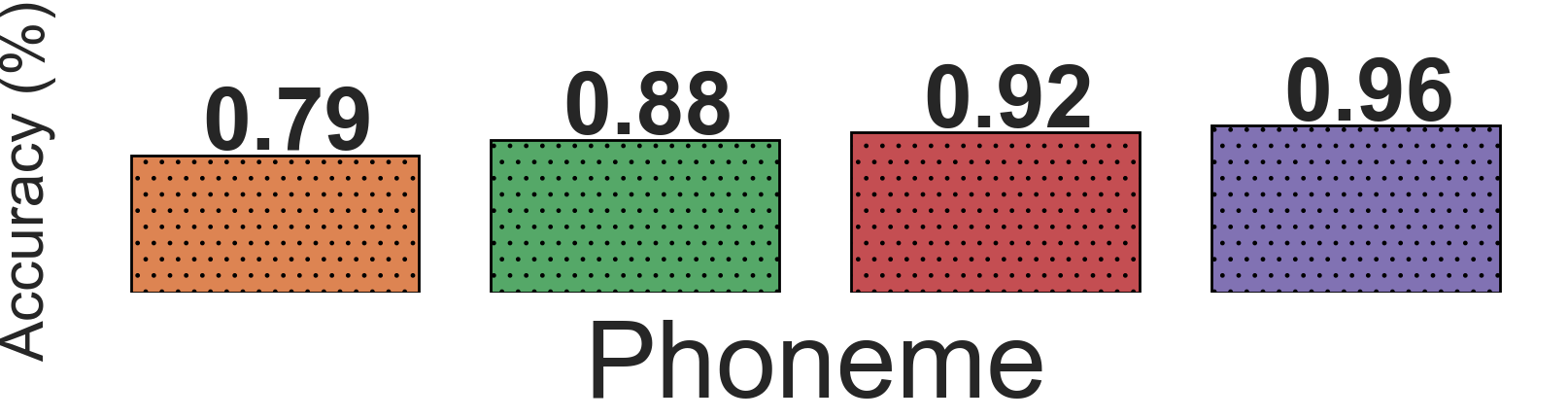}
        \end{minipage}%
\hfill 
        \begin{minipage}{0.19\linewidth}
            \centering
            \includegraphics[width=\linewidth]{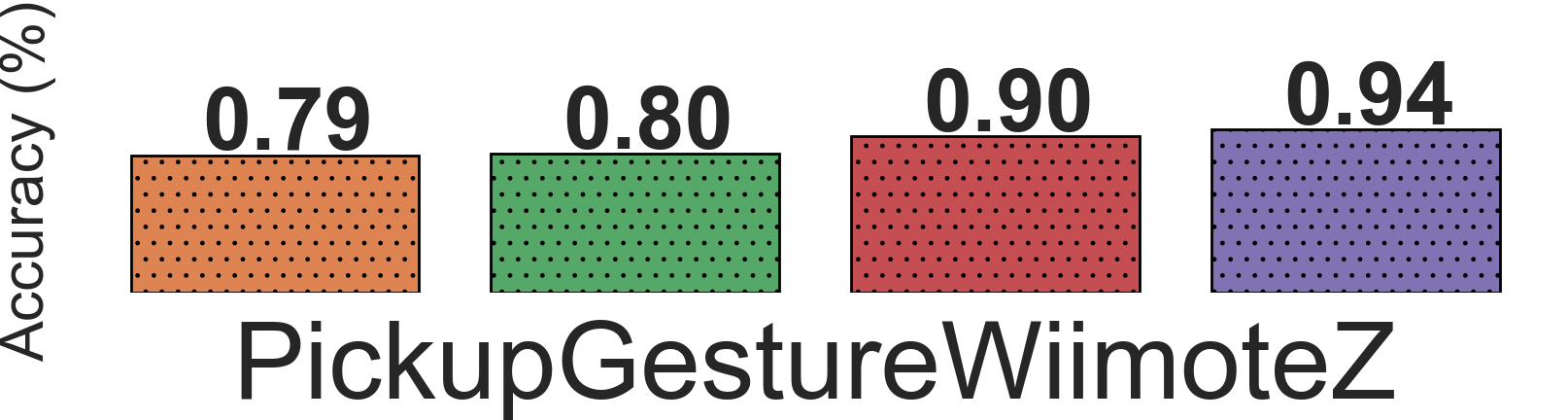}
        \end{minipage}%
\hfill 
        \begin{minipage}{0.19\linewidth}
            \centering
            \includegraphics[width=\linewidth]{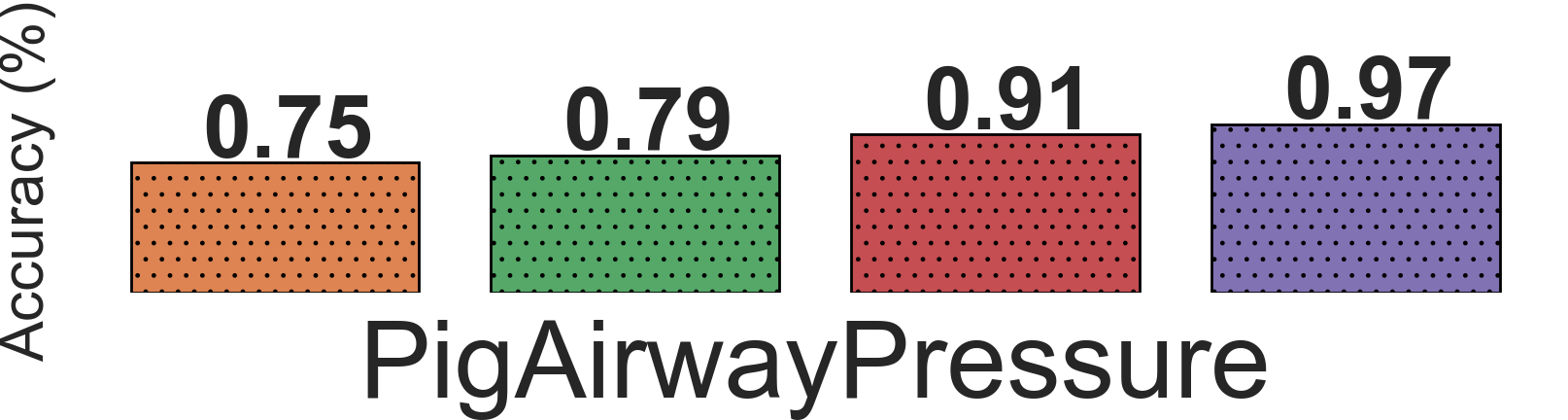}
        \end{minipage}
        \begin{minipage}{0.19\linewidth}
            \centering
            \includegraphics[width=\linewidth]{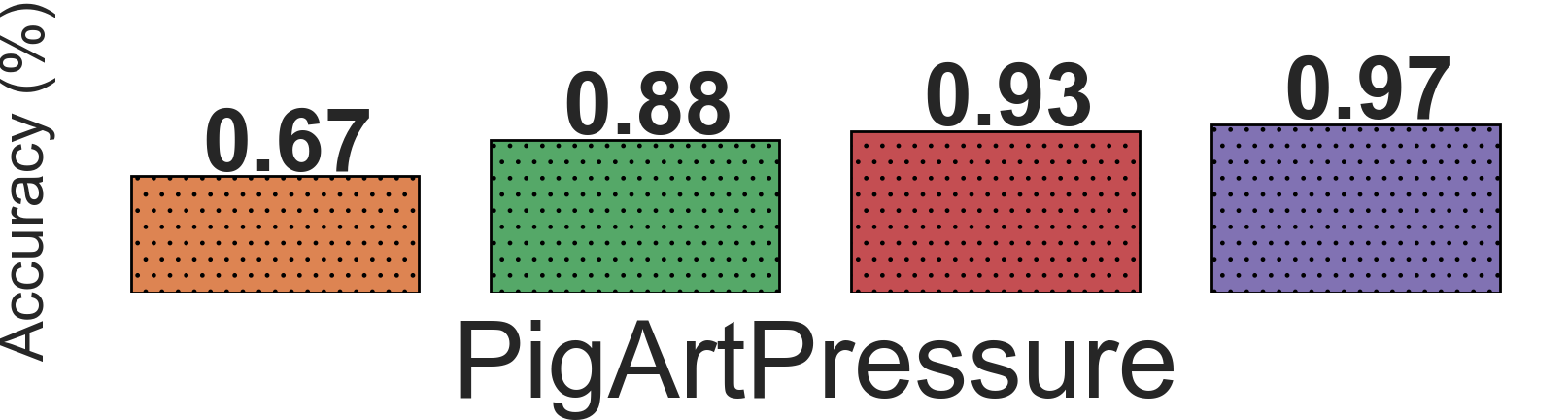}
        \end{minipage}%
\hfill 
        \begin{minipage}{0.19\linewidth}
            \centering
            \includegraphics[width=\linewidth]{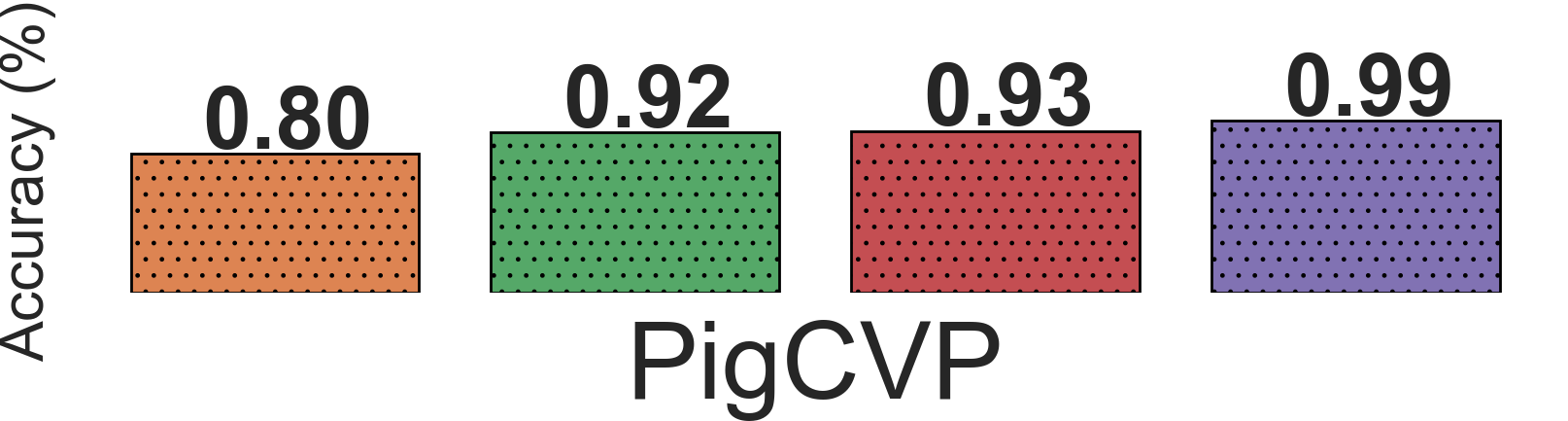}
        \end{minipage}%
\hfill 
        \begin{minipage}{0.19\linewidth}
            \centering
            \includegraphics[width=\linewidth]{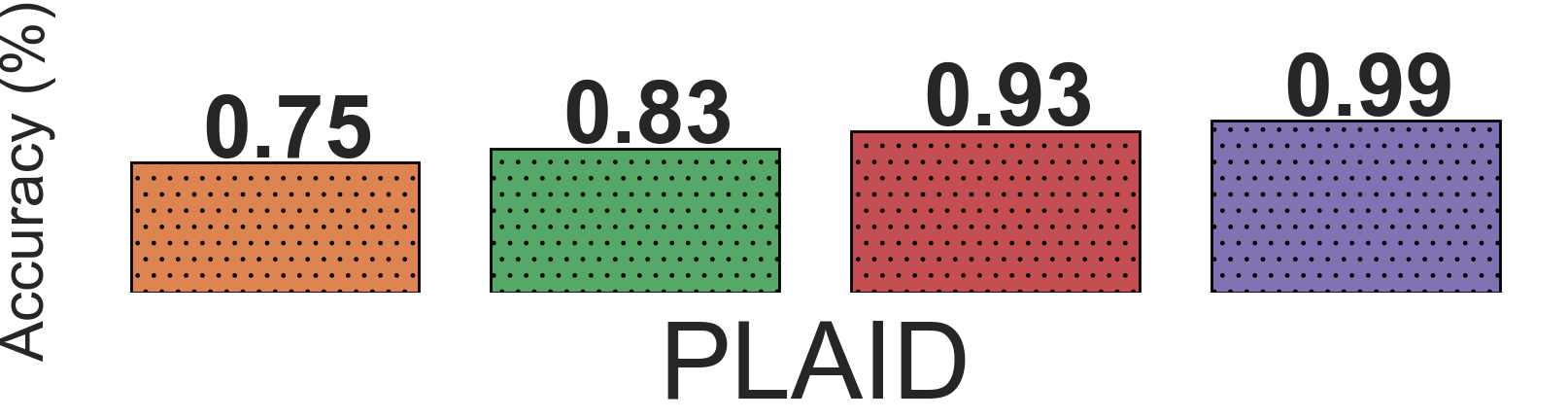}
        \end{minipage}%
\hfill 
        \begin{minipage}{0.19\linewidth}
            \centering
            \includegraphics[width=\linewidth]{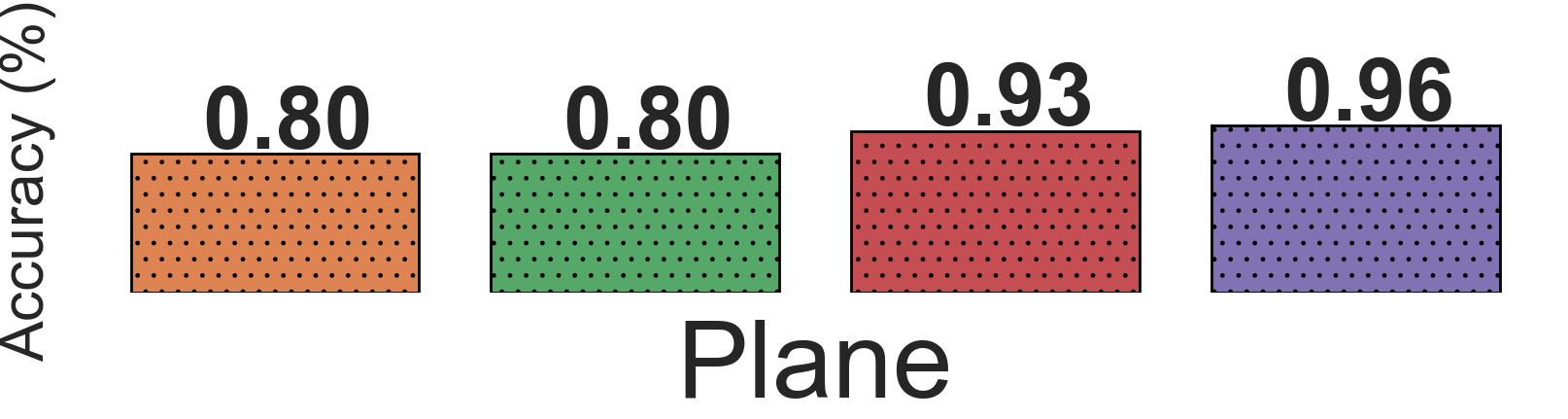}
        \end{minipage}%
\hfill 
        \begin{minipage}{0.19\linewidth}
            \centering
            \includegraphics[width=\linewidth]{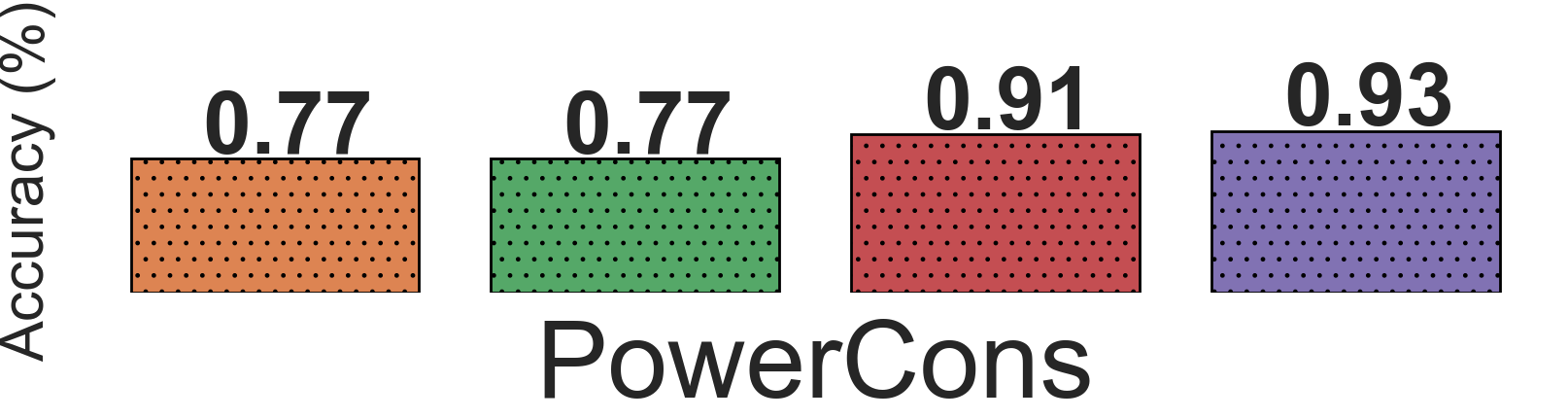}
        \end{minipage}
        \begin{minipage}{0.19\linewidth}
            \centering
            \includegraphics[width=\linewidth]{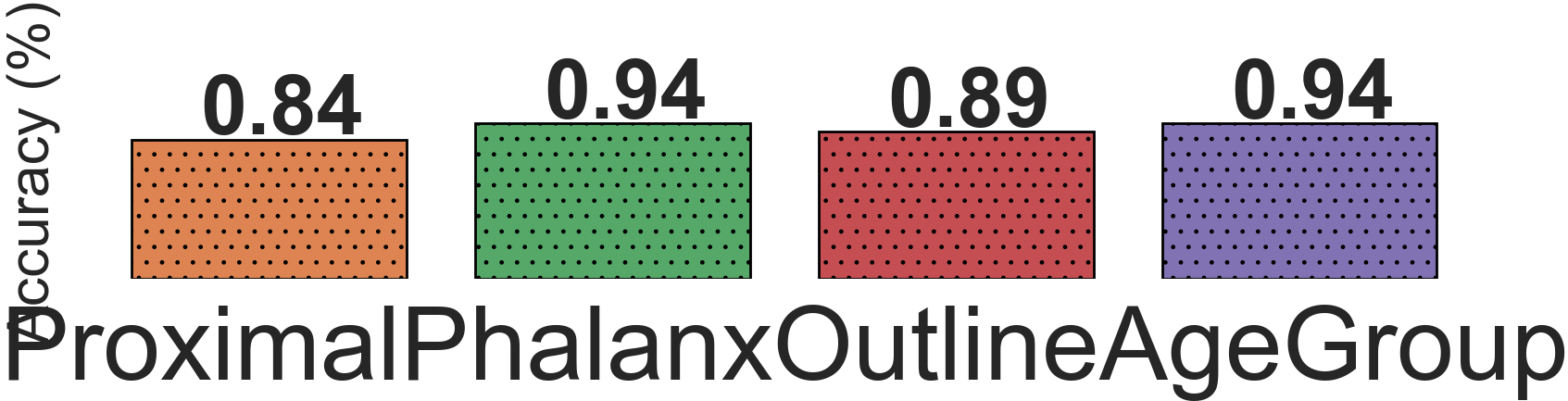}
        \end{minipage}%
\hfill 
        \begin{minipage}{0.19\linewidth}
            \centering
            \includegraphics[width=\linewidth]{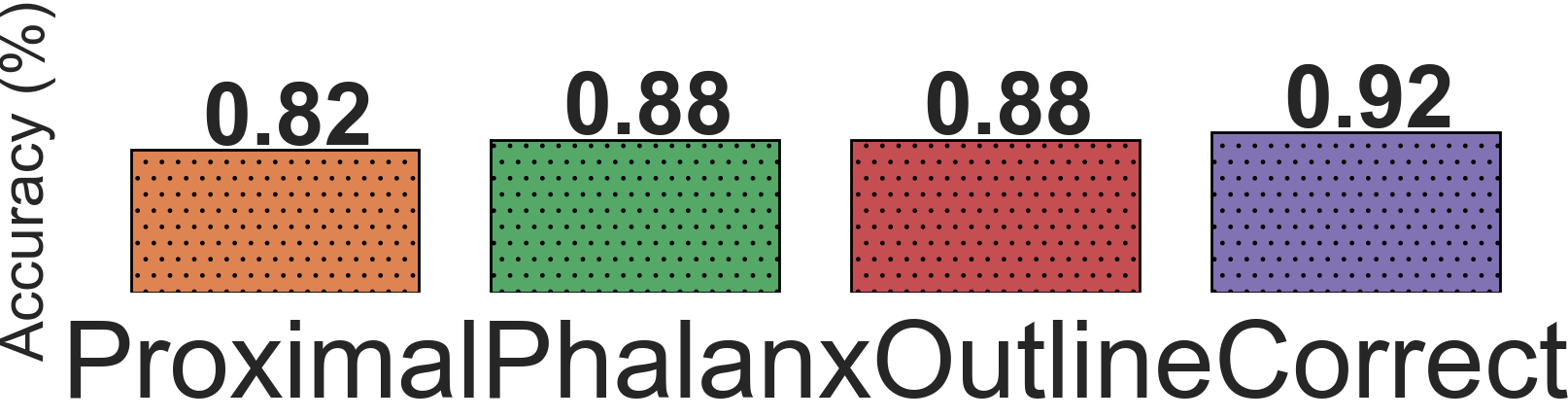}
        \end{minipage}%
\hfill 
        \begin{minipage}{0.19\linewidth}
            \centering
            \includegraphics[width=\linewidth]{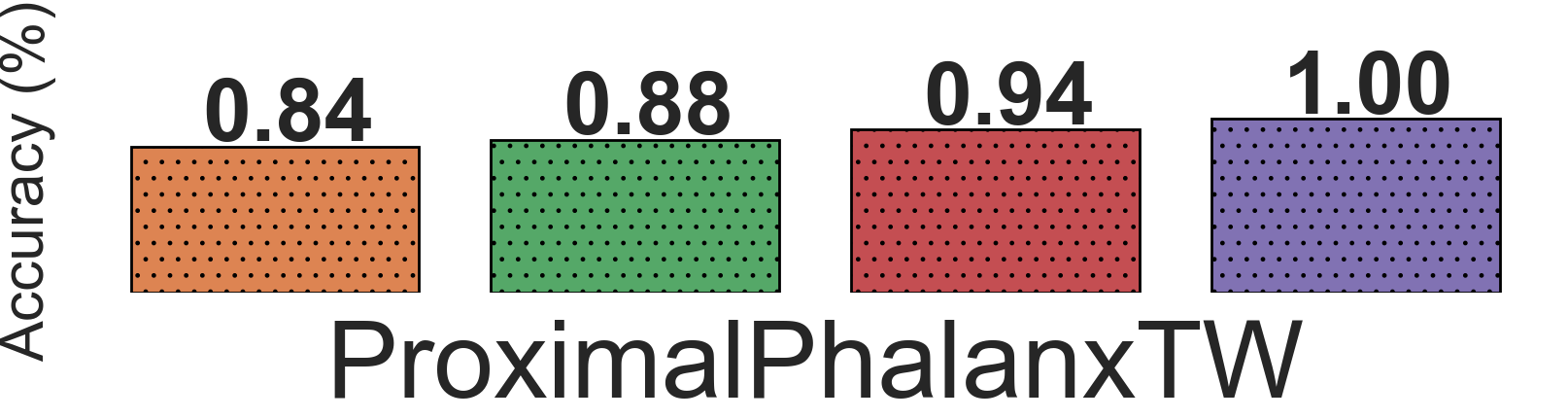}
        \end{minipage}%
\hfill 
        \begin{minipage}{0.19\linewidth}
            \centering
            \includegraphics[width=\linewidth]{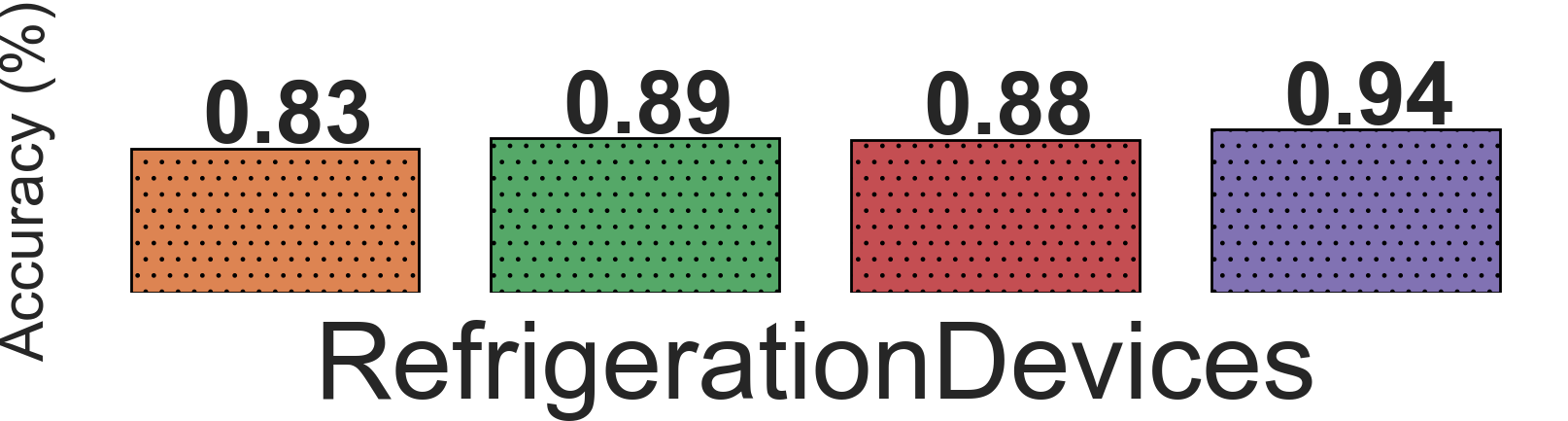}
        \end{minipage}%
\hfill 
        \begin{minipage}{0.19\linewidth}
            \centering
            \includegraphics[width=\linewidth]{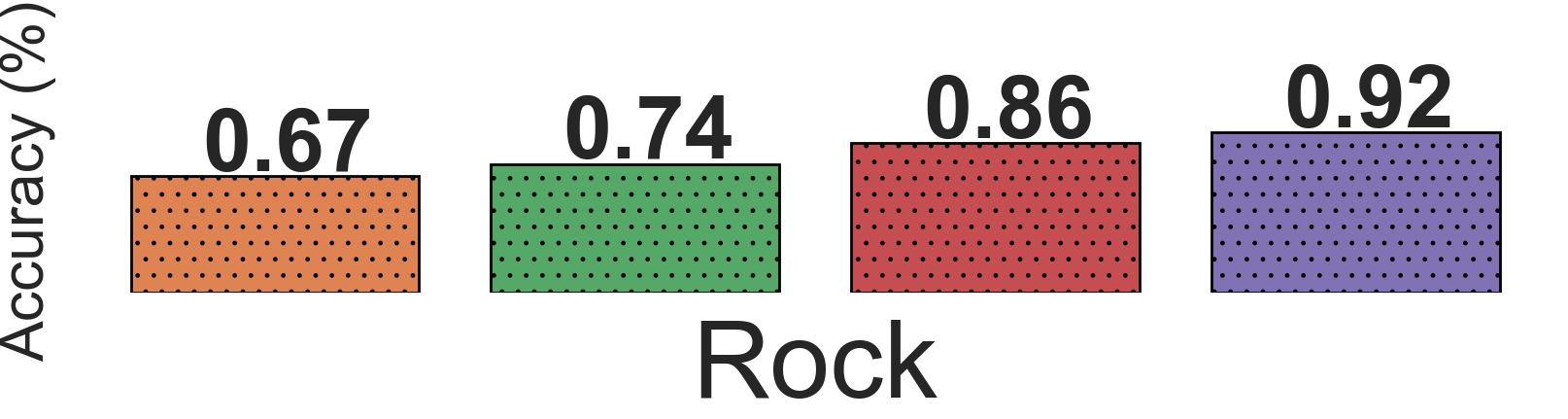}
        \end{minipage}
        \begin{minipage}{0.19\linewidth}
            \centering
            \includegraphics[width=\linewidth]{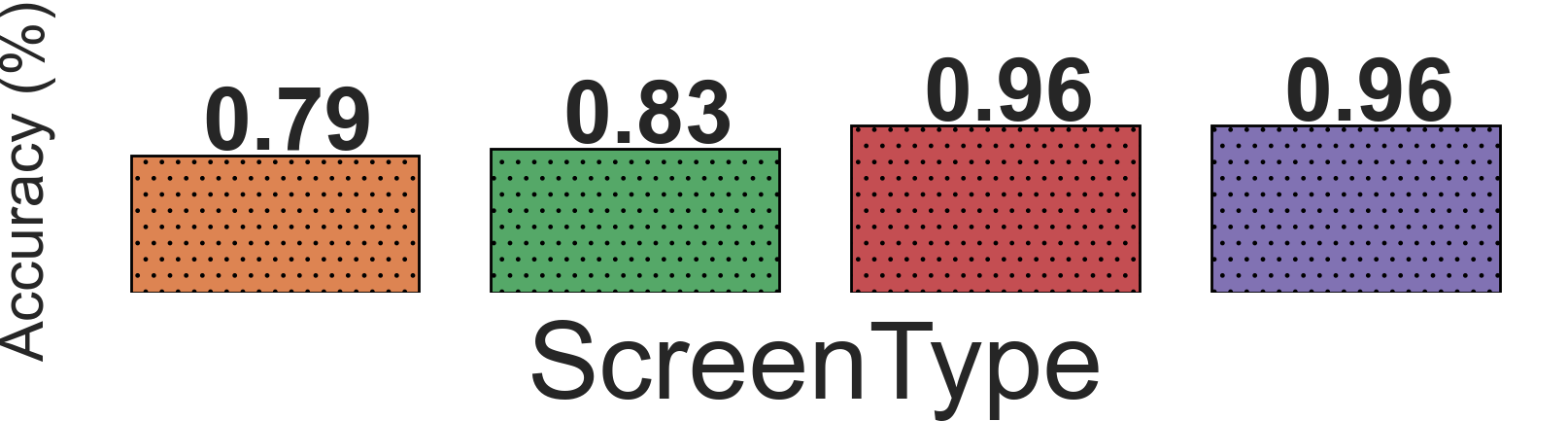}
        \end{minipage}%
\hfill 
        \begin{minipage}{0.19\linewidth}
            \centering
            \includegraphics[width=\linewidth]{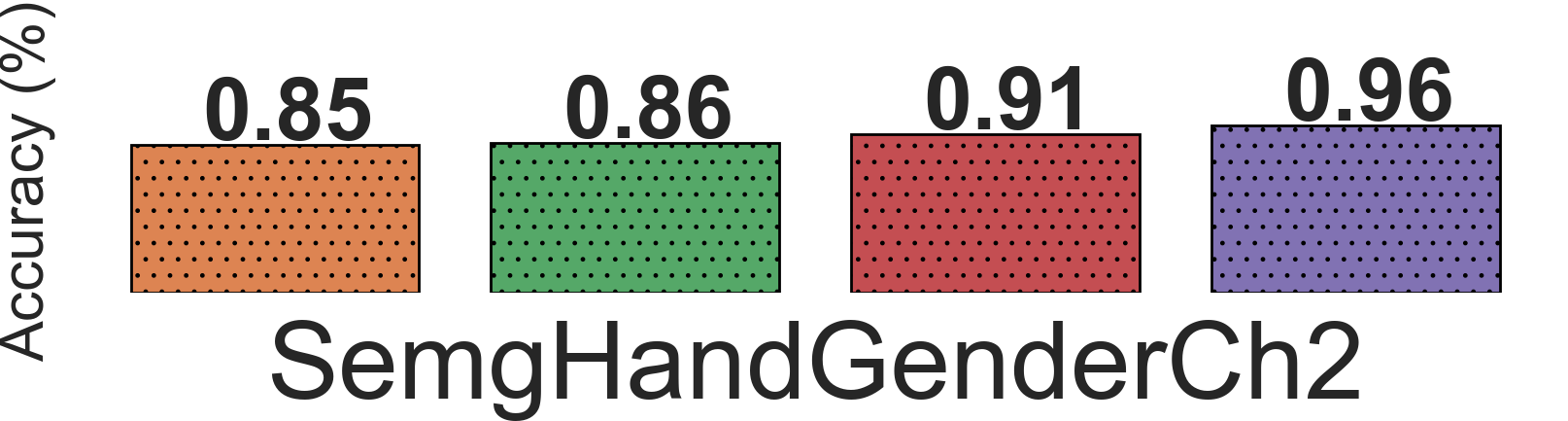}
        \end{minipage}%
\hfill 
        \begin{minipage}{0.19\linewidth}
            \centering
            \includegraphics[width=\linewidth]{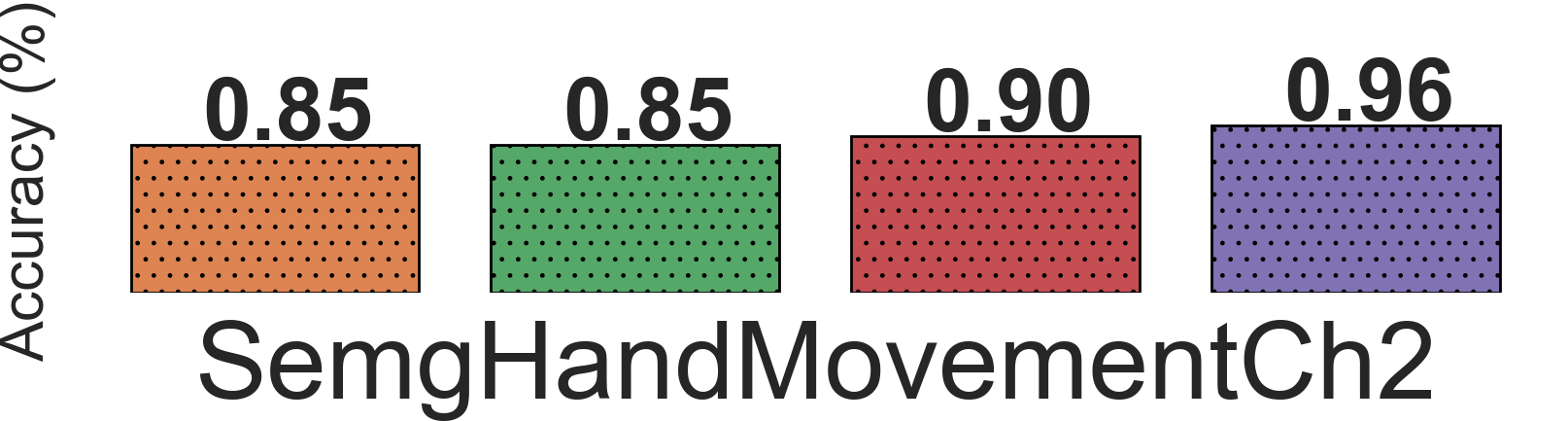}
        \end{minipage}%
\hfill 
        \begin{minipage}{0.19\linewidth}
            \centering
            \includegraphics[width=\linewidth]{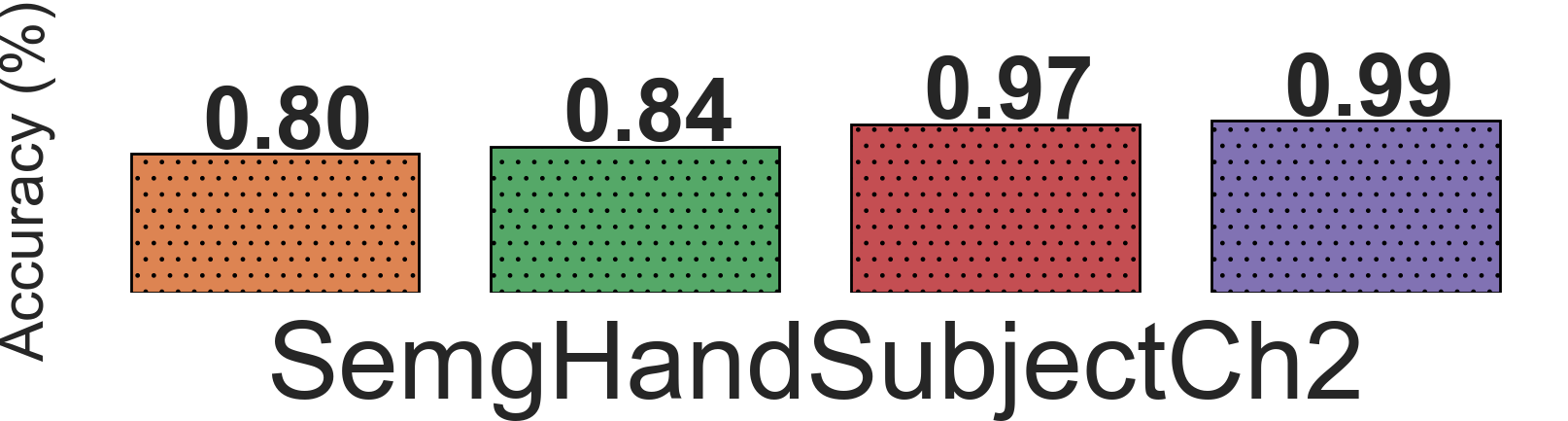}
        \end{minipage}%
\hfill 
        \begin{minipage}{0.19\linewidth}
            \centering
            \includegraphics[width=\linewidth]{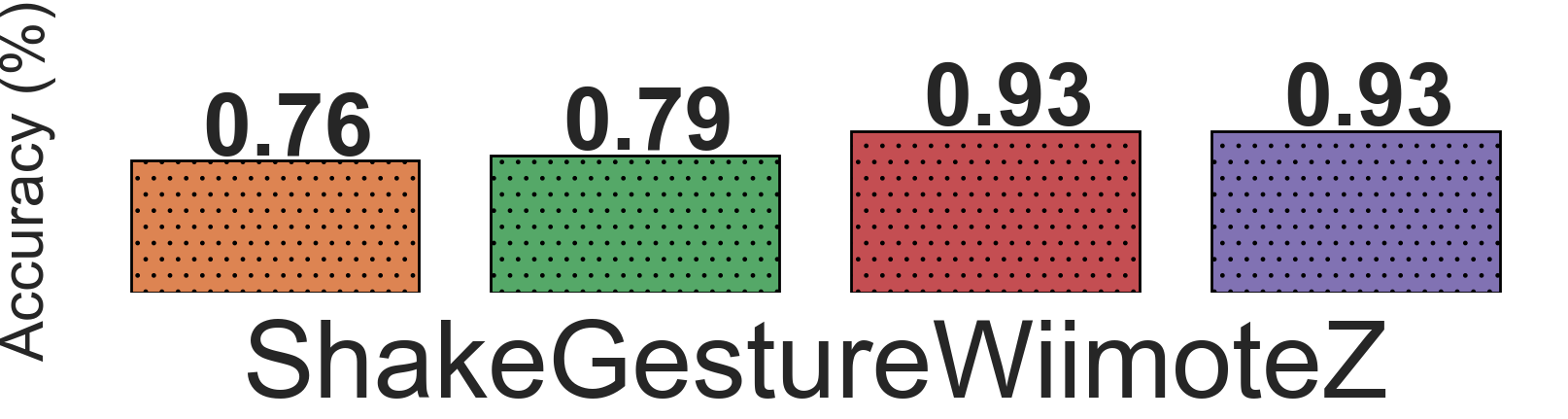}
        \end{minipage}
        \begin{minipage}{0.19\linewidth}
            \centering
            \includegraphics[width=\linewidth]{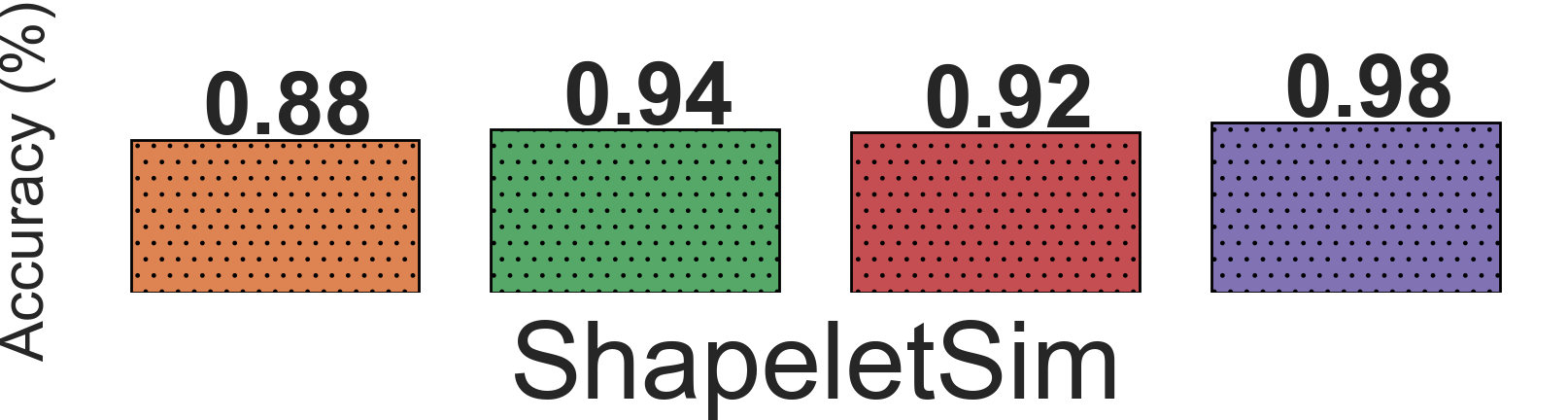}
        \end{minipage}%
\hfill 
        \begin{minipage}{0.19\linewidth}
            \centering
            \includegraphics[width=\linewidth]{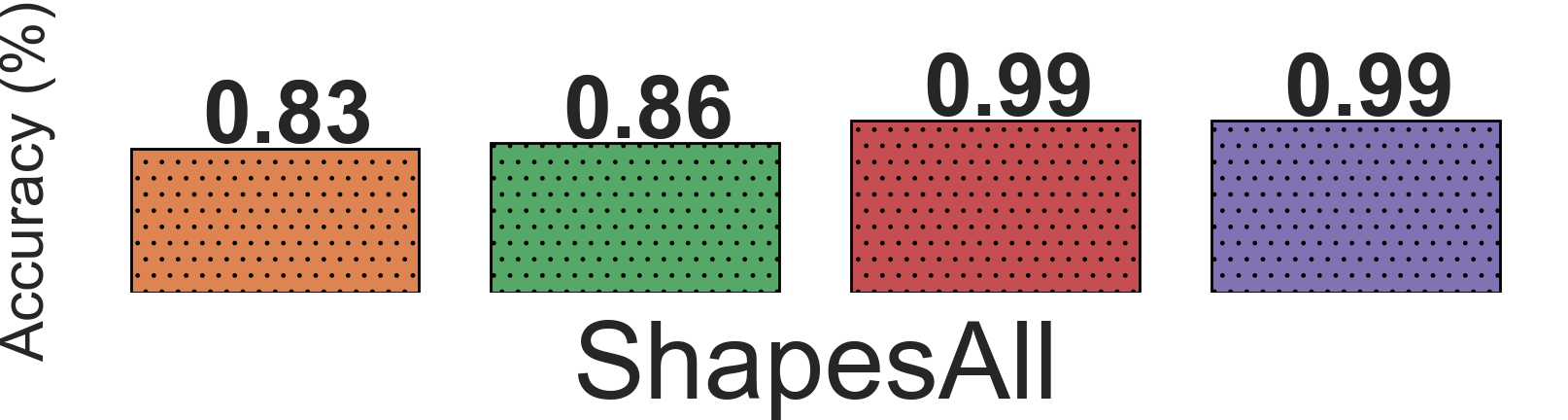}
        \end{minipage}%
\hfill 
        \begin{minipage}{0.19\linewidth}
            \centering
            \includegraphics[width=\linewidth]{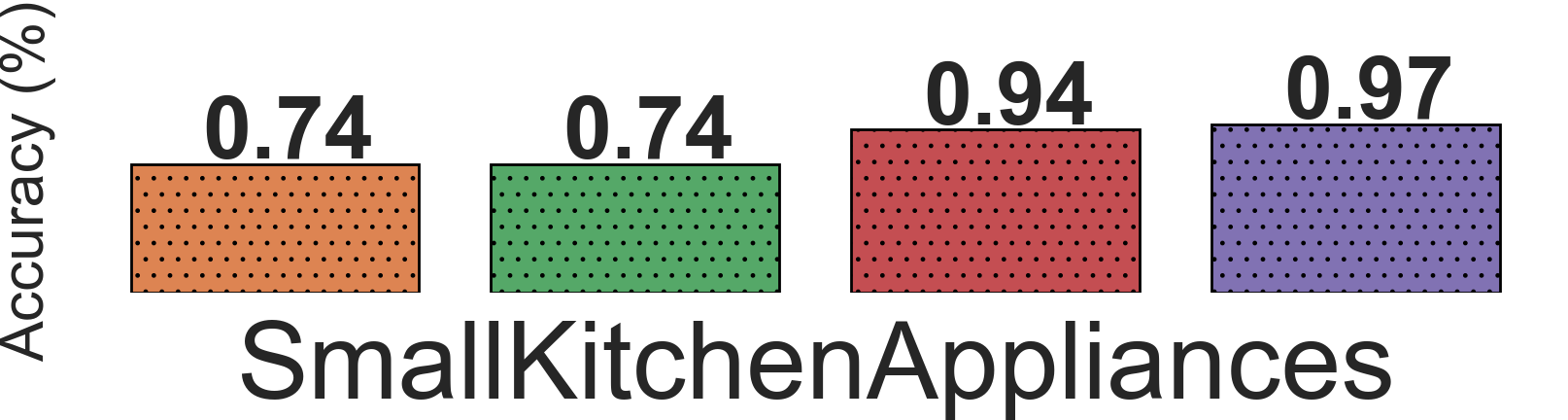}
        \end{minipage}%
\hfill 
        \begin{minipage}{0.19\linewidth}
            \centering
            \includegraphics[width=\linewidth]{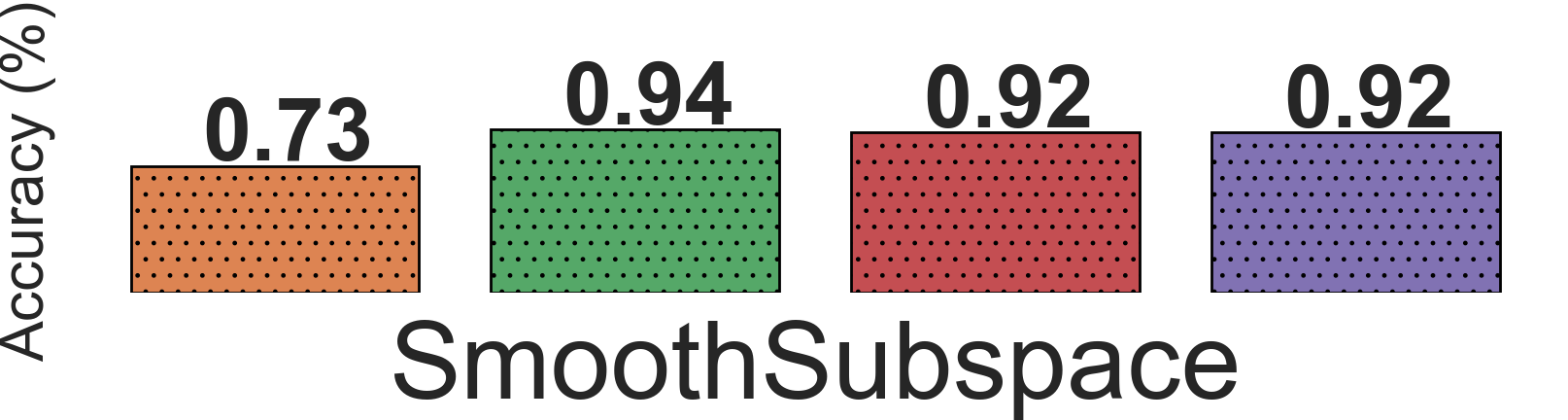}
        \end{minipage}%
\hfill 
        \begin{minipage}{0.19\linewidth}
            \centering
            \includegraphics[width=\linewidth]{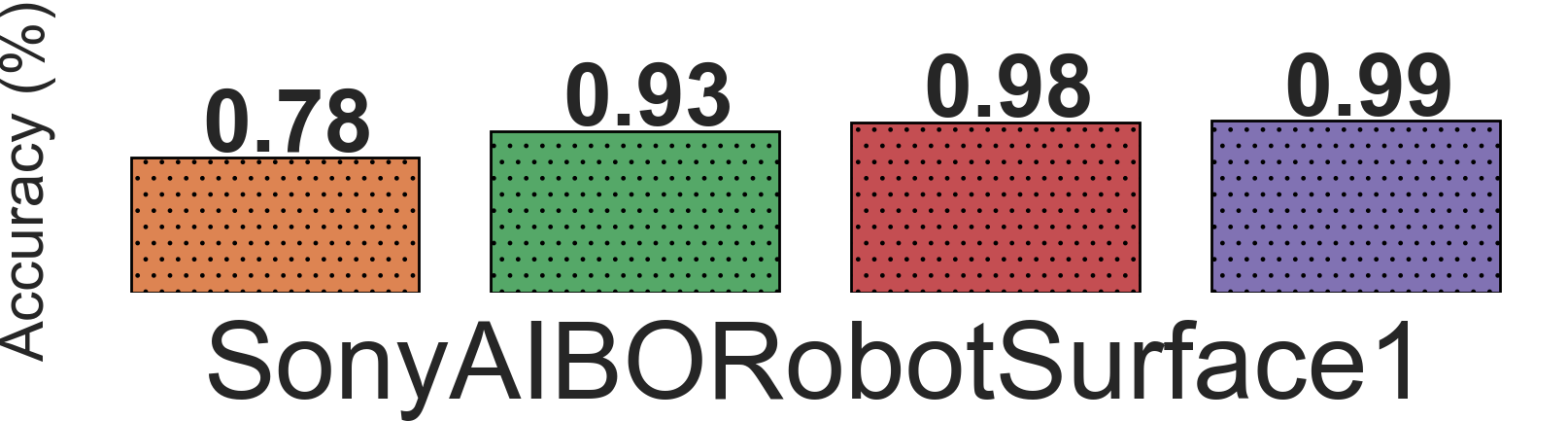}
        \end{minipage}
        \begin{minipage}{0.19\linewidth}
            \centering
            \includegraphics[width=\linewidth]{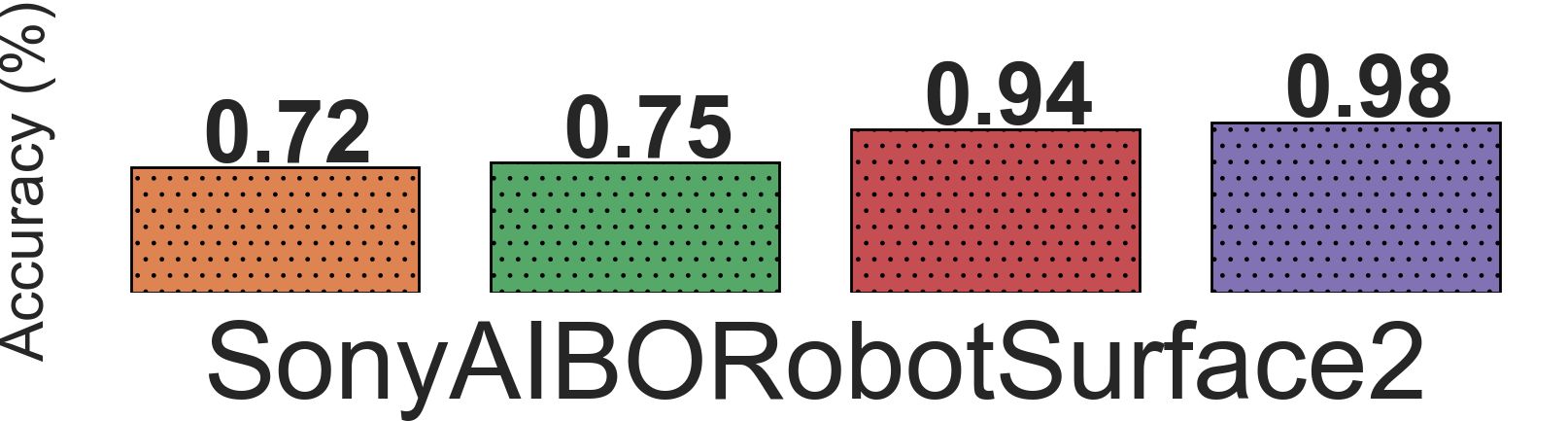}
        \end{minipage}%
\hfill 
        \begin{minipage}{0.19\linewidth}
            \centering
            \includegraphics[width=\linewidth]{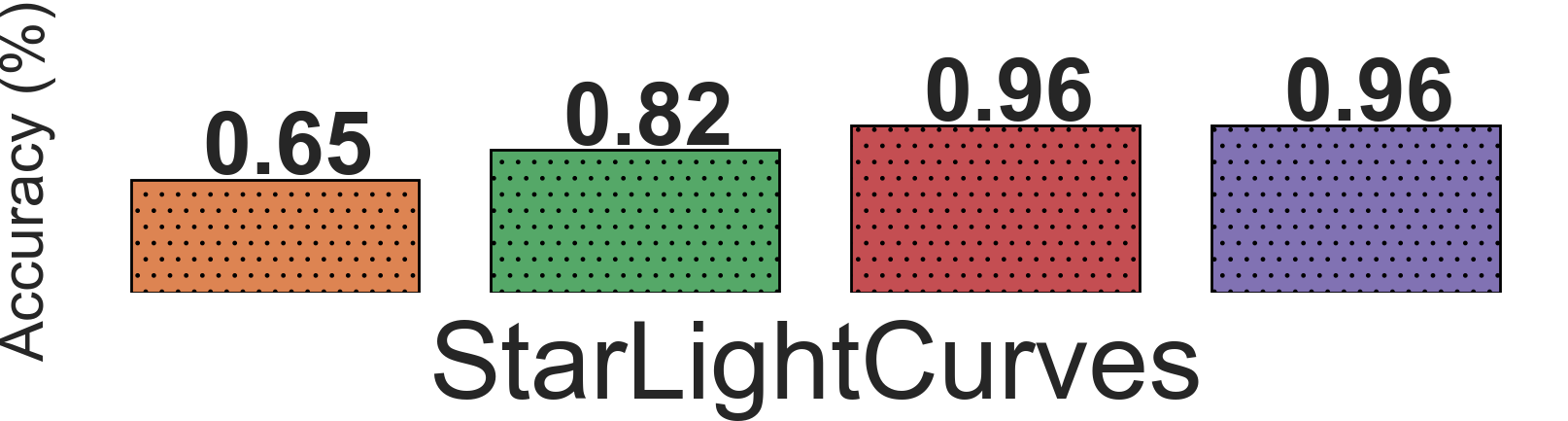}
        \end{minipage}%
\hfill 
        \begin{minipage}{0.19\linewidth}
            \centering
            \includegraphics[width=\linewidth]{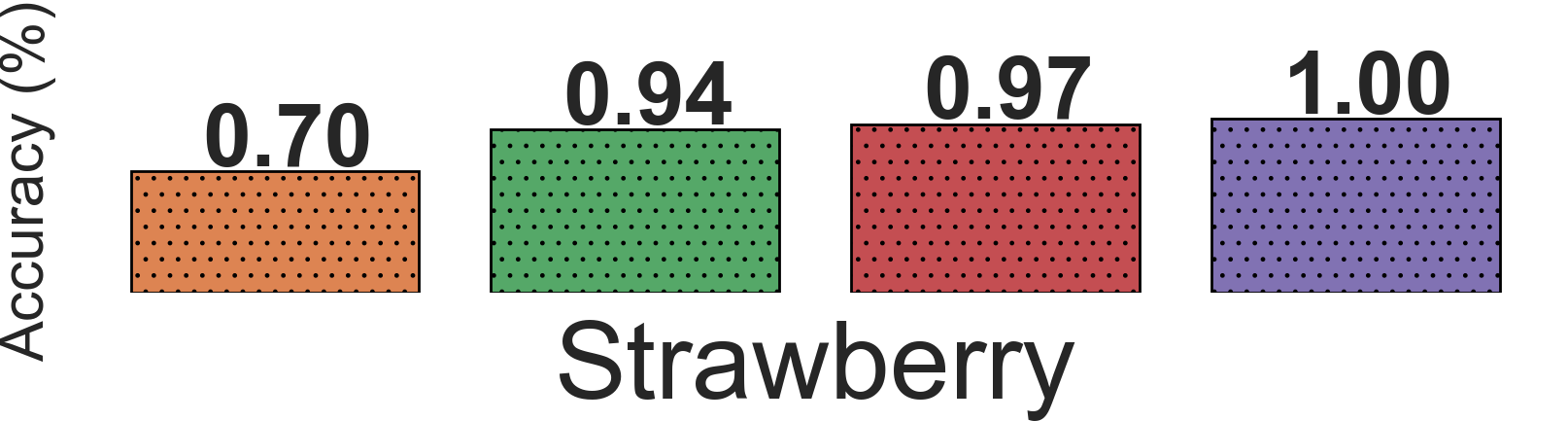}
        \end{minipage}%
\hfill 
        \begin{minipage}{0.19\linewidth}
            \centering
            \includegraphics[width=\linewidth]{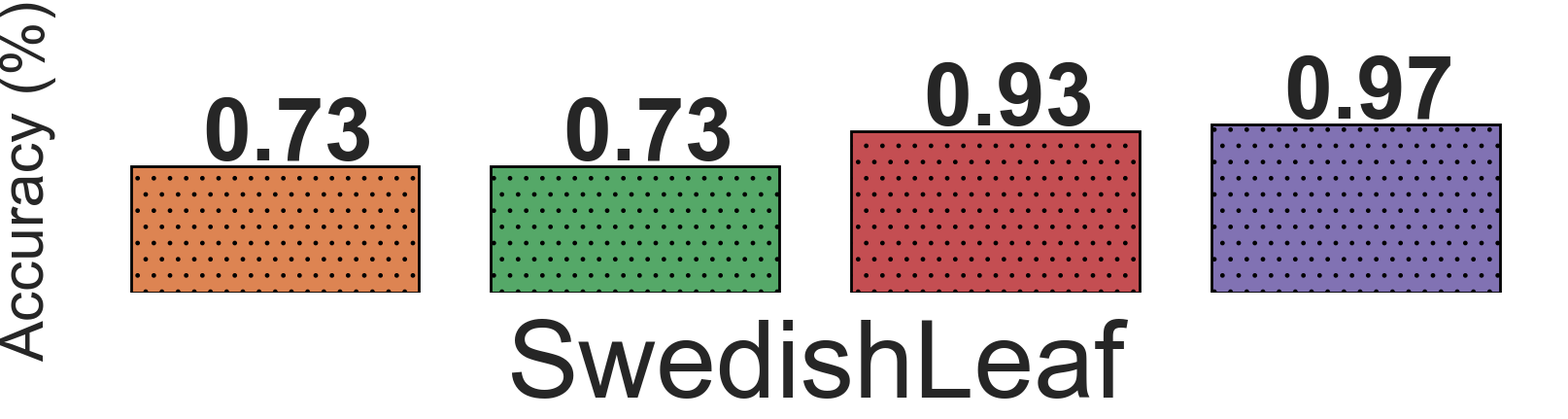}
        \end{minipage}%
\hfill 
        \begin{minipage}{0.19\linewidth}
            \centering
            \includegraphics[width=\linewidth]{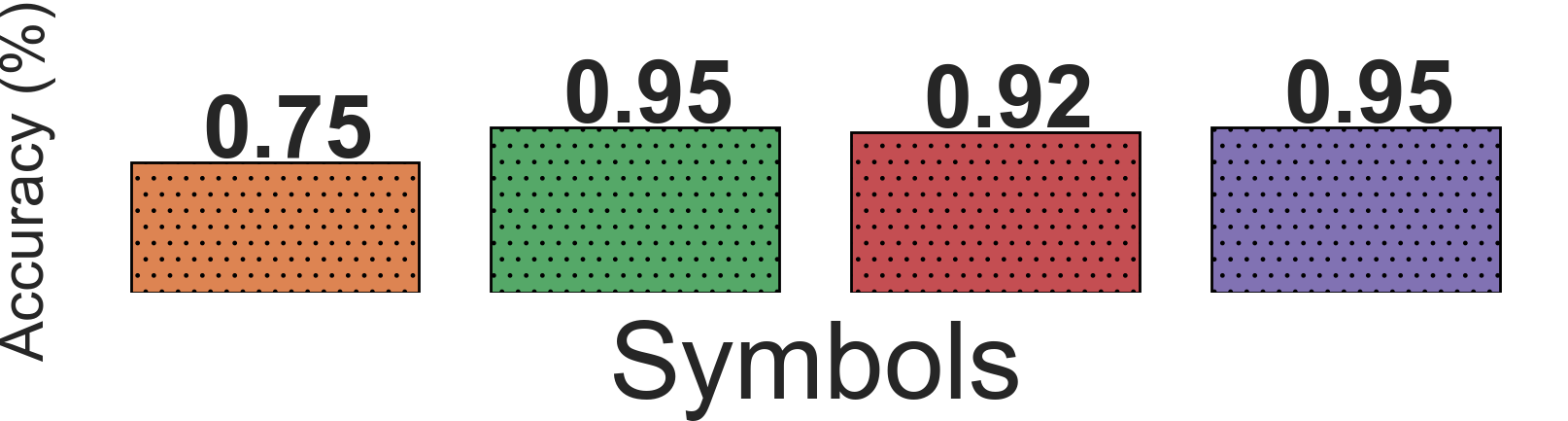}
        \end{minipage}
        \begin{minipage}{0.19\linewidth}
            \centering
            \includegraphics[width=\linewidth]{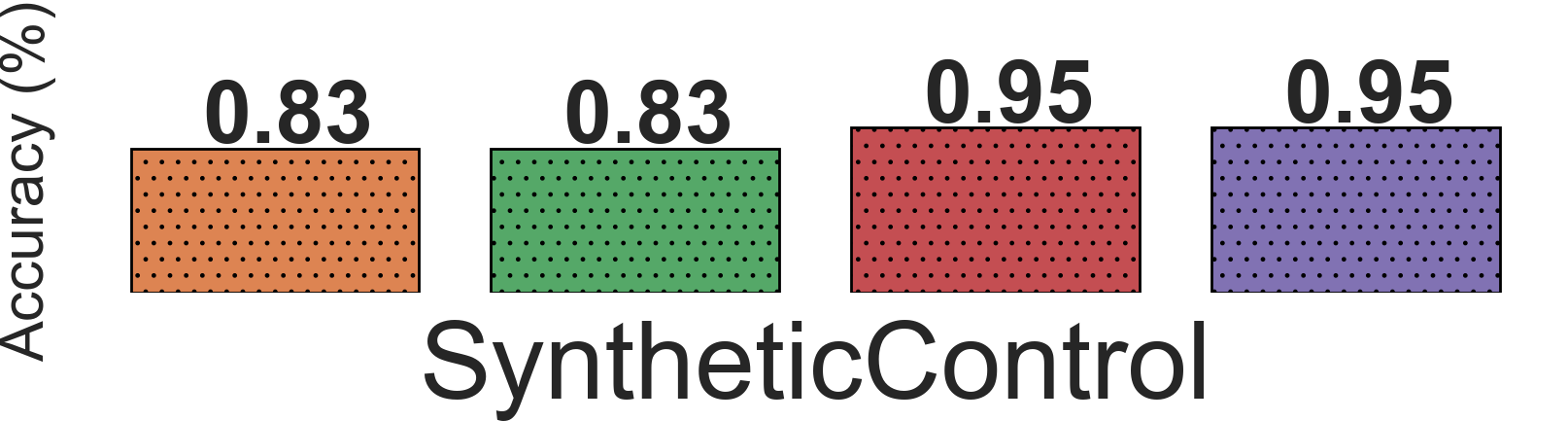}
        \end{minipage}%
\hfill 
        \begin{minipage}{0.19\linewidth}
            \centering
            \includegraphics[width=\linewidth]{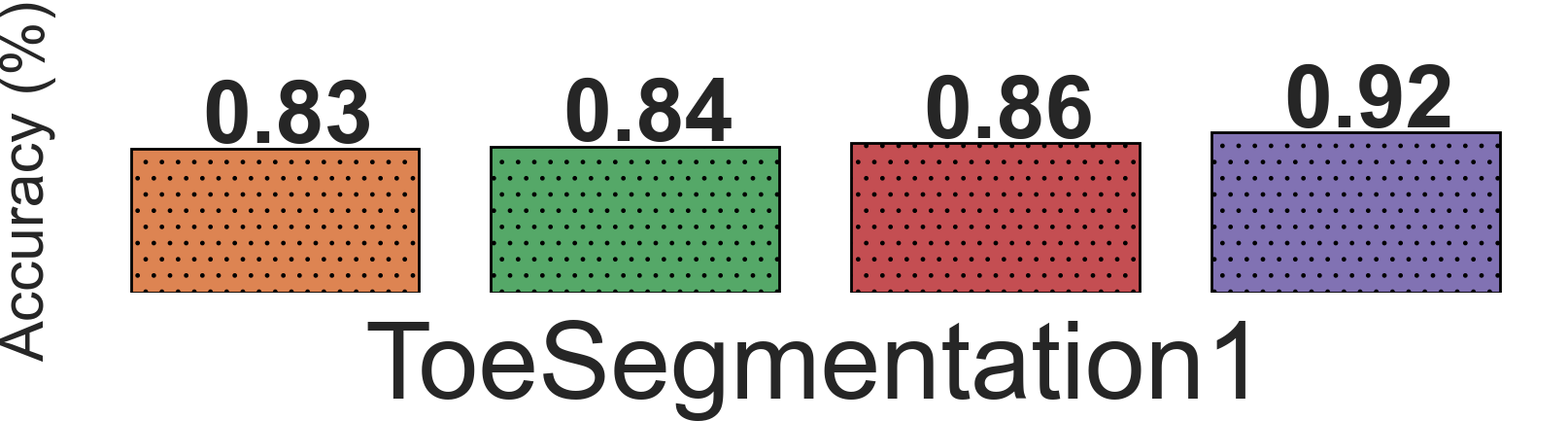}
        \end{minipage}%
\hfill 
        \begin{minipage}{0.19\linewidth}
            \centering
            \includegraphics[width=\linewidth]{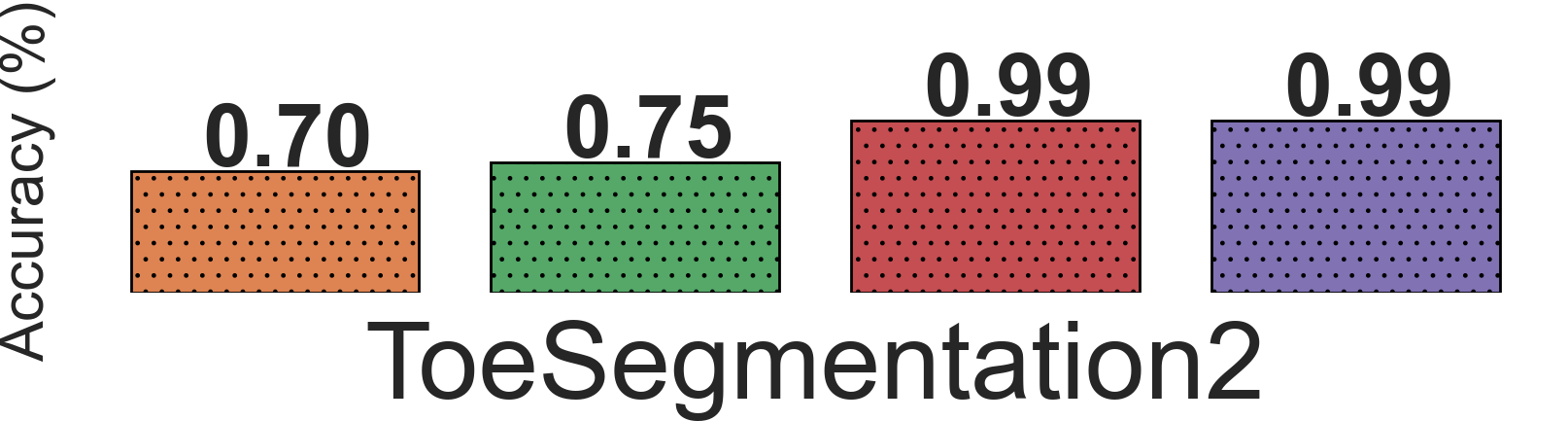}
        \end{minipage}%
\hfill 
        \begin{minipage}{0.19\linewidth}
            \centering
            \includegraphics[width=\linewidth]{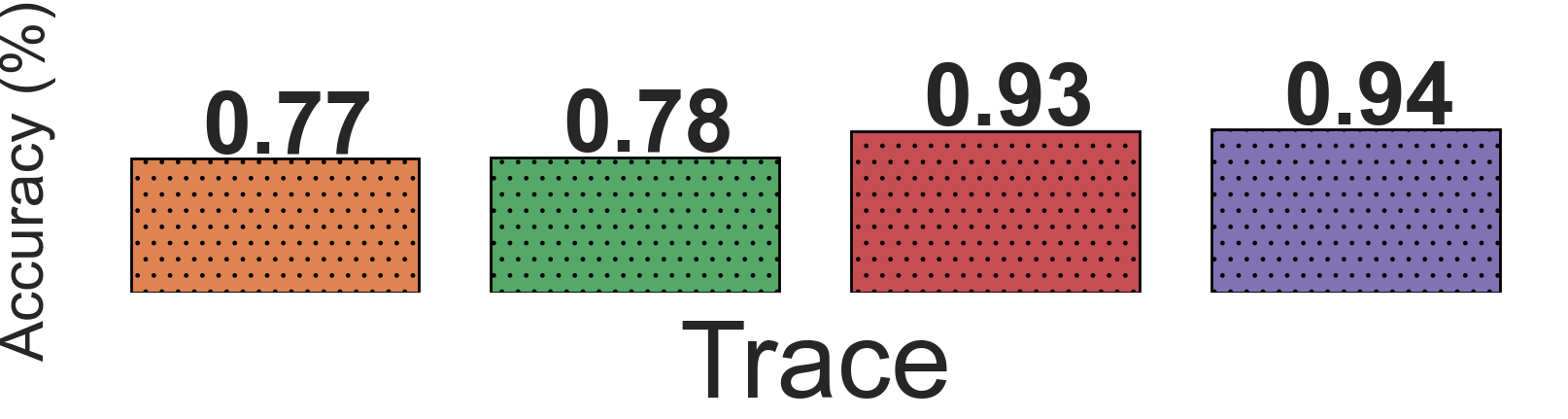}
        \end{minipage}%
\hfill 
        \begin{minipage}{0.19\linewidth}
            \centering
            \includegraphics[width=\linewidth]{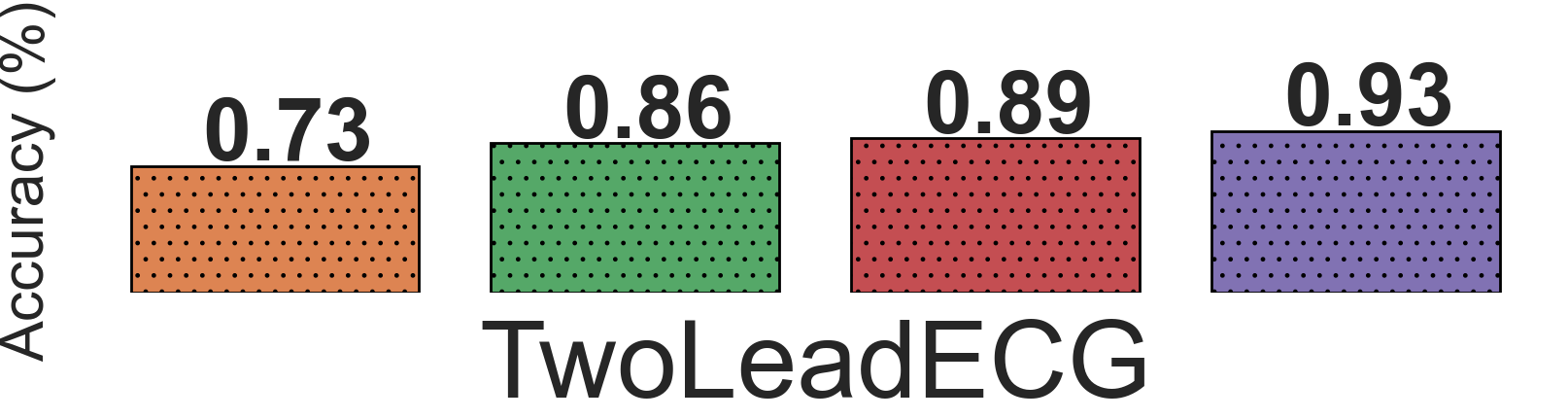}
        \end{minipage}
        \begin{minipage}{0.19\linewidth}
            \centering
            \includegraphics[width=\linewidth]{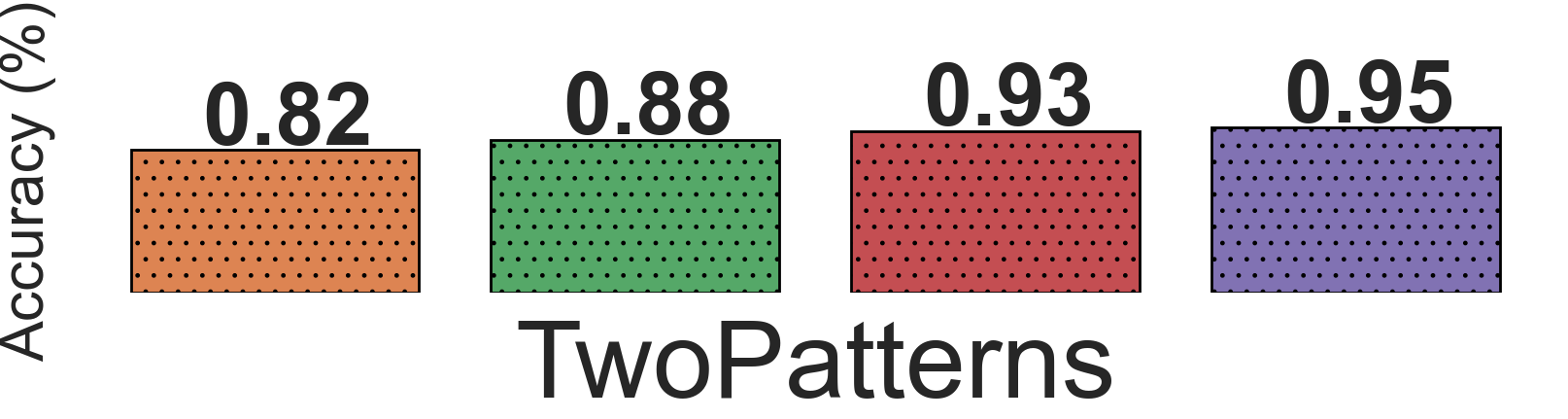}
        \end{minipage}%
\hfill 
        \begin{minipage}{0.19\linewidth}
            \centering
            \includegraphics[width=\linewidth]{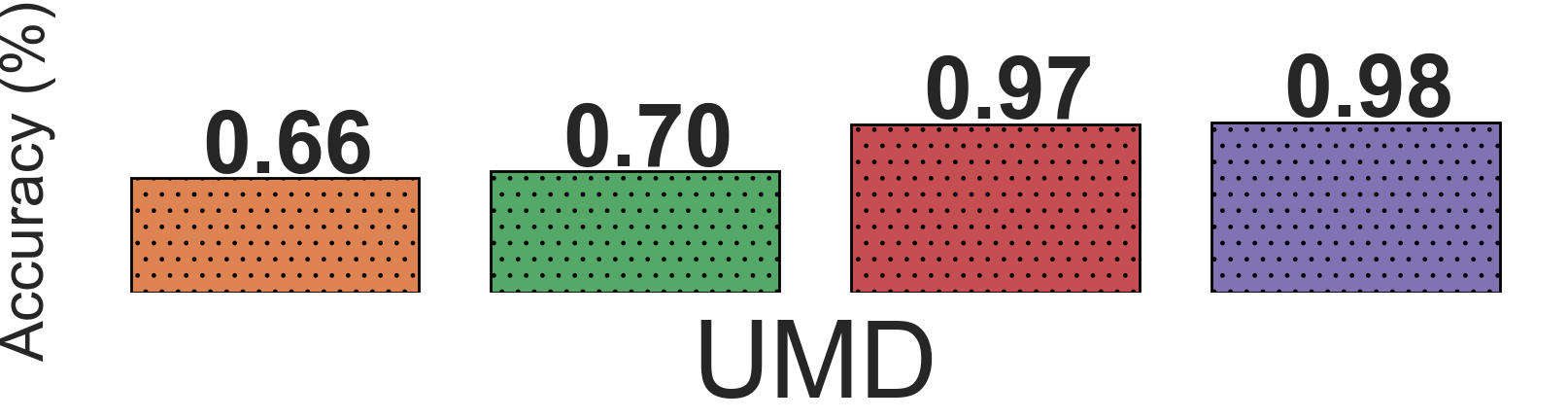}
        \end{minipage}%
\hfill 
        \begin{minipage}{0.19\linewidth}
            \centering
            \includegraphics[width=\linewidth]{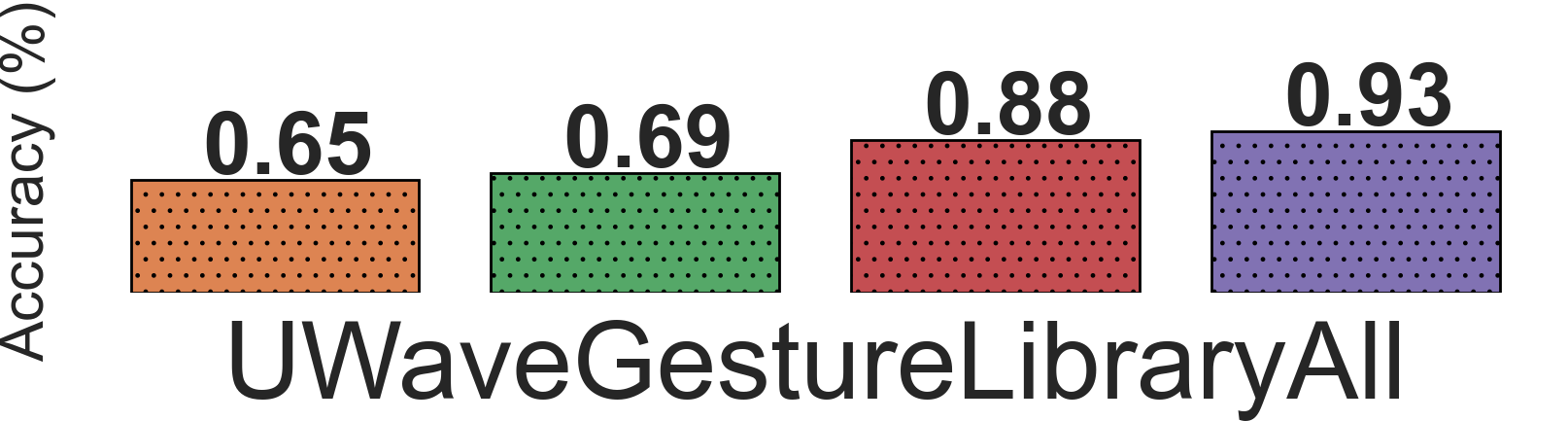}
        \end{minipage}%
\hfill 
        \begin{minipage}{0.19\linewidth}
            \centering
            \includegraphics[width=\linewidth]{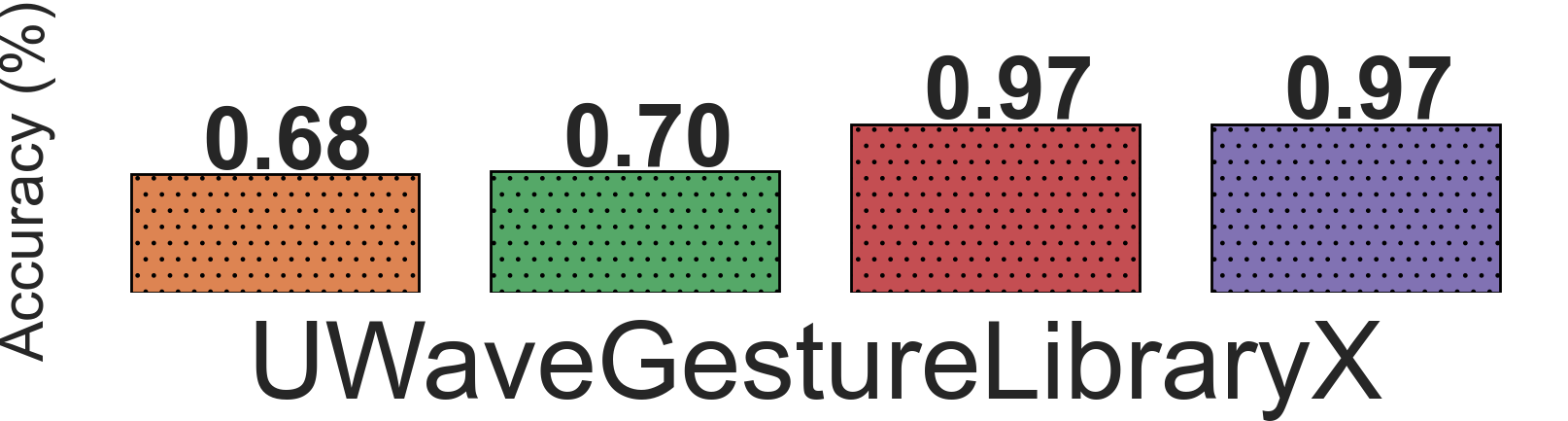}
        \end{minipage}%
\hfill 
        \begin{minipage}{0.19\linewidth}
            \centering
            \includegraphics[width=\linewidth]{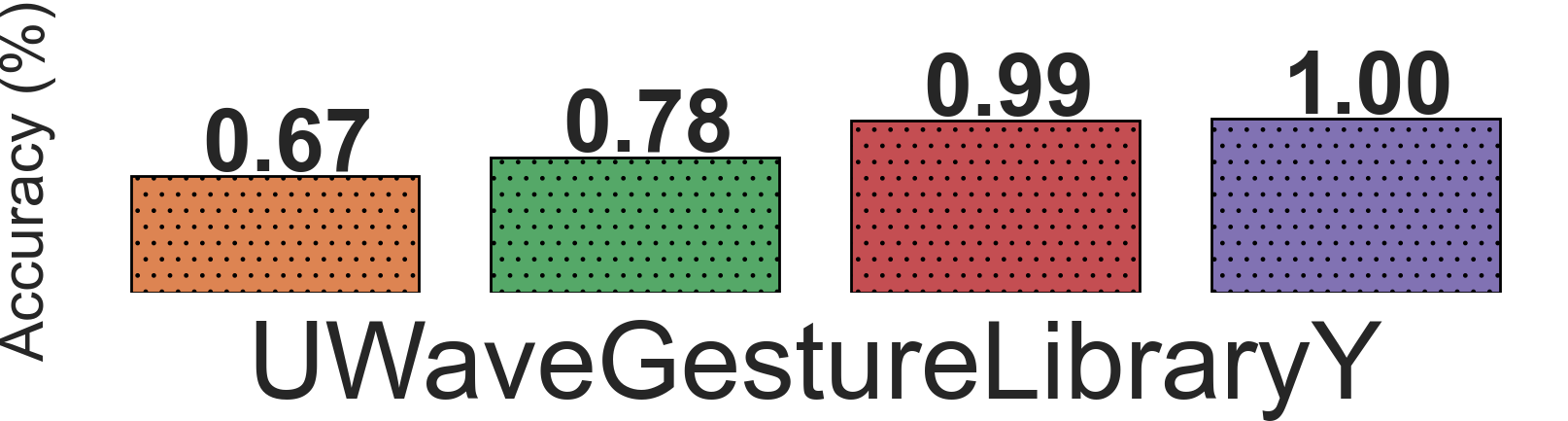}
        \end{minipage}
        \begin{minipage}{0.19\linewidth}
            \centering
            \includegraphics[width=\linewidth]{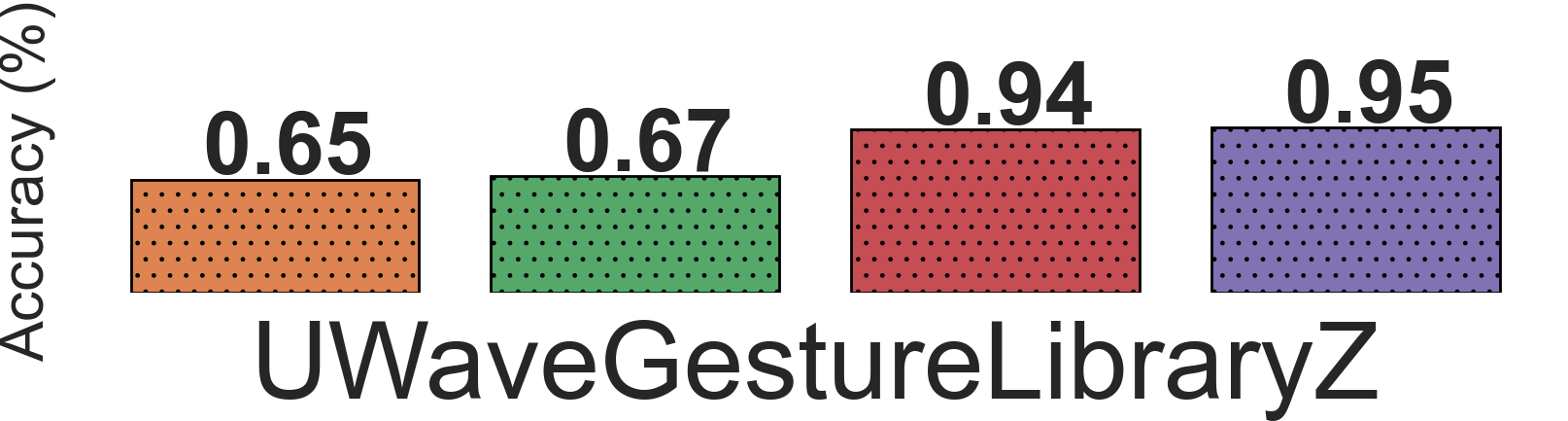}
        \end{minipage}%
\hfill 
        \begin{minipage}{0.19\linewidth}
            \centering
            \includegraphics[width=\linewidth]{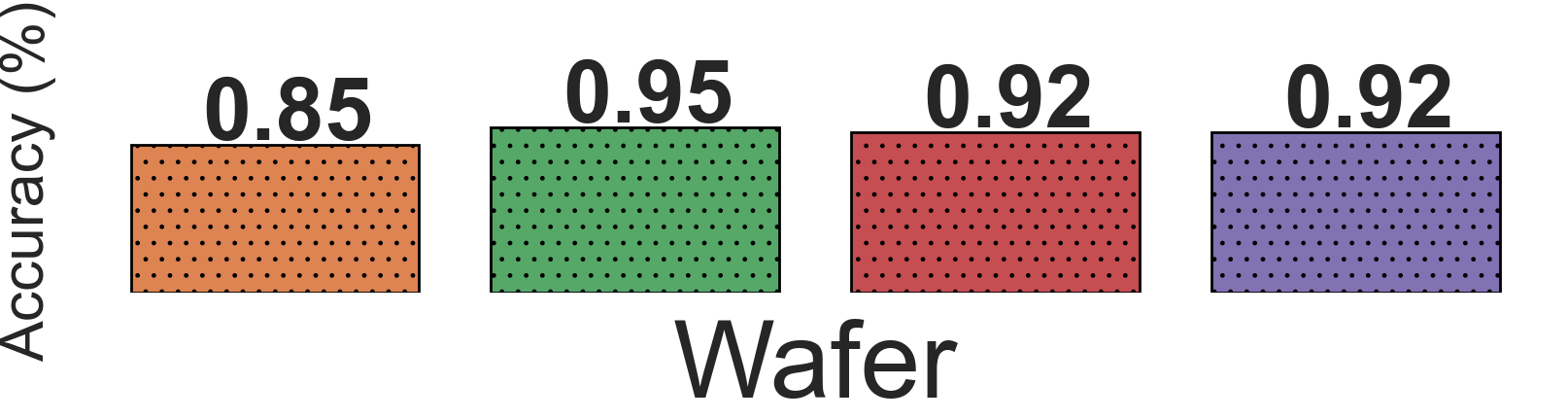}
        \end{minipage}%
\hfill 
        \begin{minipage}{0.19\linewidth}
            \centering
            \includegraphics[width=\linewidth]{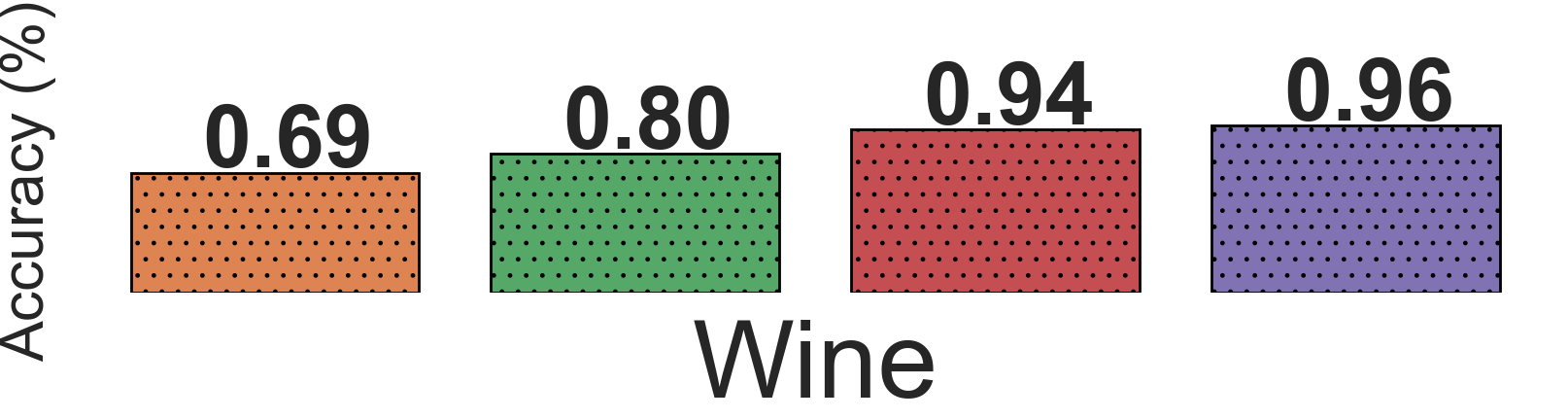}
        \end{minipage}%
\hfill 
        \begin{minipage}{0.19\linewidth}
            \centering
            \includegraphics[width=\linewidth]{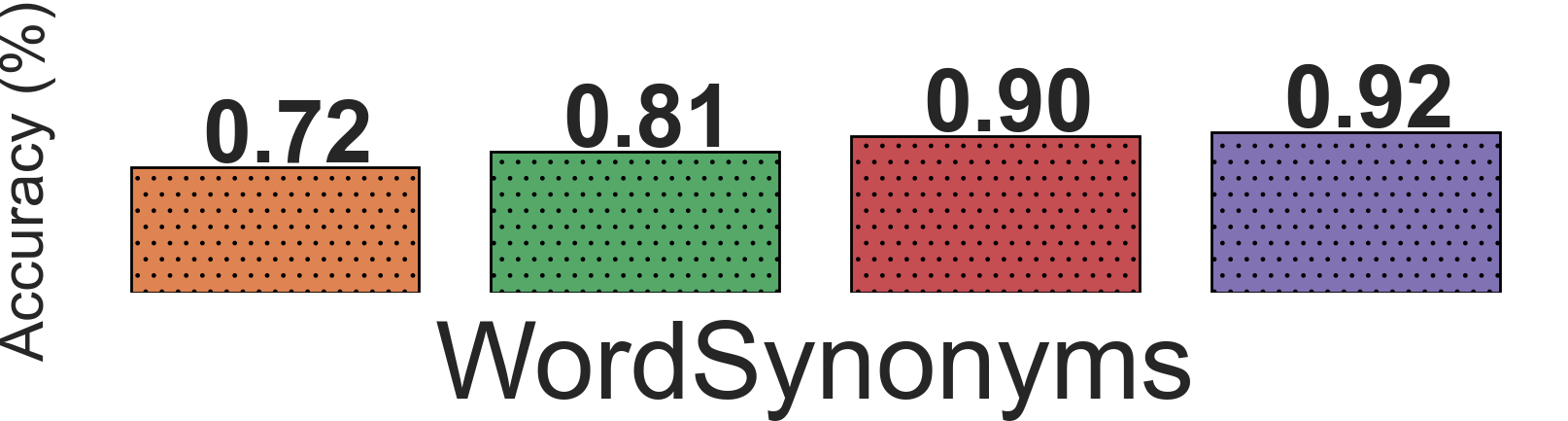}
        \end{minipage}%
\hfill 
        \begin{minipage}{0.19\linewidth}
            \centering
            \includegraphics[width=\linewidth]{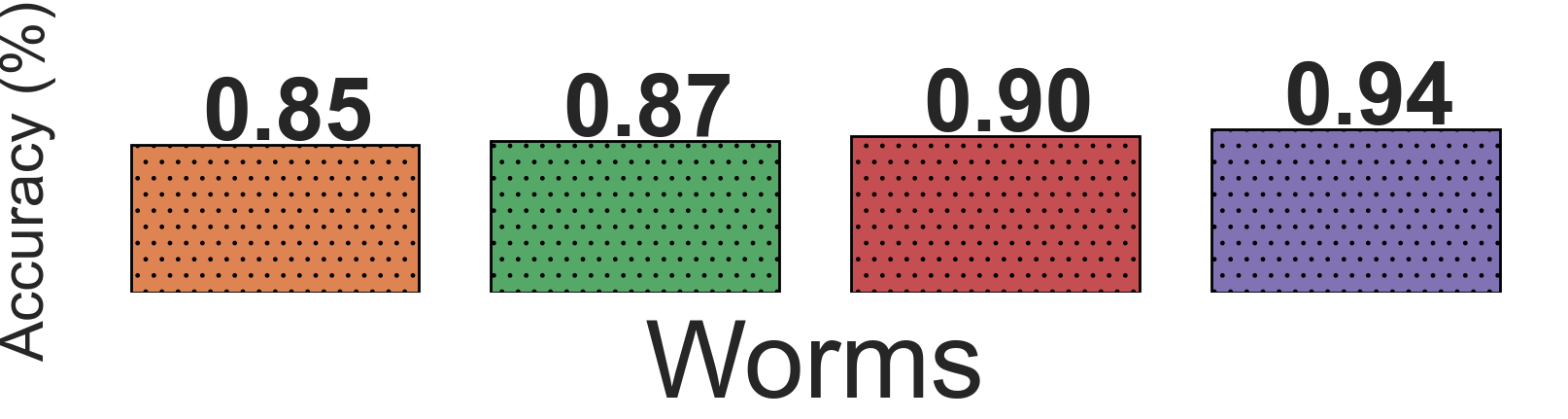}
        \end{minipage}
        \begin{minipage}{0.19\linewidth}
            \centering
            \includegraphics[width=\linewidth]{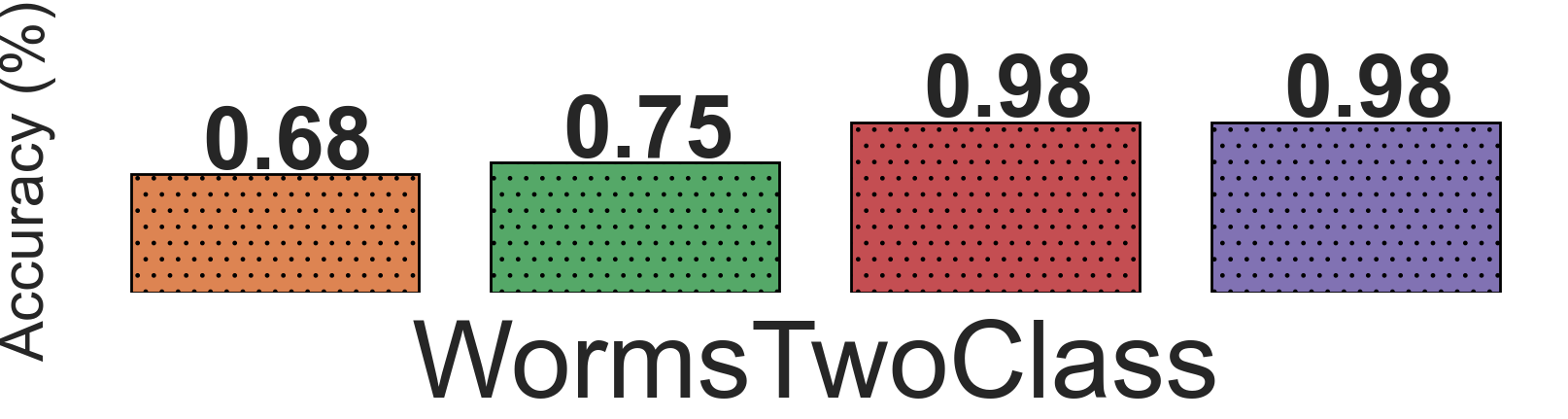}
        \end{minipage}%
        \begin{minipage}{0.19\linewidth}
            \centering
            \includegraphics[width=\linewidth]{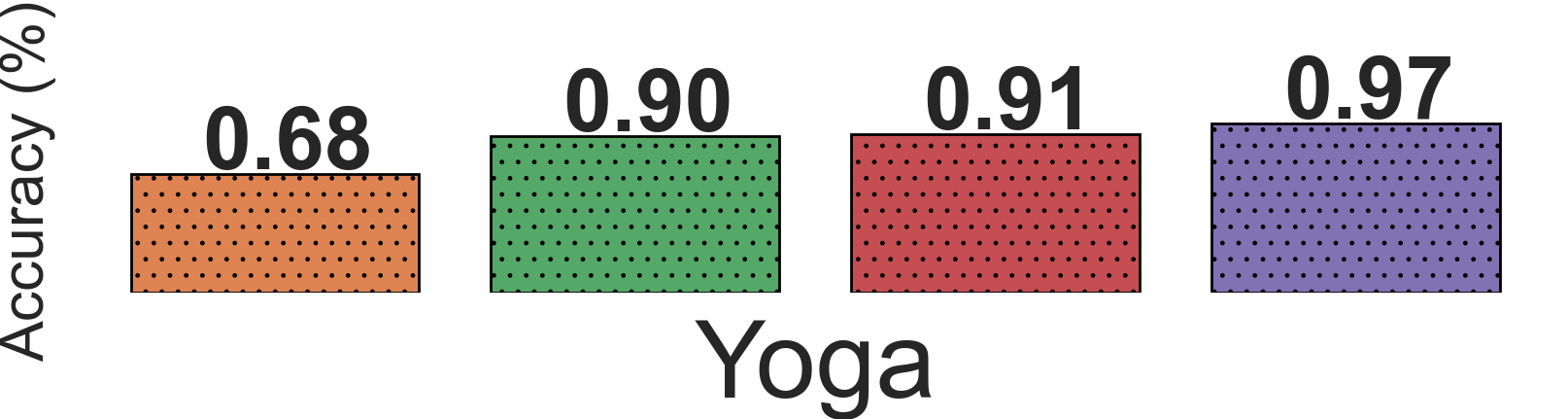}
        \end{minipage}
    \end{minipage}
\caption{Results of DTW-AR based adversarial training to predict the true labels of adversarial examples generated by DTW-AR and the baseline attack methods on the Univariate dataset. The adversarial examples considered are those that successfully fooled DNNs that do not use adversarial training.}
\label{fig:unidtwadv}
\end{figure*}

\vspace{1.0ex}

\noindent \textbf{Results on the UCR univariate datasets.}
To show the effectiveness of our proposed method, we additionally evaluate DTW-AR on  univariate datasets from the UCR time-series benchmarks repository \cite{ucrdata}. We show the results of DTW-AR based adversarial training to predict the ground-truth labels of adversarial attacks generated by DTW-AR and the baseline attack methods on the univariate datasets in Figure \ref{fig:unidtwadv}. We observe similar results as the datasets shown in the main paper. DTW-AR is successful in identifying attacks from the baseline methods by creating robust deep models. These strong results show that DTW-AR outperforms baselines for creating more effective adversarial attacks and yields to more robust DNN classifiers. We conclude that the proposed DTW-AR framework is generic, and more suitable for time-series domain to create robust deep models.


\newpage
\hspace{1ex}
\newpage
\hspace{1ex}
\newpage
\section{Theoretical Proofs}
\subsection{Proof of Observation 1}
\textit{Let $l_2$ be the equivalent of Euclidean distance using the cost matrix in the DTW space. $\forall X \in \mathbb{R}^{n \times T}$, there exists $\epsilon \in \mathbb{R}^{n \times T} $ and an alignment path $P$ such that $ dist_P(X,X+\epsilon) \le \delta$ and $l_2(X,X+\epsilon) > \delta$.}

The existence of $\epsilon$ is guaranteed as follows: We know from the nature of the DTW algorithm and the alignment paths that for two time-series signals $X$ and $X'$, the optimal alignment path is not always the diagonal path. If $\epsilon$ does not exist, it means that for all signals $X'$ that are different from $X$, the diagonal path is an optimal alignment path, which is absurd. Thus, $\epsilon=X'-X$ and it always exists for any time-series signal .
 
Let $P_{diag}$ be the diagonal alignment path in the cost matrix $C$. 

For $X \in \mathbb{R}^{n \times T}$, let $\epsilon \in \mathbb{R}^{n \times T}$ such that the optimal alignment path $P^*$ between $X$ and $X+\epsilon$ is different than $P_{diag}$.

Let us suppose that there is no alignment path $P$ between $X$ and $X+\epsilon$ such that $dist_P(X,X+\epsilon) \le \delta$ and $l_2(X,X+\epsilon) > \delta$.
The last statement is equivalent to: $dist_P(X,X+\epsilon) < dist_{P_{diag}}(X,X+\epsilon)$.

Since we assumed that there is no alignment path $P$ that satisfies this statement, this implies:
\begin{align*}
    \forall P~,~ dist_P(X,X+\epsilon) \ge dist_{P_{diag}}(X,X+\epsilon) \\
    \Rightarrow dist_{P^*}(X,X+\epsilon) \ge dist_{P_{diag}}(X,X+\epsilon) \\
    \Rightarrow DTW(X,X+\epsilon) \ge dist_{P_{diag}}(X,X+\epsilon) 
\end{align*}

Therefore, from the definition of $DTW(\cdot, \cdot)$ as a $min$ operation during backtracing of the DP process, we get:
\begin{align*}
    \Rightarrow DTW(X,X+\epsilon) = dist_{P_{diag}}(X,X+\epsilon)
\end{align*}
Hence, $P_{diag}$ = $P^*$, which contradicts our main assumption in constructing $\epsilon$ such that $P_{diag} \neq P^*$.

Therefore, we conclude that:
\begin{align*}
\exists P \; \text{s.t.} \;  dist_P(X,X+\epsilon) \le \delta ~\text{and}~ l_2(X,X+\epsilon) > \delta
\end{align*}

\subsection{Proof of Theorem 1}
\textit{For a given input space $\mathbb{R}^{n \times T}$, a constrained DTW space for adversarial examples is a strict superset of a constrained euclidean space for adversarial examples. If $X\in \mathbb{R}^{n\times T}$:
\begin{equation}
\footnotesize
    \bigg\{  X_{adv} \big|  DTW(X,X_{adv}) \le \delta\bigg\} \supset  \bigg\{ X_{adv} \big| \|X-X_{adv}\|_2^2 \le \delta \bigg\}
\label{eq:dtwspaceth}
\end{equation}}

We want to prove that a constrained DTW space allows more candidates adversarial examples than a constrained Euclidean space. Let $X \in \mathbb{R}^{n \times T}$ be an input time-series and $X_{adv}$ denote a candidate adversarial example generated from $X$. In the DTW-space, this requires that $DTW(X,X_{adv}) \le \delta$. In the Euclidean space, this requires that $\|X-X_{adv}\|_2^2 \le \delta$, which is equivalent to $dist_{P_{diag}}(X, X_{adv})  \le \delta$.

Suppose $\mathcal{A}$ be the space of all candidate adversarial examples in DTW space $\big\{X_{adv} / DTW(X,X_{adv}) \le \delta\big\}$ and $\mathcal{B}$ be the space of all candidate adversarial examples in Euclidean space  $\big\{X_{adv} \big/ \|X-X_{adv}\|_2^2 \le \delta \big\}$.

To prove $\mathcal{A} \supsetneq \mathcal{B}$, we need to prove:
\begin{enumerate}
    \item $\forall X_{adv} \in \mathcal{B}/ X_{adv} \in \mathcal{A}$
    \item $\exists X_{adv} / X_{adv} \in \mathcal{A}$ and $X_{adv} \notin \mathcal{B}$
\end{enumerate}

Statement 1: Let $X_{adv} \in \mathcal{B}$. For the optimal alignment path $P^*$ between $X$ and $X_{adv}$, if:
\begin{itemize}
    \item $P^*=P_{diag} \Rightarrow$  $DTW(X,X_{adv})=dist_{P_{diag}}(X,X_{adv})$
    
    $\Rightarrow$ $X_{adv} \in \mathcal{A}$
    \item $P^* \neq P_{diag} \Rightarrow$ According to Observation 1:
    
    $DTW(X,X+\epsilon) < dist_{P_{diag}}(X,X+\epsilon) $
    
    $\Rightarrow$ $X_{adv} \in \mathcal{A}$
\end{itemize}
Hence, we have $\forall X_{adv} \in \mathcal{B}/ X_{adv} \in \mathcal{A}$.

Statement 2: Let $X_{adv} \in \mathcal{A}$ such that $P^* \neq P_{diag}$. Consequently, according to Observation 1, $dist_{P_{diag}}(X,X+\epsilon) > DTW(X,X+\epsilon)$. 

$\Rightarrow dist_{P_{diag}}(X,X+\epsilon) > \delta$.

As the diagonal path corresponds to the Euclidean distance, we conclude that $X_{adv} \notin \mathcal{B}$.

Hence, $\exists X_{adv} / X_{adv} \in \mathcal{A}$ and $X_{adv} \notin \mathcal{B}$.

\subsection{Proof of Observation 2}
\textit{Given any alignment path $P$ and two multivariate time-series signals $X,Z \in \mathbb{R}^{n \times T}$. If we have $dist_P(X,Z) \le  \delta$, then $DTW(X,Z) \le \delta$.}

Let $P$ any given alignment path and $P^*$ be the optimal alignment path used for DTW measure along with the DTW cost matrix $C$. Let us suppose that
$dist_P(X,Z) > DTW(X,Z)$.

We denote $P$=$\{(1,1), \cdots, (i,j), \cdots, (T,T)\}$ and $P^*$=$\{(1,1), \cdots, (i^*,j^*), \cdots, (T,T)\}$. Let us denote by $k$ the index at which, $P$ and $P^*$ are not using the same cells anymore, and by $l$, the index where $P$ and $P^*$ meet again using the same cells until $(T,T)$ in a continuous way. By definition, $k>1$ and $l<\min(len(P), len(P^*))$. For example, if $P$=$\{(1,1), (1,2), (2,2), (3,3), (3,4), (4,5), (5,5)\}$ and $P^*$=$\{(1,1), (1,2), (2,3), (3,4), (4,4), (5,5)\}$, then $k$=3 and $l$=6.
\begin{itemize}
    \item If $k$=$l$, then $P$=$P^*$. Therefore, $dist_P(X,Z) > DTW(X,Z)$ is absurd.
    \item If $k\neq l$: To provide the $(k+1)^{th}$ element of $P^*$, we have $C_{(i^*_{k+1},j^*_{k+1})}$ = $d(X_{i^*_{k+1}}, Z_{j^*_{k+1}}) + C_{(i^*_{k},j^*_{k})}$. To provide the $(k+1)^{th}$ element of $P$, we have $C_{(i_{k+1},j_{k+1})} = d(X_{i_{k+1}}, Z_{j_{k+1}}) + C_{(i_{k},j_{k})}$. Using the definition of the optimal alignment path provided in Equation 1, we have $C_{(i^*_{k+1},j^*_{k+1})} \le C_{(i_{k+1},j_{k+1})}$.

If we suppose that the remaining elements of $P$ would lead to $dist_P(X,Z) <  dist_{P^*}(X,Z)$, then this would lead to $C_{T,T} < DTW(X,Z)$, which contradicts the definition of DTW. Hence, we have $dist_P(X,Z) \le  dist_{P^*}(X,Z)$ implying that $dist_P(X,Z) > DTW(X,Z)$ is absurd.
\end{itemize}
Therefore, if we upper-bound $dist_P(X,Z)$ by $\delta$ for any given $P$, then we guarantee that $DTW(X,Z) \le \delta$.

\subsection{Proof of Theorem 2}
\textit{For a given input space $\mathbb{R}^{n \times T}$ and a random alignment path $P_{rand}$, the resulting adversarial example $X_{adv}$ from the minimization over $dist_{P_{rand}}(X,X_{adv})$ is equivalent to minimizing over $DTW(X,X_{adv})$. For any $X_{adv}$ generated by DTW-AR using $P_{rand}$, we have:
\begin{equation}
\footnotesize
\begin{cases}
    \texttt{PathSim}(P_{rand}, P_{DTW}) = 0 \;\; \& \\
    dist_{P_{rand}}(X,X_{adv}) = DTW(X,X_{adv})
\end{cases}
\label{eq:dtwgap}
\end{equation}
where $P_{DTW}$ is the optimal alignment path found using DTW computation between $X$ and $X_{adv}$.}

Let $P_{rand}$ be the random alignment path over which the algorithm would minimize  $dist_{P_{rand}}(X,X_{adv})$. For the ease of notation, within this proof, we will refer to $X_{adv}$ by $X'$.

We have  $dist_{P_{rand}}(X,X') =  \sum_{(i,j)\in P_{rand}}d(X_i,X'_j)$. As $\forall i,j, d(X_i,X'_j) \ge 0$, then minimizing $dist_{P_{rand}}(X,X')$ translates to minimizing each $d(X_i,X'_j)$. 

Let us denote $\min d(X_i,X'_j)$ by $d_{min}(X_i,X'_j)$, then $\min dist_{P_{rand}}(X,X') = \sum_{(i,j)\in P_{rand}}d_{min}(X_i,X'_j)$.

Using the back-tracing approach of DTW to define the optimal alignment path, we want to verify if $\texttt{PathSim}(P_{rand}, P_{DTW})=0$.
Let $P_{rand}$ be the sequence of cells $\{c_{k, l}\}$ and $P_{DTW}$ be the sequence $\{c_{k', l'}\}$. Every cell $\{c_{k', l'}\}$ in $P_{DTW}$ is defined to be the successor of one of the cells $\{c_{k'-1, l'}\}$, $\{c_{k', l'-1}\}$, $\{c_{k'-1, l'-1}\}$ which will make the distance sum along $P_{DTW}$ until the cell $\{c_{k', l'}\}$ be the minimum distance. As we have minimized the distance over the path $P_{rand}$ to be $d_{min}(X_i,X'_j)$, the cells of $P_{DTW}$ and $P_{rand}$ will overlap. This is due to the recursive nature of DTW computation and the fact that the last cells in both sequences $P_{DTW}$ and $P_{rand}$ is the same (by the definition of DTW alignment algorithm).

Hence, $\forall (k,l) \in P_{rand} , (k',l')\in P_{DTW}$, we have $k=k'$ and $l=l'$.

Therefore, the optimal alignment between between $X$ and $X_{adv}$ will overlap with $P_{rand}$ and we obtain $\texttt{PathSim}(P_{rand}, P_{DTW})=0$.

\subsection{Proof of Corollary 1}
\textit{Let $P_1$ and $P_2$ be two alignment paths such that $\texttt{PathSim}(P_1, P_2) > 0$. If $X_{adv}^1$ and $X_{adv}^2$ are the adversarial examples generated using DTW-AR from any given time-series $X$ using paths $P_1$ and $P_2$ respectively such that $DTW(X, X_{adv}^1) = \delta$ and $DTW(X, X_{adv}^2) = \delta$, then $X_{adv}^1$ and $X_{adv}^2$ are not necessarily the same.}

Let $P_1$ and $P_2$ be two alignment paths such that $\texttt{PathSim}(P_1, P_2)>0$. We want to create an adversarial example from time-series signal $X$ using one given alignment path. When using $P_1$, we will obtain $X_{adv,1}$ such that $DTW(X, X_{adv,1}) = \delta$, and when using $P_2$, we will obtain $X_{adv,2}$ such that $DTW(X, X_{adv,2}) = \delta$.

To show that $X_{adv,1}$ and $X_{adv,2}$ are more likely to be different, let us suppose that given $P_1$ and $P_2$, we always have $X_{adv,1}=X_{adv,2}$.

Again, to simplify notations, let us notate $X_{adv,1}$ by $Z$ and $X_{adv,2}$ by $Z'$ for this proof. As we have  $DTW(X,Z) = DTW(X, Z') = \delta$, then $\sum_{(i,j)\in P_1}d(X_i,Z_j)=\sum_{(i,j)\in P_2}d(X_i,Z'_j)$. If we suppose that by construction $Z$ is always equal to $Z'$, this means that for $Z \neq Z'$, the statement $\sum_{(i,j)\in P_1}d(X_i,Z_j)=\sum_{(i,j)\in P_2}d(X_i,Z'_j)$ does not hold. The last claim is clearly incorrect. Let us suppose that $Z$ is pre-defined and we assume $Z \neq Z'$. Let the ensemble of indices $\{k\}$ refer to the indices where $Z_k \neq Z'_k$. This means that we have $k-1$ degrees of freedom to modify $Z'_k$ to fix the equality $\sum_{(i,j)\in P_1}d(X_i,Z_j)=\sum_{(i,j)\in P_2}d(X_i,Z'_j))$. 

Therefore, considering $\texttt{PathSim}(P_1, P_2)>0$, we can construct $X_{adv,1} \neq X_{adv,2}$ such that
$DTW(X, X_{adv,1}) = \delta$ and $DTW(X, X_{adv,2})$.

\end{document}